\documentclass[journal,twocolumn,10pt]{IEEEtran}
\usepackage[T1]{fontenc}
\usepackage[utf8]{inputenc}
\usepackage{amsmath,amssymb,amsfonts,amsbsy}
\usepackage{graphicx}
\usepackage[outdir=./]{epstopdf}
\usepackage[caption=false,font=footnotesize]{subfig}
\usepackage{float}               
\usepackage{stfloats}            
\usepackage{array}
\usepackage{textcomp}
\usepackage{etoolbox}
\usepackage{tikz}
\usepackage{rotating}
\usepackage{lscape}
\usepackage{placeins}
\usepackage{multirow}
\usepackage{tabularx}
\usepackage{booktabs}
\usepackage{pifont}
\usepackage[dvipsnames]{xcolor}
\usepackage{colortbl}
\usepackage{siunitx}
\usepackage{nicematrix}
\usepackage{makecell}
\usepackage{balance}
\usepackage{bibunits}

\usepackage[longnamesfirst,sort]{natbib}
\bibpunct[, ]{(}{)}{;}{a}{,}{,}

\usepackage{url}
\usepackage[hidelinks]{hyperref}
\usepackage[shortcuts,acronym]{glossaries}

\AtBeginEnvironment{tikzpicture}{%
  \fontsize{8}{12}\selectfont
  \def\tiny{\fontsize{4}{6}\selectfont}%
  \def\scriptsize{\fontsize{6}{8}\selectfont}%
  \def\footnotesize{\fontsize{7}{9.5}\selectfont}%
}

\providecommand{\tablefont}{\fontsize{8}{9}\selectfont}

\newcolumntype{C}{>{\centering\arraybackslash}X}
\newcolumntype{L}[1]{>{\centering\arraybackslash\hsize=#1\hsize\linewidth=\hsize}X}
\newcolumntype{Y}{X[c]}
\newcolumntype{Z}[1]{X[c,#1]}

\renewcommand{\thesection}{\arabic{section}}

\renewcommand{\thetable}{\arabic{table}}

\newlength{\NFwidth}
\NewDocumentCommand{\NFelement}{mmm}{\normalsize\textbf{#1} #2\hfill #3}
\NewDocumentCommand{\NFline}{O{l}m}{\footnotesize\makebox[\NFwidth][#1]{#2}}

\NewDocumentCommand{\NFentry}{sm}{%
  \makebox[.5\NFwidth][l]{\normalsize
    \IfBooleanT{#1}{\makebox[0pt][r]{\textbullet\ }}%
    #2}\ignorespaces}
\NewDocumentCommand{\NFtext}{+m}
{\parbox{\NFwidth}{\raggedright#1}}

\newcommand{\NFRULE}{\midrule[5pt]}
\newcommand{\NFRule}{\midrule[2pt]}
\newcommand{\NFrule}{\midrule}

\newacronym{cl}{CL}{Cognitive Load}
\newacronym{hci}{HCI}{Human-Computer Interaction}
\newacronym{hrv}{HRV}{Heart-Rate Variability}
\newacronym{mwl}{MWL}{Mental Workload}
\newacronym{ecg}{ECG}{Electrocardiography}
\newacronym{eda}{EDA}{Electrodermal Activity}
\newacronym{ppg}{PPG}{Photoplethysmography}
\newacronym{emg}{EMG}{Electromyography}
\newacronym{et}{ET}{Eye Tracker}
\newacronym{rsp}{RSP}{Respiration}
\newacronym{skt}{SKT}{Skin Temperature}
\newacronym{revelio}{REVELIO}{\textbf{R}obust \textbf{E}stimation \textbf{V}ia %
\textbf{E}nd-To-End \textbf{L}earning of Mult\textbf{i}modal \textbf{O}bservations}
\newacronym{eeg}{EEG}{Electroencephalography}
\newacronym{erp}{ERP}{Event-Related Potential}
\newacronym{mri}{MRI}{Magnetic Resonance Imaging}
\newacronym{fnirs}{fNIRS}{Functional Near-Infrared Spectroscopy}
\newacronym{pet}{PET}{Positron emission tomography}
\newacronym{facs}{FACS}{Facial Action Coding System}
\newacronym{xgb}{XGB}{Extreme Gradient Boosting}
\newacronym{ocean}{OCEAN}{Openness, Conscientiousness, Extraversion, Agreeableness, Neuroticism}
\newacronym{bfik}{BFI-K}{\textbf{B}ig \textbf{F}ive \textbf{I}nventory - Short (German: \textbf{K}urzform)}
\newacronym{bmi}{BMI}{Body Mass Index}
\newacronym{tlx}{TLX}{Task Load Index}
\newacronym{isa}{ISA}{Instantaneous Self-Assessment}
\newacronym{rtlx}{RTLX}{Raw Task Load Index}
\newacronym{pogqsf}{POGQ-SF}{Problematic Online Gaming Questionnaire Short Form}
\newacronym{adabase}{ADABase}{Autonomous Driving Cognitive Load Assessment Database}
\newacronym{scp}{SCP}{Standard communication protocol}
\newacronym{dtw}{DTW}{Dynamic Time Warp}
\newacronym{stft}{STFT}{Short Time Fourier Transform}
\newacronym{ema}{EMA}{Exponential Moving Average}
\newacronym{pirl}{PIRL}{Pretext-Invariant Representation Learning}
\newacronym{cam}{CAM}{Class-Activation-Maps}
\newacronym{rnn}{RNN}{recurrent neural network}
\newacronym{cnn}{CNN}{convolutional neural network}
\newacronym{lstm}{LSTM}{Long Short-Term Memory}
\newacronym{xlstm}{xLSTM}{extended \gls{lstm}}
\newacronym{slstm}{sLSTM}{scalar memory \gls{lstm}}
\newacronym{mlstm}{sLSTM}{matrix memory \gls{lstm}}
\newacronym{ffn}{FFN}{feed-forward network}
\newacronym{fcn}{FCN}{fully convolutional network}
\newacronym{flops}{FLOPs}{floating point operations}
\newacronym{auroc}{AUROC}{area under the receiver operating characteristic curve}
\newacronym{tst}{TST}{time series transformer}
\newacronym{ece}{ECE}{Expected Calibration Error}
\newacronym{nll}{NLL}{negative log-likelihood}
\newacronym{au}{AU}{Action Unit}
\newacronym{fdr}{FDR}{false discovery rate}

\begin{document}

\title{REVELIO: Multimodal Task Load Estimation for Cross-Domain Generalization}

\author{Maximilian~P.~Oppelt\textsuperscript{a,b}, Andreas~Foltyn\textsuperscript{a},
Nadine~R.~Lang-Richter\textsuperscript{a}, and Bjoern~M.~Eskofier\textsuperscript{b,c,d,e,f}%
\thanks{\textsuperscript{a}Department Digital Health and Analytics, Fraunhofer Institute for Integrated Circuits IIS,
Erlangen, Germany; \textsuperscript{b}Department Artificial Intelligence in Biomedical Engineering,
Friedrich-Alexander-Universit\"{a}t Erlangen-N\"{u}rnberg, Erlangen, Germany; \textsuperscript{c}Chair of AI-supported
Therapy Decisions, LMU M\"{u}nchen, Munich, Germany; \textsuperscript{d}Institute of AI in Medicine, LMU Hospital,
Munich, Germany; \textsuperscript{e}Institute of AI for Health, Helmholtz Zentrum M\"{u}nchen, Neuherberg, Germany;
\textsuperscript{f}Munich Center for Machine Learning (MCML), Munich, Germany.}%
\thanks{Maximilian P. Oppelt is the main contributing and corresponding author (e-mail:
maximilian.oppelt@iis.fraunhofer.de).}}

\markboth{Oppelt \MakeLowercase{\textit{et al.}}: REVELIO: Multimodal Task Load Estimation for Cross-Domain Generalization}%
{Oppelt \MakeLowercase{\textit{et al.}}: REVELIO: Multimodal Task Load Estimation for Cross-Domain Generalization}

\maketitle

\begin{abstract}
Adaptive interactive systems that respond to user mental state require task load estimation that stays reliable outside
its training context. Models built in one setting rarely transfer, limiting real-world use. This work presents
\textbf{R}obust \textbf{E}stimation \textbf{V}ia \textbf{E}nd-to-End \textbf{L}earning of Mult\textbf{i}modal
\textbf{O}bservations (REVELIO), a unified framework, dataset and evaluation protocol for cross-domain task load
estimation. Cardiac, electrodermal, muscle, respiration, skin temperature, eye tracking and facial signals were
recorded in 96 sessions, spanning a laboratory working memory task, a driving simulator and two games under consistent
hardware and annotation. Multimodal fusion reached up to 0.87 area under the receiver operating characteristic curve,
above single input eye tracking (0.83), facial action units (0.74) and biosignals (0.62). Multi-domain training stayed
calibrated on domains seen in training, whereas removing the target domain caused the classifier to collapse in some
applications. REVELIO provides a standardized foundation for measuring modality contributions and cross-domain
generalization toward deployable workload-aware systems.
\end{abstract}


\section{Introduction}
Interactive systems that adapt to user cognitive state can reduce errors, prevent adverse events, and improve
usability across everyday and safety-critical contexts \citep{longoHumanMentalWorkload2022}. Realizing such
workload-aware interfaces requires accurately estimating human mental workload from continuous biomedical and
behavioral signals. Signal-based workload measurement can guide the adaptation of interfaces and improve
performance and safety. Yet this estimation remains challenging because workload models trained in one
interaction context rarely generalize to new applications.
Scholars have developed a theoretical foundation in psychology characterizing mental workload
as a complex construct that is dynamic, nonlinear, person-specific, and multidimensional. It is closely linked to
attention and effort. However, these theories offer intertwined perspectives that cannot be disentangled when
studying mental workload.

\emph{Cognitive Load Theory} by \citet{swellerCognitiveLoadTheory2011}, for instance, distinguishes
intrinsic load from extraneous and germane load, recognizing task complexity as a
significant factor. In contrast, \emph{Flow Theory} \citep{davisBoredomAnxietyExperience1977} focuses on
the perceived demand as a critical characteristic. \emph{Arousal Theory} \citep{yerkesRelationStrengthStimulus1908}
emphasizes maintaining arousal at an optimal moderate level to ensure peak performance.
\emph{Multiple Resource Theory} \citep{wickensMultipleResourcesPerformance2002} highlights the
constraints of limited attentional resources and the finite nature of attention and cognition. An extensive
analysis of these theoretical frameworks shows that they complement each other and capture different aspects
of mental workload \citep{longoHumanMentalWorkload2022}.
In addition to these theoretical constructs, practitioners studying mental workload in experimental settings
typically employ measures drawn from four distinct groups: \emph{self-reported subjective}
measures, such as questionnaires that quantitatively report personal experiences, \emph{behavioral measures},
including gaze or certain facial expressions, \emph{physiological measures}, such as
changes in metrics like heart rate, heart rate variability, and respiration, as well as pupil
changes and \emph{performance measures}, such as hit rate or reaction times \citep{chenRobustMultimodalCognitive2016}.
These measurements have been used to explore cognitive load in various contexts. For instance, tests
like the $n$-back test are meticulously designed to induce controlled levels of task load. Practitioners have also
conducted application-specific studies aiming to enhance safety in driving or to improve long-term user engagement and
learning efficiency in video games by managing the player's task load
\citep{oppeltADABaseMultimodalDataset2022, tagalidouFindingSweetSpot2025, zhangEvaluatingPhotoplethysmogramRealTime2018}.
Although these applications all relate to cognitive load, each tends to emphasize distinct facets of what is
inherently a multidimensional construct. As a result, detection models developed for one context are often limited in
their applicability and cannot be readily transferred to other domains. One key limitation stems from inconsistencies
in the types of input data collected across different settings. For example, in video game environments, task load can
be assessed consistently without the influence of physical movement or changes in the surrounding environment.
By contrast, evaluating cognitive load in lower-limb prosthesis users during ambulatory activities requires
capturing data during diverse physical actions such as sitting, walking, and running across a range of environments
\citep{manzUsingMobileEye2024}. Participant characteristics, such as skill and experience, may also serve as
confounding variables. In driving studies, selecting drivers based on years of holding a driver's license can
significantly impact task load levels. Similarly, in video game research, the estimated number of hours of play
can influence induced mental workload. Experimental protocols may introduce further confounds, such as the induction
of emotions with varying valence and arousal. All of these factors can influence the measurement and interpretation
of task load. Throughout this paper we distinguish three related terms, as established in
Section~\ref{ssec:cognitive-load}: \emph{cognitive load} is the broad, multidimensional psychological construct,
\emph{mental workload} refers to the inclusive synthesis of \citet{longoHumanMentalWorkload2022} and \emph{task load}
denotes the demand a task imposes on the operator, controlled experimentally through task-difficulty levels. Because
cognitive load cannot be measured directly, all models in this work estimate \emph{task load}, derived from
predetermined task-difficulty levels (e.g., 1-back vs.\ 3-back), rather than cognitive load directly. Given the
multidimensional and context-dependent nature of cognitive load, this study is guided by the
following research questions: First, to what extent can commonly used continuous measurements predict task load in
levels imposing varying mental demands (\textbf{RQ1})? Second, do unimodal and multimodal inputs contribute differently
across various tasks (\textbf{RQ2})? Third, how effectively do task load detection systems generalize from one
application (e.g. trained on laboratory test) represented in the training data to other, distinct application
scenarios (e.g., real-world \gls{hci} applications) (\textbf{RQ3})?
\begin{figure*}[!t]
  \centering
  \input{./images/revelio-overview.tex}
  \caption{Overview of the \gls{revelio} evaluation framework. Models are evaluated across real-world applications and %
a well-studied laboratory experiment, estimating cross-domain generalization from uni- and multimodal continuous %
signals across multiple architectures.}
  \label{fig:number_one}
  \vspace{-0.15cm}
\end{figure*}
To address these questions, we take a step towards a more universally applicable task load detection system with our
proposed framework, \gls{revelio}. The framework depicted in Figure~\ref{fig:number_one} introduces two new practical
\gls{hci} application scenarios for task load estimation, comprising recordings from two widely played video games,
\emph{Overcooked! 2} and \emph{Hogwarts Legacy}, alongside an $n$-back task. These games were selected because, beyond
their commercial popularity, they impose qualitatively distinct cognitive demands: \emph{Overcooked! 2} taxes executive
function through planning, task switching, and dual-task coordination under time pressure, whereas
\emph{Hogwarts Legacy} taxes visuospatial processing and continuous motor coordination. Together with the $n$-back
task, the three applications of the gaming study span complementary regions of cognitive demand space \citep{swellerCognitiveLoadTheory2011},
enabling a test of whether task load models capture domain-general physiological signatures or merely task-specific
artifacts.
We extend the publicly available ADABase dataset \citep{oppeltADABaseMultimodalDataset2022}, which comprises 51
driving-simulator sessions from 51 participants including an n-back baseline, with 45 newly collected gaming sessions
from 37 participants, maintaining consistency in measurement modalities, recording equipment, and $n$-back test
protocols. In total, \gls{revelio} unifies these 96 multimodal sessions from 83 unique participants across four
applications.
Multimodal physiological recordings are time- and cost-intensive to acquire, so comparable task-load datasets
usually include only a few dozen participants \citep{chenRobustMultimodalCognitive2016, gjoreskiMachineLearningEndtoEnd2020}.
With 83 participants, \gls{revelio} is therefore one of the largest datasets in the domain.
This foundation
enables us to build the framework’s evaluation protocol, in which we first train and then evaluate state-of-the-art
classification architectures using an end-to-end learning paradigm, enabling the assessment of models across multiple
scenarios and input modalities. Finally, we systematically investigate cross-domain generalization by evaluating
systems trained on one task and tested on another. This provides a basis for assessing model behavior under domain
transfer and advances workload-monitoring applications beyond questionnaire-driven protocols.
\section{Related Work}
\label{sec:related_work}
\subsection{Cognitive Load}
\label{ssec:cognitive-load}
Cognitive load is a multifaceted concept, and \citet{longoHumanMentalWorkload2022} have developed a
comprehensive definition that synthesizes various aspects from empirical cognitive load research, integrating
several key components relevant to our work. Both \emph{environmental and situational} factors, as well as
individual \emph{internal characteristics}, such as task fluency (e.g., prior gaming or driving experience)
and the subject's intelligence \citep{leungEffectsVirtualIndustrial2010, paasCognitiveLoadMeasurement2003},
may contribute to the overall construct. Furthermore, various definitions of cognitive load account for the amount
of \emph{attention and effort} a person dedicates to a task \citep{paasCognitiveLoadMeasurement2003,%
cainReviewMentalWorkload2007}. These resources are not always static. Therefore, the execution of tasks over
\emph{time} is a vital element, especially in primary and secondary task scenarios where \emph{temporal shifts} in
attention can impact the primary task's performance \citep{rizzoModelingMentalWorkload2016,%
moustafaAssessmentMentalWorkload2017}. One must also acknowledge the \emph{solution strategies} individuals
use to manage task demands \citep{haapalainenPsychophysiologicalMeasuresAssessing2010,
hancockExperimentalEvaluationModel1993}. The cognitive system has a \emph{finite pool}
of resources with \emph{limited capacity} \citep{gavasEstimationCognitiveLoad2017, hagaEffectsTaskDifficulty2002}. 
Additionally, the degree of activation is often dependent on the individual and can be influenced by experience, such
as in driving \citep{paasCognitiveLoadMeasurement2003, gavasEstimationCognitiveLoad2017, wangUsingWirelessEEG2016}.
Finally, the nature of the \emph{task} itself plays a significant role in determining cognitive load. We utilize this
aggregated comprehensive definition to guide our work as a foundation for our dataset creation, model development, and
evaluation procedures by aiming to capture the complexity of task load, the demand component of cognitive load,
in real-world scenarios.
Throughout this paper we distinguish three related terms. \emph{Cognitive load} refers to the broad, multidimensional
psychological construct encompassing intrinsic, extraneous, and germane components as described by the theories above.
Because cognitive load cannot be measured directly, empirical studies must operationalize it through observable proxies.
\emph{Task load} denotes the demands that a task imposes on the operator and is the facet of cognitive load that is
controlled experimentally through task-difficulty levels, that is captured by instruments such as the NASA \gls{tlx}
\citep{hartDevelopmentNASATLXTask1988}. Because our binary labels are derived from predetermined task-difficulty levels
(e.g., 1-back vs.\ 3-back) rather than from moment-to-moment measurement of a participant's internal cognitive state,
the models developed in this work estimate \emph{task load}, not cognitive load in its full theoretical sense. We
use \emph{mental workload} only when referring to the inclusive definition of \citet{longoHumanMentalWorkload2022}
that synthesizes multiple theoretical perspectives.
\begin{table*}[!t]
  \centering
  \tablefont
  \caption{Empirical studies and datasets for task load estimation in various contexts with different stimuli and
  underlying psychological concepts and their respective measurements and recording setups.}
  \label{tab:related-work}
  \setlength{\tabcolsep}{3pt}
  \renewcommand{\arraystretch}{1.1}
  \begin{tabularx}{\textwidth}{@{} L{1.15} L{1.0} L{0.85} L{0.8} L{1.1} L{0.7} L{0.7} L{1.25} L{0.85} @{}}
  \toprule
  Ref. & Physiological & Subjective & Behavior & Performance & Subjects & Setup & Stimulus & Concept \\
  \midrule
    \citep{yannakakisAffectiveCameraControl2010}
    & EDA, HR
    & -
    & -
    & in game score
    & 8F, 28M
    & gaming setup
    & maze video game
    & challenge, anxiety, boredom
    \\ \hline
    \citep{linEvaluatingUsabilityBased2006}
    & ET, ECG
    & NASA-TLX
    & ET, mouse clicks
    & game score
    & 1F, 9M
    & desktop workspace
    & action puzzle video game
    & cognitive load
    \\ \hline
    \citep{karpouzisPlatformerExperienceDataset2015}
    & -
    & Scaling and forced choice test
    & AUs, head movement
    & duration per level
    & 30F, 28M
    & --
    & jump and run video game
    & engagement, frustration
    \\ \hline
    \citep{heHighCognitiveLoad2019, kumarClassificationDriverCognitive2022}
    & EEG, ECG, EDA, ET
    & NASA-TLX
    & ET
    & mean and std.\ of speed
    & 15F, 18M
    & driving simulator
    & driving tasks and modified $n$-back
    & cognitive load
    \\ \hline
  \citep{gjoreskiDatasetsCognitiveLoad2020}
    & ACC, EDA, TEMP, PPG
    & NASA-TLX
    & -
    & time on task, number of answers, game points
    & 46 (23 per task)
    & lab
    & standardized tests ($n$-back), video game
    & cognitive load
    \\ \hline
    \citep{meteierClassificationDriversWorkload2021}
    & ECG, EDA, RESP
    & NASA-TLX
    & -
    & non-driving related task performance
    & 49F, 40M, 1 other
    & driving simulator
    & driving \& verbal backward counting
    & cognitive load
    \\ \hline
  \citep{aygunInvestigatingMethodsCognitive2022}
    & EEG, ET, EDA, RSP, fNIRS, BP, SpO2
    & -
    & ET
    & stimulus onset asynchrony
    & 36\textsuperscript{$\dagger$}F, 46M
    & driving simulator
    & questions and braking events
    & cognitive load, distraction
    \\ \hline
    \citep{beaudoin-gagnonFUNiiDatabasePhysiological2019}
    & ECG, EDA, RSP, EMG, ET
    & NASA-TLX, Game experience
    & ET, AUs, head movement
    & -
    & 36F, 183M
    & gaming setup
    & action adventure video game
    & cognitive load, arousal, valence
    \\ \hline
  \citep{behMAUSDatasetMental2021}
    & ECG, EDA, PPG
    & NASA-TLX
    & -
    & invalid responses, response time
    & 2F, 20M
    & desktop workspace
    & $n$-back
    & cognitive load
    \\ \hline
    \citep{smerdovCollectionValidationPsychophysiological2021}
    & EMG, GSR, ET, HR, SpO2, TEMP, EEG
    & self-reported performance, mental workload
    & ET, body, head, mouse, keyboard
    & match outcomes, game metrics
    & 10M
    & multi-person gaming setup
    & online multiplayer game
    & mental workload
    \\ \hline
  \citep{oppeltADABaseMultimodalDataset2022}
    & ECG, EDA, EMG, ET, PPG, RESP, TEMP
    & NASA-TLX
    & ET, AUs
    & hit rate, reaction time, in-domain performance
    & 24F, 26M, 1 n/a
    & driving simulator
    & $n$-back and Multi-Tasking, Driving
    & cognitive load
    \\ \hline
    \citep{kuttBIRAFFE2MultimodalDataset2022}
    & ECG, EDA
    & emotion evaluation valence-arousal faces
    & -
    & -
    & 33F, 70M
    & gaming setup
    & emotionally-evocative images, jump/puzzle game
    & valence, arousal
    \\ \hline
    \citep{joMOCASMultimodalDataset2025}
    & BVP, GSR, HR, IBI, SKT, EEG
    & NASA-TLX, ISA, SAM
    & ET, AUs
    & detection score
    & 7F, 14M
    & gaming setup
    & CCTV observation game
    & cognitive load, emotion
    \\ \hline
    \bottomrule
  \multicolumn{9}{l}{\textsuperscript{$\dagger$}\footnotesize{Original female participant percentages converted to %
the nearest absolute integer for comparative consistency across studies.}} \\%
  \end{tabularx}
\end{table*}
\subsection{Advancing \gls{hci} Foundation}
The complex nature of \gls{hci} necessitates innovative solutions to enhance and refine the interplay between humans
and technology. Researchers continue to make significant advances in developing cognitive models based on robust
theoretical frameworks \citep{rogersHCITheoryClassical2012,colleyEffectsUrgencyCognitive2023,
dasShiftingFocusHCEye2024,tyackSelfDeterminationTheoryHCI2024}. However, many of these principles rely on knowing
the internal human state, which is commonly assessed through subjective rating approaches or task performance metrics
that carry the inherent limitations discussed above.
In addition to such methods, multimodal physiological and behavioral signals are increasingly used to analyze user
states during computer interaction. Pioneering research in \gls{hci} highlighted the challenges of ensuring
experimental setups avoid inadvertently altering user states and addressing the inherent difficulties in data
collection and accurate annotation \citep{picardAffectiveComputingHCI1999}.
Objective continuous physiological measures have been shown to reliably capture variations in task load within
controlled environments. However, accurate and continuous task load assessments require advanced computational
techniques, such as training non-linear function approximators \citep{wilsonOperatorFunctionalState2003,
haapalainenPsychophysiologicalMeasuresAssessing2010, mehlerSensitivityPhysiologicalMeasures2012, barralNoNeedLaugh2017,
maiorWorkloadAlertsUsing2018}. Advancements in another direction include the adoption of cost-effective sensor systems
and wearable devices, facilitating monitoring in more natural environments while ensuring practical implementation
\citep{gjoreskiMyWatchSays2018,oppeltComprehensiveSignalQuality2025}.
Unfortunately, a persistent challenge in this domain is the lack of model generalization. Models trained in specific
conditions often fail to perform robustly in new or unseen scenarios. Recent advancements to improve generalization
include domain adaptation, such as transfer learning techniques to align source and target domains, and data fusion
approaches integrating multimodal data sources for robustness \citep{apicellaDomainAdaptationFear2025,
foltynEvaluatingRobustnessMultimodal2024}. Furthermore, leveraging large-scale, diverse datasets not only improves
model reliability but also enhances the design of versatile feature extractors \citep{dingCLERAUnifiedModel2023}.  
Single-modality workload estimators degrade under corrupted or delayed inputs, and \citet{debieMultimodalFusionObjective2021}
report that multimodal fusion mitigates sensor-specific failures and improves real-time estimation accuracy in
\gls{hci}. Eye-tracking offer signals correlated with perceived load, enabling adaptive
interfaces in realistic settings \citep{sevcenkoTheorybasedApproachAssessing2023}, while cross-application evaluation
identifies features with a promising generalization performance \citep{mukeMachineLearningTechniques2025}. Complementarily,
behavioral measures such as keystroke dynamics, mouse tracking, and body pose, modeled via probabilistic and machine
learning estimators, yield viable low-interference workload indicators during direct \gls{hci}
\citep{elkin-frankstonAreBehavioralMeasures2017}. A systematic decision framework further maps cognitive load
measurement methods to interface types including software and robotics clarifying trade-offs and guiding reliable
usability evaluations \citep{darejehCriticalAnalysisCognitive2024}.
\subsection{Empirical Studies in Human State Detection}
The research community has built a substantial corpus of empirical studies and collected several datasets that
have facilitated the investigation of task load estimation in various contexts, focusing on different measurements,
annotation strategies, underlying concepts, and experimental \gls{hci} settings. Recent review papers have summarized
the state-of-the-art for multiple applications and stimuli. The most prominent applications are driving, healthcare,
the design of novel human-computer interfaces, and aircraft systems. Media, design, and video games are emerging
but less explored areas \citep{longoHumanMentalWorkload2022}. However, video games are extensively examined for concepts
like valence/arousal, boredom and frustration \citep{melhartArousalVideoGame2022}.
We have collected related work for our applications in driving and video gaming in Table~\ref{tab:related-work}.
The identified driving datasets are often recorded in driving simulators, utilizing physiological signals like
\gls{ecg}, \gls{eda}, \gls{emg}, \gls{et}, \gls{ppg}, \gls{rsp}, \gls{skt}
\citep{meteierClassificationDriversWorkload2021,aygunInvestigatingMethodsCognitive2022,gjoreskiDatasetsCognitiveLoad2020,%
kumarClassificationDriverCognitive2022,behMAUSDatasetMental2021,oppeltADABaseMultimodalDataset2022,%
jiangUsingMultimodalMethods2025}, while video
game datasets mostly use subjective feedback and behavioral measures like gaze from eye tracking or body 
movement and facial expressions from video recordings \citep{linEvaluatingUsabilityBased2006,%
beaudoin-gagnonFUNiiDatabasePhysiological2019,yannakakisAffectiveCameraControl2010,%
karpouzisPlatformerExperienceDataset2015,kuttBIRAFFE2MultimodalDataset2022,%
smerdovCollectionValidationPsychophysiological2021}. A commonality of these datasets is the use of the NASA-\gls{tlx}
questionnaire to assess task load \citep{hartDevelopmentNASATLXTask1988}. Similarly, they utilize the $n$-back test
as a task-load-inducing paradigm \citep{kirchnerAgeDifferencesShortterm1958} as baseline. Typically, studies are designed
to induce increasing levels of task load and utilize \emph{physiological signals} and \emph{behavioral measures} to
predict this specific task load level while utilizing \emph{performance measures} and \emph{subjective self-ratings} to
validate the predictions in their laboratory settings.
We build our experimentation setup on a multimodal dataset called \gls{adabase} where subjects participated in both a
driving simulator and an $n$-back task \citep{oppeltADABaseMultimodalDataset2022}. The study includes multiple
physiological and behavioral signals for increasing levels of task load, while the subjects were asked to rate their
task load using the NASA-\gls{tlx} questionnaire and the performance changes were recorded. Although that study
provided useful multimodal data, it is limited to a single application.
\subsection{Multimodal Domain Generalization}
\label{ssec:mmdg}
Domain generalization trains a model on several source domains and tests it on an unseen target domain
\citep{zhouDomainGeneralizationSurvey2022}. Under a common protocol, carefully tuned empirical risk minimization, that
is, standard supervised training on the pooled source domains, performs on par with dedicated domain generalization
methods \citep{gulrajaniSearchLostDomain2021}. Multimodal domain generalization extends this setting to several input
streams. SimMMDG splits the features of each modality into shared and modality-specific parts and aligns them with
supervised contrastive learning and cross-modal translation \citep{dongSimMMDGSimpleEffective2023}. Later work weights
the contrastive alignment of shared and modality-specific features adaptively
\citep{liMultimodalDomainGeneralization2026}, handles unknown classes in the target domain
\citep{dongMultimodalOpenSetDomain2025}, learns from few labels \citep{liMultimodalDomainGeneralization2026a}, and
balances the gradients of strong and weak modalities \citep{liBalancingMultimodalDomain2026}. A recent benchmark found
that, under fair comparison, these methods improve only marginally over empirical risk minimization, and that most
evaluations address action recognition \citep{dongAreWeMaking2026}.
In physiological computing, generalization is mostly studied as transfer to unseen subjects, for example with
adversarial or contrastive subject-invariant features for \gls{eeg} \citep{maReducingSubjectVariability2019,
shenContrastiveLearningSubjectInvariant2023}, with domain generalization methods adapted to \gls{ecg} and \gls{eeg}
\citep{ballasDomainGeneralizationECG2023}, or with domain adaptation within a single task
\citep{apicellaDomainAdaptationFear2025}. Transfer across interaction contexts has received less attention, for example
between desktop and virtual reality platforms for \gls{eeg}-based cognitive load assessment
\citep{zhuCrossPlatformCognitiveLoad2026}. \gls{revelio} connects the two lines of work. It does not propose a new
generalization method. Instead, it provides a standardized protocol for multimodal biosignals in which subject, cohort, and task shift occur together, and it reports empirical risk minimization baselines
under subject-wise, cross-domain, and leave-one-domain-out splits. Multimodal domain generalization methods are
therefore complementary to this work and can be evaluated on \gls{revelio} directly.
\subsection{Research Gaps}
Across task load, \gls{hci} foundations, and empirical sensing studies, a persistent gap remains between tightly
controlled lab paradigms and the noisy realities of \gls{hci} applications. Many works optimize for specialized
laboratory tasks, limiting validity and transfer. Others focus on real-world deployments with application-specific
sensors, obscuring mechanisms and hindering comparability to canonical tests such as $n$-back. Datasets frequently
include only narrow subsets of modalities and features, impeding context-agnostic prediction and
making it difficult to disentangle the genuinely multifaceted nature of cognitive load from task- or context-induced
artifacts (e.g., gaze patterns driven by the interface rather than workload). Critically for \gls{hci}, there is no
unified experimental and benchmarking framework that links real-world interface use with well-established lab
baselines, standardizes annotation and protocols, and evaluates cross-domain generalization, calibration, and
robustness (Section~\ref{ssec:mmdg}). A broad, multimodal foundation that combines physiological and
behavioral measurements under consistent procedures is needed to isolate spurious correlations, attribute modality
contributions, and support reliable, near-continuous task-load estimation that can inform adaptive,
real-world \gls{hci} systems. These gaps motivate a universal, multimodal framework that spans lab and application
settings to enable principled, comparable, and deployable workload detection.
\section{\acrlong{revelio}}
We collected a multimodal dataset for task load estimation from a sample of healthy colleagues and students
at our institute. Participation was voluntary and anonymous, with no compensation beyond regular salary. The study was
conducted in accordance with the Declaration of Helsinki and was approved prior to data collection by the Ethics
Committee of Friedrich-Alexander-Universität Erlangen-Nürnberg (FAU), Germany (approval no.\ \textit{129\_21\,B},
granted on \textit{21 April 2021}). Prior to participation, all participants received written and verbal information
about the purpose of the study, the modalities recorded (physiological signals, eye tracking, and video), and how their
data would be anonymized, stored, and used, including the conditions for sharing it with the research community. All
participants gave written informed consent, including separate written consent for the use of their data in scientific
publications and presentations and participants who did not consent to data sharing are excluded from the released
subset. Building on \citet{oppeltADABaseMultimodalDataset2022}, we extended a prior driving-simulator dataset with
application-oriented gaming experiments and included the established $n$-back test
\citep{kirchnerAgeDifferencesShortterm1958}, maintaining consistency in measurement modalities and equipment. For the
application tasks, task-load labels (low vs.\ high) were defined in close alignment with $n$-back difficulty and
verified using objective performance indicators and post-level subjective NASA-\gls{tlx} ratings
\citep{hartDevelopmentNASATLXTask1988}.
\subsection{Protocol}
We use three terms throughout. A \emph{cohort} denotes one data collection, namely the driving study (51
sessions from 51 participants) or the gaming study (45 sessions from 37 participants). An \emph{application} denotes one
task, namely the laboratory $n$-back, recorded in both cohorts, driving, \emph{Overcooked! 2}, or \emph{Hogwarts
Legacy}. A \emph{domain} denotes an evaluation subset formed from one or more applications and cohorts, for example the
gaming-study $n$-back or both games combined. Supplementary Table~\ref{tab:domain-composition} lists the sessions
and participants per domain.
Applications were selected on three grounds: \emph{ecological validity} (commercial games create sustained,
naturalistic engagement closer to real-world \gls{hci} than purpose-built laboratory stimuli
\citep{longoHumanMentalWorkload2022}), \emph{community-recognized demand} (\emph{Overcooked! 2} is widely associated
with frantic multitasking under time pressure, \emph{Hogwarts Legacy} with fast-paced visuospatial navigation, 
increasing the likelihood that difficulty levels produce genuine load variation) and \emph{complementary cognitive
profiles} (the two games differ in primary demand type, ensuring cross-domain evaluation spans qualitatively distinct
tasks).
Figure~\ref{app:study-protocol} presents the experiment task schedule, detailing the sequence and structure of all
study tasks. Each participant completed three main experiments: $n$-back, \emph{Hogwarts Legacy}, and \emph{Overcooked! 2},
administered in randomized order. The protocol incorporated structured introduction phases for health assessment,
sensor setup, and eye tracker calibration, followed by baseline or training periods to ensure participant
familiarity with each task. After every major task segment, participants completed the NASA-\gls{tlx} questionnaire to
assess subjective workload. Regularly scheduled breaks, including computer-guided meditation or relaxation, were
provided between sessions to mitigate fatigue and maintain baseline physiological states. Additional questionnaires
were administered during breaks and at the end of the session for comprehensive data collection. The task schedule also
indicates opportunities for task repetition and variable time intervals, ensuring that all participants could fully
understand and engage with each experiment. The full procedure, including setup and preparation, lasted approximately
$2-3$ hours.
\begin{figure*}
  \centering
  \noindent
\input{./images/study-protocol.tex}

\vspace{-0.05cm}
  \caption{Task schedule illustrating the randomized experimentation plan across three tasks, including potential %
repetitions, breaks, and debriefing phases. Breaks, meditation sessions, and the debriefing protocol were included. %
Meditation was computer-guided with controlled breathing. Dashed lines indicate variable time intervals for phases %
allowing participants to pause the training flight, cooking sessions, or repeat $n$-back tasks to ensure adequate %
understanding.}
  \label{app:study-protocol}
  \vspace{-0.15cm}
\end{figure*}
\begin{figure}
  \centering
  \subfloat[\centering Flight class]{\label{sfig:hogwarts-flight-class}\includegraphics[width=0.49\linewidth]{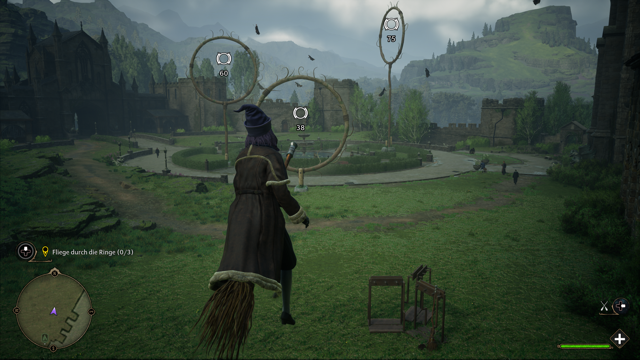}} \hfill%
  \subfloat[\centering Hogwarts gardens]{\label{sfig:hogwarts-gardens}\includegraphics[width=0.49\linewidth]{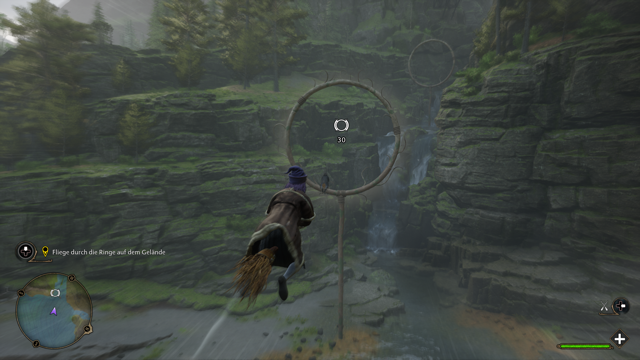}}%
\quad %
  \subfloat[\centering Flight race]{\label{sfig:hogwarts-race}\includegraphics[width=0.49\linewidth]{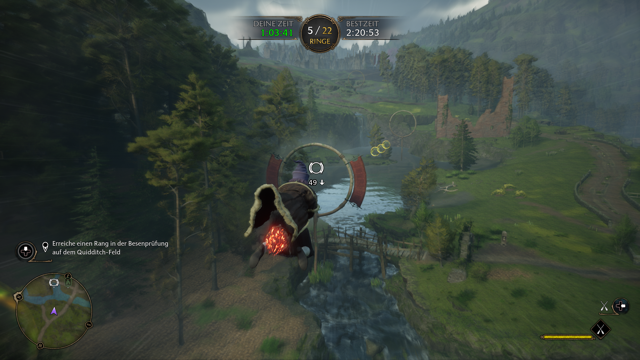}} \hfill%
  \subfloat[\centering Score]{\label{sfig:hogwarts-score}\includegraphics[width=0.49\linewidth]{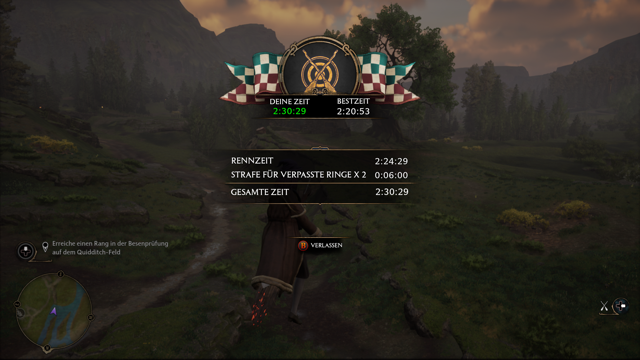}}%
  \caption{Screenshots recorded during the \emph{Hogwarts Legacy} gaming scenario. \emph{Hogwarts Legacy} \textcopyright{} Warner Bros. Games / Avalanche Software. Screenshots captured by the corresponding author.}
  \label{app:screenshots-hogwarts}
  \vspace{-0.2cm}
\end{figure}
The first experiment employed the $n$-back test, a widely used paradigm for assessing working memory and cognitive load
\citep{kirchnerAgeDifferencesShortterm1958, oppeltADABaseMultimodalDataset2022}. Three difficulty levels were
implemented, corresponding to $n \in \{1, 2, 3\}$. The driving study comprised six levels, a single-task
variant with visual stimuli (levels 1--3) and a dual-task variant with simultaneous visual and auditory stimuli
(levels 4--6), which is evaluated as a separate domain. The gaming
study used only the three single-task levels. Prior to the main experiment, participants were given the opportunity
to practice each difficulty level to ensure comprehension of the task, with additional training sessions provided upon
request. After each test session, participants evaluated their perceived task load using the NASA-\ac{tlx}
questionnaire \citep{hartDevelopmentNASATLXTask1988}. Performance was quantified by measuring hits, misses, and
false positives, as well as event times and response times.
During the \emph{Hogwarts Legacy} sessions, participants completed a series of structured gaming tasks, as illustrated
in Fig.~\ref{app:screenshots-hogwarts}. The first baseline required participants to watch a prerecorded broom-race
video while holding the controller and pressing buttons at random. This reproduced the motor activity and visual
stimulation of gameplay without introducing additional cognitive demands, thereby reducing confounds from movement
artifacts and sensory differences. In the second baseline, participants practiced broom flight around the Hogwarts
Quidditch stadium to familiarize themselves with the controls and to better match the sensorimotor and perceptual
context of subsequent tasks. A passive resting baseline was deliberately avoided to minimize spurious correlations.
Beforehand, the controller layout and functions were explained using a schematic illustration. Study personnel provided
brief verbal guidance and requested specific flying maneuvers to ensure adequate task proficiency.
Following the baselines, participants completed the ``Flying Class'' level, which involved attending an in-game lecture,
navigating rings in the Hogwarts Garden, and then approximately two minutes of guided flight through pre-positioned
rings around Hogwarts buildings while following an in-game instructor (Fig.~\ref{sfig:hogwarts-flight-class},
\subref*{sfig:hogwarts-gardens}). After completing this level, participants filled out a NASA-\gls{tlx} questionnaire.
The game did not provide performance metrics for this segment. In the final level, participants competed in a ring
race, during which the number of missed rings and the completion time were recorded as performance metrics
(Fig.~\ref{sfig:hogwarts-race}, \subref*{sfig:hogwarts-score}). A second NASA-\gls{tlx} questionnaire was administered
after the race.

\begin{figure}
  \centering
  \subfloat[\centering Game modes/levels.]{\label{sfig:overcooked-modes}%
\includegraphics[width=0.49\linewidth]{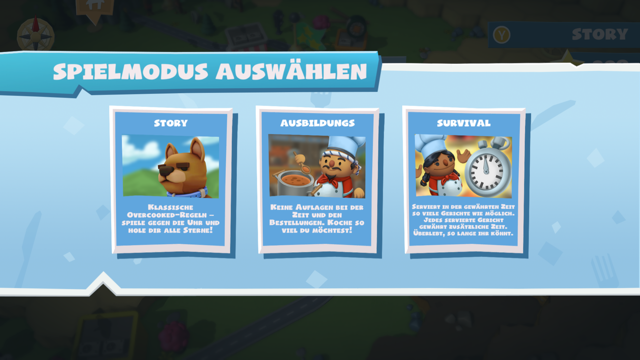}} \hfill%
  \subfloat[\centering Level 1 (education mode)]{\label{sfig:overcooked-education}%
  \includegraphics[width=0.49\linewidth]{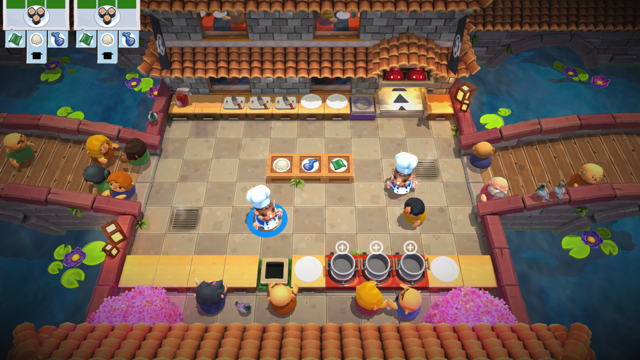}}%
\quad %
  \subfloat[\centering Sushi Level 2 (survival mode)]{\label{sfig:overcooked-survival}%
  \includegraphics[width=0.49\linewidth]{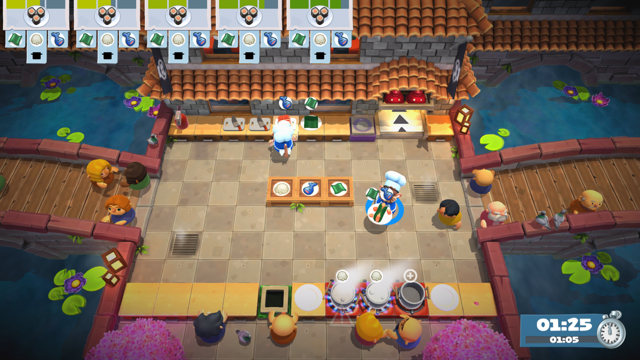}} \hfill%
  \subfloat[\centering Game performance metrics]{\label{sfig:overcooked-score}%
  \includegraphics[width=0.49\linewidth]{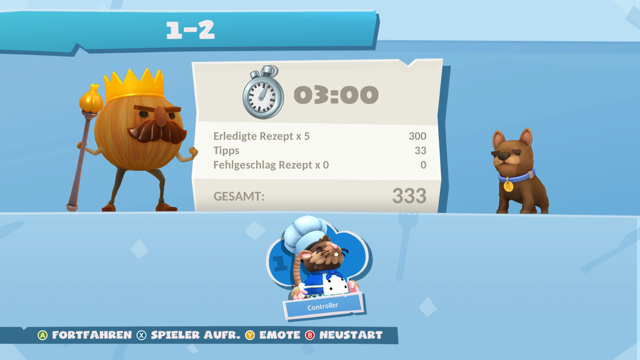}}%
  \caption{Screenshots recorded during the \emph{Overcooked! 2} gaming scenarios. \emph{Overcooked! 2} \textcopyright{} Ghost Town Games / Team17. Screenshots captured by the corresponding author.}
  \vspace{-0.2cm}
\label{app:screenshots-overcooked}
\end{figure}
The \emph{Overcooked! 2} scenarios followed a structured protocol shown in Figure~\ref{app:study-protocol}. Screenshots
are presented in Figure~\ref{app:screenshots-overcooked}. Participants first watched a prerecorded gameplay video while
randomly pressing controller buttons to replicate sensorimotor engagement as a baseline, then completed a brief
training session preparing a simple recipe to familiarize themselves with the controls. In \emph{Overcooked! 2},
players fulfill incoming food orders by picking up ingredients, processing them (e.g., chopping, cooking), assembling
the dish, and serving it before the order expires. All levels were selected from the game's mode menu
(Fig.~\ref{sfig:overcooked-modes}). The first level was conducted in education mode without a recipe time limit and
required controlling a single cook (Fig.~\ref{sfig:overcooked-education}). The second level used survival mode with one
cook (time limit, recipe queue, Fig.~\ref{sfig:overcooked-survival}), and the third enabled switching between two cooks
to support parallel task management. Performance was quantified using in-game metrics
(Fig.~\ref{sfig:overcooked-score}), and subjective workload was assessed after each level with the NASA-\ac{tlx}
questionnaire.
Table~\ref{tab:demand-profiles} summarizes the cognitive demand profiles of the three applications of the
gaming study.
\begin{table}[h]
  \centering
  \tablefont
  \caption{Cognitive demand profiles of the three applications of the gaming study.}
  \label{tab:demand-profiles}
  \begin{tabularx}{\linewidth}{lCCC}
    \toprule
      Dimension & $n$-back & Overcooked! 2 & Hogwarts Legacy \\
    \midrule
      Primary process & working memory updating & planning, task switching, monitoring & visuospatial tracking, motor coordination \\
      Input modality & visual (symbols) & visual (2D scene) & visual (3D first-person) \\
      Response type & button press & button sequences & continuous joystick \\
      Time pressure & fixed-pace & none to survival & none to race \\
      Divided attention & single-task & baseline to dual-task & dual visual routes \\
      Visuospatial complexity & minimal & 2D kitchen & 3D open-world \\
      Intrinsic load & single rule & recipes, locations, cooks & rings, obstacles \\
      Motor demands & minimal & 2D controls & 3D flight control \\
    \bottomrule
  \end{tabularx}
\end{table}
\subsection{Acquisition Setup}
\begin{figure}[!htb]
  \centering
  \subfloat[\centering Facial landmarks with image coordinates.]%
{\label{sfig:video-face}\includegraphics[width=0.49\linewidth]{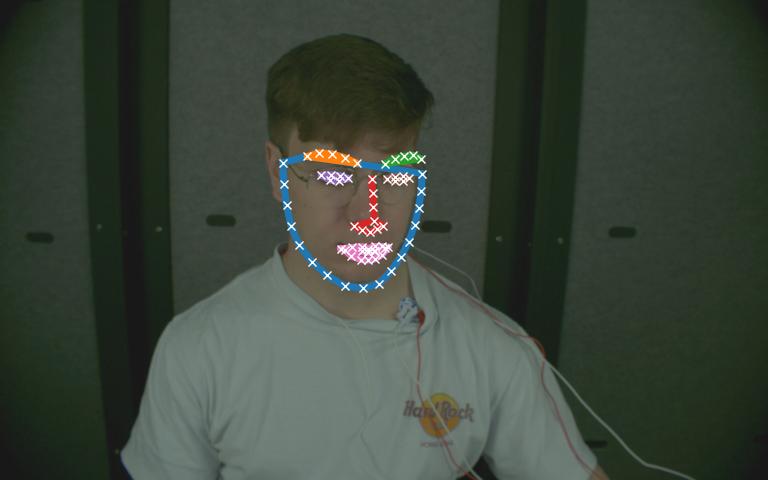}} \hfill%
  \subfloat[\centering Pose landmarks with x, y and approximated z coordinates for 3d body pose estimation of eye, ear,
nose and shoulders.]{\label{sfig:video-pose}\includegraphics[width=0.49\linewidth]{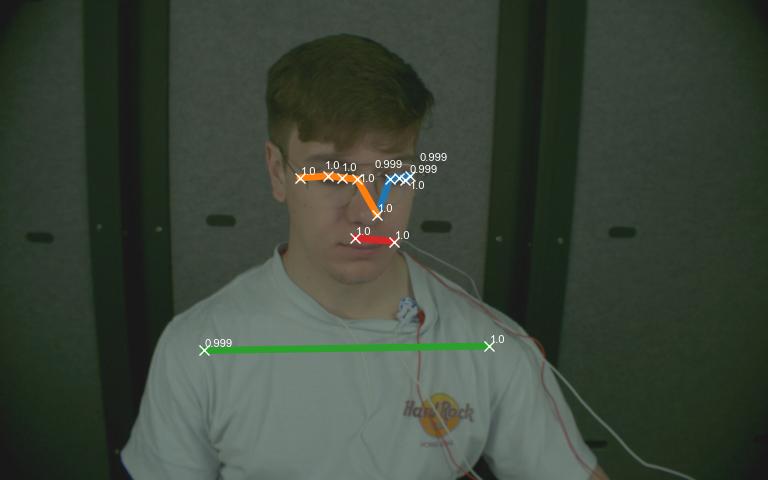}}%
\caption{Facial expressions and pose landmarks extracted from RGB camera videos for facial expression and movement
analysis.}
\label{app:video-examples}
  \vspace{-0.15cm}
\end{figure}

\begin{figure}[!htb]
  \centering
  \subfloat[\centering Controlled study setup to prevent confounding factors such as noise or active study personnel with %
constant lighting conditions for optimal video recordings.]{\label{sfig:room-lab}\includegraphics[width=0.49\linewidth]{%
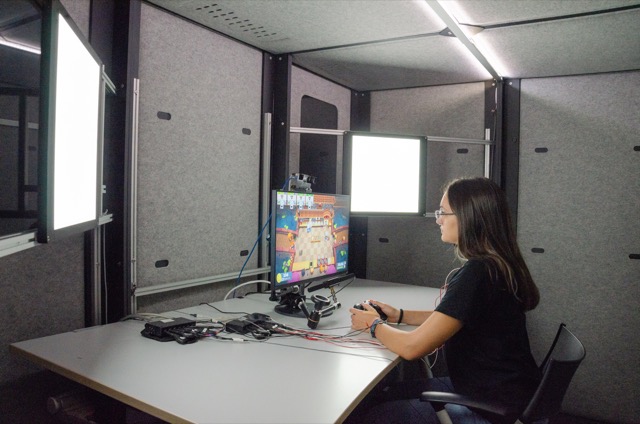}} \hfill%
  \subfloat[\centering \emph{Overcooked! 2} used with a Wireless Xbox controller as input device, subject connected %
to a BIOPAC system for biosignal monitoring, a Tobii eye tracker and a camera for facial videos.]{\label{sfig:room-overcooked}%
\includegraphics[width=0.49\linewidth]{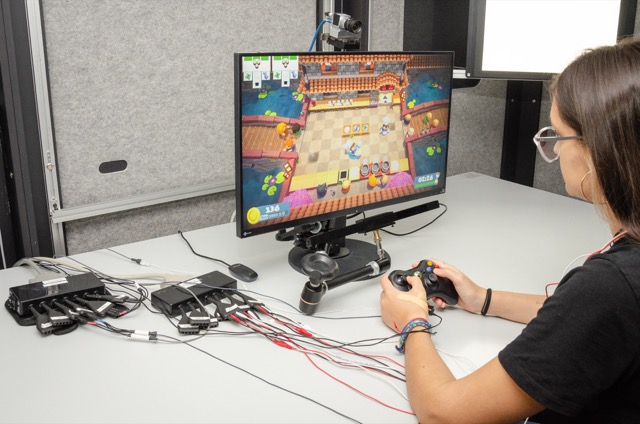}}%
\quad %
  \subfloat[\centering \emph{Hogwarts Legacy} broom flying.]{\label{sfig:room-hogwarts}\includegraphics[width=0.49\linewidth]{%
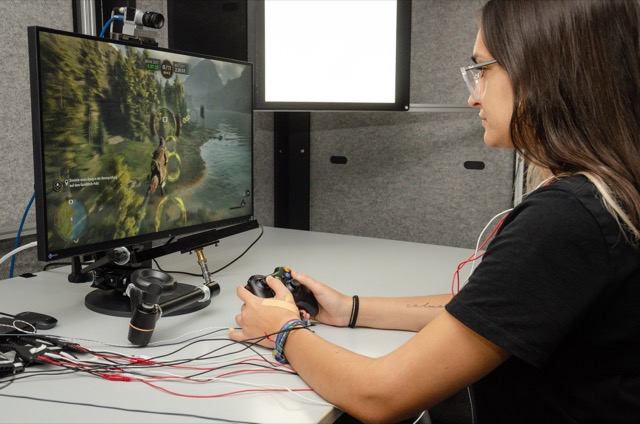}} \hfill%
  \subfloat[\centering Custom questionnaires controlled by mouse.]{\label{sfig:room-questionnaire}\includegraphics[width=0.49\linewidth]{%
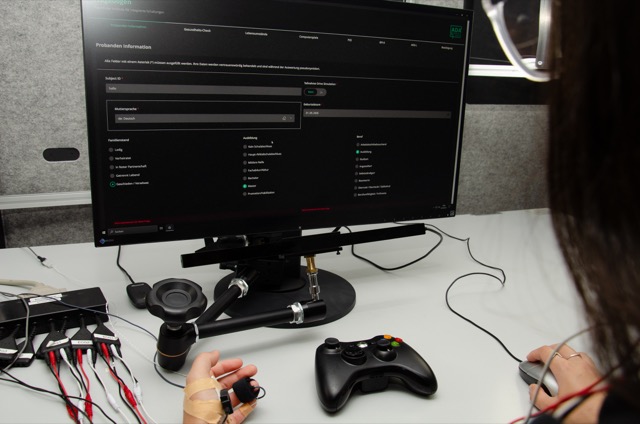}}%
  \caption{Experimentation setup in a controlled study environment enabling data recording without confounding factors and %
constant environmental conditions.}
\label{app:images-room}
\end{figure}

Consistent with prior studies \citep{oppeltADABaseMultimodalDataset2022}, we collected physiological signals using a
BIOPAC MP160 system at a sampling rate of 2000 Hz, capturing cardiac electrical and mechanical activity (\gls{ecg}, 
\gls{ppg}), activity of the sweat glands on the skin (\gls{eda}), trapezius muscle activity (\gls{emg}), respiration via a
chest belt, and skin temperature from the pinky finger. Eye movements and pupil diameter were monitored with a Tobii Pro
Fusion I5S at 250 Hz. Facial videos were acquired using a Basler acA1920 RGB camera, triggered at 25 Hz with constant 
exposure time to ensure temporal alignment with other data streams.
Video data were processed using two primary approaches, facial analysis and body pose estimation
(Fig.~\ref{sfig:video-face}, \subref*{sfig:video-pose}). Facial action units
introduced by \citet{ekmanpaulFacialActionCoding1978} and previously used in this configuration in
related research were extracted with the py-feat library \citep{cheongPyFeatPythonFacial2023}. Face detection utilized
the RetinaFace algorithm by \citet{dengRetinaFaceSingleShotMultiLevel2020}, while facial landmarks were
identified using MobileFaceNet \citep{chenMobileFaceNetsEfficientCNNs2018}. For head pose estimation, we employed the
img2pose model \citep{albieroImg2poseFaceAlignment2021}. Movement detection was performed using the MoveNet
\emph{Thunder} model \citep{belettiMoveNetUltraFast2021}, extracting key landmarks such as the eyes, ears, shoulders,
and nose. Missing landmark data, which were infrequent, were interpolated linearly, while eye tracking data missing due
to blinks were set to zero.
Ambient brightness during each task was measured with a calibrated Lux-Meter (PeakTech 5065), which approximates the
CIE spectral sensitivity, over two minutes at five-second intervals. Mean illuminance values were $53.88\pm0.78$ lux
for $n$-back, $60.17\pm0.62$ lux for \emph{Overcooked! 2}, and $46.17\pm1.57$ lux for \emph{Hogwarts Legacy},
indicating slight variations that may affect pupil diameter. While these measurements help account for potential
effects on pupil size, gaze direction and temporal dynamics must also be considered to distinguish lighting-related
changes from psychophysiological responses.
Device synchronization was achieved using a common clock. The experimental environment was controlled to minimize
confounding influences, including stable ambient lighting, soundproofing, and absence of active study personnel in
the room (Fig.~\ref{sfig:room-lab}). All tasks were performed using an Xbox controller, with the participant seated
in a comfortable chair at a fixed monitor distance and instructed to maintain a stable head position. Study personnel
monitored the session remotely via a mirrored screen. Figure~\ref{app:images-room} shows the controlled recording
room and the setup for each task, namely the two games and the questionnaires, which participants answered by mouse
(Fig.~\ref{sfig:room-overcooked}--\subref*{sfig:room-questionnaire}).
\subsection{Target Operationalization for Task-Load Estimation}
\label{ssec:target_operationalization}
Accurate task load prediction from continuous physiological and behavioral signals requires reliable ground-truth
data. However, as outlined above, cognitive load cannot be directly operationalized since individual factors preclude
any absolute quantification. We therefore use task load, as defined by the experimental design, as the annotation basis
and validate it through secondary indicators. This annotation process nonetheless presents several challenges: the
subjective and context-sensitive nature of task load, its dynamic variability, the potential intrusiveness of
continuous subjective feedback, and biases introduced by manual labeling. To address these, we propose combining
established task designs from literature with validated objective performance metrics and periodic subjective feedback
to improve annotation precision and reliability.
\paragraph{Subjective Assessment}
Subjective measures remain a widely applied approach for evaluating perceived task load, despite susceptibility to
individual differences, task complexity, and contextual factors
\citep{paasTrainingStrategiesAttaining1992,hartDevelopmentNASATLXTask1988,chenComparisonFourMethods2011}. In this study,
participants provided post-task feedback
via the NASA-\gls{tlx} questionnaire, which evaluates perceived task load across six dimensions: mental, physical, and
temporal demand, perceived performance, effort, and frustration. While self-reported measures can be affected by
biases, such as prior experience and personality traits, they are nonetheless valuable as verification tools,
particularly where objective metrics are unavailable. We collected unweighted NASA-\gls{rtlx} values for task load
levels alongside the non-aggregated six constituent dimensions, allowing for a nuanced analysis of load increases across
tasks and levels. Furthermore, noting the influence of personality traits, such as Neuroticism and Extraversion, on
perceived task load in specific contexts \citep{roseRoleBigFive2002}, we evaluated self-reported rating
variations as task complexity increased and correlated them in the supplementary material.
\paragraph{Performance-Based Metrics}
Prior research firmly associates task-induced cognitive overload with declines in performance
\citep{paasInstructionalControlCognitive1994}. \citet{veltmanRoleOperatorState2006} refined these ideas,
observing a pattern of performance enhancement during lower loads, stabilization under normal conditions, and
degradation upon overload. Objective performance metrics were extracted for each experimental task.
For \emph{Overcooked! 2}, these included successful recipes, tips, and scores.
For \emph{Hogwarts Legacy}, missed rings, race completion time, and penalties were recorded.
For the $n$-back task, precision, recall, and reaction time were measured. Beyond psychology-based
paradigms like the $n$-back test, where traditional metrics (e.g., F1 score and reaction time) are readily computable,
real-world applications occasionally fall short by design in providing robust measurements across load levels. 
\paragraph{Annotation for Supervised Estimation}
Task load annotation poses challenges, particularly in achieving alignment between experimental tasks and
real-world validity. Many studies employ the $n$-back paradigm to systematically vary task difficulty, mapping
low $n$-back levels to baseline conditions and higher levels ($n \in \{2,3\}$) to high task load
\citep{oppeltADABaseMultimodalDataset2022, jaeggiConcurrentValidityNBack2010}. Building on
\citet{oppeltADABaseMultimodalDataset2022}, our study mapped domain-specific load demands such as driving complexity
transitions (single events to multi-task scenarios) and dual-task user interface operations (e.g., song playlist
management alongside navigation). Real-world tasks like \emph{Overcooked! 2} ranged across low-load education modes
to high-load time-critical dual-task cooking, while \emph{Hogwarts Legacy} included low-load free-flight phases and
high-load guided race challenges, respectively yielding balanced annotation sets for model training.
\paragraph{Binary Classification of Task Load}
Classifying tasks into binary low and high task load categories is practical for early-stage deployment in
adaptive systems, despite notable limitations. Continuously varying scales and arbitrary thresholds may obscure finer
behavioral or neural subtleties at boundaries. This can compromise model generalizability under intermediate loads.
While binary classification simplifies system development, these frameworks must incrementally evolve toward
multi-class schemes. The widespread adoption of binary classification in \gls{hci} underscores its utility, especially
in adaptive systems where model responsiveness is prioritized.
\paragraph{Task-Load Monotonicity: Held Out Experiments}
A key question for binary-labeled task load estimation is whether the output of a binary classifier orders difficulty
levels beyond the two trained classes. To test this, we trained supervised models exclusively on the two extreme
task load levels, low and high anchors, while withholding an intermediate level entirely from training. After training,
each model was deployed on all levels, including the held-out intermediate condition, using only test-fold subjects 
unseen during training (pooled across the five subject-wise folds). Monotonicity holds if the classifier's predicted
high-load probability $\hat{p}$ for the withheld intermediate level falls strictly between the two training anchors.
For the $n$-back paradigm, 1-back and 3-back served as anchors with 2-back held out, for \emph{Overcooked! 2}, easy
and hard levels were anchors with medium held out and for \emph{Hogwarts Legacy}, resting baseline and the highest-load
gameplay phase served as anchors with the lower-load gameplay phase held out.
\paragraph{Prediction-Label Monotonicity}
Complementary to the held-out protocol, we assessed whether classifiers trained on the full binary label set encode an
ordered difficulty gradient extending beyond the training dichotomy. Predictions were taken directly from the
supervised baseline deploy runs without retraining. For each analysis experiment we mapped levels to an ordered
difficulty rank sequence spanning resting baselines, intermediate load conditions, and all test levels, including
levels that share an identical binary training label. Per-subject $\hat{p}$ was computed at each rank and tested for
monotone increase.
\subsection{Task Load Estimation}
Stable and reliable prediction of task load remains a central challenge in task load research,
requiring models that accurately infer load levels from ongoing physiological and behavioral signals. To this end,
we first train and evaluate automated task-load detection systems using unimodal inputs, and then progress to
multimodal models that integrate multiple input channels. We further assess robustness across diverse tasks and
architectures, addressing a key deployment issue, namely generalization to unseen subjects and to \gls{hci}
applications not previously encountered.
Rather than optimizing for the highest in-domain predictive performance on binary task-load classification, our
objective with \gls{revelio} is to enable the \gls{hci} community to build reliable, transferable models across tasks
and, at minimum, to rigorously evaluate their cross-domain performance. Accordingly, a primary goal of \gls{revelio} is
to establish a principled basis for cross-domain evaluation. We achieve this through two integral components:
\paragraph{Evaluation Protocol}
To ensure fair and reproducible model evaluation, we implement a subject-wise (grouped) 5-fold cross-validation
strategy. This approach assigns all data from a given participant, including multiple experiments and repeat sessions,
to a single fold, carefully avoiding subject-related data leakage. Five folds keep the computational cost of
repeatedly training modern state-of-the-art architectures (ConvNeXt, Transformer, and xLSTM encoders) tractable. Statistical
inference is made on participants (see \emph{Statistical Inference} below). This avoids the
prohibitive expense of leave-one-subject-out evaluation without compromising the strict subject-wise grouping that
prevents data leakage. In each split, one fold is held out for testing,
three folds are used for training, and one serves as a validation set. The stratification is further extended across
experiments, grouping task contexts to balance experimental conditions. During training, the inner splits facilitate
processes such as early stopping, while the held-out folds serve to rigorously assess generalization on unseen
participants and tasks. Performance metrics, including mean and standard deviation of \gls{auroc}, are aggregated across
folds to report reliable statistics. This protocol ensures robust evaluations while mitigating potential biases caused
by overfitting to individuals or specific datasets. For transfer to an unseen domain, the target domain is
also removed from the training and validation folds (Section~\ref{ssec:lodo}). \gls{ece} uses 15
equal-width bins over the predicted probabilities of both classes. It is computed over all test windows of a fold, without class weighting, and averaged
over the five folds. The models were not recalibrated. \gls{ece} depends on binning and sample size, and overlapping
windows of one participant are correlated. The supplement therefore adds two proper scoring rules, the Brier score and
the \gls{nll}, with participant-clustered intervals, together with reliability diagrams (Supplementary
Section~\ref{ssec:supp-calibration}).
\paragraph{Statistical Inference}
All tests treat the participant as the sampling unit. Participants who took part in both cohorts form a single
cluster. Each model scored every test window, and comparisons were made on the same windows. For results reported per
fold, the statistic is the difference of the fold-mean \gls{auroc} (also averaged over encoders where stated). For the leave-one-domain-out and leave-one-cohort-out analyses, the out-of-fold predictions were pooled, so that every
participant was scored once by a model that had not seen them. Intervals (95\,\%) come from a cluster bootstrap that
resamples the 83 participants with all their windows across folds and encoders. Two-sided $p$-values
come from a sign-flip test that exchanges the two compared models for all windows of a participant jointly (10{,}000
draws each). Families of tests were corrected with Holm's method, the 42 source-target pairs with the
Benjamini-Hochberg procedure. Both procedures condition on the trained models and therefore do not include the
variability of retraining, which the result tables show descriptively as the standard deviation over folds.
\paragraph{Data Availability}
Accompanying the evaluation framework, we release 30 randomly sampled complete sessions of the gaming study, each with
the $n$-back and both games, as an extension of the \gls{adabase} dataset \citep{oppeltADABaseMultimodalDataset2022}.
Each recording includes synchronized multimodal physiological signals, performance measures, subjective feedback, and
meta-information such as personality traits, gaming experience, and action units and movement extracted from video
data. Together with the 30 driving recordings of \gls{adabase}, 60 of the 96 sessions are available, covering both
cohorts and all four applications. The remaining recordings are retained as an undisclosed test set for benchmarking
challenges. Access conditions and persistent identifiers are given in the Data Availability Statement.
\subsubsection{Input Data}
\begin{figure*}[htb!]
\centering
\input{./images/inputs.tex}
  \caption{Schematic visualization of time-series inputs.}
\label{fig:inputs}
\end{figure*}
All input modalities were resampled to a uniform rate of $100\,\mathrm{Hz}$ and segmented into overlapping sliding windows,
following established protocols \citep{oppeltADABaseMultimodalDataset2022,foltynEvaluatingRobustnessMultimodal2024,
gjoreskiMachineLearningEndtoEnd2020}. Normalization was applied to the input signals, not to derived features
such as heart rate or heart rate variability. Each channel was shifted and scaled with statistics computed separately
for each application within a session, once before any data split and without labels. Each session belongs to one participant 
and all splits are subject-wise, so no statistic combines data of training and test participants. Normalization
removes amplitude differences unrelated to task load but preserves the temporal structure, so heart rate and its variability
remain available to the encoder. The \gls{ecg} amplitude, for example, depends on electrode placement, body habitus and
the position of the heart \citep{rautaharjuStandardizedProcedureLocating1998}, and the skin conductance level differs
strongly between individuals \citep{boucseinElectrodermalActivity2012}. Physiological signals and pupil diameters were
therefore standardized to zero mean and unit variance. Gaze coordinates are defined relative to the display by the
eye-tracker calibration. They were only median-centered and divided by the spread between the 10th and 90th percentile,
which limits outliers from misdetections. Head and body movement signals were scaled in the same way. Facial expression
activations, such as action units, lie on a common scale between zero and one and were left unaltered. Because the
statistics cover the complete application, they require offline access to it. Normalizing each input window with its
own statistics instead requires no data outside the window. It changed the \gls{auroc} of ConvNeXt-small with biosignals
and eye tracking on the complete dataset from $0.83$ to $0.81$ (Supplementary
Section~\ref{ssec:supp-window-normalization}). Representative visualizations of these raw data signals are provided
in Fig.~\ref{fig:inputs}. All processed input modalities were concatenated along the channel dimension.
The proposed acquisition setup captures measurements from multiple modalities, namely biosignals, eye tracking, head
and shoulder pose, and facial action units (Fig.~\ref{sfig:inputs-biosignals}--\subref*{sfig:inputs-action-units}).
Recognizing that real-world deployment may limit access to all modalities, we systematically evaluated model performance
using both unimodal inputs and various multimodal subsets. Unimodal models were trained on respiration (\gls{rsp}),
muscle activity (\gls{emg}), electrodermal activity (\gls{eda}), mechanical and two-lead electrical cardiac activity
(\gls{ppg}, \gls{ecg}), pupil diameter from both eyes, eye movement, head position, head rotation, shoulder
position, and facial expressions via action units. For multimodal evaluation, we assessed models on combined cardiac
signals (\gls{ecg} and \gls{ppg}), all biosignals from the BIOPAC system, complete eye tracking features (gaze
coordinates and pupil diameter), head rotation and position, all facial action units, and all movement markers.
Further, we explored comprehensive combinations, such as integrating biosignals, eye tracker data, movement, and
action units, as well as a baseline configuration combining biosignals, eye tracking, and action units. This evaluation
protocol enables us to identify the most critical modalities and assess the effectiveness of various unimodal and
multimodal combinations for different applications and scenarios.
\subsection{Model}
\label{sec:model}
The input modalities are stacked as separate channels and processed by an end-to-end classification model. All models
share a common architecture. The encoder $f(\cdot)$ receives multichannel time series input of shape
$x \in \mathbb{R}^{\mathcal{B}\times\mathcal{C}_{\mathrm{in}}\times\mathcal{S}}$, where $\mathcal{B}$ is the batch size,
$\mathcal{C}_{\mathrm{in}}$ the number of input channels, and $\mathcal{S}$ the sequence length. The encoder produces
latent representations, which are averaged along the sequence dimension to yield
$z \in \mathbb{R}^{\mathcal{B}\times\mathcal{C}_{\mathrm{latent}}}$, with $\mathcal{C}_{\mathrm{latent}}$ denoting the
latent dimensionality. These representations are then passed to a multi-layer perceptron $g(\cdot)$ with a final
sigmoid activation to obtain the output $y \in \mathbb{R}^{\mathcal{B}\times\mathcal{C}}$. The network is trained
using binary cross-entropy loss and the Adam optimizer \citep{kingmaAdamMethodStochastic2015}.
\paragraph{Encoder}
Encoding time series data presents challenges due to potentially long sequences and complex temporal dependencies. We
select three principal encoder architectures based on their recency and diversity of approach
: a convolutional model (ConvNeXt), a recurrent model (\gls{xlstm}), and an attention-based model (transformer),
sketched in Fig.~\ref{sfig:encoder-cnn}, \subref*{sfig:encoder-rnn}, and \subref*{sfig:encoder-transformer}.
\begin{figure}[!htb]
  \centering
\subfloat[\centering Convolutional Neural Networks]{\label{sfig:encoder-cnn}
\begin{minipage}[t][82pt][c]{0.32\linewidth}\centering\scalebox{0.9}{\tikzset{every picture/.style={line width=0.75pt}} 
\begin{tikzpicture}[x=0.75pt,y=0.75pt,yscale=-1,xscale=1] 
\draw  [fill={rgb, 255:red, 255; green, 255; blue, 255 }  ,fill opacity=1 ][line width=0.75]  (84.06,51.28) -- (89.06,51.28) -- (89.06,56.28) -- (84.06,56.28) -- cycle ; %
\draw  [fill={rgb, 255:red, 255; green, 255; blue, 255 }  ,fill opacity=1 ][line width=0.75]  (86.56,53.78) -- (91.56,53.78) -- (91.56,58.78) -- (86.56,58.78) -- cycle ; %
\draw  [fill={rgb, 255:red, 255; green, 255; blue, 255 }  ,fill opacity=1 ][line width=0.75]  (57.43,42) -- (62.43,42) -- (62.43,59) -- (57.43,59) -- cycle ; %
\draw  [fill={rgb, 255:red, 255; green, 255; blue, 255 }  ,fill opacity=1 ][line width=0.75]  (59.93,44.5) -- (64.93,44.5) -- (64.93,61.5) -- (59.93,61.5) -- cycle ; %
\draw  [fill={rgb, 255:red, 255; green, 255; blue, 255 }  ,fill opacity=1 ][line width=0.75]  (37.43,32.89) -- (42.43,32.89) -- (42.43,67.89) -- (37.43,67.89) -- cycle ; %
\draw  [fill={rgb, 255:red, 255; green, 255; blue, 255 }  ,fill opacity=1 ][line width=0.75]  (39.93,35.39) -- (44.93,35.39) -- (44.93,70.39) -- (39.93,70.39) -- cycle ; %
\draw  [fill={rgb, 255:red, 255; green, 255; blue, 255 }  ,fill opacity=1 ][line width=0.75]  (17.43,15.39) -- (22.43,15.39) -- (22.43,85.39) -- (17.43,85.39) -- cycle ; %
\draw  [fill={rgb, 255:red, 255; green, 255; blue, 255 }  ,fill opacity=1 ][line width=0.75]  (19.93,17.89) -- (24.93,17.89) -- (24.93,87.89) -- (19.93,87.89) -- cycle ; %
\draw    (21.42,6.71) -- (36.04,20.05) ; %
\draw [shift={(37.52,21.39)}, rotate = 222.36] [color={rgb, 255:red, 0; green, 0; blue, 0 }  ][line width=0.75]    (4.37,-1.32) .. controls (2.78,-0.56) and (1.32,-0.12) .. (0,0) .. controls (1.32,0.12) and (2.78,0.56) .. (4.37,1.32)   ; %
\draw  [fill={rgb, 255:red, 255; green, 255; blue, 255 }  ,fill opacity=1 ][line width=0.75]  (22.43,20.39) -- (27.43,20.39) -- (27.43,90.39) -- (22.43,90.39) -- cycle ; %
\draw    (12.43,14.91) -- (12.43,46.59) ; %
\draw [shift={(12.43,48.59)}, rotate = 270] [color={rgb, 255:red, 0; green, 0; blue, 0 }  ][line width=0.75]    (4.37,-1.32) .. controls (2.78,-0.56) and (1.32,-0.12) .. (0,0) .. controls (1.32,0.12) and (2.78,0.56) .. (4.37,1.32)   ; %
\draw  [fill={rgb, 255:red, 255; green, 255; blue, 255 }  ,fill opacity=1 ][line width=0.75]  (22.43,20.39) -- (27.43,20.39) -- (27.43,25.39) -- (22.43,25.39) -- cycle ; %
\draw  [fill={rgb, 255:red, 255; green, 255; blue, 255 }  ,fill opacity=1 ][line width=0.75]  (22.43,25.39) -- (27.43,25.39) -- (27.43,30.39) -- (22.43,30.39) -- cycle ; %
\draw  [fill={rgb, 255:red, 255; green, 255; blue, 255 }  ,fill opacity=1 ][line width=0.75]  (22.43,30.39) -- (27.43,30.39) -- (27.43,35.39) -- (22.43,35.39) -- cycle ; %
\draw  [fill={rgb, 255:red, 255; green, 255; blue, 255 }  ,fill opacity=1 ][line width=0.75]  (22.43,85.39) -- (27.43,85.39) -- (27.43,90.39) -- (22.43,90.39) -- cycle ; %
\draw  [fill={rgb, 255:red, 255; green, 255; blue, 255 }  ,fill opacity=1 ][line width=0.75]  (22.43,80.39) -- (27.43,80.39) -- (27.43,85.39) -- (22.43,85.39) -- cycle ; %
\draw  [fill={rgb, 255:red, 255; green, 255; blue, 255 }  ,fill opacity=1 ][line width=0.75]  (22.43,75.39) -- (27.43,75.39) -- (27.43,80.39) -- (22.43,80.39) -- cycle ; %
\draw  [fill={rgb, 255:red, 255; green, 255; blue, 255 }  ,fill opacity=1 ][line width=0.75]  (42.43,37.89) -- (47.43,37.89) -- (47.43,72.89) -- (42.43,72.89) -- cycle ; %
\draw  [fill={rgb, 255:red, 255; green, 255; blue, 255 }  ,fill opacity=1 ][line width=0.75]  (42.43,37.89) -- (47.43,37.89) -- (47.43,42.89) -- (42.43,42.89) -- cycle ; %
\draw  [fill={rgb, 255:red, 255; green, 255; blue, 255 }  ,fill opacity=1 ][line width=0.75]  (42.43,42.89) -- (47.43,42.89) -- (47.43,47.89) -- (42.43,47.89) -- cycle ; %
\draw  [fill={rgb, 255:red, 255; green, 255; blue, 255 }  ,fill opacity=1 ][line width=0.75]  (42.43,67.89) -- (47.43,67.89) -- (47.43,72.89) -- (42.43,72.89) -- cycle ; %
\draw  [fill={rgb, 255:red, 255; green, 255; blue, 255 }  ,fill opacity=1 ][line width=0.75]  (42.43,62.89) -- (47.43,62.89) -- (47.43,67.89) -- (42.43,67.89) -- cycle ; %
\draw  [fill={rgb, 255:red, 255; green, 255; blue, 255 }  ,fill opacity=1 ][line width=0.75]  (62.43,47) -- (67.43,47) -- (67.43,64) -- (62.43,64) -- cycle ; %
\draw  [fill={rgb, 255:red, 255; green, 255; blue, 255 }  ,fill opacity=1 ][line width=0.75]  (62.43,47) -- (67.43,47) -- (67.43,52) -- (62.43,52) -- cycle ; %
\draw  [fill={rgb, 255:red, 255; green, 255; blue, 255 }  ,fill opacity=1 ][line width=0.75]  (62.43,59) -- (67.43,59) -- (67.43,64) -- (62.43,64) -- cycle ; %
\draw [line width=0.75]    (27.43,20.39) -- (42.43,37.89) ; %
\draw [line width=0.75]    (27.43,35.39) -- (42.43,42.89) ; %
\draw    (27.43,20.39) -- (27.43,35.39) ; %
\draw    (42.43,37.89) -- (42.43,42.89) ; %
\draw [line width=0.75]    (42.43,72.89) -- (27.43,90.39) ; %
\draw [line width=0.75]    (42.43,67.89) -- (27.43,75.39) ; %
\draw  [fill={rgb, 255:red, 255; green, 255; blue, 255 }  ,fill opacity=1 ][line width=0.75]  (42.43,47.89) -- (47.43,47.89) -- (47.43,52.89) -- (42.43,52.89) -- cycle ; %
\draw  [fill={rgb, 255:red, 255; green, 255; blue, 255 }  ,fill opacity=1 ][line width=0.75]  (42.43,57.89) -- (47.43,57.89) -- (47.43,62.89) -- (42.43,62.89) -- cycle ; %
\draw [line width=0.75]    (47.43,37.89) -- (62.43,47) ; %
\draw [line width=0.75]    (47.43,52.89) -- (62.43,52) ; %
\draw [line width=0.75]    (47.43,72.89) -- (62.43,64) ; %
\draw [line width=0.75]    (47.43,57.89) -- (62.43,59) ; %
\draw  [fill={rgb, 255:red, 255; green, 255; blue, 255 }  ,fill opacity=1 ][line width=0.75]  (89.06,56.28) -- (94.06,56.28) -- (94.06,61.28) -- (89.06,61.28) -- cycle ; %
\draw    (68.22,54.23) -- (80.74,54.35) ; %
\draw [shift={(82.74,54.37)}, rotate = 180.52] [color={rgb, 255:red, 0; green, 0; blue, 0 }  ][line width=0.75]    (4.37,-1.32) .. controls (2.78,-0.56) and (1.32,-0.12) .. (0,0) .. controls (1.32,0.12) and (2.78,0.56) .. (4.37,1.32)   ; %
\draw (36.75,5.5) node  [font=\scriptsize] [align=left] {\begin{minipage}[lt]{10.33pt}\setlength\topsep{0pt} %
\begin{flushright} %
{\scriptsize  $C_{\mathrm{in}}$} %
\end{flushright} %
\end{minipage}}; %
\draw (5,31.31) node  [font=\scriptsize] [align=left] {\begin{minipage}[lt]{6.99pt}\setlength\topsep{0pt} %
\begin{flushright} %
{\scriptsize  $S$} %
\end{flushright} %
\end{minipage}}; %
\draw (69.4,41.5) node [anchor=north west][inner sep=0.75pt]  [font=\tiny] [align=left] {{\tiny $\sum $}}; %
\end{tikzpicture}}\end{minipage}} \hfill%
\subfloat[\centering Recurrent Neural Network]{\label{sfig:encoder-rnn}
\begin{minipage}[t][82pt][c]{0.32\linewidth}\centering\scalebox{0.9}{\tikzset{every picture/.style={line width=0.75pt}} 

\begin{tikzpicture}[x=0.75pt,y=0.75pt,yscale=-1,xscale=1]

\draw  [fill={rgb, 255:red, 255; green, 255; blue, 255 }  ,fill opacity=1 ][line width=0.75]  (33.2,38.7) -- (38.2,38.7) -- (38.2,43.7) -- (33.2,43.7) -- cycle ;
\draw  [fill={rgb, 255:red, 255; green, 255; blue, 255 }  ,fill opacity=1 ][line width=0.75]  (31.7,41.2) -- (36.7,41.2) -- (36.7,46.2) -- (31.7,46.2) -- cycle ;
\draw  [fill={rgb, 255:red, 255; green, 255; blue, 255 }  ,fill opacity=1 ][line width=0.75]  (30.16,43.75) -- (35.16,43.75) -- (35.16,48.75) -- (30.16,48.75) -- cycle ;
\draw  [fill={rgb, 255:red, 255; green, 255; blue, 255 }  ,fill opacity=1 ][line width=0.75]  (50.16,43.75) -- (55.16,43.75) -- (55.16,48.75) -- (50.16,48.75) -- cycle ;
\draw  [fill={rgb, 255:red, 255; green, 255; blue, 255 }  ,fill opacity=1 ][line width=0.75]  (70.16,43.75) -- (75.16,43.75) -- (75.16,48.75) -- (70.16,48.75) -- cycle ;
\draw    (35.7,46.25) -- (48.16,46.25) ;
\draw [shift={(50.16,46.25)}, rotate = 180] [color={rgb, 255:red, 0; green, 0; blue, 0 }  ][line width=0.75]    (4.37,-1.32) .. controls (2.78,-0.56) and (1.32,-0.12) .. (0,0) .. controls (1.32,0.12) and (2.78,0.56) .. (4.37,1.32)   ;
\draw    (55.16,46.25) -- (67.61,46.25) ;
\draw [shift={(69.61,46.25)}, rotate = 180] [color={rgb, 255:red, 0; green, 0; blue, 0 }  ][line width=0.75]    (4.37,-1.32) .. controls (2.78,-0.56) and (1.32,-0.12) .. (0,0) .. controls (1.32,0.12) and (2.78,0.56) .. (4.37,1.32)   ;
\draw    (75.16,46.25) -- (87.61,46.25) ;
\draw [shift={(89.61,46.25)}, rotate = 180] [color={rgb, 255:red, 0; green, 0; blue, 0 }  ][line width=0.75]    (4.37,-1.32) .. controls (2.78,-0.56) and (1.32,-0.12) .. (0,0) .. controls (1.32,0.12) and (2.78,0.56) .. (4.37,1.32)   ;
\draw    (32.61,62.75) -- (32.54,51.43) ;
\draw [shift={(32.52,49.43)}, rotate = 89.61] [color={rgb, 255:red, 0; green, 0; blue, 0 }  ][line width=0.75]    (4.37,-1.32) .. controls (2.78,-0.56) and (1.32,-0.12) .. (0,0) .. controls (1.32,0.12) and (2.78,0.56) .. (4.37,1.32)   ;
\draw    (52.61,62.75) -- (52.54,51.43) ;
\draw [shift={(52.52,49.43)}, rotate = 89.61] [color={rgb, 255:red, 0; green, 0; blue, 0 }  ][line width=0.75]    (4.37,-1.32) .. controls (2.78,-0.56) and (1.32,-0.12) .. (0,0) .. controls (1.32,0.12) and (2.78,0.56) .. (4.37,1.32)   ;
\draw    (72.61,62.75) -- (72.54,51.43) ;
\draw [shift={(72.52,49.43)}, rotate = 89.61] [color={rgb, 255:red, 0; green, 0; blue, 0 }  ][line width=0.75]    (4.37,-1.32) .. controls (2.78,-0.56) and (1.32,-0.12) .. (0,0) .. controls (1.32,0.12) and (2.78,0.56) .. (4.37,1.32)   ;
\draw  [fill={rgb, 255:red, 255; green, 255; blue, 255 }  ,fill opacity=1 ][line width=0.75]  (90.16,43.75) -- (95.16,43.75) -- (95.16,48.75) -- (90.16,48.75) -- cycle ;
\draw    (92.61,62.75) -- (92.54,51.43) ;
\draw [shift={(92.52,49.43)}, rotate = 89.61] [color={rgb, 255:red, 0; green, 0; blue, 0 }  ][line width=0.75]    (4.37,-1.32) .. controls (2.78,-0.56) and (1.32,-0.12) .. (0,0) .. controls (1.32,0.12) and (2.78,0.56) .. (4.37,1.32)   ;
\draw  [fill={rgb, 255:red, 255; green, 255; blue, 255 }  ,fill opacity=1 ][line width=0.75]  (30.16,23.75) -- (35.16,23.75) -- (35.16,28.75) -- (30.16,28.75) -- cycle ;
\draw  [fill={rgb, 255:red, 255; green, 255; blue, 255 }  ,fill opacity=1 ][line width=0.75]  (50.16,23.75) -- (55.16,23.75) -- (55.16,28.75) -- (50.16,28.75) -- cycle ;
\draw  [fill={rgb, 255:red, 255; green, 255; blue, 255 }  ,fill opacity=1 ][line width=0.75]  (70.16,23.75) -- (75.16,23.75) -- (75.16,28.75) -- (70.16,28.75) -- cycle ;
\draw    (35.7,26.25) -- (48.16,26.25) ;
\draw [shift={(50.16,26.25)}, rotate = 180] [color={rgb, 255:red, 0; green, 0; blue, 0 }  ][line width=0.75]    (4.37,-1.32) .. controls (2.78,-0.56) and (1.32,-0.12) .. (0,0) .. controls (1.32,0.12) and (2.78,0.56) .. (4.37,1.32)   ;
\draw    (55.16,26.25) -- (67.61,26.25) ;
\draw [shift={(69.61,26.25)}, rotate = 180] [color={rgb, 255:red, 0; green, 0; blue, 0 }  ][line width=0.75]    (4.37,-1.32) .. controls (2.78,-0.56) and (1.32,-0.12) .. (0,0) .. controls (1.32,0.12) and (2.78,0.56) .. (4.37,1.32)   ;
\draw    (75.16,26.25) -- (87.61,26.25) ;
\draw [shift={(89.61,26.25)}, rotate = 180] [color={rgb, 255:red, 0; green, 0; blue, 0 }  ][line width=0.75]    (4.37,-1.32) .. controls (2.78,-0.56) and (1.32,-0.12) .. (0,0) .. controls (1.32,0.12) and (2.78,0.56) .. (4.37,1.32)   ;
\draw    (32.52,41.11) -- (32.52,31.43) ;
\draw [shift={(32.52,29.43)}, rotate = 90] [color={rgb, 255:red, 0; green, 0; blue, 0 }  ][line width=0.75]    (4.37,-1.32) .. controls (2.78,-0.56) and (1.32,-0.12) .. (0,0) .. controls (1.32,0.12) and (2.78,0.56) .. (4.37,1.32)   ;
\draw    (52.61,42.75) -- (52.54,31.43) ;
\draw [shift={(52.52,29.43)}, rotate = 89.61] [color={rgb, 255:red, 0; green, 0; blue, 0 }  ][line width=0.75]    (4.37,-1.32) .. controls (2.78,-0.56) and (1.32,-0.12) .. (0,0) .. controls (1.32,0.12) and (2.78,0.56) .. (4.37,1.32)   ;
\draw    (72.61,42.75) -- (72.54,31.43) ;
\draw [shift={(72.52,29.43)}, rotate = 89.61] [color={rgb, 255:red, 0; green, 0; blue, 0 }  ][line width=0.75]    (4.37,-1.32) .. controls (2.78,-0.56) and (1.32,-0.12) .. (0,0) .. controls (1.32,0.12) and (2.78,0.56) .. (4.37,1.32)   ;
\draw  [fill={rgb, 255:red, 255; green, 255; blue, 255 }  ,fill opacity=1 ][line width=0.75]  (90.16,23.75) -- (95.16,23.75) -- (95.16,28.75) -- (90.16,28.75) -- cycle ;
\draw    (92.61,42.75) -- (92.54,31.43) ;
\draw [shift={(92.52,29.43)}, rotate = 89.61] [color={rgb, 255:red, 0; green, 0; blue, 0 }  ][line width=0.75]    (4.37,-1.32) .. controls (2.78,-0.56) and (1.32,-0.12) .. (0,0) .. controls (1.32,0.12) and (2.78,0.56) .. (4.37,1.32)   ;
\draw    (92.66,23.75) -- (92.58,12.43) ;
\draw [shift={(92.57,10.43)}, rotate = 89.61] [color={rgb, 255:red, 0; green, 0; blue, 0 }  ][line width=0.75]    (4.37,-1.32) .. controls (2.78,-0.56) and (1.32,-0.12) .. (0,0) .. controls (1.32,0.12) and (2.78,0.56) .. (4.37,1.32)   ;
\draw  [fill={rgb, 255:red, 255; green, 255; blue, 255 }  ,fill opacity=1 ][line width=0.75]  (93.2,0) -- (98.2,0) -- (98.2,5) -- (93.2,5) -- cycle ;
\draw  [fill={rgb, 255:red, 255; green, 255; blue, 255 }  ,fill opacity=1 ][line width=0.75]  (91.7,2.5) -- (96.7,2.5) -- (96.7,7.5) -- (91.7,7.5) -- cycle ;
\draw  [fill={rgb, 255:red, 255; green, 255; blue, 255 }  ,fill opacity=1 ][line width=0.75]  (90.16,5.05) -- (95.16,5.05) -- (95.16,10.05) -- (90.16,10.05) -- cycle ;
\draw    (24.56,82.88) -- (70.19,82.87) ;
\draw [shift={(72.19,82.87)}, rotate = 179.99] [color={rgb, 255:red, 0; green, 0; blue, 0 }  ][line width=0.75]    (4.37,-1.32) .. controls (2.78,-0.56) and (1.32,-0.12) .. (0,0) .. controls (1.32,0.12) and (2.78,0.56) .. (4.37,1.32)   ;
\draw  [fill={rgb, 255:red, 255; green, 255; blue, 255 }  ,fill opacity=1 ][line width=0.75]  (24.51,78.77) -- (24.51,73.77) -- (94.51,73.77) -- (94.51,78.77) -- cycle ;
\draw  [fill={rgb, 255:red, 255; green, 255; blue, 255 }  ,fill opacity=1 ][line width=0.75]  (27.01,76.27) -- (27.01,71.27) -- (97.01,71.27) -- (97.01,76.27) -- cycle ;
\draw  [fill={rgb, 255:red, 255; green, 255; blue, 255 }  ,fill opacity=1 ][line width=0.75]  (29.51,73.77) -- (29.51,68.77) -- (99.51,68.77) -- (99.51,73.77) -- cycle ;
\draw  [fill={rgb, 255:red, 255; green, 255; blue, 255 }  ,fill opacity=1 ][line width=0.75]  (29.51,73.77) -- (29.51,68.77) -- (34.51,68.77) -- (34.51,73.77) -- cycle ;
\draw  [fill={rgb, 255:red, 255; green, 255; blue, 255 }  ,fill opacity=1 ][line width=0.75]  (34.51,73.77) -- (34.51,68.77) -- (39.51,68.77) -- (39.51,73.77) -- cycle ;
\draw  [fill={rgb, 255:red, 255; green, 255; blue, 255 }  ,fill opacity=1 ][line width=0.75]  (39.51,73.77) -- (39.51,68.77) -- (44.51,68.77) -- (44.51,73.77) -- cycle ;
\draw  [fill={rgb, 255:red, 255; green, 255; blue, 255 }  ,fill opacity=1 ][line width=0.75]  (94.51,73.77) -- (94.51,68.77) -- (99.51,68.77) -- (99.51,73.77) -- cycle ;
\draw  [fill={rgb, 255:red, 255; green, 255; blue, 255 }  ,fill opacity=1 ][line width=0.75]  (89.51,73.77) -- (89.51,68.77) -- (94.51,68.77) -- (94.51,73.77) -- cycle ;
\draw  [fill={rgb, 255:red, 255; green, 255; blue, 255 }  ,fill opacity=1 ][line width=0.75]  (84.51,73.77) -- (84.51,68.77) -- (89.51,68.77) -- (89.51,73.77) -- cycle ;
\draw    (29.51,68.77) -- (44.51,68.77) ;
\draw    (13,79.95) -- (26.03,67.3) ;
\draw [shift={(27.47,65.91)}, rotate = 135.85] [color={rgb, 255:red, 0; green, 0; blue, 0 }  ][line width=0.75]    (4.37,-1.32) .. controls (2.78,-0.56) and (1.32,-0.12) .. (0,0) .. controls (1.32,0.12) and (2.78,0.56) .. (4.37,1.32)   ;

\draw (28.83,89.92) node  [font=\scriptsize] [align=left] {\begin{minipage}[lt]{6.99pt}\setlength\topsep{0pt}
\begin{flushright}
{\scriptsize  $S$}
\end{flushright}

\end{minipage}};
\draw (7.89,64.78) node  [font=\scriptsize] [align=left] {\begin{minipage}[lt]{10.33pt}\setlength\topsep{0pt}
\begin{flushright}
{\scriptsize  $C_{\mathrm{in}}$}
\end{flushright}

\end{minipage}};

\end{tikzpicture}}\end{minipage}} \hfill%
\subfloat[\centering Transformer Networks]{\label{sfig:encoder-transformer}
\begin{minipage}[t][82pt][c]{0.32\linewidth}\centering\scalebox{0.9}{\tikzset{every picture/.style={line width=0.75pt}} 
\begin{tikzpicture}[x=0.75pt,y=0.75pt,yscale=-1,xscale=1] %
\draw  [fill={rgb, 255:red, 255; green, 255; blue, 255 }  ,fill opacity=1 ][line width=0.75]  (55.68,81.66) -- (105.68,81.66) -- (105.68,101.66) -- (55.68,101.66) -- cycle ; %
\draw  [fill={rgb, 255:red, 255; green, 255; blue, 255 }  ,fill opacity=1 ][line width=0.75]  (55.68,61.66) -- (105.68,61.66) -- (105.68,71.66) -- (55.68,71.66) -- cycle ; %
\draw    (80.77,80.98) -- (80.71,74.79) ; %
\draw [shift={(80.68,72.79)}, rotate = 89.36] [color={rgb, 255:red, 0; green, 0; blue, 0 }  ][line width=0.75]    (4.37,-1.32) .. controls (2.78,-0.56) and (1.32,-0.12) .. (0,0) .. controls (1.32,0.12) and (2.78,0.56) .. (4.37,1.32)   ; %
\draw  [fill={rgb, 255:red, 255; green, 255; blue, 255 }  ,fill opacity=1 ][line width=0.75]  (55.68,25.77) -- (105.68,25.77) -- (105.68,35.77) -- (55.68,35.77) -- cycle ; %
\draw    (80.65,45.09) -- (80.59,38.89) ; %
\draw [shift={(80.56,36.89)}, rotate = 89.36] [color={rgb, 255:red, 0; green, 0; blue, 0 }  ][line width=0.75]    (4.37,-1.32) .. controls (2.78,-0.56) and (1.32,-0.12) .. (0,0) .. controls (1.32,0.12) and (2.78,0.56) .. (4.37,1.32)   ; %
\draw  [fill={rgb, 255:red, 255; green, 255; blue, 255 }  ,fill opacity=1 ][line width=0.75]  (55.68,6.02) -- (105.68,6.02) -- (105.68,16.02) -- (55.68,16.02) -- cycle ; %
\draw    (80.68,25.34) -- (80.61,19.14) ; %
\draw [shift={(80.59,17.14)}, rotate = 89.36] [color={rgb, 255:red, 0; green, 0; blue, 0 }  ][line width=0.75]    (4.37,-1.32) .. controls (2.78,-0.56) and (1.32,-0.12) .. (0,0) .. controls (1.32,0.12) and (2.78,0.56) .. (4.37,1.32)   ; %
\draw    (80.68,61.66) -- (80.61,55.47) ; %
\draw [shift={(80.59,53.47)}, rotate = 89.36] [color={rgb, 255:red, 0; green, 0; blue, 0 }  ][line width=0.75]    (4.37,-1.32) .. controls (2.78,-0.56) and (1.32,-0.12) .. (0,0) .. controls (1.32,0.12) and (2.78,0.56) .. (4.37,1.32)   ; %
\draw   (76.97,49.16) .. controls (76.97,47.15) and (78.59,45.53) .. (80.59,45.53) .. controls (82.59,45.53) and (84.22,47.15) .. (84.22,49.16) .. controls (84.22,51.16) and (82.59,52.78) .. (80.59,52.78) .. controls (78.59,52.78) and (76.97,51.16) .. (76.97,49.16) -- cycle ; %
\draw    (80.59,45.53) -- (80.59,52.78) ; %
\draw    (76.97,49.16) -- (84.22,49.16) ; %
\draw    (80.95,111.24) -- (49.68,111.02) -- (49.68,49.17) -- (74.97,49.16) ; %
\draw [shift={(76.97,49.16)}, rotate = 179.97] [color={rgb, 255:red, 0; green, 0; blue, 0 }  ][line width=0.75]    (4.37,-1.32) .. controls (2.78,-0.56) and (1.32,-0.12) .. (0,0) .. controls (1.32,0.12) and (2.78,0.56) .. (4.37,1.32)   ; %
\draw    (80.95,111.24) -- (80.78,104.55) ; %
\draw [shift={(80.73,102.55)}, rotate = 88.58] [color={rgb, 255:red, 0; green, 0; blue, 0 }  ][line width=0.75]    (4.37,-1.32) .. controls (2.78,-0.56) and (1.32,-0.12) .. (0,0) .. controls (1.32,0.12) and (2.78,0.56) .. (4.37,1.32)   ; %
\draw  [fill={rgb, 255:red, 255; green, 255; blue, 255 }  ,fill opacity=1 ][line width=0.75]  (25.64,57.98) -- (30.64,57.98) -- (30.64,107.98) -- (25.64,107.98) -- cycle ; %
\draw  [fill={rgb, 255:red, 255; green, 255; blue, 255 }  ,fill opacity=1 ][line width=0.75]  (28.14,60.48) -- (33.14,60.48) -- (33.14,110.48) -- (28.14,110.48) -- cycle ; %
\draw  [fill={rgb, 255:red, 255; green, 255; blue, 255 }  ,fill opacity=1 ][line width=0.75]  (30.64,62.98) -- (35.64,62.98) -- (35.64,112.98) -- (30.64,112.98) -- cycle ; %
\draw  [fill={rgb, 255:red, 255; green, 255; blue, 255 }  ,fill opacity=1 ][line width=0.75]  (30.64,62.98) -- (35.64,62.98) -- (35.64,67.98) -- (30.64,67.98) -- cycle ; %
\draw  [fill={rgb, 255:red, 255; green, 255; blue, 255 }  ,fill opacity=1 ][line width=0.75]  (30.64,67.98) -- (35.64,67.98) -- (35.64,72.98) -- (30.64,72.98) -- cycle ; %
\draw  [fill={rgb, 255:red, 255; green, 255; blue, 255 }  ,fill opacity=1 ][line width=0.75]  (30.64,72.98) -- (35.64,72.98) -- (35.64,77.98) -- (30.64,77.98) -- cycle ; %
\draw  [fill={rgb, 255:red, 255; green, 255; blue, 255 }  ,fill opacity=1 ][line width=0.75]  (30.64,107.98) -- (35.64,107.98) -- (35.64,112.98) -- (30.64,112.98) -- cycle ; %
\draw  [fill={rgb, 255:red, 255; green, 255; blue, 255 }  ,fill opacity=1 ][line width=0.75]  (30.64,102.98) -- (35.64,102.98) -- (35.64,107.98) -- (30.64,107.98) -- cycle ; %
\draw  [fill={rgb, 255:red, 255; green, 255; blue, 255 }  ,fill opacity=1 ][line width=0.75]  (30.64,97.98) -- (35.64,97.98) -- (35.64,102.98) -- (30.64,102.98) -- cycle ; %
\draw    (35.64,62.98) -- (35.64,77.98) ; %
\draw    (35.68,65.49) -- (44.18,65.52) -- (44.18,110.58) ; %
\draw    (35.68,70.58) -- (43.68,70.58) ; %
\draw    (35.68,75.58) -- (43.68,75.58) ; %
\draw    (36.18,100.58) -- (44.18,100.58) ; %
\draw    (36.18,105.63) -- (44.18,105.63) ; %
\draw    (36.18,110.63) -- (44.18,110.63) ; %
\draw  [fill={rgb, 255:red, 255; green, 255; blue, 255 }  ,fill opacity=1 ][line width=0.75]  (6.44,44.7) -- (11.44,44.7) -- (11.44,49.7) -- (6.44,49.7) -- cycle ; %
\draw  [fill={rgb, 255:red, 255; green, 255; blue, 255 }  ,fill opacity=1 ][line width=0.75]  (8.94,47.2) -- (13.94,47.2) -- (13.94,52.2) -- (8.94,52.2) -- cycle ; %
\draw  [fill={rgb, 255:red, 255; green, 255; blue, 255 }  ,fill opacity=1 ][line width=0.75]  (11.44,49.7) -- (16.44,49.7) -- (16.44,54.7) -- (11.44,54.7) -- cycle ; %
\draw    (28.68,52.33) -- (20.09,52.38) ; %
\draw [shift={(18.09,52.39)}, rotate = 359.65] [color={rgb, 255:red, 0; green, 0; blue, 0 }  ][line width=0.75]    (4.37,-1.32) .. controls (2.78,-0.56) and (1.32,-0.12) .. (0,0) .. controls (1.32,0.12) and (2.78,0.56) .. (4.37,1.32)   ; %
\draw    (48.38,85.58) -- (45.36,85.56) ; %
\draw  [fill={rgb, 255:red, 255; green, 255; blue, 255 }  ,fill opacity=1 ][line width=0.75]  (25.68,0) -- (30.68,0) -- (30.68,50) -- (25.68,50) -- cycle ; %
\draw  [fill={rgb, 255:red, 255; green, 255; blue, 255 }  ,fill opacity=1 ][line width=0.75]  (28.18,2.5) -- (33.18,2.5) -- (33.18,52.5) -- (28.18,52.5) -- cycle ; %
\draw  [fill={rgb, 255:red, 255; green, 255; blue, 255 }  ,fill opacity=1 ][line width=0.75]  (30.68,5) -- (35.68,5) -- (35.68,55) -- (30.68,55) -- cycle ; %
\draw  [fill={rgb, 255:red, 255; green, 255; blue, 255 }  ,fill opacity=1 ][line width=0.75]  (30.68,5) -- (35.68,5) -- (35.68,10) -- (30.68,10) -- cycle ; %
\draw  [fill={rgb, 255:red, 255; green, 255; blue, 255 }  ,fill opacity=1 ][line width=0.75]  (30.68,10) -- (35.68,10) -- (35.68,15) -- (30.68,15) -- cycle ; %
\draw  [fill={rgb, 255:red, 255; green, 255; blue, 255 }  ,fill opacity=1 ][line width=0.75]  (30.68,15) -- (35.68,15) -- (35.68,20) -- (30.68,20) -- cycle ; %
\draw  [fill={rgb, 255:red, 255; green, 255; blue, 255 }  ,fill opacity=1 ][line width=0.75]  (30.68,50) -- (35.68,50) -- (35.68,55) -- (30.68,55) -- cycle ; %
\draw  [fill={rgb, 255:red, 255; green, 255; blue, 255 }  ,fill opacity=1 ][line width=0.75]  (30.68,45) -- (35.68,45) -- (35.68,50) -- (30.68,50) -- cycle ; %
\draw  [fill={rgb, 255:red, 255; green, 255; blue, 255 }  ,fill opacity=1 ][line width=0.75]  (30.68,40) -- (35.68,40) -- (35.68,45) -- (30.68,45) -- cycle ; %
\draw    (35.68,5) -- (35.68,20) ; %
\draw    (35.72,7.51) -- (44.22,7.54) -- (44.22,52.6) ; %
\draw    (35.72,12.6) -- (43.72,12.6) ; %
\draw    (35.72,17.6) -- (43.72,17.6) ; %
\draw    (36.22,42.6) -- (44.22,42.6) ; %
\draw    (36.22,47.65) -- (44.22,47.65) ; %
\draw    (36.22,52.65) -- (44.22,52.65) ; %
\draw    (48.42,27.6) -- (45.4,27.58) ; %
\draw    (49.68,1.08) -- (49.68,28.39) ; %
\draw    (80.42,5.24) -- (80.53,1.14) -- (49.68,1.08) ; %
\draw    (21.68,99.53) -- (21.68,65.75) ; %
\draw [shift={(21.68,63.75)}, rotate = 90] [color={rgb, 255:red, 0; green, 0; blue, 0 }  ][line width=0.75]    (4.37,-1.32) .. controls (2.78,-0.56) and (1.32,-0.12) .. (0,0) .. controls (1.32,0.12) and (2.78,0.56) .. (4.37,1.32)   ; %
\draw    (12.17,99.22) -- (26.8,112.55) ; %
\draw [shift={(28.27,113.9)}, rotate = 222.36] [color={rgb, 255:red, 0; green, 0; blue, 0 }  ][line width=0.75]    (4.37,-1.32) .. controls (2.78,-0.56) and (1.32,-0.12) .. (0,0) .. controls (1.32,0.12) and (2.78,0.56) .. (4.37,1.32)   ; %
\draw (80.68,91.66) node  [font=\tiny] [align=left] {\begin{minipage}[lt]{34pt}\setlength\topsep{0pt} %
\begin{center} %
{\tiny Multi-Head Attention} %
\end{center} %
\end{minipage}}; %
\draw (80.68,66.66) node  [font=\tiny] [align=left] {\begin{minipage}[lt]{34pt}\setlength\topsep{0pt} %
\begin{center} %
{\tiny LN} %
\end{center} %
\end{minipage}}; %
\draw (80.68,30.77) node  [font=\tiny] [align=left] {\begin{minipage}[lt]{34pt}\setlength\topsep{0pt} %
\begin{center} %
{\tiny MLP} %
\end{center} %
\end{minipage}}; %
\draw (80.68,11.02) node  [font=\tiny] [align=left] {\begin{minipage}[lt]{34pt}\setlength\topsep{0pt} %
\begin{center} %
{\tiny LN} %
\end{center} %
\end{minipage}}; %
\draw (13.75,71.54) node  [font=\scriptsize] [align=left] {\begin{minipage}[lt]{6.99pt}\setlength\topsep{0pt} %
\begin{flushright} %
{\scriptsize  $S$} %
\end{flushright} %
\end{minipage}}; %
\draw (7.5,107.76) node  [font=\scriptsize] [align=left] {\begin{minipage}[lt]{10.33pt}\setlength\topsep{0pt} %
\begin{flushright} %
{\scriptsize  $C_{\mathrm{in}}$} %
\end{flushright} %
\end{minipage}}; %
\end{tikzpicture}}\end{minipage}}%
  \caption{Principal types of encoder architectures for time series classification evaluated in this study.}
\label{fig:encoder-architectures}
  \vspace{-0.15cm}
\end{figure}
ConvNeXt \citep{liuConvNet2020s2022} incorporates advanced normalization, larger kernels, and depthwise convolutions.
\gls{xlstm} \citep{beckXLSTMExtendedLong2024} extends the standard \gls{lstm} with enhanced memory mechanisms and memory
mixing. The transformer encoder \citep{vaswaniAttentionAllYou2017} leverages self-attention
\citep{bahdanauNeuralMachineTranslation2014}.
As \gls{xlstm} and transformer encoders are sensitive to sequence length, we prepend a \gls{cnn} front-end to increase
channel dimensionality and downsample inputs. Following \citet{liuConvNet2020s2022}, we implement the \emph{tiny}, \emph{small},
and \emph{base} ConvNeXt variants with one-dimensional convolutions and omit the larger variants as a computational
trade-off. We adopt the same \emph{tiny}, \emph{small}, and \emph{base} nomenclature for all three encoders. Each
configuration follows the respective original publication \citep{liuConvNet2020s2022,beckXLSTMExtendedLong2024,%
vaswaniAttentionAllYou2017}, and the exact settings are reported in the supplementary material. Parameter
counts and inference times therefore differ across architectures, and the labels are not intended for a one-to-one
comparison of equally named variants. They instead provide three consistent capacity settings per encoder, offering a
practical basis for reproducible comparison in future work rather than a size-controlled benchmark. Our objective is
the estimation of task load, not a one-to-one comparison of architecture sizes. Hyperparameters follow established
literature and the original implementations, with full architectural details in the supplementary material.
\section{Results}
Building on the 51 driving-simulator sessions from ADABase \citep{oppeltADABaseMultimodalDataset2022}, we newly
collected 45 gaming sessions from 37 participants, eight of whom completed two sessions, yielding the combined
\gls{revelio} corpus of 96 sessions from 83 unique participants, because five people took part in both the
driving and the gaming study. The new cohort included 15 females and 22 males, aged 20--59 years
(mean $30\,\pm\,9$ years). Individuals reporting health-related issues were excluded. To mitigate circadian 
effects, session times were randomized between morning and afternoon.
For the eight participants with two gaming sessions, the answers of the first session are reported.
Gaming experience was assessed using a five-point Likert scale: 16 participants reported no or little experience,
9 identified as casual gamers, 3 as regular gamers, 8 as experienced, and 1 as professional. Reported gaming frequency
was: 15 almost never, 4 less than once a month, 6 less than once a week, 4 one to three times per week, 3 three to five
times per week, and 5 almost daily.
Familiarity with the remaining applications was documented as well: 50 of 51 driving-study participants held a
driver's license, for $8.0\,\pm\,6.0$ years, and 8 of 45 gaming sessions came from five participants who had
completed the $n$-back in the driving study. Previously played titles were reported in 24 of 45 gaming sessions. Among
these, \emph{Hogwarts Legacy} was selected twice and \emph{Overcooked! 2} not at all, indicating that both titles were
novel or at least not among the most recently played for nearly all participants.
The included participants from the driving study had a mean age of $26\,\pm\,6$ years (24 female, 26 male, 1 unknown).
Across both studies, the 83 participants comprised 37 females, 45 males, and one participant who did not
report sex, aged 18--59 years at first participation (mean $27\,\pm\,8$ years). Participants who took part more than
once are counted once, with the data of their first session.
\subsection{Subjective Measures}
\begin{figure}[!htb]
  \centering
  \subfloat[\centering $n$-back]{\label{sfig:tlx-dimensions-n-back}\includegraphics[width=0.48\linewidth]{%
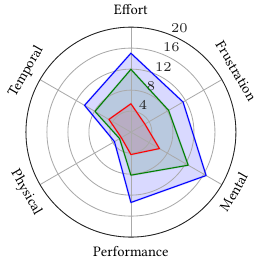}}%
  \hfill
  \subfloat[\centering \emph{Hogwarts Legacy}]{\label{sfig:tlx-dimensions-hogwarts}\includegraphics[width=0.48\linewidth]{%
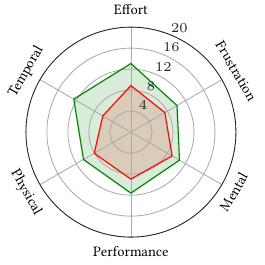}}%
  \\[4pt]
  \subfloat[\centering \emph{Overcooked! 2}]{\label{sfig:tlx-dimensions-overcooked}\includegraphics[width=0.48\linewidth]{%
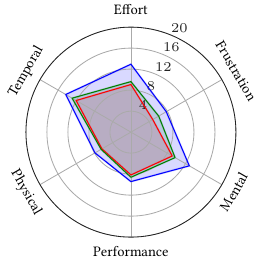}}%
  \caption{Self-reported task load indices for each dimension from our $n$-back test, alongside experiences
in the gaming scenarios of \emph{Hogwarts Legacy} and \emph{Overcooked! 2}. Each line and colored area represents a
different level, with red as the first level, green as the second level, and blue as the third level.}
  \label{fig:nasa-rtlx-all}
\end{figure}

\begin{figure}[!htb]
  \centering
  \subfloat[\centering $n$-back]{\label{sfig:rtlx-n-back}\includegraphics[width=0.48\linewidth]{%
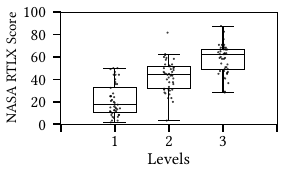}}%
  \hfill
  \subfloat[\centering \emph{Hogwarts Legacy}]{\label{sfig:rtlx-hogwarts}\includegraphics[width=0.48\linewidth]{%
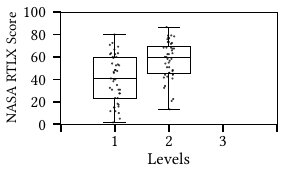}}%
  \\[4pt]
  \subfloat[\centering \emph{Overcooked! 2}]{\label{sfig:rtlx-overcooked}\includegraphics[width=0.48\linewidth]{%
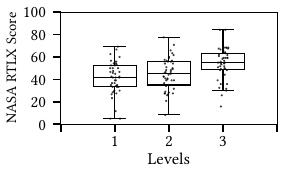}}%
  \caption{Aggregated \gls{rtlx} for $n$-back, \emph{Hogwarts Legacy} and \emph{Overcooked! 2} for increasing
levels.}
  \label{fig:nasa-rtlx}
\end{figure}
Subjective task load was measured after each task level using the NASA-\gls{tlx} questionnaire
\citep{hartDevelopmentNASATLXTask1988}. The $n$-back task was associated with higher performance and mental demand
(Fig.~\ref{sfig:tlx-dimensions-n-back}), while \emph{Hogwarts Legacy} and \emph{Overcooked! 2} showed greater temporal
demand (Fig.~\ref{sfig:tlx-dimensions-hogwarts}, \subref*{sfig:tlx-dimensions-overcooked}).
Repeated-measures ANOVA confirmed a significant main effect of task level on NASA-\gls{rtlx} scores for all three task
domains, with detailed statistics and Holm-corrected post-hoc comparisons are reported in
Table~\ref{tab:labels-subjective-measurements-levels}.
\begin{table*}[!t]
  \centering
  \setlength{\tabcolsep}{4pt}
  \tablefont
  \caption{Repeated-measures ANOVA results and Holm-corrected post-hoc comparisons for NASA-RTLX scores across task difficulty levels ($n = 37$).}
  \label{tab:labels-subjective-measurements-levels}
  \begin{tabularx}{\linewidth}{llccX}
    \toprule
    Application & $F(df_1, df_2)$ & $p$ & $\eta^2_G$ & Post-hoc pairwise (Holm-corrected) \\
    \midrule
    $n$-back
      & $F(2, 72) = 170.45$ & ${<}.001$ & $0.590$
      & $\mathrm{L1}{<}\mathrm{L2}:p<.001$, $\mathrm{L1}{<}\mathrm{L3}:p<.001$, $\mathrm{L2}{<}\mathrm{L3}:p<.001$ \\[2pt]
    \emph{Overcooked! 2}
      & $F(2, 72) = 15.90$ & ${<}.001$ & $0.113$
      & $\mathrm{L1}{<}\mathrm{L2}:p=0.138$, $\mathrm{L1}{<}\mathrm{L3}:p<.001$, $\mathrm{L2}{<}\mathrm{L3}:p<.001$ \\[2pt]
    \emph{Hogwarts L.}
      & $F(1, 36) = 78.80$ & ${<}.001$ & $0.168$
      & $\mathrm{L1}{<}\mathrm{L2}:p<.001$ \\
    \bottomrule
  \end{tabularx}
  \begin{minipage}{\linewidth}
    \vspace{2pt}\footnotesize\textit{Note.}
    $\eta^2_G$ is the generalized eta squared.
    Mauchly's test rejected sphericity for $n$-back ($\varepsilon = 0.71$) and \emph{Overcooked! 2} ($\varepsilon = 0.79$). Greenhouse-Geisser-corrected $p$-values are all ${<}.001$, the table reports uncorrected values.
    Post-hoc $p$-values are Holm-corrected pairwise comparisons.
    \emph{Hogwarts Legacy} has only two difficulty levels (L1, L2).
  \end{minipage}
\end{table*}

These results are consistent with previous findings and confirm that subjective task load scores increase with task
difficulty in all three tasks (Fig.~\ref{sfig:rtlx-n-back}, \subref*{sfig:rtlx-hogwarts}, and
\subref*{sfig:rtlx-overcooked}). Compared to the $n$-back, real-world gaming tasks exhibited greater variability
in self-reported load, reflecting the individualized nature of these experiences. Overall, the results highlight the
importance of collecting diverse real-world data to comprehensively assess task load, while supporting our task load
annotation setting.
\subsection{Performance Metrics}
\begin{figure}[!htb]
  \centering
  \subfloat[\centering{$n$-back precision}]{\label{sfig:n-back-performance-precision}%
\includegraphics[width=0.48\linewidth]{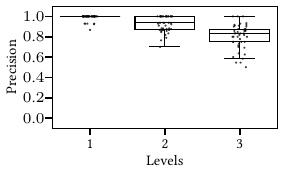}}%
  \hfill
  \subfloat[\centering{$n$-back recall}]{\label{sfig:n-back-performance-recall}%
\includegraphics[width=0.48\linewidth]{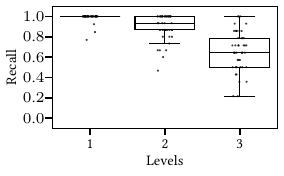}}%
  \\[4pt]
  \subfloat[\centering{$n$-back reaction time}]{\label{sfig:n-back-performance-reaction-time}%
\includegraphics[width=0.48\linewidth]{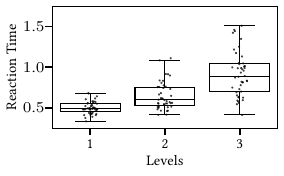}}%
  \caption{$n$-back performance measures.}
  \label{fig:performance-ratings-n-back}
\end{figure}

Performance results for the $n$-back test confirm that increasing task difficulty lowers precision and recall
(Fig.~\ref{sfig:n-back-performance-precision}, \subref*{sfig:n-back-performance-recall}) and lengthens reaction times
(Fig.~\ref{sfig:n-back-performance-reaction-time}), consistent with previous findings
\citep{oppeltADABaseMultimodalDataset2022}. At lower levels, participants responded quickly and accurately, while higher
levels resulted in reduced and more variable performance.
Friedman tests confirmed a significant main effect of task level on all three $n$-back performance metrics, with
performance consistently degrading across difficulty levels. Detailed statistics and Holm-corrected post-hoc
comparisons (Wilcoxon signed-rank) are reported in Table~\ref{tab:labels-performance-measurements-n-back}.
\begin{table*}[!t]
  \centering
  \setlength{\tabcolsep}{4pt}
  \tablefont
  \caption{Friedman test results and Holm-corrected post-hoc comparisons for $n$-back performance metrics across task difficulty levels ($n = 37$).}
  \label{tab:labels-performance-measurements-n-back}
  \begin{tabularx}{\linewidth}{llccX}
    \toprule
    Metric & $\chi^2(df)$ & $p$ & $W$ & Post-hoc pairwise (Holm-corrected) \\
    \midrule
    Precision
      & $\chi^2(2) = 43.26$ & $< .001$ & $0.585$
      & $\mathrm{L1}>\mathrm{L2}$: $p < .001$, $\mathrm{L1}>\mathrm{L3}$: $p < .001$, $\mathrm{L2}>\mathrm{L3}$: $p < .001$ \\[2pt]
    Recall
      & $\chi^2(2) = 55.13$ & $< .001$ & $0.745$
      & $\mathrm{L1}>\mathrm{L2}$: $p < .001$, $\mathrm{L1}>\mathrm{L3}$: $p < .001$, $\mathrm{L2}>\mathrm{L3}$: $p < .001$ \\[2pt]
    Reaction Time
      & $\chi^2(2) = 56.00$ & $< .001$ & $0.757$
      & $\mathrm{L1}<\mathrm{L2}$: $p < .001$, $\mathrm{L1}<\mathrm{L3}$: $p < .001$, $\mathrm{L2}<\mathrm{L3}$: $p < .001$ \\
    \bottomrule
  \end{tabularx}
  \begin{minipage}{\linewidth}
    \vspace{2pt}\footnotesize\textit{Note.}
    Friedman test; post-hoc: Wilcoxon signed-rank test with Holm correction.
    $W$ = Kendall's coefficient of concordance.
  \end{minipage}
\end{table*}

\begin{figure}[!htb]
  \centering
  \subfloat[\centering{\emph{Overcooked! 2} successful recipes per cook}]{\label{sfig:s-cook-performance-successful-recipes}%
\includegraphics[width=0.48\linewidth]{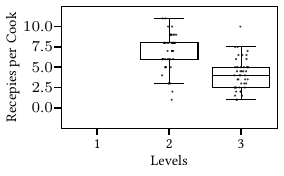}}%
  \hfill
  \subfloat[\centering{\emph{Overcooked! 2} tip per cook}]{\label{sfig:s-cook-performance-tip}%
\includegraphics[width=0.48\linewidth]{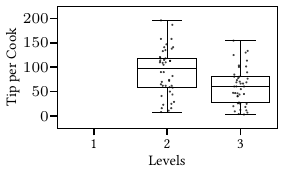}}%
  \\[4pt]
  \subfloat[\centering{\emph{Overcooked! 2} recipe points per cook}]{\label{sfig:s-cook-performance-recipes_points}%
\includegraphics[width=0.48\linewidth]{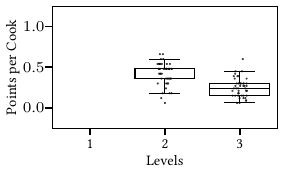}}%
  \caption{\emph{Overcooked! 2} performance measures.}
  \label{fig:performance-ratings-s-cook}
\end{figure}

In \emph{Overcooked! 2}, performance metrics including successful recipes, tips, and recipe points were corrected for
dual-cook levels (two cooks can be controlled independently). Score variance increased at higher loads, indicating
greater individual differences (Fig.~\ref{sfig:s-cook-performance-successful-recipes},
\subref*{sfig:s-cook-performance-tip}, and \subref*{sfig:s-cook-performance-recipes_points}). Friedman tests confirmed a significant main
effect of task level on all three metrics, with detailed statistics and Holm-corrected post-hoc comparisons (Wilcoxon
signed-rank) reported in Table~\ref{tab:labels-performance-measurements-s-cook}.
\begin{table*}[!t]
  \centering
  \setlength{\tabcolsep}{4pt}
  \tablefont
  \caption{Friedman test results and Holm-corrected post-hoc comparisons for \emph{Overcooked!~2} performance metrics across task difficulty levels ($n = 37$; Tip: $n = 35$).}
  \label{tab:labels-performance-measurements-s-cook}
  \begin{tabularx}{\linewidth}{llccX}
    \toprule
    Metric & $\chi^2(df)$ & $p$ & $W$ & Post-hoc pairwise (Holm-corrected) \\
    \midrule
    Successful Recipes per Cook
      & $\chi^2(1) = 32.11$ & $< .001$ & $0.868$
      & $\mathrm{L2}>\mathrm{L3}$: $p < .001$ \\[2pt]
    Tip per Cook
      & $\chi^2(1) = 15.11$ & $< .001$ & $0.432$
      & $\mathrm{L2}>\mathrm{L3}$: $p < .001$ \\[2pt]
    Recipe Points per Cook [k]
      & $\chi^2(1) = 32.11$ & $< .001$ & $0.868$
      & $\mathrm{L2}>\mathrm{L3}$: $p < .001$ \\
    \bottomrule
  \end{tabularx}
  \begin{minipage}{\linewidth}
    \vspace{2pt}\footnotesize\textit{Note.}
    Friedman test; post-hoc: Wilcoxon signed-rank test with Holm correction.
    $W$ = Kendall's coefficient of concordance.
    Scores are per-cook values, corrected for dual-operator levels.
    Scores are recorded for levels 2 and 3 only.
    A participant missing a metric at one level is excluded from that metric only.
  \end{minipage}
\end{table*}

\begin{figure}[!htb]
  \centering
  \subfloat[\centering{\emph{Hogwarts Legacy} missed rings}]{\label{sfig:u-harry-performance-missed-rings}%
\includegraphics[width=0.48\linewidth]{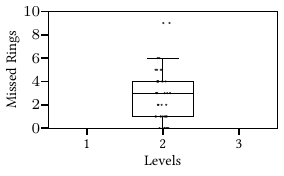}}%
  \hfill
  \subfloat[\centering{\emph{Hogwarts Legacy} race time [s]}]{\label{sfig:u-harry-performance-time-seconds}%
\includegraphics[width=0.48\linewidth]{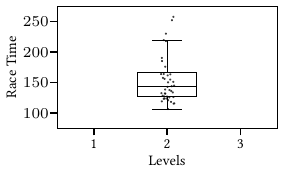}}%
  \\[4pt]
  \subfloat[\centering{\emph{Hogwarts Legacy} total time [s]}]{\label{sfig:u-harry-performance-total-time-seconds}%
\includegraphics[width=0.48\linewidth]{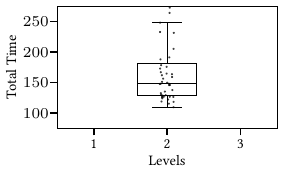}}%
  \caption{\emph{Hogwarts Legacy} performance measures.}
  \label{fig:performance-ratings-u-harry}
\end{figure}

For \emph{Hogwarts Legacy}, only the racing level provided formal performance data (missed rings, race time, and
total time with penalties), as lower levels lacked such metrics. Descriptive results are shown in
Fig.~\ref{sfig:u-harry-performance-missed-rings}, \subref*{sfig:u-harry-performance-time-seconds}, and
\subref*{sfig:u-harry-performance-total-time-seconds}.
\subsection{Task Load Estimation}
\label{ssec:task-load-estimation}
We evaluated the three encoders introduced above (ConvNeXt, \gls{xlstm}, and transformer) on binary task-load
classification across domains. Following established practice \citep{oppeltADABaseMultimodalDataset2022,%
foltynEvaluatingRobustnessMultimodal2024}, the baselines used physiological signals (two-lead \gls{ecg}, \gls{ppg},
\gls{eda}, \gls{emg}, \gls{rsp}, \gls{skt}) together with pupil diameter and gaze from the eye tracker. Models
were trained on 40 second input windows extracted with an overlapping sliding window of $20$ second step, retaining
only sequences with at least $90\,\%$ overlap with the low or high task-load labels
\citep{foltynEvaluatingRobustnessMultimodal2024}. A prediction is made every 20 seconds and integrates the
preceding 40 seconds. The step is a free parameter at inference and can be reduced at the cost of additional forward 
passes. With per-window normalization, which requires no data outside the input window, this setting can be applied
online at a slightly lower \gls{auroc} (Supplementary Section~\ref{ssec:supp-window-normalization}).

\begin{figure*}[!t]
  \centering
  \includegraphics{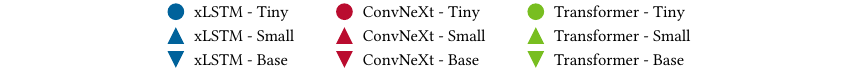} \vspace{-0.75cm} \\
  \subfloat[\centering $n$-back (Driving Study)]{\label{sfig:baselines-n-back-drive}\includegraphics[width=0.33\linewidth]{%
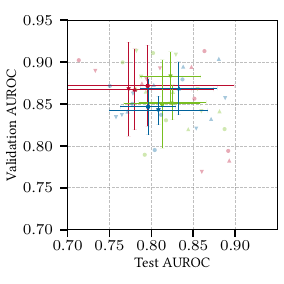}}%
  \subfloat[\centering $n$-back (Gaming Study)]{\label{sfig:baselines-n-back-game}\includegraphics[width=0.33\linewidth]{%
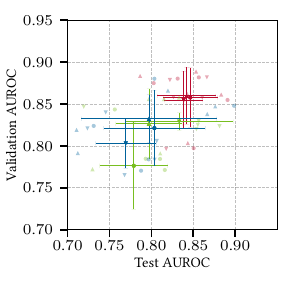}}%
  \subfloat[\centering $n$-back (Combined)]{\label{sfig:baselines-n-back-combined}\includegraphics[width=0.33\linewidth]{%
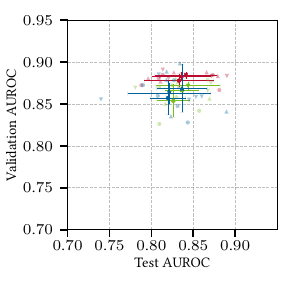}} \\ %
  \subfloat[\centering Driving Application]{\label{sfig:baselines-drive}\includegraphics[width=0.33\linewidth]{%
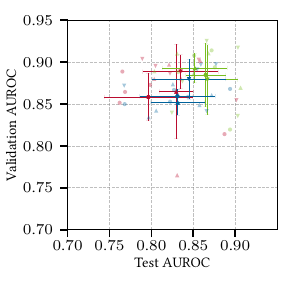}}%
  \subfloat[\centering Hogwarts Legacy]{\label{sfig:baselines-hogwarts}\includegraphics[width=0.33\linewidth]{%
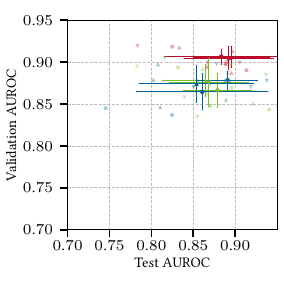}}%
  \subfloat[\centering Overcooked! 2]{\label{sfig:baselines-overcooked}\includegraphics[width=0.33\linewidth]{%
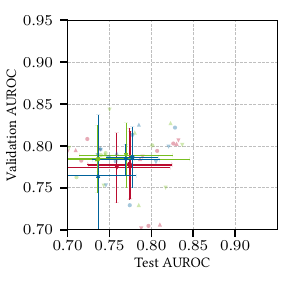}} \\ %
  \subfloat[\centering Gaming]{\label{sfig:baselines-game}\includegraphics[width=0.33\linewidth]{%
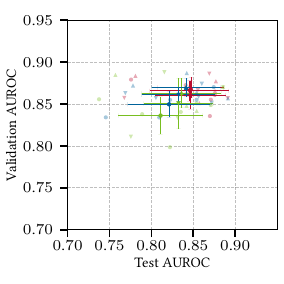}}%
  \subfloat[\centering Combined Applications]{\label{sfig:baselines-applications}\includegraphics[width=0.33\linewidth]{%
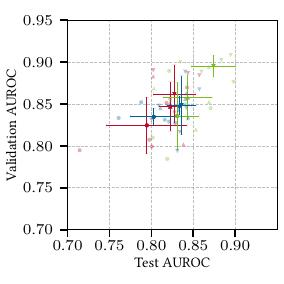}}%
  \subfloat[\centering Complete]{\label{sfig:baselines-complete}\includegraphics[width=0.33\linewidth]{%
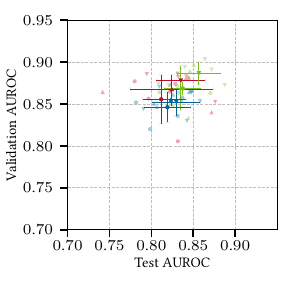}}%
  \caption{Performance of the encoder architectures on the different datasets and modalities. The colors indicate the %
network architecture and the columns show the different datasets. Marker shapes encode network size. Performance is %
shown as the mean and standard deviation of the test \gls{auroc} on the x-axis and the validation \gls{auroc} on the %
y-axis. Every computed metric is also shown as a transparent marker to expose the underlying distribution.}
  \label{fig:supervised-baselines}
  \vspace{-0.15cm}
\end{figure*}

Figure~\ref{fig:supervised-baselines} summarizes the architectural comparison across domains, from the three $n$-back
variants (Fig.~\ref{sfig:baselines-n-back-drive}--\subref*{sfig:baselines-n-back-combined}) over the single applications
(Fig.~\ref{sfig:baselines-drive}--\subref*{sfig:baselines-overcooked}) to the pooled datasets
(Fig.~\ref{sfig:baselines-game}--\subref*{sfig:baselines-complete}). We report the mean and
standard deviation of validation and test \gls{auroc} over five subject-wise folds, with test \gls{auroc} on the x-axis
and validation \gls{auroc} on the y-axis. Validation and test scores align closely, with a mean validation-to-test gap
of $0.03$ \gls{auroc}. This indicates no overfitting to the training participants. It does not rule out
cohort- or setup-specific features, since validation and test data come from the same cohorts
(Section~\ref{ssec:cohort-shift}). Across datasets the
three encoders performed comparably, and the transformer reached the highest test \gls{auroc} on the combined domains,
for example $0.86 \pm 0.03$ on the complete dataset against $0.84 \pm 0.03$ for ConvNeXt and $0.83 \pm 0.03$ for
\gls{xlstm}, each in its base configuration.

We report \gls{auroc} in the figures because it is threshold and prevalence agnostic, summarizes the full sensitivity
and specificity trade-off, and enables fair comparisons across folds and domains. Accuracy depends on a chosen decision
threshold and therefore serves as a complementary metric. We computed accuracy at a fixed decision threshold of $0.5$
on the predicted high-load probability $\hat{p}$. For readability we summarize it here as the mean and standard deviation
across the five subject-wise folds, with all configurations using the \emph{base} model. On the complete dataset the 
encoders reached  $0.75 \pm 0.03$ (ConvNeXt), $0.76 \pm 0.03$ (transformer), and $0.74 \pm 0.03$ (\gls{xlstm}). On the
combined $n$-back condition they reached $0.76 \pm 0.03$, $0.77 \pm 0.04$, and $0.74 \pm 0.03$, on the driving application
$0.76 \pm 0.05$, $0.79 \pm 0.03$, and $0.78 \pm 0.06$, and on the gaming application $0.74 \pm 0.06$, $0.74 \pm 0.04$,
and $0.75 \pm 0.05$, in the same order. The accuracy ordering is consistent with the \gls{auroc} results and confirms
that the three encoders perform comparably across domains.

We also examined how each encoder responds to its own capacity settings, using the \emph{tiny}, \emph{small}, and
\emph{base} configurations described in Section~\ref{sec:model}. Within every encoder the larger configuration
performed at least as well as the smaller ones when pooled across domains. The transformer benefited most, improving
from $0.83$ to $0.84$ test \gls{auroc} between its tiny and base variants, whereas ConvNeXt and \gls{xlstm}
were almost insensitive to their size, varying by at most $0.01$ test \gls{auroc}. Consistent with the nomenclature, we
do not interpret these labels as a size-matched comparison across architectures, having a descriptive comparison for
future studies, without using it as selection criterion for subsequent analyses.

\begin{figure}[!htb]
  \centering
  \subfloat[\centering ConvNeXt]{\label{sfig:input-length-convnext}\includegraphics[width=0.48\linewidth]{%
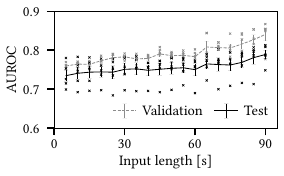}}%
  \hfill
  \subfloat[\centering \gls{xlstm}]{\label{sfig:input-length-xlstm}\includegraphics[width=0.48\linewidth]{%
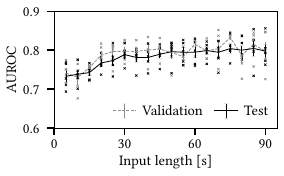}}%
  \\[4pt]
  \subfloat[\centering Transformer]{\label{sfig:input-length-transformer}\includegraphics[width=0.48\linewidth]{%
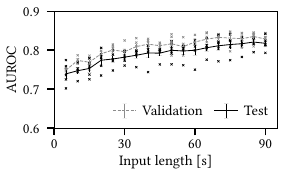}}%
  \caption{Task-load prediction \gls{auroc} for the three encoder architectures
  over increasing input lengths. Validation (dashed) and test (solid) \gls{auroc} show the mean over five
  subject-wise folds with the standard error. Dots mark the individual folds.}
  \label{fig:supervised-input-length}
\end{figure}

We further explored alternative window lengths for the ConvNeXt, \gls{xlstm}, and transformer backbones from 5 seconds
to 1 minute 30 seconds, as shown in Fig.~\ref{sfig:input-length-convnext}, \subref*{sfig:input-length-xlstm}, and
\subref*{sfig:input-length-transformer}, respectively. Test \gls{auroc} increased with longer
inputs for all three encoders. The transformer improved from $0.74$ at 5 seconds to $0.79$ at 40 seconds and $0.82$ at
90 seconds, and the \gls{xlstm} followed a similar trajectory from $0.73$ to $0.78$ and $0.80$. Both backbones gained
most below roughly 30 seconds and then largely plateaued, while ConvNeXt continued to improve up to the longest window.
Validation \gls{auroc} showed the same increase with input length and lay $0.01$--$0.04$ above test \gls{auroc}
(Fig.~\ref{fig:supervised-input-length}).
The 40-second window used throughout this work follows established practice
\citep{oppeltADABaseMultimodalDataset2022,foltynEvaluatingRobustnessMultimodal2024} and the timescale of heart rate
variability, which changes over 30 seconds or more \citep{phamHeartRateVariability2021}, while keeping the delay
acceptable for adaptive \gls{hci}. The analysis confirms this setting as reasonable. On the test data, windows of 30
seconds reached a comparable \gls{auroc} (at most $0.01$ lower), whereas windows of 80 seconds raised the \gls{auroc}
by about $0.02$ for all three encoders at the cost of a longer delay.

Our baselines establish standardized reference points across architectures, modalities, and domains, ensuring
reproducibility and comparability for future work. They establish the expected performance for commonly used inputs,
namely biosignals and eye tracking, under a consistent protocol \citep{oppeltADABaseMultimodalDataset2022,%
foltynEvaluatingRobustnessMultimodal2024}. They disentangle domain effects from architecture effects by holding the
training and evaluation procedure constant, and they identify ceiling and floor effects that indicate where
multimodal fusion or task redesign is warranted.
\subsection{Unimodal and Multimodal Performance}
\label{sssec:uni-multi-modal}
We quantified the contribution of each input using the \emph{small} ConvNeXt as the representative architecture
(\gls{xlstm} and transformer in Supplementary
Tables~\ref{tab:supp-modalities-single-xlstm}-\ref{tab:modalities-single}).
Among the physiological signals we distinguish the classical biosignals captured with dedicated sensing equipment
(\gls{ecg}, \gls{ppg}, \gls{eda}, \gls{emg}, \gls{rsp}, and \gls{skt}) from the eye-tracker-derived pupil diameter.
Among the classical biosignals, \gls{ecg} was the strongest biosignal on the complete dataset for all three encoders.
In driving it led for ConvNeXt ($0.77$, Table~\ref{tab:modalities-single-convnext}) and was matched or slightly
exceeded by \gls{emg} for \gls{xlstm} ($0.74$ vs.\ $0.74$) and the transformer ($0.75$ vs.\ $0.72$). \gls{eda} was the
strongest biosignal in Hogwarts Legacy ($0.77$), Overcooked!~2 ($0.66$) and the combined $n$-back ($0.65$), and on par
with \gls{ecg} in Gaming overall ($0.70$ vs.\ $0.71$). Combining \gls{ecg} and \gls{ppg} into a joint cardiac input
degraded performance relative to \gls{ecg} alone (e.g., $0.55$ vs.\ $0.73$ on the complete dataset
Table~\ref{tab:modalities-multiple-convnext}), suggesting that \gls{ppg} introduces redundant or noisy channels rather
than complementary information when processed end-to-end. Pooling all six biosignals improved over the cardiac-only
combination (e.g., from $0.56$ to $0.69$ in driving) yet generally did not exceed the best individual biosignal within a
given domain. Pupil diameter was highly predictive in the cognitive $n$-back task (\gls{auroc} $0.80$ on the combined
$n$-back Table~\ref{tab:modalities-single-convnext}) but weaker in driving ($0.64$). Among the behavioral signals,
eye-tracker-derived eye movement achieved its highest single-modality performance in the driving scenarios ($0.83$),
while facial action units were the strongest predictor overall in driving ($0.88$) and in the driving-study $n$-back
($0.85$). \glspl{au} were near chance, however, in the gaming-study $n$-back ($0.54$), Hogwarts Legacy ($0.48$), and
Overcooked!~2 ($0.50$), indicating that their informativeness is study-setup or domain dependent. Combining the two
eye-tracking channels, pupil diameter and gaze, was consistently among the best inputs across all data subsets
(Table~\ref{tab:modalities-multiple-convnext}). Head rotation was informative in driving and in the driving-study
$n$-back ($0.78$ in both), plausibly reflecting postural adjustments under load, whereas absolute head and body
positions were uninformative ($\leq 0.63$). Single modalities are summarised in
Table~\ref{tab:modalities-single-convnext} and multimodal combinations in Table~\ref{tab:modalities-multiple-convnext},
with values per architecture shown in Table~\ref{tab:modalities-multiple}.

Across all data subsets, the fusion of biosignals and eye tracking provided the most robust baseline combination
(\gls{auroc} $0.83$ on the complete dataset Table~\ref{tab:modalities-multiple-convnext}), and adding facial action units
yielded the largest gains in driving-related domains (from $0.85$ to $0.90$ in the Driving Application) and smaller,
inconsistent changes in gaming (from $0.81$ to $0.85$ in Gaming, but from $0.90$ to $0.89$ in Hogwarts Legacy). The full
fusion of biosignals, eye tracking, and \glspl{au} reached the highest overall \gls{auroc} ($0.87$ on the complete dataset).
Adding movement features to the full combination did not improve and, in some gaming settings, reduced performance
($0.83$ vs.\ $0.87$ on the complete dataset and $0.78$ vs.\ $0.85$ in Gaming). Fusion of biosignals and eye
tracking exceeded the individual biosignal and eye-tracking modalities in 61 of the 64 domain $\times$ modality
comparisons across the eight domains, and the advantage was significant in 56 of them after Holm correction
(Table~\ref{tab:modality-multimodal-vs-unimodal}). In the remaining three it trailed non-significantly, and all three
involved pupil diameter ($-0.012$, $-0.014$, and $-0.013$ on the driving-study and gaming-study $n$-back and in
\emph{Overcooked! 2}). Pupil diameter matched fusion in every $n$-back and gaming domain (six non-significant
comparisons), and eye movement was also not significantly worse in \emph{Overcooked! 2} ($+0.062$) and in the driving
application ($+0.030$). Tested against the best single modality in each domain and per encoder (Supplementary
Table~\ref{tab:supp-multimodal-vs-unimodal}), fusion was ahead in eleven of the twelve comparisons of the four main
domains (the complete dataset, the combined $n$-back, driving, and gaming, each with three encoders), with a mean
difference of $+0.033$. The exception was the x\gls{lstm} on the combined $n$-back ($-0.005$, n.s.). Four of these were
significant after Holm correction across all 24 domain and encoder pairs, namely the complete dataset with every
encoder ($+0.057$ to $+0.092$, $p_{\mathrm{Holm}} = .002$) and driving with the x\gls{lstm}
($+0.067$, $p_{\mathrm{Holm}} = .048$). The best single input was eye movement in these two domains and pupil diameter
in all others. Fusion therefore helped clearly where eye movement was the best single input, whereas on the $n$-back and
in gaming, pupil diameter alone performed as well as fusion. No comparison favored a unimodal input significantly.

The most informative single modality, however, depended on the task, which we examined through three pre-specified
contrasts (Table~\ref{tab:modality-contrasts}). Pupil diameter outperformed \gls{ecg} in every cognitively demanding
condition, including both $n$-back cohorts and all gaming tasks ($\Delta\mathrm{AUROC}$ between $+0.10$ and
$+0.30$, $p_{\mathrm{Holm}} \leq .005$), the two were equivalent on the complete dataset
($+0.010$, n.s.), and in driving \gls{ecg} was significantly better ($\Delta\mathrm{AUROC} = -0.066$, 95\,\% CI
$[-0.12, -0.02]$, $p_{\mathrm{Holm}} = .023$).
Pupillary dilation is thus a reliable index of cognitive task load but not of the load induced by driving. Facial action
units surpassed pupil diameter only in driving ($\Delta\mathrm{AUROC} = +0.205$, $p_{\mathrm{Holm}} =
.001$). In gaming and in the gaming-study $n$-back the ranking reversed in favor of pupil diameter
($\Delta\mathrm{AUROC}$ between $-0.25$ and $-0.37$, $p_{\mathrm{Holm}} < .001$), which was also significantly
better on the combined $n$-back ($\Delta\mathrm{AUROC} = -0.067$, $[-0.12, -0.02]$, $p_{\mathrm{Holm}} = .015$), 
indicating that facial expressions capture driving-specific demand rather than a domain-general load signature.
Finally, pupil diameter exceeded eye movement throughout the cognitive and gaming conditions ($p_{\mathrm{Holm}}$
from $< .001$ to $.004$), whereas eye movement was significantly better in driving ($\Delta\mathrm{AUROC}
= -0.162$, $p_{\mathrm{Holm}} < .001$) and on the complete dataset ($-0.040$, $[-0.08, -0.00]$, $p_{\mathrm{Holm}} =
.041$). This is consistent with eye movement reflecting visual scanning demands during driving rather than cognitive
effort per se. In sum, single modalities are diagnostic in a task-specific way, while multimodal fusion is
never significantly worse than any single input, and significantly better than the best one on the complete
dataset with every encoder and in driving with the x\gls{lstm}.

\begin{table*}[!t]

\caption{Mean and standard deviation of the AUROC for single modalities (ConvNeXt).}
  \label{tab:modalities-single-convnext}
  \centering
  \setlength{\tabcolsep}{3pt}
  \renewcommand{\arraystretch}{1.25}
  \begin{NiceTabular}{r|c|c|c|c|c|c|c|c|c|c}
  \CodeBefore
      \cellcolor[RGB]{255, 185, 0}{2-2,6-4}
      \cellcolor[RGB]{255, 189, 0}{2-3,9-2}
      \cellcolor[RGB]{255, 183, 0}{2-4}
      \cellcolor[RGB]{255, 203, 0}{2-5,4-2,5-8,7-8}
      \cellcolor[RGB]{202, 202, 13}{2-6,4-5,8-11}
      \cellcolor[RGB]{255, 171, 0}{2-7}
      \cellcolor[RGB]{255, 193, 0}{2-8,6-6}
      \cellcolor[RGB]{237, 206, 4}{2-9,5-2}
      \cellcolor[RGB]{211, 203, 10}{2-10,3-11,4-10,8-6,9-9}
      \cellcolor[RGB]{205, 202, 12}{2-11}
      \cellcolor[RGB]{255, 202, 0}{3-2,5-9}
      \cellcolor[RGB]{199, 201, 13}{3-3,3-10,4-8,8-10}
      \cellcolor[RGB]{255, 184, 0}{3-4,6-5,10-8}
      \cellcolor[RGB]{255, 208, 0}{3-5,4-4}
      \cellcolor[RGB]{151, 194, 25}{3-6,7-9}
      \cellcolor[RGB]{234, 206, 5}{3-7}
      \cellcolor[RGB]{255, 192, 0}{3-8}
      \cellcolor[RGB]{255, 195, 0}{3-9,5-4,5-5,5-11}
      \cellcolor[RGB]{180, 198, 18}{4-3}
      \cellcolor[RGB]{218, 204, 9}{4-6,6-9}
      \cellcolor[RGB]{102, 181, 36}{4-7,8-8}
      \cellcolor[RGB]{160, 196, 22}{4-9,9-5}
      \cellcolor[RGB]{192, 200, 15}{4-11}
      \cellcolor[RGB]{255, 204, 0}{5-3}
      \cellcolor[RGB]{170, 197, 20}{5-6}
      \cellcolor[RGB]{255, 178, 0}{5-7,10-7}
      \cellcolor[RGB]{255, 206, 0}{5-10}
      \cellcolor[RGB]{255, 201, 0}{6-2}
      \cellcolor[RGB]{255, 170, 0}{6-3}
      \cellcolor[RGB]{215, 203, 10}{6-7}
      \cellcolor[RGB]{240, 207, 4}{6-8}
      \cellcolor[RGB]{255, 197, 0}{6-10,10-4}
      \cellcolor[RGB]{255, 187, 0}{6-11,7-4}
      \cellcolor[RGB]{255, 205, 0}{7-2}
      \cellcolor[RGB]{255, 200, 0}{7-3,10-9}
      \cellcolor[RGB]{243, 207, 3}{7-5}
      \cellcolor[RGB]{108, 184, 35}{7-6,8-2}
      \cellcolor[RGB]{189, 200, 16}{7-7}
      \cellcolor[RGB]{113, 187, 34}{7-10}
      \cellcolor[RGB]{138, 193, 28}{7-11}
      \cellcolor[RGB]{51, 157, 49}{8-3}
      \cellcolor[RGB]{62, 162, 46}{8-4}
      \cellcolor[RGB]{79, 170, 42}{8-5}
      \cellcolor[RGB]{45, 154, 50}{8-7,10-2}
      \cellcolor[RGB]{88, 174, 40}{8-9}
      \cellcolor[RGB]{77, 169, 43}{9-3}
      \cellcolor[RGB]{255, 191, 0}{9-4}
      \cellcolor[RGB]{60, 161, 47}{9-6}
      \cellcolor[RGB]{183, 199, 17}{9-7}
      \cellcolor[RGB]{208, 202, 11}{9-8}
      \cellcolor[RGB]{111, 185, 34}{9-10}
      \cellcolor[RGB]{135, 192, 28}{9-11}
      \cellcolor[RGB]{17, 140, 57}{10-3}
      \cellcolor[RGB]{116, 188, 33}{10-5}
      \cellcolor[RGB]{26, 144, 55}{10-6}
      \cellcolor[RGB]{141, 193, 27}{10-10}
      \cellcolor[RGB]{129, 191, 30}{10-11}
  \Body
      \hline
      ~ & \scriptsize{\makecell{n-back\\(Driving)}} & \scriptsize{\makecell{n-back\\Dual Task}} & \scriptsize{\makecell{n-back\\(Gaming)}} & \scriptsize{\makecell{n-back\\(Combined)}} & \scriptsize{\makecell{Driving\\Application}} & \scriptsize{\makecell{Hogwarts\\Legacy}} & \scriptsize{Overcooked! 2} & \scriptsize{Gaming} & \scriptsize{\makecell{Combined\\Applications}} & \scriptsize{Complete} \\
      \cline{2-11}
      ~ \scriptsize{RSP} & \scriptsize{0.50\,$\pm$\,0.059} & \scriptsize{0.52\,$\pm$\,0.052} & \scriptsize{0.50\,$\pm$\,0.073} & \scriptsize{0.57\,$\pm$\,0.047} & \scriptsize{0.65\,$\pm$\,0.063} & \scriptsize{0.45\,$\pm$\,0.090} & \scriptsize{0.53\,$\pm$\,0.110} & \scriptsize{0.61\,$\pm$\,0.054} & \scriptsize{0.64\,$\pm$\,0.028} & \scriptsize{0.65\,$\pm$\,0.013} \\ 
      ~ \scriptsize{EMG} & \scriptsize{0.57\,$\pm$\,0.021} & \scriptsize{0.66\,$\pm$\,0.033} & \scriptsize{0.50\,$\pm$\,0.021} & \scriptsize{0.58\,$\pm$\,0.031} & \scriptsize{0.71\,$\pm$\,0.032} & \scriptsize{0.61\,$\pm$\,0.065} & \scriptsize{0.53\,$\pm$\,0.028} & \scriptsize{0.54\,$\pm$\,0.034} & \scriptsize{0.66\,$\pm$\,0.030} & \scriptsize{0.64\,$\pm$\,0.016} \\ 
      ~ \scriptsize{EDA} & \scriptsize{0.57\,$\pm$\,0.047} & \scriptsize{0.68\,$\pm$\,0.036} & \scriptsize{0.58\,$\pm$\,0.072} & \scriptsize{0.65\,$\pm$\,0.040} & \scriptsize{0.63\,$\pm$\,0.042} & \scriptsize{0.77\,$\pm$\,0.078} & \scriptsize{0.66\,$\pm$\,0.073} & \scriptsize{0.70\,$\pm$\,0.078} & \scriptsize{0.64\,$\pm$\,0.035} & \scriptsize{0.66\,$\pm$\,0.038} \\ 
      ~ \scriptsize{SKT} & \scriptsize{0.61\,$\pm$\,0.110} & \scriptsize{0.57\,$\pm$\,0.074} & \scriptsize{0.54\,$\pm$\,0.052} & \scriptsize{0.54\,$\pm$\,0.076} & \scriptsize{0.69\,$\pm$\,0.028} & \scriptsize{0.48\,$\pm$\,0.116} & \scriptsize{0.57\,$\pm$\,0.111} & \scriptsize{0.57\,$\pm$\,0.123} & \scriptsize{0.58\,$\pm$\,0.035} & \scriptsize{0.54\,$\pm$\,0.048} \\ 
      ~ \scriptsize{PPG} & \scriptsize{0.56\,$\pm$\,0.094} & \scriptsize{0.45\,$\pm$\,0.104} & \scriptsize{0.50\,$\pm$\,0.075} & \scriptsize{0.50\,$\pm$\,0.048} & \scriptsize{0.53\,$\pm$\,0.085} & \scriptsize{0.64\,$\pm$\,0.085} & \scriptsize{0.61\,$\pm$\,0.073} & \scriptsize{0.64\,$\pm$\,0.041} & \scriptsize{0.55\,$\pm$\,0.048} & \scriptsize{0.51\,$\pm$\,0.043} \\ 
      ~ \scriptsize{ECG} & \scriptsize{0.58\,$\pm$\,0.060} & \scriptsize{0.56\,$\pm$\,0.058} & \scriptsize{0.51\,$\pm$\,0.022} & \scriptsize{0.60\,$\pm$\,0.042} & \scriptsize{0.77\,$\pm$\,0.032} & \scriptsize{0.67\,$\pm$\,0.085} & \scriptsize{0.57\,$\pm$\,0.054} & \scriptsize{0.71\,$\pm$\,0.032} & \scriptsize{0.76\,$\pm$\,0.023} & \scriptsize{0.73\,$\pm$\,0.026} \\ 
      ~ \scriptsize{ET Pupil Diameter} & \scriptsize{0.76\,$\pm$\,0.076} & \scriptsize{0.84\,$\pm$\,0.030} & \scriptsize{0.83\,$\pm$\,0.045} & \scriptsize{0.80\,$\pm$\,0.034} & \scriptsize{0.64\,$\pm$\,0.038} & \scriptsize{0.85\,$\pm$\,0.099} & \scriptsize{0.78\,$\pm$\,0.044} & \scriptsize{0.79\,$\pm$\,0.063} & \scriptsize{0.66\,$\pm$\,0.064} & \scriptsize{0.65\,$\pm$\,0.037} \\ 
      ~ \scriptsize{ET Move.} & \scriptsize{0.52\,$\pm$\,0.024} & \scriptsize{0.81\,$\pm$\,0.022} & \scriptsize{0.53\,$\pm$\,0.051} & \scriptsize{0.70\,$\pm$\,0.020} & \scriptsize{0.83\,$\pm$\,0.035} & \scriptsize{0.67\,$\pm$\,0.072} & \scriptsize{0.64\,$\pm$\,0.056} & \scriptsize{0.64\,$\pm$\,0.046} & \scriptsize{0.76\,$\pm$\,0.028} & \scriptsize{0.73\,$\pm$\,0.019} \\ 
      ~ \scriptsize{AUs} & \scriptsize{0.85\,$\pm$\,0.050} & \scriptsize{0.89\,$\pm$\,0.052} & \scriptsize{0.54\,$\pm$\,0.103} & \scriptsize{0.76\,$\pm$\,0.041} & \scriptsize{0.88\,$\pm$\,0.051} & \scriptsize{0.48\,$\pm$\,0.075} & \scriptsize{0.50\,$\pm$\,0.054} & \scriptsize{0.56\,$\pm$\,0.075} & \scriptsize{0.72\,$\pm$\,0.054} & \scriptsize{0.74\,$\pm$\,0.043} \\ 
\hline
  \end{NiceTabular}

\end{table*}
\begin{table*}[!t]

\caption{Mean and standard deviation of the AUROC for multi-modality combinations (ConvNeXt).}
  \label{tab:modalities-multiple-convnext}
  \centering
  \setlength{\tabcolsep}{3pt}
  \renewcommand{\arraystretch}{1.25}
  \begin{NiceTabular}{r|c|c|c|c|c|c|c|c|c|c}
  \CodeBefore
      \cellcolor[RGB]{255, 200, 0}{2-2,2-11,8-3}
      \cellcolor[RGB]{255, 193, 0}{2-3}
      \cellcolor[RGB]{255, 174, 0}{2-4,2-5}
      \cellcolor[RGB]{255, 202, 0}{2-6,9-9}
      \cellcolor[RGB]{255, 191, 0}{2-7}
      \cellcolor[RGB]{240, 207, 4}{2-8}
      \cellcolor[RGB]{234, 206, 5}{2-9,2-10,3-8,8-7}
      \cellcolor[RGB]{221, 204, 8}{3-2,3-3,3-11,7-5}
      \cellcolor[RGB]{255, 190, 0}{3-4}
      \cellcolor[RGB]{255, 205, 0}{3-5}
      \cellcolor[RGB]{167, 197, 21}{3-6,5-10,13-8}
      \cellcolor[RGB]{148, 194, 25}{3-7}
      \cellcolor[RGB]{164, 196, 22}{3-9,12-8}
      \cellcolor[RGB]{231, 206, 6}{3-10}
      \cellcolor[RGB]{85, 173, 40}{4-2,11-8}
      \cellcolor[RGB]{23, 143, 56}{4-3,10-3}
      \cellcolor[RGB]{40, 151, 51}{4-4,10-4,11-4}
      \cellcolor[RGB]{37, 150, 52}{4-5,11-9}
      \cellcolor[RGB]{43, 153, 51}{4-6}
      \cellcolor[RGB]{14, 139, 58}{4-7}
      \cellcolor[RGB]{91, 176, 39}{4-8,5-6,10-2}
      \cellcolor[RGB]{34, 148, 53}{4-9,10-6,13-3}
      \cellcolor[RGB]{48, 155, 49}{4-10,4-11,11-5}
      \cellcolor[RGB]{94, 177, 38}{5-2,10-8}
      \cellcolor[RGB]{157, 195, 23}{5-3,9-2,12-10}
      \cellcolor[RGB]{173, 197, 19}{5-4,9-4}
      \cellcolor[RGB]{199, 201, 13}{5-5}
      \cellcolor[RGB]{255, 206, 0}{5-7,6-11,9-8}
      \cellcolor[RGB]{246, 208, 2}{5-8,8-10}
      \cellcolor[RGB]{255, 208, 0}{5-9,6-10}
      \cellcolor[RGB]{183, 199, 17}{5-11}
      \cellcolor[RGB]{255, 195, 0}{6-2,6-5}
      \cellcolor[RGB]{255, 184, 0}{6-3}
      \cellcolor[RGB]{218, 204, 9}{6-4}
      \cellcolor[RGB]{253, 209, 1}{6-6,8-5,8-6}
      \cellcolor[RGB]{237, 206, 4}{6-7}
      \cellcolor[RGB]{255, 197, 0}{6-8,6-9}
      \cellcolor[RGB]{135, 192, 28}{7-2}
      \cellcolor[RGB]{215, 203, 10}{7-3}
      \cellcolor[RGB]{202, 202, 13}{7-4,9-10,9-11}
      \cellcolor[RGB]{119, 190, 32}{7-6}
      \cellcolor[RGB]{208, 202, 11}{7-7}
      \cellcolor[RGB]{255, 188, 0}{7-8}
      \cellcolor[RGB]{255, 201, 0}{7-9,8-8,8-11}
      \cellcolor[RGB]{205, 202, 12}{7-10,7-11,9-3}
      \cellcolor[RGB]{255, 209, 0}{8-2}
      \cellcolor[RGB]{255, 192, 0}{8-4}
      \cellcolor[RGB]{255, 199, 0}{8-9}
      \cellcolor[RGB]{192, 200, 15}{9-5}
      \cellcolor[RGB]{129, 191, 30}{9-6,12-4,12-7}
      \cellcolor[RGB]{227, 205, 7}{9-7}
      \cellcolor[RGB]{51, 157, 49}{10-5,12-3}
      \cellcolor[RGB]{0, 132, 61}{10-7,11-6}
      \cellcolor[RGB]{62, 162, 46}{10-9,13-2}
      \cellcolor[RGB]{57, 159, 47}{10-10,13-7}
      \cellcolor[RGB]{54, 158, 48}{10-11,12-6,13-11}
      \cellcolor[RGB]{68, 165, 45}{11-2}
      \cellcolor[RGB]{11, 137, 58}{11-3}
      \cellcolor[RGB]{6, 135, 60}{11-7,13-6}
      \cellcolor[RGB]{31, 147, 53}{11-10}
      \cellcolor[RGB]{17, 140, 57}{11-11}
      \cellcolor[RGB]{96, 179, 38}{12-2}
      \cellcolor[RGB]{79, 170, 42}{12-5}
      \cellcolor[RGB]{154, 195, 24}{12-9}
      \cellcolor[RGB]{141, 193, 27}{12-11}
      \cellcolor[RGB]{116, 188, 33}{13-4}
      \cellcolor[RGB]{65, 164, 45}{13-5}
      \cellcolor[RGB]{88, 174, 40}{13-9}
      \cellcolor[RGB]{77, 169, 43}{13-10}
  \Body
      \hline
      ~ & \scriptsize{\makecell{n-back\\(Driving)}} & \scriptsize{\makecell{n-back\\Dual Task}} & \scriptsize{\makecell{n-back\\(Gaming)}} & \scriptsize{\makecell{n-back\\(Combined)}} & \scriptsize{\makecell{Driving\\Application}} & \scriptsize{\makecell{Hogwarts\\Legacy}} & \scriptsize{Overcooked! 2} & \scriptsize{Gaming} & \scriptsize{\makecell{Combined\\Applications}} & \scriptsize{Complete} \\
      \cline{2-11}
      ~ \scriptsize{Heart Comb.} & \scriptsize{0.55\,$\pm$\,0.099} & \scriptsize{0.53\,$\pm$\,0.121} & \scriptsize{0.47\,$\pm$\,0.080} & \scriptsize{0.47\,$\pm$\,0.070} & \scriptsize{0.56\,$\pm$\,0.053} & \scriptsize{0.52\,$\pm$\,0.104} & \scriptsize{0.60\,$\pm$\,0.059} & \scriptsize{0.61\,$\pm$\,0.057} & \scriptsize{0.61\,$\pm$\,0.037} & \scriptsize{0.55\,$\pm$\,0.051} \\ 
      ~ \scriptsize{Biosignals} & \scriptsize{0.62\,$\pm$\,0.054} & \scriptsize{0.62\,$\pm$\,0.072} & \scriptsize{0.52\,$\pm$\,0.084} & \scriptsize{0.57\,$\pm$\,0.067} & \scriptsize{0.69\,$\pm$\,0.007} & \scriptsize{0.71\,$\pm$\,0.040} & \scriptsize{0.61\,$\pm$\,0.067} & \scriptsize{0.69\,$\pm$\,0.059} & \scriptsize{0.61\,$\pm$\,0.034} & \scriptsize{0.62\,$\pm$\,0.034} \\ 
      ~ \scriptsize{Eye Comb.} & \scriptsize{0.78\,$\pm$\,0.072} & \scriptsize{0.87\,$\pm$\,0.033} & \scriptsize{0.84\,$\pm$\,0.028} & \scriptsize{0.85\,$\pm$\,0.026} & \scriptsize{0.84\,$\pm$\,0.036} & \scriptsize{0.88\,$\pm$\,0.080} & \scriptsize{0.78\,$\pm$\,0.077} & \scriptsize{0.85\,$\pm$\,0.052} & \scriptsize{0.83\,$\pm$\,0.031} & \scriptsize{0.83\,$\pm$\,0.031} \\ 
      ~ \scriptsize{Head Rotation} & \scriptsize{0.78\,$\pm$\,0.054} & \scriptsize{0.70\,$\pm$\,0.029} & \scriptsize{0.68\,$\pm$\,0.037} & \scriptsize{0.65\,$\pm$\,0.024} & \scriptsize{0.78\,$\pm$\,0.033} & \scriptsize{0.58\,$\pm$\,0.032} & \scriptsize{0.59\,$\pm$\,0.075} & \scriptsize{0.58\,$\pm$\,0.037} & \scriptsize{0.68\,$\pm$\,0.027} & \scriptsize{0.67\,$\pm$\,0.015} \\ 
      ~ \scriptsize{Head Position} & \scriptsize{0.53\,$\pm$\,0.042} & \scriptsize{0.50\,$\pm$\,0.043} & \scriptsize{0.63\,$\pm$\,0.070} & \scriptsize{0.54\,$\pm$\,0.041} & \scriptsize{0.59\,$\pm$\,0.031} & \scriptsize{0.60\,$\pm$\,0.073} & \scriptsize{0.54\,$\pm$\,0.085} & \scriptsize{0.54\,$\pm$\,0.067} & \scriptsize{0.58\,$\pm$\,0.050} & \scriptsize{0.58\,$\pm$\,0.028} \\ 
      ~ \scriptsize{Head Comb.} & \scriptsize{0.72\,$\pm$\,0.075} & \scriptsize{0.63\,$\pm$\,0.053} & \scriptsize{0.64\,$\pm$\,0.038} & \scriptsize{0.62\,$\pm$\,0.060} & \scriptsize{0.74\,$\pm$\,0.045} & \scriptsize{0.64\,$\pm$\,0.053} & \scriptsize{0.51\,$\pm$\,0.042} & \scriptsize{0.56\,$\pm$\,0.057} & \scriptsize{0.64\,$\pm$\,0.041} & \scriptsize{0.64\,$\pm$\,0.035} \\ 
      ~ \scriptsize{Body Position} & \scriptsize{0.58\,$\pm$\,0.034} & \scriptsize{0.55\,$\pm$\,0.066} & \scriptsize{0.53\,$\pm$\,0.050} & \scriptsize{0.59\,$\pm$\,0.048} & \scriptsize{0.58\,$\pm$\,0.018} & \scriptsize{0.61\,$\pm$\,0.018} & \scriptsize{0.56\,$\pm$\,0.074} & \scriptsize{0.55\,$\pm$\,0.078} & \scriptsize{0.59\,$\pm$\,0.024} & \scriptsize{0.56\,$\pm$\,0.021} \\ 
      ~ \scriptsize{Move. Comb.} & \scriptsize{0.70\,$\pm$\,0.074} & \scriptsize{0.64\,$\pm$\,0.051} & \scriptsize{0.68\,$\pm$\,0.052} & \scriptsize{0.65\,$\pm$\,0.047} & \scriptsize{0.73\,$\pm$\,0.035} & \scriptsize{0.62\,$\pm$\,0.075} & \scriptsize{0.57\,$\pm$\,0.051} & \scriptsize{0.56\,$\pm$\,0.046} & \scriptsize{0.65\,$\pm$\,0.039} & \scriptsize{0.65\,$\pm$\,0.033} \\ 
      ~ \scriptsize{Biosignals, Eye} & \scriptsize{0.78\,$\pm$\,0.096} & \scriptsize{0.87\,$\pm$\,0.039} & \scriptsize{0.85\,$\pm$\,0.035} & \scriptsize{0.83\,$\pm$\,0.033} & \scriptsize{0.85\,$\pm$\,0.018} & \scriptsize{0.90\,$\pm$\,0.046} & \scriptsize{0.77\,$\pm$\,0.050} & \scriptsize{0.81\,$\pm$\,0.051} & \scriptsize{0.82\,$\pm$\,0.014} & \scriptsize{0.83\,$\pm$\,0.045} \\ 
      ~ \scriptsize{Bio., AUs, ET} & \scriptsize{0.81\,$\pm$\,0.080} & \scriptsize{0.88\,$\pm$\,0.031} & \scriptsize{0.85\,$\pm$\,0.026} & \scriptsize{0.83\,$\pm$\,0.043} & \scriptsize{0.90\,$\pm$\,0.028} & \scriptsize{0.89\,$\pm$\,0.056} & \scriptsize{0.79\,$\pm$\,0.059} & \scriptsize{0.85\,$\pm$\,0.048} & \scriptsize{0.86\,$\pm$\,0.042} & \scriptsize{0.87\,$\pm$\,0.022} \\ 
      ~ \scriptsize{Move., AUs, ET} & \scriptsize{0.77\,$\pm$\,0.088} & \scriptsize{0.83\,$\pm$\,0.030} & \scriptsize{0.73\,$\pm$\,0.053} & \scriptsize{0.79\,$\pm$\,0.036} & \scriptsize{0.82\,$\pm$\,0.046} & \scriptsize{0.73\,$\pm$\,0.072} & \scriptsize{0.69\,$\pm$\,0.057} & \scriptsize{0.70\,$\pm$\,0.044} & \scriptsize{0.70\,$\pm$\,0.025} & \scriptsize{0.72\,$\pm$\,0.019} \\ 
      ~ \scriptsize{Bio, AUs, ET, Move.} & \scriptsize{0.82\,$\pm$\,0.058} & \scriptsize{0.85\,$\pm$\,0.055} & \scriptsize{0.74\,$\pm$\,0.043} & \scriptsize{0.81\,$\pm$\,0.040} & \scriptsize{0.89\,$\pm$\,0.026} & \scriptsize{0.82\,$\pm$\,0.042} & \scriptsize{0.69\,$\pm$\,0.036} & \scriptsize{0.78\,$\pm$\,0.074} & \scriptsize{0.79\,$\pm$\,0.018} & \scriptsize{0.83\,$\pm$\,0.038} \\ 
\hline
  \end{NiceTabular}

\end{table*}
\begin{table*}[!t]
\caption{Mean and standard deviation of the AUROC for multi-modality combinations across datasets. The complete dataset consists of $n$-back, driving and gaming experiments.}
  \label{tab:modalities-multiple}
  \centering
  \setlength{\tabcolsep}{5.8pt}
  \renewcommand{\arraystretch}{1.1}
  \begin{NiceTabular}{r|ccc|ccc|ccc|ccc}
  \CodeBefore 
      \cellcolor[RGB]{192, 200, 15}{3-2,3-4,8-4,13-11}
      \cellcolor[RGB]{255, 199, 0}{3-3}
      \cellcolor[RGB]{255, 197, 0}{3-5,7-5,9-8,9-11}
      \cellcolor[RGB]{255, 174, 0}{3-6}
      \cellcolor[RGB]{255, 188, 0}{3-7}
      \cellcolor[RGB]{215, 203, 10}{3-8,3-10,10-7}
      \cellcolor[RGB]{255, 201, 0}{3-9,10-12}
      \cellcolor[RGB]{218, 204, 9}{3-11,8-7}
      \cellcolor[RGB]{240, 207, 4}{3-12}
      \cellcolor[RGB]{234, 206, 5}{3-13}
      \cellcolor[RGB]{157, 195, 23}{4-2}
      \cellcolor[RGB]{227, 205, 7}{4-3}
      \cellcolor[RGB]{176, 198, 19}{4-4}
      \cellcolor[RGB]{224, 205, 7}{4-5,8-6}
      \cellcolor[RGB]{255, 204, 0}{4-6,7-4,7-8,10-11}
      \cellcolor[RGB]{246, 208, 2}{4-7}
      \cellcolor[RGB]{138, 193, 28}{4-8,6-5,8-10,10-9}
      \cellcolor[RGB]{173, 197, 19}{4-9}
      \cellcolor[RGB]{148, 194, 25}{4-10,4-13,13-3}
      \cellcolor[RGB]{141, 193, 27}{4-11,6-2,14-13}
      \cellcolor[RGB]{170, 197, 20}{4-12,10-8}
      \cellcolor[RGB]{51, 157, 49}{5-2,12-4}
      \cellcolor[RGB]{57, 159, 47}{5-3,6-10,11-5}
      \cellcolor[RGB]{43, 153, 51}{5-4}
      \cellcolor[RGB]{45, 154, 50}{5-5,5-12,11-9,12-12,14-8}
      \cellcolor[RGB]{48, 155, 49}{5-6}
      \cellcolor[RGB]{37, 150, 52}{5-7,5-11}
      \cellcolor[RGB]{62, 162, 46}{5-8,14-3}
      \cellcolor[RGB]{54, 158, 48}{5-9,11-7,12-11,12-13}
      \cellcolor[RGB]{28, 146, 54}{5-10,12-3}
      \cellcolor[RGB]{40, 151, 51}{5-13,12-2}
      \cellcolor[RGB]{189, 200, 16}{6-3}
      \cellcolor[RGB]{135, 192, 28}{6-4,13-2}
      \cellcolor[RGB]{205, 202, 12}{6-6,10-4}
      \cellcolor[RGB]{144, 193, 26}{6-7,13-5,14-11}
      \cellcolor[RGB]{60, 161, 47}{6-8,11-4,11-6,12-6,12-7}
      \cellcolor[RGB]{99, 180, 37}{6-9}
      \cellcolor[RGB]{253, 209, 1}{6-11,7-2,8-11}
      \cellcolor[RGB]{255, 206, 0}{6-12}
      \cellcolor[RGB]{255, 200, 0}{6-13,8-12,9-2,9-3}
      \cellcolor[RGB]{255, 205, 0}{7-3,7-10}
      \cellcolor[RGB]{255, 193, 0}{7-6}
      \cellcolor[RGB]{255, 190, 0}{7-7,9-7}
      \cellcolor[RGB]{255, 209, 0}{7-9,9-6}
      \cellcolor[RGB]{255, 202, 0}{7-11}
      \cellcolor[RGB]{255, 196, 0}{7-12,9-5}
      \cellcolor[RGB]{250, 208, 1}{7-13}
      \cellcolor[RGB]{199, 201, 13}{8-2,10-6}
      \cellcolor[RGB]{211, 203, 10}{8-3}
      \cellcolor[RGB]{208, 202, 11}{8-5,10-3}
      \cellcolor[RGB]{125, 191, 31}{8-8,8-9}
      \cellcolor[RGB]{255, 192, 0}{8-13,9-10}
      \cellcolor[RGB]{255, 195, 0}{9-4}
      \cellcolor[RGB]{255, 208, 0}{9-9}
      \cellcolor[RGB]{255, 198, 0}{9-12}
      \cellcolor[RGB]{255, 187, 0}{9-13}
      \cellcolor[RGB]{202, 202, 13}{10-2}
      \cellcolor[RGB]{221, 204, 8}{10-5}
      \cellcolor[RGB]{154, 195, 24}{10-10}
      \cellcolor[RGB]{255, 191, 0}{10-13}
      \cellcolor[RGB]{65, 164, 45}{11-2,11-3,11-11,13-9}
      \cellcolor[RGB]{31, 147, 53}{11-8,12-10}
      \cellcolor[RGB]{26, 144, 55}{11-10}
      \cellcolor[RGB]{74, 168, 43}{11-12,14-6}
      \cellcolor[RGB]{71, 166, 44}{11-13}
      \cellcolor[RGB]{68, 165, 45}{12-5}
      \cellcolor[RGB]{34, 148, 53}{12-8}
      \cellcolor[RGB]{11, 137, 58}{12-9}
      \cellcolor[RGB]{132, 192, 29}{13-4}
      \cellcolor[RGB]{88, 174, 40}{13-6}
      \cellcolor[RGB]{113, 187, 34}{13-7}
      \cellcolor[RGB]{102, 181, 36}{13-8}
      \cellcolor[RGB]{96, 179, 38}{13-10,14-12}
      \cellcolor[RGB]{164, 196, 22}{13-12}
      \cellcolor[RGB]{186, 199, 16}{13-13}
      \cellcolor[RGB]{82, 172, 41}{14-2,14-7}
      \cellcolor[RGB]{77, 169, 43}{14-4,14-5}
      \cellcolor[RGB]{17, 140, 57}{14-9,14-10}
  \Body
      \hline
      ~ & \multicolumn{3}{c}{Complete} & \multicolumn{3}{c}{n-Back} & \multicolumn{3}{c}{Driving} & \multicolumn{3}{c}{Gaming} \\
      \hline
      ~ & \scriptsize{xLSTM} & \scriptsize{\makecell{Conv-\\Next}} & \scriptsize{\makecell{Trans-\\former}} & \scriptsize{xLSTM} & \scriptsize{\makecell{Conv-\\Next}} & \scriptsize{\makecell{Trans-\\former}} & \scriptsize{xLSTM} & \scriptsize{\makecell{Conv-\\Next}} & \scriptsize{\makecell{Trans-\\former}} & \scriptsize{xLSTM} & \scriptsize{\makecell{Conv-\\Next}} & \scriptsize{\makecell{Trans-\\former}} \\
      \cline{2-13}
      ~ \scriptsize{Heart Comb.} & \scriptsize{\makecell{0.67\\$\pm$\,0.026}} & \scriptsize{\makecell{0.55\\$\pm$\,0.051}} & \scriptsize{\makecell{0.66\\$\pm$\,0.066}} & \scriptsize{\makecell{0.55\\$\pm$\,0.046}} & \scriptsize{\makecell{0.47\\$\pm$\,0.070}} & \scriptsize{\makecell{0.51\\$\pm$\,0.018}} & \scriptsize{\makecell{0.64\\$\pm$\,0.044}} & \scriptsize{\makecell{0.56\\$\pm$\,0.053}} & \scriptsize{\makecell{0.64\\$\pm$\,0.056}} & \scriptsize{\makecell{0.63\\$\pm$\,0.046}} & \scriptsize{\makecell{0.61\\$\pm$\,0.057}} & \scriptsize{\makecell{0.62\\$\pm$\,0.059}} \\ 
      ~ \scriptsize{Biosignals} & \scriptsize{\makecell{0.71\\$\pm$\,0.015}} & \scriptsize{\makecell{0.62\\$\pm$\,0.034}} & \scriptsize{\makecell{0.68\\$\pm$\,0.033}} & \scriptsize{\makecell{0.63\\$\pm$\,0.004}} & \scriptsize{\makecell{0.57\\$\pm$\,0.067}} & \scriptsize{\makecell{0.60\\$\pm$\,0.042}} & \scriptsize{\makecell{0.73\\$\pm$\,0.028}} & \scriptsize{\makecell{0.69\\$\pm$\,0.007}} & \scriptsize{\makecell{0.72\\$\pm$\,0.033}} & \scriptsize{\makecell{0.72\\$\pm$\,0.068}} & \scriptsize{\makecell{0.69\\$\pm$\,0.059}} & \scriptsize{\makecell{0.72\\$\pm$\,0.073}} \\ 
      ~ \scriptsize{Eye Comb.} & \scriptsize{\makecell{0.84\\$\pm$\,0.021}} & \scriptsize{\makecell{0.83\\$\pm$\,0.031}} & \scriptsize{\makecell{0.85\\$\pm$\,0.028}} & \scriptsize{\makecell{0.85\\$\pm$\,0.023}} & \scriptsize{\makecell{0.85\\$\pm$\,0.026}} & \scriptsize{\makecell{0.86\\$\pm$\,0.022}} & \scriptsize{\makecell{0.83\\$\pm$\,0.053}} & \scriptsize{\makecell{0.84\\$\pm$\,0.036}} & \scriptsize{\makecell{0.87\\$\pm$\,0.031}} & \scriptsize{\makecell{0.86\\$\pm$\,0.054}} & \scriptsize{\makecell{0.85\\$\pm$\,0.052}} & \scriptsize{\makecell{0.86\\$\pm$\,0.053}} \\ 
      ~ \scriptsize{Head Rotation} & \scriptsize{\makecell{0.73\\$\pm$\,0.047}} & \scriptsize{\makecell{0.67\\$\pm$\,0.015}} & \scriptsize{\makecell{0.73\\$\pm$\,0.033}} & \scriptsize{\makecell{0.73\\$\pm$\,0.028}} & \scriptsize{\makecell{0.65\\$\pm$\,0.024}} & \scriptsize{\makecell{0.72\\$\pm$\,0.044}} & \scriptsize{\makecell{0.83\\$\pm$\,0.055}} & \scriptsize{\makecell{0.78\\$\pm$\,0.033}} & \scriptsize{\makecell{0.84\\$\pm$\,0.053}} & \scriptsize{\makecell{0.59\\$\pm$\,0.026}} & \scriptsize{\makecell{0.58\\$\pm$\,0.037}} & \scriptsize{\makecell{0.56\\$\pm$\,0.058}} \\ 
      ~ \scriptsize{Head Position} & \scriptsize{\makecell{0.59\\$\pm$\,0.060}} & \scriptsize{\makecell{0.58\\$\pm$\,0.028}} & \scriptsize{\makecell{0.57\\$\pm$\,0.048}} & \scriptsize{\makecell{0.55\\$\pm$\,0.044}} & \scriptsize{\makecell{0.54\\$\pm$\,0.041}} & \scriptsize{\makecell{0.52\\$\pm$\,0.025}} & \scriptsize{\makecell{0.57\\$\pm$\,0.030}} & \scriptsize{\makecell{0.59\\$\pm$\,0.031}} & \scriptsize{\makecell{0.58\\$\pm$\,0.041}} & \scriptsize{\makecell{0.56\\$\pm$\,0.048}} & \scriptsize{\makecell{0.54\\$\pm$\,0.067}} & \scriptsize{\makecell{0.60\\$\pm$\,0.073}} \\ 
      ~ \scriptsize{Head Comb.} & \scriptsize{\makecell{0.66\\$\pm$\,0.036}} & \scriptsize{\makecell{0.64\\$\pm$\,0.035}} & \scriptsize{\makecell{0.67\\$\pm$\,0.040}} & \scriptsize{\makecell{0.65\\$\pm$\,0.041}} & \scriptsize{\makecell{0.62\\$\pm$\,0.060}} & \scriptsize{\makecell{0.63\\$\pm$\,0.048}} & \scriptsize{\makecell{0.74\\$\pm$\,0.065}} & \scriptsize{\makecell{0.74\\$\pm$\,0.045}} & \scriptsize{\makecell{0.73\\$\pm$\,0.050}} & \scriptsize{\makecell{0.59\\$\pm$\,0.049}} & \scriptsize{\makecell{0.56\\$\pm$\,0.057}} & \scriptsize{\makecell{0.53\\$\pm$\,0.053}} \\ 
      ~ \scriptsize{Body Position} & \scriptsize{\makecell{0.56\\$\pm$\,0.035}} & \scriptsize{\makecell{0.56\\$\pm$\,0.021}} & \scriptsize{\makecell{0.54\\$\pm$\,0.019}} & \scriptsize{\makecell{0.54\\$\pm$\,0.051}} & \scriptsize{\makecell{0.59\\$\pm$\,0.048}} & \scriptsize{\makecell{0.52\\$\pm$\,0.045}} & \scriptsize{\makecell{0.54\\$\pm$\,0.034}} & \scriptsize{\makecell{0.58\\$\pm$\,0.018}} & \scriptsize{\makecell{0.53\\$\pm$\,0.040}} & \scriptsize{\makecell{0.55\\$\pm$\,0.080}} & \scriptsize{\makecell{0.55\\$\pm$\,0.078}} & \scriptsize{\makecell{0.51\\$\pm$\,0.075}} \\ 
      ~ \scriptsize{Move. Comb.} & \scriptsize{\makecell{0.65\\$\pm$\,0.045}} & \scriptsize{\makecell{0.65\\$\pm$\,0.033}} & \scriptsize{\makecell{0.65\\$\pm$\,0.042}} & \scriptsize{\makecell{0.63\\$\pm$\,0.060}} & \scriptsize{\makecell{0.65\\$\pm$\,0.047}} & \scriptsize{\makecell{0.64\\$\pm$\,0.033}} & \scriptsize{\makecell{0.69\\$\pm$\,0.022}} & \scriptsize{\makecell{0.73\\$\pm$\,0.035}} & \scriptsize{\makecell{0.71\\$\pm$\,0.027}} & \scriptsize{\makecell{0.57\\$\pm$\,0.037}} & \scriptsize{\makecell{0.56\\$\pm$\,0.046}} & \scriptsize{\makecell{0.53\\$\pm$\,0.027}} \\ 
      ~ \scriptsize{Biosignals, Eye} & \scriptsize{\makecell{0.82\\$\pm$\,0.019}} & \scriptsize{\makecell{0.83\\$\pm$\,0.045}} & \scriptsize{\makecell{0.83\\$\pm$\,0.021}} & \scriptsize{\makecell{0.84\\$\pm$\,0.028}} & \scriptsize{\makecell{0.83\\$\pm$\,0.033}} & \scriptsize{\makecell{0.84\\$\pm$\,0.018}} & \scriptsize{\makecell{0.87\\$\pm$\,0.028}} & \scriptsize{\makecell{0.85\\$\pm$\,0.018}} & \scriptsize{\makecell{0.88\\$\pm$\,0.028}} & \scriptsize{\makecell{0.82\\$\pm$\,0.043}} & \scriptsize{\makecell{0.81\\$\pm$\,0.051}} & \scriptsize{\makecell{0.82\\$\pm$\,0.055}} \\ 
      ~ \scriptsize{Bio., AUs, ET} & \scriptsize{\makecell{0.86\\$\pm$\,0.026}} & \scriptsize{\makecell{0.87\\$\pm$\,0.022}} & \scriptsize{\makecell{0.84\\$\pm$\,0.048}} & \scriptsize{\makecell{0.82\\$\pm$\,0.037}} & \scriptsize{\makecell{0.83\\$\pm$\,0.043}} & \scriptsize{\makecell{0.83\\$\pm$\,0.021}} & \scriptsize{\makecell{0.87\\$\pm$\,0.031}} & \scriptsize{\makecell{0.90\\$\pm$\,0.028}} & \scriptsize{\makecell{0.87\\$\pm$\,0.012}} & \scriptsize{\makecell{0.84\\$\pm$\,0.044}} & \scriptsize{\makecell{0.85\\$\pm$\,0.048}} & \scriptsize{\makecell{0.84\\$\pm$\,0.052}} \\ 
      ~ \scriptsize{Move., AUs, ET} & \scriptsize{\makecell{0.73\\$\pm$\,0.022}} & \scriptsize{\makecell{0.72\\$\pm$\,0.019}} & \scriptsize{\makecell{0.73\\$\pm$\,0.025}} & \scriptsize{\makecell{0.72\\$\pm$\,0.048}} & \scriptsize{\makecell{0.79\\$\pm$\,0.036}} & \scriptsize{\makecell{0.76\\$\pm$\,0.042}} & \scriptsize{\makecell{0.78\\$\pm$\,0.032}} & \scriptsize{\makecell{0.82\\$\pm$\,0.046}} & \scriptsize{\makecell{0.78\\$\pm$\,0.047}} & \scriptsize{\makecell{0.66\\$\pm$\,0.052}} & \scriptsize{\makecell{0.70\\$\pm$\,0.044}} & \scriptsize{\makecell{0.67\\$\pm$\,0.061}} \\ 
      ~ \scriptsize{Bio, AUs, ET, Move.} & \scriptsize{\makecell{0.80\\$\pm$\,0.023}} & \scriptsize{\makecell{0.83\\$\pm$\,0.038}} & \scriptsize{\makecell{0.81\\$\pm$\,0.034}} & \scriptsize{\makecell{0.81\\$\pm$\,0.031}} & \scriptsize{\makecell{0.81\\$\pm$\,0.040}} & \scriptsize{\makecell{0.80\\$\pm$\,0.030}} & \scriptsize{\makecell{0.85\\$\pm$\,0.047}} & \scriptsize{\makecell{0.89\\$\pm$\,0.026}} & \scriptsize{\makecell{0.89\\$\pm$\,0.016}} & \scriptsize{\makecell{0.72\\$\pm$\,0.059}} & \scriptsize{\makecell{0.78\\$\pm$\,0.074}} & \scriptsize{\makecell{0.72\\$\pm$\,0.079}} \\ 
\hline
  \end{NiceTabular}
\end{table*}

\begin{table*}[!t]
  \centering
  \setlength{\tabcolsep}{4pt}
  \tablefont
  \caption{Task-specific modality contrasts in binary task-load classification (test \gls{auroc}, mean over ConvNeXt, x\gls{lstm} and transformer). A positive $\Delta$ favours the first modality. Brackets give participant-clustered 95\,\% confidence intervals.}
  \label{tab:modality-contrasts}
  \begin{tabularx}{\linewidth}{Xcccc c}
    \toprule
    Domain & \gls{auroc}$_1$ & \gls{auroc}$_2$ & $\Delta$ & 95\,\% CI & $p_{\mathrm{Holm}}$ \\
    \midrule
    \multicolumn{6}{@{}l}{\emph{Pupil diameter vs.\ ECG}} \\
    Complete & 0.717 & 0.708 & ${+}0.010$ & $[{-}0.03,\,{+}0.04]$ & $.574$ \\
    $n$-back (Combined) & 0.825 & 0.597 & ${+}0.228$ & $[{+}0.20,\,{+}0.26]$ & ${<}.001$ \\
    $n$-back (Driving) & 0.824 & 0.580 & ${+}0.245$ & $[{+}0.19,\,{+}0.29]$ & ${<}.001$ \\
    $n$-back (Gaming) & 0.825 & 0.526 & ${+}0.299$ & $[{+}0.25,\,{+}0.35]$ & ${<}.001$ \\
    Driving Application & 0.674 & 0.739 & ${-}0.066$ & $[{-}0.12,\,{-}0.02]$ & $.023$ \\
    Gaming & 0.793 & 0.689 & ${+}0.104$ & $[{+}0.04,\,{+}0.15]$ & $.005$ \\
    \emph{Hogwarts L.} & 0.853 & 0.694 & ${+}0.159$ & $[{+}0.09,\,{+}0.21]$ & ${<}.001$ \\
    \emph{Overcooked! 2} & 0.763 & 0.578 & ${+}0.185$ & $[{+}0.13,\,{+}0.23]$ & ${<}.001$ \\
    \addlinespace
    \multicolumn{6}{@{}l}{\emph{Facial action units vs.\ pupil diameter}} \\
    Complete & 0.748 & 0.717 & ${+}0.030$ & $[{-}0.03,\,{+}0.08]$ & $.591$ \\
    $n$-back (Combined) & 0.757 & 0.825 & ${-}0.067$ & $[{-}0.12,\,{-}0.02]$ & $.015$ \\
    $n$-back (Driving) & 0.852 & 0.824 & ${+}0.028$ & $[{-}0.06,\,{+}0.09]$ & $.591$ \\
    $n$-back (Gaming) & 0.506 & 0.825 & ${-}0.319$ & $[{-}0.40,\,{-}0.25]$ & ${<}.001$ \\
    Driving Application & 0.879 & 0.674 & ${+}0.205$ & $[{+}0.11,\,{+}0.27]$ & $.001$ \\
    Gaming & 0.544 & 0.793 & ${-}0.249$ & $[{-}0.29,\,{-}0.20]$ & ${<}.001$ \\
    \emph{Hogwarts L.} & 0.483 & 0.853 & ${-}0.369$ & $[{-}0.42,\,{-}0.29]$ & ${<}.001$ \\
    \emph{Overcooked! 2} & 0.511 & 0.763 & ${-}0.251$ & $[{-}0.30,\,{-}0.20]$ & ${<}.001$ \\
    \addlinespace
    \multicolumn{6}{@{}l}{\emph{Pupil diameter vs.\ eye movement}} \\
    Complete & 0.717 & 0.758 & ${-}0.040$ & $[{-}0.08,\,{-}0.00]$ & $.041$ \\
    $n$-back (Combined) & 0.825 & 0.727 & ${+}0.098$ & $[{+}0.07,\,{+}0.13]$ & ${<}.001$ \\
    $n$-back (Driving) & 0.824 & 0.520 & ${+}0.305$ & $[{+}0.25,\,{+}0.35]$ & ${<}.001$ \\
    $n$-back (Gaming) & 0.825 & 0.596 & ${+}0.229$ & $[{+}0.18,\,{+}0.29]$ & ${<}.001$ \\
    Driving Application & 0.674 & 0.836 & ${-}0.162$ & $[{-}0.21,\,{-}0.11]$ & ${<}.001$ \\
    Gaming & 0.793 & 0.663 & ${+}0.130$ & $[{+}0.08,\,{+}0.17]$ & ${<}.001$ \\
    \emph{Hogwarts L.} & 0.853 & 0.699 & ${+}0.154$ & $[{+}0.09,\,{+}0.20]$ & ${<}.001$ \\
    \emph{Overcooked! 2} & 0.763 & 0.688 & ${+}0.075$ & $[{+}0.04,\,{+}0.11]$ & $.004$ \\
    \bottomrule
  \end{tabularx}
  \begin{minipage}{\linewidth}
    \vspace{2pt}\footnotesize\textit{Note.} \gls{auroc}$_1$ and \gls{auroc}$_2$ are the first and second modality of the contrast, per subject-wise fold, averaged over the five folds and the three encoders. Intervals from a participant-level cluster bootstrap, $p$ from a two-sided participant-level sign-flip test (10{,}000 draws each), Holm-corrected across the eight domains within each contrast.
  \end{minipage}
\end{table*}

\begin{table*}[!t]
  \centering
  \setlength{\tabcolsep}{4pt}
  \tablefont
  \caption{Advantage of multimodal fusion (biosignals $+$ eye tracking) over each single modality. Cells report the difference in test \gls{auroc} (fusion minus modality), averaged over the three encoders. A positive value indicates that fusion outperforms the modality.}
  \label{tab:modality-multimodal-vs-unimodal}
  \begin{tabularx}{\linewidth}{X r@{}l r@{}l r@{}l r@{}l r@{}l r@{}l r@{}l r@{}l}
    \toprule
    Domain & \multicolumn{2}{c}{ECG} & \multicolumn{2}{c}{PPG} & \multicolumn{2}{c}{EDA} & \multicolumn{2}{c}{EMG} & \multicolumn{2}{c}{RSP} & \multicolumn{2}{c}{SKT} & \multicolumn{2}{c}{Eye Mv.} & \multicolumn{2}{c}{Pupil} \\
    \midrule
    Complete & ${+}0$&$.119$\textsuperscript{*} & ${+}0$&$.289$\textsuperscript{*} & ${+}0$&$.159$\textsuperscript{*} & ${+}0$&$.165$\textsuperscript{*} & ${+}0$&$.176$\textsuperscript{*} & ${+}0$&$.281$\textsuperscript{*} & ${+}0$&$.069$\textsuperscript{*} & ${+}0$&$.109$\textsuperscript{*} \\
    $n$-back (Combined) & ${+}0$&$.238$\textsuperscript{*} & ${+}0$&$.374$\textsuperscript{*} & ${+}0$&$.190$\textsuperscript{*} & ${+}0$&$.244$\textsuperscript{*} & ${+}0$&$.267$\textsuperscript{*} & ${+}0$&$.303$\textsuperscript{*} & ${+}0$&$.108$\textsuperscript{*} & ${+}0$&$.010$ \\
    $n$-back (Driving) & ${+}0$&$.233$\textsuperscript{*} & ${+}0$&$.280$\textsuperscript{*} & ${+}0$&$.193$\textsuperscript{*} & ${+}0$&$.217$\textsuperscript{*} & ${+}0$&$.306$\textsuperscript{*} & ${+}0$&$.192$\textsuperscript{*} & ${+}0$&$.292$\textsuperscript{*} & ${-}0$&$.012$ \\
    $n$-back (Gaming) & ${+}0$&$.285$\textsuperscript{*} & ${+}0$&$.337$\textsuperscript{*} & ${+}0$&$.259$\textsuperscript{*} & ${+}0$&$.288$\textsuperscript{*} & ${+}0$&$.317$\textsuperscript{*} & ${+}0$&$.293$\textsuperscript{*} & ${+}0$&$.216$\textsuperscript{*} & ${-}0$&$.014$ \\
    Driving Application & ${+}0$&$.126$\textsuperscript{*} & ${+}0$&$.317$\textsuperscript{*} & ${+}0$&$.222$\textsuperscript{*} & ${+}0$&$.134$\textsuperscript{*} & ${+}0$&$.223$\textsuperscript{*} & ${+}0$&$.152$\textsuperscript{*} & ${+}0$&$.030$ & ${+}0$&$.192$\textsuperscript{*} \\
    Gaming & ${+}0$&$.128$\textsuperscript{*} & ${+}0$&$.157$\textsuperscript{*} & ${+}0$&$.102$\textsuperscript{*} & ${+}0$&$.290$\textsuperscript{*} & ${+}0$&$.234$\textsuperscript{*} & ${+}0$&$.232$\textsuperscript{*} & ${+}0$&$.154$\textsuperscript{*} & ${+}0$&$.025$ \\
    \emph{Hogwarts L.} & ${+}0$&$.178$\textsuperscript{*} & ${+}0$&$.247$\textsuperscript{*} & ${+}0$&$.111$\textsuperscript{*} & ${+}0$&$.275$\textsuperscript{*} & ${+}0$&$.362$\textsuperscript{*} & ${+}0$&$.365$\textsuperscript{*} & ${+}0$&$.173$\textsuperscript{*} & ${+}0$&$.019$ \\
    \emph{Overcooked! 2} & ${+}0$&$.171$\textsuperscript{*} & ${+}0$&$.140$\textsuperscript{*} & ${+}0$&$.097$\textsuperscript{*} & ${+}0$&$.236$\textsuperscript{*} & ${+}0$&$.202$\textsuperscript{*} & ${+}0$&$.176$\textsuperscript{*} & ${+}0$&$.062$ & ${-}0$&$.013$ \\
    \bottomrule
  \end{tabularx}
  \begin{minipage}{\linewidth}
    \vspace{2pt}\footnotesize\textit{Note.} \gls{auroc} per subject-wise fold, averaged over the five folds. Two-sided participant-level sign-flip test (10{,}000 draws), Holm-corrected across the 64 cells, \textsuperscript{*}$p_{\mathrm{Holm}} < .05$. Eye Mv.\ $=$ eye movement, Pupil $=$ pupil diameter.
  \end{minipage}
\end{table*}

\subsection{Task Load Generalization}
\label{ssec:task-load-generalization}
\begin{figure*}[!htb]
  \centering
  \includegraphics[width=0.95\linewidth]{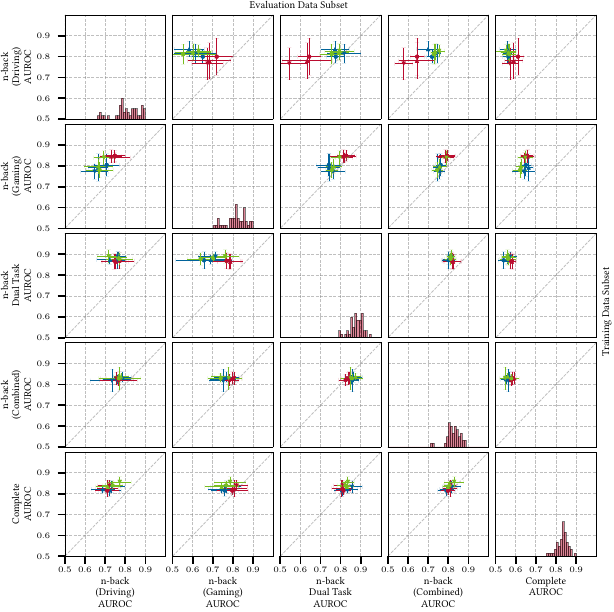}
  \caption{Cross-domain test \gls{auroc} across five $n$-back variants and the complete dataset. Each panel reports
the mean and standard deviation of test \gls{auroc} (x-axis) against in-domain validation \gls{auroc} (y-axis)
over five subject-wise folds. Rows denote the training subset, columns the evaluation subset. Diagonal cells show
the in-domain score distribution as a histogram. Colors and markers follow the network legend in
Fig.~\ref{fig:supervised-baselines}.}
  \label{fig:supervised-distribution-shifts}
\end{figure*}
\begin{figure*}[!htb]
  \centering
  \includegraphics[width=0.95\linewidth]{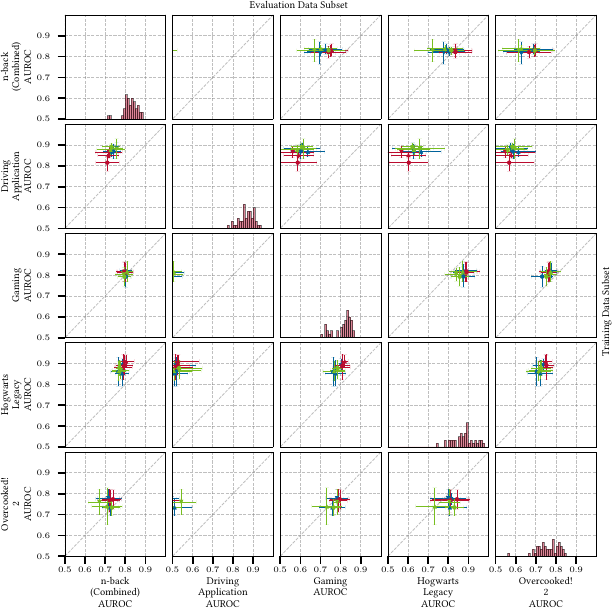}
  \caption{Transfer from the laboratory $n$-back paradigm to the three application targets (Driving, Gaming,
\emph{Hogwarts Legacy}, \emph{Overcooked! 2}). Panels report test \gls{auroc} vs.\ validation \gls{auroc} (mean
$\pm$ SD over five folds). Diagonal cells show the in-domain histogram. Colors and markers follow the network
legend in Fig.~\ref{fig:supervised-baselines}.}
  \label{fig:supervised-distribution-shifts-transfer}
\end{figure*}
\begin{figure*}[!htb]
  \centering
  \includegraphics[width=0.95\linewidth]{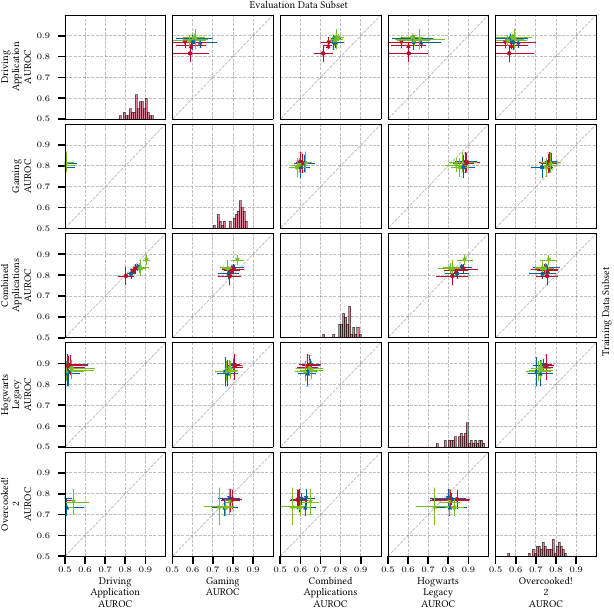}
  \caption{Cross-application transfer between Driving, Gaming, their union, and the individual gaming titles
\emph{Hogwarts Legacy} and \emph{Overcooked! 2}. Panels report test \gls{auroc} vs.\ validation \gls{auroc}
(mean $\pm$ SD over five folds). Diagonal cells show the in-domain histogram. Colors and markers follow the
network legend in Fig.~\ref{fig:supervised-baselines}.}
  \label{fig:supervised-distribution-shifts-applications}
\end{figure*}
We examined cross-domain generalization with a train-by-evaluate matrix over five training configurations, namely
complete (all domains), applications (driving and gaming), driving-only, gaming-only, and $n$-back-only. In this
matrix the diagonal ($X\,\to X$) is the in-domain specialist, the single-domain off-diagonal entries ($X\,\to Y$
with $X \neq Y$) measure narrow transfer, and the complete and application rows are the diverse-training
conditions. Figures~\ref{fig:supervised-distribution-shifts}, \ref{fig:supervised-distribution-shifts-transfer},
and \ref{fig:supervised-distribution-shifts-applications} report \gls{auroc} for the $n$-back variants, the
laboratory-to-application transfer, and the cross-application setting, respectively.
Table~\ref{tab:distribution-shift-ece} reports \gls{ece} under the same shifts. All results are reported on
validation/test sets with subject that are unseen during training. The complete configuration, and the
applications configuration on driving and gaming, include other participants of the evaluated domain in training.
These entries measure transfer to unseen participants, not to an unseen domain, which Section~\ref{ssec:lodo} tests
separately.

Training on the complete dataset produced the most stable behaviour. \gls{auroc} remained competitive across $n$-back,
driving, gaming, and their combinations, and calibration stayed good throughout, with \gls{ece} between $0.06$ and
$0.11$ across all evaluation subsets (Table~\ref{tab:distribution-shift-ece}). Models trained only on $n$-back did not
transfer to driving or to the combined applications, where \gls{auroc} dropped and \gls{ece} rose to $0.30$ or above
($0.48$ to $0.54$ on driving). Transfer from $n$-back to gaming was far milder, with a smaller \gls{auroc} loss and
\gls{ece} of $0.09$ to $0.14$, which suggests closer similarity between the $n$-back paradigm and gaming in our
setting, partly because the $n$-back training data include the gaming cohort (Section~\ref{ssec:cohort-shift}).
Cross-application transfer between driving and gaming was weak in both directions and was accompanied by elevated
\gls{ece}, consistent with application-specific cues that do not carry across domains. Training jointly on the two
application tasks improved robustness within applications relative to single-application training and transferred to
$n$-back with only moderate degradation. Training-set size alone is an unlikely explanation, since more
training data at constant source diversity had no consistent effect (Section~\ref{ssec:lodo}). Across all of these
settings, the \gls{xlstm}, ConvNeXt, and transformer encoders showed similar calibration without systematic \gls{ece}
differences (Table~\ref{tab:distribution-shift-ece}).

\gls{auroc} and \gls{ece} provide complementary evidence. \gls{auroc} is threshold-agnostic and therefore enables fair
comparison across domains, whereas \gls{ece} measures how well predicted confidence matches observed accuracy. Under
distribution shift the two moved together. \gls{ece} increased wherever \gls{auroc} dropped, most clearly for transfer
to and from driving, and remained low when the training data were diverse. Calibration is thus a practical indicator of 
out-of-domain risk alongside discriminative accuracy. For ConvNeXt-small evaluated on $n$-back and gaming, the
Brier score and the \gls{nll} confirm that diverse training is better calibrated than transfer from
another domain. A model trained on the evaluated domain itself, however, reached a slightly lower Brier score than the
complete model (Supplementary Section~\ref{ssec:supp-calibration}).

To quantify the benefit of diverse training, we used \emph{small} ConvNeXt as the representative architecture and
compared, for each domain $Y$ or the complete-dataset model (complete\,$\to Y$) against the mean of the
single-other-domain models ($X\,\to Y$) on the same test windows. We applied the participant-level bootstrap
and sign-flip test per domain (two-sided) and Holm-corrected across the seven domains
(Table~\ref{tab:generalization-complete-vs-single}). Complete-dataset training generalized significantly better than
single-domain transfer in six of the seven evaluation domains. The gains were largest for the application targets,
including driving ($\Delta\mathrm{AUROC} = +0.361$, 95\,\% CI $[+0.29, +0.42]$,
$p_{\mathrm{Holm}} < .001$) and gaming ($\Delta\mathrm{AUROC} = +0.086$, $[+0.06, +0.11]$,
$p_{\mathrm{Holm}} < .001$). At the same time, complete-dataset training stayed close to the domain-specialist
($Y\,\to Y$), trailing it by $0.006$ to $0.033$ \gls{auroc} on the applications and by $0.023$ to $0.067$ on
the $n$-back variants, which reflects transfer to unseen participants, because the complete dataset contains other
participants of the target domain. With the target domain removed from training, the \gls{auroc} fell by $0.05$--$0.48$
(Section~\ref{ssec:lodo}). One non-significant case was the narrow $n$-back (Driving Study) cohort, where a single
related source already transferred well ($\Delta\mathrm{AUROC} = +0.013$, $p_{\mathrm{Holm}} = .422$). The full
per-source breakdown, with Benjamini-Hochberg \gls{fdr} correction across the $42$ source-target pairs, is reported in
Supplementary Table~\ref{tab:supp-generalization-detail}. Complete-dataset training was significantly
better than 20 of the 42 single sources, mostly for laboratory-to-application and cross-application pairs ($+0.26$ to
$+0.47$ on the driving application). Two single sources were significantly better, both on the driving-study $n$-back,
namely the combined $n$-back ($-0.057$, $p_{\mathrm{BH}} = .005$) and the driving application ($-0.065$,
$p_{\mathrm{BH}} = .022$), which both contain data of the driving cohort. Overall, diverse training yields more robust
and better-calibrated models for domains represented in training, whereas single-task training does not generalize
across domains.
\begin{table}[!htb]
  \centering
  \setlength{\tabcolsep}{4pt}
  \tablefont
  \caption{Expected calibration error (\gls{ece}, mean $\pm$ SD over five subject-wise folds) under distribution shift. Rows give the training subset, columns the encoder, and each block the evaluation subset. Lower is better. In-domain cells (training subset $=$ evaluation subset) are in bold.}
  \label{tab:distribution-shift-ece}
  \begin{tabularx}{\linewidth}{cl*{3}{>{\centering\arraybackslash}X}}
    \toprule
    Trained on & Evaluated on & xLSTM & ConvNeXt & Transformer \\
    \midrule
    \multirow{5}{*}{\rotatebox[origin=c]{90}{Complete}} & Complete & \textbf{0.08 $\pm$ 0.05} & \textbf{0.06 $\pm$ 0.06} & \textbf{0.09 $\pm$ 0.03} \\
     & $n$-back & 0.09 $\pm$ 0.05 & 0.06 $\pm$ 0.05 & 0.11 $\pm$ 0.02 \\
     & Driving & 0.10 $\pm$ 0.05 & 0.06 $\pm$ 0.05 & 0.09 $\pm$ 0.04 \\
     & Gaming & 0.08 $\pm$ 0.05 & 0.08 $\pm$ 0.07 & 0.09 $\pm$ 0.04 \\
     & D.\,\&\,G. & 0.08 $\pm$ 0.04 & 0.06 $\pm$ 0.06 & 0.09 $\pm$ 0.03 \\
    \midrule
    \multirow{5}{*}{\rotatebox[origin=c]{90}{$n$-back}} & Complete & 0.23 $\pm$ 0.03 & 0.18 $\pm$ 0.02 & 0.24 $\pm$ 0.03 \\
     & $n$-back & \textbf{0.05 $\pm$ 0.03} & \textbf{0.04 $\pm$ 0.01} & \textbf{0.06 $\pm$ 0.02} \\
     & Driving & 0.54 $\pm$ 0.06 & 0.48 $\pm$ 0.06 & 0.54 $\pm$ 0.06 \\
     & Gaming & 0.13 $\pm$ 0.07 & 0.09 $\pm$ 0.06 & 0.14 $\pm$ 0.07 \\
     & D.\,\&\,G. & 0.36 $\pm$ 0.04 & 0.30 $\pm$ 0.03 & 0.36 $\pm$ 0.04 \\
    \midrule
    \multirow{5}{*}{\rotatebox[origin=c]{90}{Driving}} & Complete & 0.13 $\pm$ 0.05 & 0.09 $\pm$ 0.04 & 0.16 $\pm$ 0.04 \\
     & $n$-back & 0.10 $\pm$ 0.04 & 0.07 $\pm$ 0.03 & 0.12 $\pm$ 0.05 \\
     & Driving & \textbf{0.08 $\pm$ 0.06} & \textbf{0.04 $\pm$ 0.01} & \textbf{0.09 $\pm$ 0.04} \\
     & Gaming & 0.23 $\pm$ 0.09 & 0.25 $\pm$ 0.07 & 0.32 $\pm$ 0.05 \\
     & D.\,\&\,G. & 0.15 $\pm$ 0.07 & 0.12 $\pm$ 0.05 & 0.19 $\pm$ 0.03 \\
    \midrule
    \multirow{5}{*}{\rotatebox[origin=c]{90}{Gaming}} & Complete & 0.20 $\pm$ 0.02 & 0.19 $\pm$ 0.03 & 0.19 $\pm$ 0.03 \\
     & $n$-back & 0.14 $\pm$ 0.02 & 0.14 $\pm$ 0.02 & 0.14 $\pm$ 0.02 \\
     & Driving & 0.41 $\pm$ 0.09 & 0.38 $\pm$ 0.06 & 0.34 $\pm$ 0.08 \\
     & Gaming & \textbf{0.06 $\pm$ 0.04} & \textbf{0.07 $\pm$ 0.03} & \textbf{0.08 $\pm$ 0.05} \\
     & D.\,\&\,G. & 0.25 $\pm$ 0.04 & 0.23 $\pm$ 0.04 & 0.22 $\pm$ 0.05 \\
    \midrule
    \multirow{5}{*}{\rotatebox[origin=c]{90}{D.\,\&\,G.}} & Complete & 0.08 $\pm$ 0.01 & 0.07 $\pm$ 0.04 & 0.13 $\pm$ 0.05 \\
     & $n$-back & 0.13 $\pm$ 0.03 & 0.11 $\pm$ 0.04 & 0.17 $\pm$ 0.05 \\
     & Driving & 0.06 $\pm$ 0.03 & 0.06 $\pm$ 0.03 & 0.10 $\pm$ 0.04 \\
     & Gaming & 0.06 $\pm$ 0.02 & 0.07 $\pm$ 0.03 & 0.12 $\pm$ 0.06 \\
     & D.\,\&\,G. & \textbf{0.05 $\pm$ 0.01} & \textbf{0.06 $\pm$ 0.02} & \textbf{0.10 $\pm$ 0.05} \\
    \bottomrule
  \end{tabularx}
  \begin{minipage}{\linewidth}
    \vspace{2pt}\footnotesize\textit{Note.} D.\,\&\,G.\ $=$ driving and gaming combined. \gls{ece} rises sharply off the diagonal, most strongly for transfer to or from driving, and stays low for models trained on the complete dataset.
  \end{minipage}
\end{table}

\begin{table*}[!t]
  \centering
  \setlength{\tabcolsep}{4pt}
  \tablefont
  \caption{Generalization of complete-dataset training to domains (ConvNeXt-small, test \gls{auroc}). For each domain $Y$, complete-dataset training (complete\,$\to Y$) is compared with the mean of the single-other-domain models ($X\,\to Y$, $X\neq Y$) on the same test windows. The in-domain model ($Y\,\to Y$) is shown as the specialist reference. A positive $\Delta$ favours complete-dataset training. Brackets give participant-clustered 95\,\% confidence intervals.}
  \label{tab:generalization-complete-vs-single}
  \begin{tabularx}{\linewidth}{Xccccc}
    \toprule
    Domain & $Y\,{\to}\,Y$ & compl.\,${\to}\,Y$ & $X\,{\to}\,Y$ & $\Delta$ & $p_{\mathrm{Holm}}$ \\
    \midrule
    $n$-back (Driving Study) & 0.780 & 0.713 & 0.700 & ${+}0.013$ $[{-}0.02,\,{+}0.05]$ & $.422$ \\
    $n$-back (Gaming Study) & 0.845 & 0.807 & 0.752 & ${+}0.055$ $[{+}0.04,\,{+}0.08]$ & $.002$ \\
    $n$-back (Combined) & 0.831 & 0.808 & 0.746 & ${+}0.063$ $[{+}0.04,\,{+}0.08]$ & ${<}.001$ \\
    Driving Application & 0.851 & 0.845 & 0.484 & ${+}0.361$ $[{+}0.29,\,{+}0.42]$ & ${<}.001$ \\
    \emph{Hogwarts Legacy} & 0.896 & 0.863 & 0.765 & ${+}0.097$ $[{+}0.07,\,{+}0.12]$ & ${<}.001$ \\
    \emph{Overcooked! 2} & 0.774 & 0.759 & 0.687 & ${+}0.072$ $[{+}0.04,\,{+}0.09]$ & ${<}.001$ \\
    Gaming & 0.813 & 0.805 & 0.718 & ${+}0.086$ $[{+}0.06,\,{+}0.11]$ & ${<}.001$ \\
    \bottomrule
  \end{tabularx}
  \begin{minipage}{\linewidth}
    \vspace{2pt}\footnotesize\textit{Note.} \gls{auroc} per subject-wise fold, averaged over the five folds. Intervals from a participant-level cluster bootstrap, $p$ from a two-sided participant-level sign-flip test (10{,}000 draws each), Holm-corrected across the seven domains. $X\,{\to}\,Y$ is the mean over the single-other-domain models.
  \end{minipage}
\end{table*}

\subsection{Leave-One-Domain-Out Generalization}
\label{ssec:lodo}
The complete configuration contains other participants of every evaluated domain, so complete\,$\to Y$
measures transfer to unseen participants, not to an unseen domain. To test the latter, we removed the target domain
from training and model selection and trained on all remaining domains (leave-one-domain-out, LODO). The four targets
were the single-task $n$-back, with the entire $n$-back study held out, the driving application, \emph{Overcooked! 2},
and \emph{Hogwarts Legacy}. The in-domain reference was trained on the target domain itself. To separate source
diversity from data quantity, we subsampled the multi-source training and validation sets to the windows per class of
each single-source model (size-matched). The diversity contrast compares the size-matched multi-source model with the
single source. The quantity contrast compares the full with the size-matched multi-source model. All models used
biosignals and eye tracking with ConvNeXt-base and ConvNeXt-small on the participant-wise folds of the main analysis.
Intervals (95\,\%) and $p$-values come from a participant-level cluster bootstrap and sign-flip test (10{,}000 draws
each).
\begin{table*}[!t]
  \centering
  \setlength{\tabcolsep}{4pt}
  \tablefont
  \caption{Leave-one-domain-out transfer. \gls{auroc} of ConvNeXt-base trained on the target domain (in-domain) and on all other domains with the target removed from training and model selection (LODO), and the paired difference LODO minus in-domain for ConvNeXt-base and ConvNeXt-small. Brackets give participant-clustered 95\,\% confidence intervals.}
  \label{tab:lodo}
  \begin{tabularx}{\linewidth}{Xcccccc}
    \toprule
    Target & In-domain & LODO & $\Delta$ Base & $p$ & $\Delta$ Small & $p$ \\
    \midrule
    $n$-back & 0.766 & 0.713 & ${-}0.053$ $[{-}0.09,\,{-}0.01]$ & $.015$ & ${-}0.049$ $[{-}0.09,\,{-}0.01]$ & $.017$ \\
    Driving Application & 0.879 & 0.411 & ${-}0.468$ $[{-}0.53,\,{-}0.39]$ & ${<}.001$ & ${-}0.482$ $[{-}0.55,\,{-}0.41]$ & ${<}.001$ \\
    \emph{Overcooked! 2} & 0.784 & 0.719 & ${-}0.065$ $[{-}0.11,\,{-}0.02]$ & $.014$ & ${-}0.059$ $[{-}0.11,\,{-}0.02]$ & $.019$ \\
    \emph{Hogwarts Legacy} & 0.878 & 0.786 & ${-}0.092$ $[{-}0.14,\,{-}0.05]$ & ${<}.001$ & ${-}0.070$ $[{-}0.11,\,{-}0.03]$ & ${<}.001$ \\
    \bottomrule
  \end{tabularx}
  \begin{minipage}{\linewidth}
    \vspace{2pt}\footnotesize\textit{Note.} Out-of-fold predictions of five participant-wise folds, pooled per target. Intervals from a participant-level cluster bootstrap, $p$ from a two-sided participant-level sign-flip test (10{,}000 draws each). $n$-back is the single-task $n$-back of both studies, with the whole $n$-back study held out.
  \end{minipage}
\end{table*}

Removing the target domain lowered the \gls{auroc} on all four targets with both architectures
(Table~\ref{tab:lodo}), by $0.05$--$0.09$ on the $n$-back and the two games and by $0.47$--$0.48$ on the driving
application. On driving, the \gls{auroc} fell below chance ($0.41$) and the \gls{ece} rose from $0.07$ to $0.45$
(Section~\ref{ssec:cohort-shift}). On the other three targets, the \gls{ece} rose from at most $0.07$ to
$0.08$--$0.20$.
More source domains did not improve generalization in general (Supplementary
Table~\ref{tab:lodo-sources}). At equal training size, they raised the \gls{auroc} by $0.25$--$0.51$ when the
driving application, which transfers poorly, was the single source (six of six comparisons significant after Holm
correction). They lowered it by $0.10$--$0.17$ on the driving target (four of six). In the remaining comparisons the
effect was small (mean $+0.02$ and $+0.03$, two of 12 significant). More training data at constant diversity had no
consistent effect (mean $+0.004$ with both architectures). The multi-source model outperformed the best single source
significantly in only one of eight comparisons and was significantly worse on the driving target. Diverse training thus
protects against a poorly transferring source but does not by itself ensure transfer to an unseen domain. These results
concern transfer to unseen domains, whereas transfer to unseen participants and cohorts is examined in
Sections~\ref{ssec:task-load-generalization} and~\ref{ssec:cohort-shift}.

\subsection{Cohort and Participant Shift}
\label{ssec:cohort-shift}
Transfer between the driving and gaming applications also changes the cohort, and with it the participants,
recording period, and setup. To separate these factors, we trained on one cohort and tested on the other, with the
target cohort excluded from training and model selection (leave-one-cohort-out). Checkpoints were therefore selected
on validation data of the source cohort, whereas the in-cohort reference used validation data of the target cohort.
The $n$-back, shared by both cohorts,
serves as a negative control because only the cohort changes. Intervals (95\,\%) and $p$-values come from a participant-level cluster
bootstrap and sign-flip test (10{,}000 draws each).
\begin{table*}[!t]
  \centering
  \setlength{\tabcolsep}{4pt}
  \tablefont
  \caption{Leave-one-cohort-out transfer. \gls{auroc} of ConvNeXt-base trained on the target cohort (in-cohort) and on the other cohort with the target cohort removed from training and model selection (LOCO), and the paired difference LOCO minus in-cohort for ConvNeXt-base and ConvNeXt-small. Brackets give participant-clustered 95\,\% confidence intervals.}
  \label{tab:cohort-shift}
  \begin{tabularx}{\linewidth}{Xcccccc}
    \toprule
    Target & In-cohort & LOCO & $\Delta$ Base & $p$ & $\Delta$ Small & $p$ \\
    \midrule
    \multicolumn{7}{@{}l}{\emph{Driving study}} \\
    $n$-back & 0.787 & 0.689 & ${-}0.098$ $[{-}0.14,\,{-}0.05]$ & ${<}.001$ & ${-}0.094$ $[{-}0.14,\,{-}0.05]$ & ${<}.001$ \\
    Driving Application & 0.888 & 0.429 & ${-}0.459$ $[{-}0.52,\,{-}0.39]$ & ${<}.001$ & ${-}0.442$ $[{-}0.51,\,{-}0.37]$ & ${<}.001$ \\
    Pooled & 0.859 & 0.490 & ${-}0.369$ $[{-}0.41,\,{-}0.32]$ & ${<}.001$ & ${-}0.364$ $[{-}0.40,\,{-}0.32]$ & ${<}.001$ \\
    \addlinespace
    \multicolumn{7}{@{}l}{\emph{Gaming study}} \\
    $n$-back & 0.796 & 0.579 & ${-}0.216$ $[{-}0.29,\,{-}0.14]$ & ${<}.001$ & ${-}0.135$ $[{-}0.21,\,{-}0.06]$ & $.001$ \\
    \emph{Overcooked! 2} & 0.778 & 0.517 & ${-}0.261$ $[{-}0.32,\,{-}0.20]$ & ${<}.001$ & ${-}0.211$ $[{-}0.27,\,{-}0.15]$ & ${<}.001$ \\
    \emph{Hogwarts Legacy} & 0.884 & 0.581 & ${-}0.303$ $[{-}0.36,\,{-}0.24]$ & ${<}.001$ & ${-}0.255$ $[{-}0.31,\,{-}0.21]$ & ${<}.001$ \\
    Pooled & 0.851 & 0.560 & ${-}0.291$ $[{-}0.33,\,{-}0.25]$ & ${<}.001$ & ${-}0.229$ $[{-}0.27,\,{-}0.19]$ & ${<}.001$ \\
    \bottomrule
  \end{tabularx}
  \begin{minipage}{\linewidth}
    \vspace{2pt}\footnotesize\textit{Note.} Out-of-fold predictions of five participant-wise folds, pooled per target. Intervals from a participant-level cluster bootstrap, $p$ from a two-sided participant-level sign-flip test (10{,}000 draws each). $n$-back is the single-task $n$-back of the cohort, Pooled every labeled window of the cohort.
  \end{minipage}
\end{table*}

Holding out the cohort lowered the \gls{auroc} on all seven targets ($p < .002$, Table~\ref{tab:cohort-shift}),
by $0.09$--$0.22$ on the shared $n$-back and by up to $0.46$ on the applications. The driving application fell below
chance ($0.43$). For comparison, removing the target domain from training cost up to $0.48$ (Table~\ref{tab:lodo}). Reducing the driving training data to the size of the gaming cohort had no effect on the
driving targets ($|\Delta| \le 0.011$, n.s.). On the gaming targets, the same reduction changed the \gls{auroc} by
$-0.042$ to $+0.060$, with opposite signs for ConvNeXt-base and ConvNeXt-small. Both are small against the cohort loss,
so the cohort loss is not a matter of data quantity here.
Cross-cohort transfer relies on the shared paradigm. Removing the source cohort's $n$-back from training lowered
the \gls{auroc} on the target cohort's $n$-back by a further $0.05$--$0.07$ (driving) and $0.22$--$0.35$ (gaming,
$p < .002$). To quantify the loss beyond the cohort effect, we subtracted the $n$-back loss of the same cohort from the
application loss. On the driving application, a loss of $0.35$--$0.36$ remained, so the cohort effect explains only
about a fifth of it. On the games, only $0.05$--$0.12$ remained, with zero inside the interval for \emph{Overcooked! 2}.
The gaming loss is thus largely a cohort effect, whereas the driving loss is mostly specific to the application.
Finally, we tested whether the trained workload models encode the cohort. For the models trained on the
complete dataset, we took the features each model passes to its classification head, reduced them to 50 dimensions
with principal component analysis, and trained a logistic regression to predict the cohort. We used only the shared
$n$-back of participants unseen by the workload model, so the task was identical and only the cohort differed, and
cross-validated the logistic regression by participant. The cohort was identified with an \gls{auroc} of $0.77$--$0.81$
across the three models (chance $0.50$), so the models encode the cohort in addition to task load. As
participants, recording period, and setup all differ between the cohorts, the source of this difference cannot be
determined. Additional controls are reported in Supplementary Section~\ref{ssec:supp-domain-confounds}.
As an additional ablation, independent of the driving and gaming cohorts, we tested whether task-load models
transfer between male and female participants. We applied the same design to two groups defined by participant sex,
training on one group and testing on the other. Transfer between the sexes had no significant effect in 22 of 24
participant-level comparisons ($|\Delta| \le 0.031$), and only one comparison remained significant after Holm
correction (Supplementary Section~\ref{ssec:supp-sex-ablation}). We therefore found no evidence of a group effect
between male and female participants.

\subsection{Graded Tracking of Predicted Load}
\label{ssec:label-monotonicity}
\begin{figure*}[!t]
  \centering
  \subfloat[\centering Combined $n$-back]{\label{sfig:monotonicity-n-back}\includegraphics[width=0.33\linewidth]{%
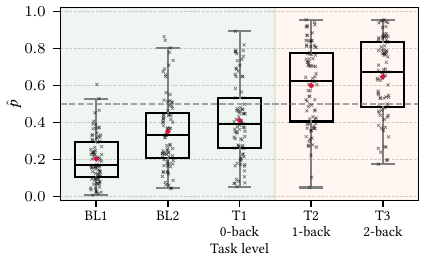}}%
  \subfloat[\centering \emph{Overcooked! 2}]{\label{sfig:monotonicity-overcooked}\includegraphics[width=0.33\linewidth]{%
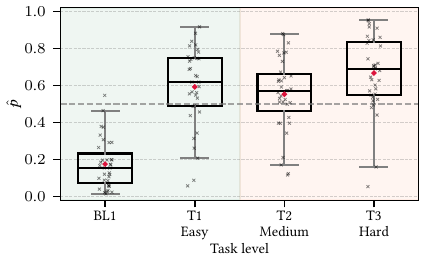}}%
  \subfloat[\centering \emph{Hogwarts Legacy}]{\label{sfig:monotonicity-hogwarts}\includegraphics[width=0.33\linewidth]{%
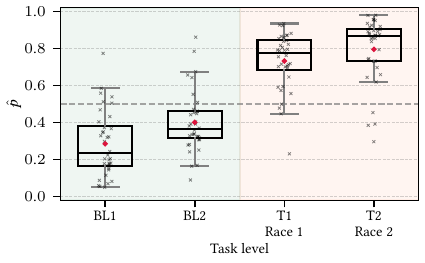}} \\ %
  \subfloat[\centering Combined $n$-back (held-out)]{\label{sfig:monotonicity-heldout-n-back}\includegraphics[width=0.33\linewidth]{%
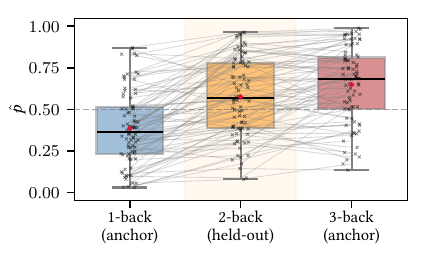}}%
  \subfloat[\centering \emph{Overcooked! 2} (held-out)]{\label{sfig:monotonicity-heldout-overcooked}\includegraphics[width=0.33\linewidth]{%
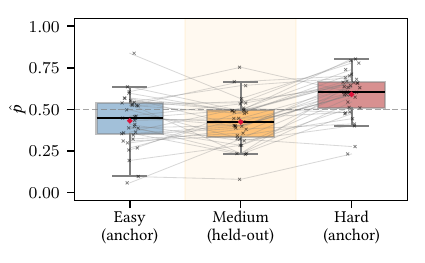}}%
  \subfloat[\centering \emph{Hogwarts Legacy} (held-out)]{\label{sfig:monotonicity-heldout-hogwarts}\includegraphics[width=0.33\linewidth]{%
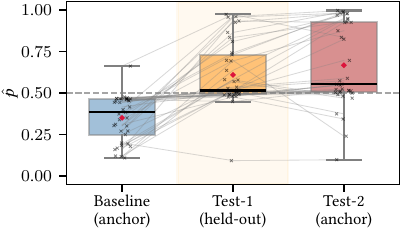}}%
  \caption{Predicted task load $\hat{p}$ per ordered difficulty level (\emph{small} ConvNeXt). Top row: in-distribution %
graded tracking, where the $n$-back and gaming models are evaluated on all levels without retraining. Bottom %
row: held-out generalization, where the model is trained on the extreme anchors only and the shaded intermediate level
is withheld from training. Boxes show the per-subject distribution with individual subjects overlaid and the dashed %
line marks the $0.5$ decision boundary.}
  \label{fig:monotonicity}
  \vspace{-0.15cm}
\end{figure*}

The task-load classifier is trained only on the binary low/high targets, as commonly done in the research community
\citep{oppeltADABaseMultimodalDataset2022}. Task load, however, is a continuous construct. As an internal state,
it cannot be observed directly, and our annotation provides only binary labels. We therefore use the ordered
experimental difficulty as a surrogate. A useful estimator should also resolve this graded difficulty, which
the binary label collapses. We tested this on the \emph{small} ConvNeXt models trained on all $n$-back data
(for the $n$-back domains) and on the gaming data without any retraining, evaluating their predicted high-load
probability $\hat{p}$ across every ordered task level, including intermediate levels and resting baselines that
share a binary label with neighboring levels (Table~\ref{tab:label-monotonicity},
Fig.~\ref{sfig:monotonicity-n-back}--\subref*{sfig:monotonicity-hogwarts}). The low/high boundary is set per protocol (Section~\ref{ssec:target_operationalization}). For
\emph{Hogwarts Legacy} and the driving application, the low class contains only resting baselines. The trained contrast
is therefore task versus rest, and an ordering of $\hat{p}$ can reflect task-phase differences such as visual
stimulation, motor demand or interaction as well as load.
Predicted load increased with difficulty in all domains, with one exception in \emph{Overcooked! 2} described
below. Group-mean $\hat{p}$ rose from the lowest to the highest level in every case, for example from $0.202$ to $0.647$
on the combined $n$-back, from $0.175$ to $0.666$ on \emph{Overcooked! 2}, and from $0.284$ to $0.795$ on
\emph{Hogwarts Legacy}, and Page's trend test was significant throughout ($p < .001$). The mean per-subject Spearman
correlation between level rank and $\hat{p}$ ranged from $\bar\rho = .64$ to $\bar\rho = .77$, and the mean
$\hat{p}$ over the low-difficulty levels lay below that over the high-difficulty levels in $95$--$97\,\%$ of subjects
(Supplementary Table~\ref{tab:supp-label-monotonicity}). Notably, the two highest $n$-back levels carry the same
\emph{high} binary label, yet $\hat{p}$ still separated them, showing that the classifier encodes a finer difficulty
gradient than its training signal.
The predicted gradient also aligned with subjective experience: at the subject $\times$ level granularity where ratings
were available, $\hat{p}$ correlated with NASA-\gls{rtlx} ($\rho = .45$, $p < .001$ for the gaming $n$-back and
$\rho = .29$, $p < .01$ for \emph{Overcooked! 2}). A participant-level mixed-effects analysis on the $n$-back
confirmed this association within participants (standardized $\beta = 0.28$ and $0.55$, $p < .001$), but not beyond
the difficulty level (Supplementary Section~\ref{ssec:supp-mixed-effects}). The exception was
\emph{Overcooked! 2}, where $\hat{p}$ did not increase from easy to medium ($0.59$ and $0.55$) although the two levels
carry different binary labels in training. This is consistent with the non-significant easy-to-medium difference in the
subjective ratings, but the design cannot distinguish similar load from a failure of the model on this contrast.
Overall, a classifier trained on a binary label recovers an ordinal difficulty axis in-distribution.
\begin{table}[!htb]
  \centering
  \setlength{\tabcolsep}{4pt}
  \tablefont
  \caption{In-distribution graded tracking of predicted task load. ConvNeXt-small models were trained on the $n$-back data ($n$-back domains) and on the gaming data (games), using the binary low/high label only; per-subject mean predicted high-load probability $\hat p$ is then evaluated across all $K$ ordered task levels. $\hat p_{\text{low}}$ and $\hat p_{\text{high}}$ are the group means at the lowest and highest level; Page's $L$ tests the ascending trend; $\bar\rho$ is the mean per-subject Spearman correlation between level rank and $\hat p$; $\rho_{\text{RTLX}}$ is the Spearman correlation between $\hat p$ and the NASA-\gls{rtlx} rating at the subject $\times$ level granularity.}
  \label{tab:label-monotonicity}
  \begin{tabularx}{\linewidth}{Xccccccc}
    \toprule
    Domain & $n$ & $K$ & $\hat p_{\text{low}}$ & $\hat p_{\text{high}}$ & Page's $p$ & $\bar\rho$ & $\rho_{\text{RTLX}}$ \\
    \midrule
    Game $n$-back & 37 & 5 & 0.171 & 0.591 & ${<}.001$ & $.64$ & $.45$ \\
    Drive $n$-back & 51 & 5 & 0.224 & 0.688 & ${<}.001$ & $.77$ & -- \\
    Combined $n$-back & 83 & 5 & 0.202 & 0.647 & ${<}.001$ & $.71$ & -- \\
    \emph{Overcooked! 2} & 35 & 4 & 0.175 & 0.666 & ${<}.001$ & $.71$ & $.29$ \\
    \emph{Hogwarts Legacy} & 36 & 4 & 0.284 & 0.795 & ${<}.001$ & $.77$ & -- \\
    \bottomrule
  \end{tabularx}
  \begin{minipage}{\linewidth}
    \vspace{2pt}\footnotesize\textit{Note.} Predictions are read from the deploy runs without retraining (eye tracking and biosignals modalities). Page's trend test is one-sided (ascending). $\rho_{\text{RTLX}}$ is reported for the gaming $n$-back and \emph{Overcooked! 2}. Dashes mark domains for which it was not computed. Per-subject monotonicity shares (the share with a positive rank-$\hat p$ correlation and with mean $\hat p$ over the low levels below that over the high levels) are reported in Supplementary Table~\ref{tab:supp-label-monotonicity}.
  \end{minipage}
\end{table}

\subsection{Monotonic Generalization to Held-Out Levels}
\label{ssec:heldout-monotonicity}
The preceding analysis evaluated task levels that were present during training, reduced into the binary targets. To
test whether the learned difficulty axis interpolates to task levels the model has never seen, we
retrained on the extreme anchors alone, the lowest and highest level, and withheld an intermediate level entirely from
training, then measured $\hat{p}$ on that held-out level for unseen subjects across five subject-wise folds 
(Table~\ref{tab:heldout-monotonicity},
Fig.~\ref{sfig:monotonicity-heldout-n-back}--\subref*{sfig:monotonicity-heldout-hogwarts}).
In four of the five domains the held-out level fell strictly between the two anchors. On the combined $n$-back, mean
$\hat{p}$ rose from $0.39$ (low anchor) to $0.58$ (held-out) to $0.65$ (high anchor) with both
adjacent steps significant ($p < .001$), and \emph{Hogwarts Legacy} showed the strongest ordering
(up to $\bar\rho = .84$). Page's trend test was significant in every domain ($p < .005$).
The exception was \emph{Overcooked! 2}, where only the high anchor separated cleanly: the held-out medium level was
statistically indistinguishable from the easy anchor (one-sided Wilcoxon $p_{1\to2}$ between $.599$ and
$.845$), even though the overall ascending trend remained significant (Page's $p < .005$).
This is consistent with the missing easy-to-medium increase in-distribution and with the absence of an
easy-to-medium difference in NASA-\gls{rtlx}. Whether the two lower difficulties evoke comparable load or the model
fails to separate them cannot be decided with this design. Taken together, the binary classifier learns an
ordinal task-difficulty representation that tracks difficulty in-distribution and, in four of five domains,
interpolates to difficulty levels held out from training. Interpolation within a protocol does not exclude task-phase
cues.
\begin{table}[!htb]
  \centering
  \setlength{\tabcolsep}{2pt}
  \tablefont
  \caption{Monotonic generalization to a held-out task level. Models were trained on the extreme anchors only (lowest and highest level, $\hat p_{1}$ and $\hat p_{3}$); the intermediate level ($\hat p_{2}^{\dagger}$) was withheld from training entirely and the model was evaluated on it after training. Cells report the group mean per-subject predicted high-load probability $\hat p$ on unseen subjects. Page's $L$ tests the ascending trend across the three ordered levels, $\bar\rho$ is the mean per-subject Spearman correlation, and $p_{1\to2}$/$p_{2\to3}$ are one-sided Wilcoxon signed-rank tests for the two adjacent steps. Mono.\ marks domains where both adjacent steps increase.}
  \label{tab:heldout-monotonicity}
  \begin{tabularx}{\linewidth}{Xcccccccc}
    \toprule
    Domain & $\hat p_{1}$ & $\hat p_{2}^{\dagger}$ & $\hat p_{3}$ & Page's $p$ & $\bar\rho$ & $p_{1\to2}$ & $p_{2\to3}$ & Mono. \\
    \midrule
    \multicolumn{9}{@{}l}{\emph{ConvNeXt}} \\
    Game $n$-back & 0.295 & 0.626 & 0.713 & ${<}.001$ & $.76$ & ${<}.001$ & $.009$ & Yes \\
    Drive $n$-back & 0.342 & 0.545 & 0.595 & ${<}.001$ & $.69$ & ${<}.001$ & $.030$ & Yes \\
    Combined $n$-back & 0.386 & 0.576 & 0.649 & ${<}.001$ & $.64$ & ${<}.001$ & ${<}.001$ & Yes \\
    \emph{Overcooked! 2} & 0.432 & 0.425 & 0.591 & ${<}.001$ & $.60$ & $.599$ & ${<}.001$ & No \\
    \emph{Hogwarts Legacy} & 0.350 & 0.609 & 0.667 & ${<}.001$ & $.77$ & ${<}.001$ & $.003$ & Yes \\
    \addlinespace
    \multicolumn{9}{@{}l}{\emph{x\gls{lstm}}} \\
    Game $n$-back & 0.367 & 0.590 & 0.655 & ${<}.001$ & $.62$ & ${<}.001$ & $.030$ & Yes \\
    Drive $n$-back & 0.338 & 0.558 & 0.667 & ${<}.001$ & $.69$ & ${<}.001$ & $.001$ & Yes \\
    Combined $n$-back & 0.394 & 0.598 & 0.682 & ${<}.001$ & $.68$ & ${<}.001$ & ${<}.001$ & Yes \\
    \emph{Overcooked! 2} & 0.533 & 0.516 & 0.648 & ${<}.001$ & $.46$ & $.845$ & ${<}.001$ & No \\
    \emph{Hogwarts Legacy} & 0.328 & 0.586 & 0.740 & ${<}.001$ & $.84$ & ${<}.001$ & ${<}.001$ & Yes \\
    \addlinespace
    \multicolumn{9}{@{}l}{\emph{Transformer}} \\
    Game $n$-back & 0.274 & 0.606 & 0.641 & ${<}.001$ & $.65$ & ${<}.001$ & $.223$ & Yes \\
    Drive $n$-back & 0.335 & 0.594 & 0.673 & ${<}.001$ & $.60$ & ${<}.001$ & $.032$ & Yes \\
    Combined $n$-back & 0.376 & 0.598 & 0.677 & ${<}.001$ & $.58$ & ${<}.001$ & ${<}.001$ & Yes \\
    \emph{Overcooked! 2} & 0.529 & 0.514 & 0.623 & $.004$ & $.31$ & $.765$ & ${<}.001$ & No \\
    \emph{Hogwarts Legacy} & 0.259 & 0.621 & 0.790 & ${<}.001$ & $.80$ & ${<}.001$ & ${<}.001$ & Yes \\
    \bottomrule
  \end{tabularx}
  \begin{minipage}{\linewidth}
    \vspace{2pt}\footnotesize\textit{Note.} $^{\dagger}$Held-out level, never seen during training. Participants per domain: 37 (Game $n$-back), 51 (Drive $n$-back), 83 (Combined $n$-back), 35 (\emph{Overcooked! 2}), 37 (\emph{Hogwarts Legacy}), all 37 because the low anchor pools both baselines. Adjacent Wilcoxon tests are one-sided (ascending) and Bonferroni-corrected within each domain ($\alpha = .025$ per step); Page's trend test is one-sided. Mean and statistics are pooled over five subject-wise folds.
  \end{minipage}
\end{table}

\subsection{Validity of Task-Load Measures}
The binary labels assume that a higher $n$-back level means more task load. We checked this assumption
against three independent references, namely the NASA-\gls{rtlx} rating, the reaction time and the F1-score of the
participant's responses to the $n$-back targets (the harmonic mean of precision and recall). We also compared
the labels with the model estimate $\hat{p}$, the predicted high-load probability of the \emph{small} ConvNeXt
trained on the complete dataset. Each participant was scored by the fold model that had not seen them. All measures
were averaged per participant and level, shown in Table~\ref{tab:correlation_n_nback}. We report Pearson correlations,
which for the binary label equal the point-biserial correlation. Each participant contributed three levels, each being one
observation, so the $p$-values in the table treat dependent observations as independent and should be read as
descriptive.
The references behaved as expected for an increase in load. Higher-load levels were rated as more demanding
($r = .69$), and participants responded more slowly ($r = .48$) and less accurately ($r = -.54$). The three
references also agreed with each other. Levels rated as more demanding also had slower ($r = .49$) and less accurate
($r = -.64$) responses.
The model estimate pointed in the same direction but was weaker. It correlated with the label ($r = .37$), the
rating ($r = .24$), the reaction time ($r = .23$) and the F1-score ($r = -.19$). Two reasons are likely. First, the
rating and the performance measures respond directly to the task, whereas $\hat{p}$ is inferred from physiological
and eye-tracking signals alone. Second, the baseline of $\hat{p}$ differs between participants, and pooling across
participants mixes these differences with the change between levels. Within participants, a mixed model on the
$n$-back confirmed that $\hat{p}$ rose with the rating, but not beyond what the difficulty level already explained
(Section~\ref{ssec:label-monotonicity} and Supplementary Section~\ref{ssec:supp-mixed-effects}). Together,
the rating and the performance measures support the binary labels as a change in task load that participants
noticed and that affected their performance. This change is partly recovered by the model.
\begin{table*}[!t]
  \centering
  \setlength{\tabcolsep}{4pt}
  \tablefont
  \caption{Correlations between the task-load label, the model estimate, the subjective rating and task performance on the $n$-back data (Pearson's $r$, one observation per participant and level).}
  \label{tab:correlation_n_nback}
  \begin{tabularx}{\linewidth}{>{\hsize=1.5\hsize}X*{5}{>{\hsize=0.84\hsize\centering\arraybackslash}X}}
    \toprule
    Measure & Task Load & Predicted~load & NASA-\gls{rtlx} & Reaction~time & F1-Score \\
    \midrule
    Task Load & -- &  &  &  &  \\
    Predicted~load & $\hphantom{-}.37$\textsuperscript{***} & -- &  &  &  \\
    NASA-\gls{rtlx} & $\hphantom{-}.69$\textsuperscript{***} & $\hphantom{-}.24$\textsuperscript{***} & -- &  &  \\
    Reaction~time & $\hphantom{-}.48$\textsuperscript{***} & $\hphantom{-}.23$\textsuperscript{***} & $\hphantom{-}.49$\textsuperscript{***} & -- &  \\
    F1-score & ${-}.54$\textsuperscript{***} & ${-}.19$\textsuperscript{**\hphantom{*}} & ${-}.64$\textsuperscript{***} & ${-}.52$\textsuperscript{***} & -- \\
    \bottomrule
  \end{tabularx}
  \begin{minipage}{\linewidth}
    \vspace{2pt}\footnotesize\textit{Note.} $N = 83$ participants $\times$ 3 $n$-back levels. For the binary task-load label, Pearson's $r$ equals the point-biserial correlation. Predicted load is the predicted high-load probability $\hat{p}$ of the ConvNeXt-small model trained on the complete dataset, scored on participants it had not seen. The $p$-values treat the three observations per participant as independent and are descriptive. \textsuperscript{*}$p<.05$, \textsuperscript{**}$p<.01$, \textsuperscript{***}$p<.001$.
  \end{minipage}
\end{table*}

\section{Discussion}
This work introduces \gls{revelio}, a unified multimodal framework, dataset, and evaluation protocol for cross-domain
task-load estimation. Three key contributions distinguish this work from prior efforts. First, \gls{revelio} is one of 
the largest synchronized multimodal task-load datasets reported to date (96 sessions from 83 participants
across four applications), enabling the participant-level statistical testing and cross-domain contrasts that underpin
our conclusions. Second, the framework provides the first systematic evaluation of multi-domain training, which
yields better generalization and calibration than transfer from a single other domain on domains
represented in training but does not reliably transfer to a domain removed from training (Section~\ref{ssec:lodo}). Third, the results
demonstrate that a classifier trained on coarse binary labels recovers an ordinal task-difficulty representation that
generalizes to difficulty levels withheld from training and correlates with both subjective ratings and behavioral performance.
\subsection{Summary of Findings}
\textbf{RQ1} (predictive utility of continuous measures): Multimodal fusion of biosignals and eye tracking
was ahead of every single input in 61 of 64 domain $\times$ modality comparisons (56 significant after Holm
correction) and ahead of the best single input in 11 of the 12 main domain-by-encoder comparisons, four of them significantly
(the complete dataset with every encoder and driving with the x\gls{lstm}, where eye movement was the best single
input). On the $n$-back and in gaming, pupil diameter alone performed as well as fusion. Among individual biosignals,
\gls{ecg} exhibited the broadest predictive utility across domains, whereas \gls{eda} was the strongest single
biosignal in both games. Pupil diameter was
highly predictive in the $n$-back task (\gls{auroc} 0.80) but degraded in driving (\gls{auroc} 0.65), while eye movement features
excelled in driving (\gls{auroc} 0.78). These findings establish that combining biosignals with eye tracking forms the most
robust default for deployable workload estimation, since it was never significantly worse than the best single
input, which itself differed between tasks.

\textbf{RQ2} (modality contributions by task): Contributions were task dependent. In $n$-back, pupil diameter,
eye tracking, facial \gls{au}s, and head rotation were informative. In $n$-back, the biosignals were weak
(\gls{auroc} $\leq 0.68$). \gls{eda} led in the $n$-back application. In driving, eye movement features, head
rotation, \gls{au}s, and \gls{ecg} dominated, whereas pupil diameter was unreliable. In gaming, \gls{eda} and
\gls{ecg} contributed among biosignals and pupil diameter provided additional signal, but eye movement and
\gls{au}s were weaker than in driving. Absolute body and head position were uninformative across all domains.
Combining \gls{ecg} with \gls{ppg} did not improve over \gls{ecg} alone, indicating that end-to-end learning already
captures the shared electromechanical information.

\textbf{RQ3} (cross-domain generalization): Single-domain training degraded both discrimination (\gls{auroc})
and calibration (\gls{ece}) on unseen domains. The $n$-back to driving transfer showed the largest drop. Gaming was
closer to $n$-back than to driving, consistent with the shared seated, controller-based interaction context described
in Table~\ref{tab:demand-profiles}. This closeness is partly a cohort effect, because the $n$-back training
data include the gaming cohort (Section~\ref{ssec:cohort-shift}). Training on diverse data (complete set or the union
of applications) maintained \gls{auroc} $\geq$ 0.79 (averaged over encoders) and \gls{ece} $\leq$
0.12 across the main domains, all of which were part of its training data.
Participant-level tests confirmed a significant advantage of complete-dataset training over single-domain
transfer in six of seven domains (Table~\ref{tab:generalization-complete-vs-single}) and over 20 of 42 individual
single sources. With the target domain removed from training, the \gls{auroc} fell by $0.05$--$0.09$ on the
$n$-back and the games and below chance on driving. Additional source domains helped mainly when the single source
transferred poorly (Section~\ref{ssec:lodo}).

Beyond the binary classification objective, the trained models recovered an ordinal difficulty axis. Predicted high-load
probability increased with difficulty rank in all domains (Page's trend test $p < .001$), including
levels sharing an identical binary label, with one non-increasing step from easy to medium in
\emph{Overcooked! 2}. On held-out intermediate levels never seen during training, the predicted
probability fell between the two training anchors in four of five domains. Model outputs correlated with NASA-\gls{rtlx} ($\rho = .45$ for the gaming $n$-back,
$\rho = .29$ for \emph{Overcooked! 2}) and, in the convergent validity analysis, with reaction time ($r = .23$) and
task $F_1$-score ($r = -.19$) on the $n$-back data.

\subsection{Relation to Theory and Prior Work}
Our findings extend the conclusions drawn by foundational work on the \gls{adabase} dataset.
\citet{oppeltADABaseMultimodalDataset2022} demonstrated that multimodal fusion of physiological and behavioral signals
improves cognitive load classification, with eye-tracker data emerging as the most predictive single modality in
their combined $n$-back and driving evaluation. \gls{revelio} confirms the strong role of pupil-based features in
cognitively demanding laboratory tasks and further reveals that this dominance is task-specific: pupil diameter
degrades markedly in driving, where eye movement and facial action units become the leading indicators.
\citet{foltynEvaluatingRobustnessMultimodal2024} were the first to systematically investigate the robustness of task
load models under distribution shift using \gls{adabase}, showing that models trained on n-back and evaluated on
driving suffer pronounced performance and calibration drops and that late fusion provides a more stable compromise
than feature-level concatenation. \gls{revelio} extends this robustness analysis from two scenarios to four
applications and confirms their central observation: single-domain training fails to generalize, while diverse
multi-domain training maintains \gls{auroc} $\geq 0.79$ (averaged over encoders) across the main domains when these are represented in training.
\citet{foltynEvaluatingRobustnessMultimodal2024} evaluated on limited data and a single application transfer pair,
we leverage the substantially larger and more heterogeneous \gls{revelio} corpus to achieve competitive cross-domain
performance, suggesting that data diversity can narrow the generalization gap previously observed for domains
represented in training. Under leave-one-domain-out, more data at constant diversity had no consistent effect, and a
domain removed from training lost $0.05$--$0.48$ \gls{auroc} (Section~\ref{ssec:lodo}).
However, on the shared $n$-back, changing only the cohort cost
$0.09$--$0.22$ \gls{auroc}, and the workload models encoded the cohort (Section~\ref{ssec:cohort-shift}).
Beyond this cohort effect, little loss remained on the games, but about $0.35$ remained on driving. Workload models
should therefore be validated on the target population and recording setup.

Beyond replicating and extending these cross-domain findings, \gls{revelio} departs from the majority of prior work
listed in Table~\ref{tab:related-work} in four further respects.
First, the framework adopts a fully end-to-end learning paradigm: raw multichannel time series are fed directly
into the encoder without manual feature extraction, hand-crafted normalization schemes, or domain-guided feature
selection, thereby removing the implicit bias introduced when practitioners pre-select features, such as specific
\gls{hrv} indices or \gls{eda} features, on the basis of prior domain knowledge with other  modalities that are less
explored and therefore lack in baseline performance.
Second, with 96 sessions from 83 participants spanning four applications, \gls{revelio} is one of the
largest synchronized multimodal task-load collection reported to date, scaling not only in the \emph{versatility}
direction by adding qualitatively new interaction domains and variety of collected modalities but also in the
\emph{participant} direction, providing the statistical power required for subject-wise cross-validation and
participant-level inference across architectures and domains.
Third, every task level is grounded in a convergent validation protocol combining objective performance indicators
with post-task subjective self-report (NASA-TLX), establishing a solid psychometric basis that many purely
signal-driven studies lack~\citep{longoHumanMentalWorkload2022}.
Fourth, whereas most related systems are limited to predicting a discrete binary or categorical load label,
\gls{revelio} demonstrates through the held-out anchor protocol that binary-trained classifiers recover an ordinal task-difficulty axis, with predicted probabilities
interpolating to difficulty levels never seen during training in four of five domains. The evidence is relative to
each protocol and does not exclude task-phase cues.
\subsection{Practical Guidance for Adaptive System Design}

A practical interpretation emerges from the observed domain differences. A modality is informative when it is
directly related to the demand manipulation of the respective task, but becomes less reliable when it is affected by the
recording conditions. The autonomic modalities showed moderate yet comparatively stable performance across domains:
\gls{ecg} achieved an \gls{auroc} of 0.73 on the complete dataset and 0.77 in driving, while \gls{eda} was the
strongest biosignal in the combined $n$-back and in both games, ranging from 0.65 to 0.77. By contrast, the
best-performing unimodal predictor depended on the task. Pupil diameter was most informative in the seated,
low-movement, and fixed-illumination conditions of the gaming domains and the combined $n$-back, reaching 0.80 in the
latter. In the driving task, eye movements and head rotation were the strongest non-facial inputs (0.78 each), whereas
pupil diameter reached only 0.65, presumably
because variable illumination and continuous visual scanning reduced its reliability. The results for facial \glspl{au}
further indicate a substantial influence of the recording setup. Their performance reached 0.85 in both the driving
task and the driving cohort's $n$-back task, but only 0.54 in the corresponding $n$-back task of the gaming cohort. This
difference suggests that camera framing and illumination contributed more strongly to the prediction than cognitive load
itself. More generally, the modality ranking followed the cohort more closely than the task. \glspl{au} were
the strongest single modality in every driving-cohort domain and pupil diameter in every gaming-cohort domain. On the $n$-back task,
which both cohorts share, the workload models also encoded the cohort (Section~\ref{ssec:cohort-shift}). This test
does not show which modality carries the cohort information. The advantage of pupil diameter in the gaming cohort may
therefore partly reflect the recording setup rather than task load. A
similar limitation was observed for movement features: absolute head and body positions remained close to
chance level ($\leq$ 0.63), and adding movement features to the full multimodal model reduced the \gls{auroc} from 0.87
to 0.83. These features therefore appear to capture participant-specific posture rather than graded task demand. For
system design, the findings support an autonomic core comprising \gls{ecg} and \gls{eda}, as these modalities showed
comparatively robust performance across contexts. Within this benchmark, however, adding the biosignals to eye
tracking changed the \gls{auroc} by at most 0.04 in any domain (Table~\ref{tab:modalities-multiple-convnext}), so the
autonomic core does not add measurable accuracy. Whether it also depends less on illumination and camera setup
remains a hypothesis that this benchmark does not test. This core may be
complemented by a task-specific modality, such as pupil diameter under controlled illumination or gaze and head
rotation in visuomotor tasks. Appearance-and position-based modalities should, however, only be included after
validation across cohorts and recording conditions. Their association with task demand must be demonstrated for the
intended deployment context rather than assumed from benchmark performance.

The results translate into actionable recommendations for designers of workload-aware systems, such as games that
adjust their interaction to the player's monitored cognitive load \citep{zhangEnhancingUserExperience2026}. For
\textbf{adaptive gaming interfaces}, \gls{eda} and eye tracking should be prioritized over \gls{emg}, which provides
negligible discriminative signal. For \textbf{driving-related applications}, eye movement features and head rotation
are the most informative behavioral indicators, and designers should avoid relying on pupil diameter
alone in environments with variable illumination, where both eye movement and \gls{ecg} were significantly more
informative (Table~\ref{tab:modality-contrasts}). For \textbf{general-purpose adaptive systems} operating across
multiple interaction contexts, biosignals combined with eye tracking provide the most robust and computationally
efficient default configuration.

On the modeling side, three architectural families (ConvNeXt, Transformer, and \gls{xlstm}) attained comparable
performance under a controlled protocol, indicating that domain and modality selection dominate over architecture
choice. Longer input windows generally improved performance, and our 40\,s window provided a practical trade-off
between latency and accuracy, with updates every 20\,s, each integrating the preceding 40\,s. This timing suits
adaptation at the session or shift level, such as break scheduling, task reallocation, or difficulty adjustment, but
not adaptation to individual events, such as suppressing a single warning. The finding that
diverse training data improved both \gls{auroc} and calibration over transfer from a single other domain
suggests that developers should invest in heterogeneous training corpora rather than optimizing models for narrow
application scenarios, provided the corpus includes the target application. A model should not be expected to
transfer to an application absent from its training data (Section~\ref{ssec:lodo}). Calibration (\gls{ece}) should be
treated as a relevant evaluation criterion alongside discrimination, because overconfident predictions under domain
shift can trigger inappropriate responses.

Binary classification is pragmatic for early-stage deployment. The monotonicity results, however, demonstrate that
even binary-trained models encode a finer-grained difficulty gradient.

\subsection{Limitations and Future Work}

External validity is constrained by a sample of colleagues and students from a single institute, controlled
laboratory environments, specific hardware, and two game titles selected for ecological validity and demand diversity 
rather than for systematic factorial manipulation of cognitive dimensions. A factorial design crossing specific demand 
types would enable stronger causal claims about which cognitive dimensions drive cross-domain transfer. Differences in 
participant skill and gaming experience may confound load induction, as indicated by the significant effect of 
self-reported experience on \emph{Overcooked! 2} performance, whereas the same test was non-significant for
\emph{Hogwarts Legacy} and the $n$-back (Supplementary Section~\ref{app:ssec:skill-performance}).
Familiarity was assessed by self-report rather than a validated expertise instrument. Construct validity is 
limited by the coarse binary mapping of task levels to load. Objective performance metrics were unavailable in some
task segments, and emotional arousal or prior experience may influence perceived or physiological load.

Measurement limitations include slight lighting differences across tasks affecting pupil diameter, \gls{au} extraction
at 25\,Hz and eye tracking at 250\,Hz limiting temporal fidelity for fast dynamics, and potential residual label noise
despite multi-indicator verification. Cohort, recording period, and setup are confounded in this corpus
(Section~\ref{ssec:cohort-shift}), so their effects cannot be separated.

Future work should extend \gls{revelio} in several directions: (1) moving beyond binary labels to multi-class or
continuous load estimation to capture the inherently multidimensional nature of cognitive load, (2) applying domain
adaptation, self-supervised pretraining \citep{oppeltLUMOSSelfsupervisedPretraining2026}, and personalization techniques to improve transfer under distribution shift,
(3) assessing robustness to missing or corrupted modalities for real-world deployment, (4) broadening the task
diversity to include mobile and field studies with practical wearable sensors, (5) expanding participant demographics
and investigating fairness across gender, age, and skill groups, (6) developing uncertainty-aware adaptation
policies that leverage calibrated confidence estimates, and (7) measuring how quickly the predictions follow
the onset and offset of task load. Ethical considerations include privacy and informed consent
for continuous biosignal, gaze, and video capture, transparency about adaptive system behaviors, and equitable
performance across diverse user populations.
\subsection{Conclusion}
\gls{revelio} provides a standardized foundation for measuring modality contributions, quantifying cross-domain
generalization, and advancing workload-aware adaptive systems. The framework unifies 96 multimodal sessions across
four applications under consistent hardware, synchronization, and annotation. Our results establish
three central findings: multimodal fusion of biosignals and eye tracking is never significantly worse than
the best single modality and significantly better on the complete dataset, modality informativeness is task-specific
rather than universal, and diverse multi-domain training produces robust and well-calibrated models for the
domains represented in training, but not for a domain removed from training. The public release of
60 of the 96 synchronized recordings with metadata, combined with a hidden test set for community
benchmarking, provides the infrastructure for reproducible progress toward deployable workload estimation in adaptive
\gls{hci}.
\section*{Disclosure Statement}
The authors report there are no competing interests to declare.
\section*{Funding}
The authors acknowledge the partial (data quality) funding by the EU TEFHealth project which is part of the Digital
Europe Program of the EU (DIGITAL-2022-CLOUD-AI-02-TEFHEALTH) under grant agreement no. 101100700 and the Berlin Senate
and the partial (training) funding by a grant of the Bavarian Research Society under the project “FOR Social Robots”,
AZ-1594-23. The data collection was financed exclusively through Fraunhofer institutional self-funding and
Fraunhofer-Gesellschaft e.V. is therefore the sole owner of the resulting dataset, and the aforementioned grants confer
no rights therein.
\section*{Data Availability Statement}
The data that support the findings of this study are available under restricted access from Zenodo
(\url{https://doi.org/10.5281/zenodo.24060698}, \url{https://doi.org/10.5281/zenodo.24051552}),
or at \texttt{adabase@iis.fraunhofer.de}, upon reasonable request. The data are not openly available because of the
sensitive nature of the continuous physiological, eye-tracking, and behavioral signals recorded from human
participants: such biosignals cannot be fully anonymized and are not under the conscious control of the individual,
so unrestricted release could enable misuse. Access to the data files is granted to researchers at academic
institutions, free of charge and for non-commercial academic research only, after signing an End User License Agreement
(EULA). The EULA restricts use to academic research, prohibits commercial, military, governmental-surveillance, and
interrogation-related applications, and requires verification of the requester's academic affiliation and only data
from participants who gave written consent to data sharing are included. The release comprises 60 of the 96
sessions, namely the 30 driving recordings of the earlier \gls{adabase} release and 30 new gaming recordings, and covers
both cohorts and all four applications. Within each cohort, the public recordings were sampled at random from
the participants who consented to data sharing. Python code to read, select, and preprocess the released data is
provided together with documentation of its use. The remaining 36 sessions are retained as a hidden test set
for benchmarking challenges. The released data support the complete evaluation protocol at a reduced sample size.

\bibliographystyle{apalike}
\bibliography{bibliography}
%

\twocolumn[%
\begin{center}
  {\fontsize{20}{24}\selectfont Supplementary Material for:\\[2pt]
  REVELIO: Multimodal Task Load Estimation for\\[2pt] Cross-Domain Generalization\par}
  \vspace{10pt}
  {\large Maximilian P. Oppelt, Andreas Foltyn, Nadine R. Lang-Richter, and Bjoern M. Eskofier\par}
\end{center}
\vspace{14pt}]

\setcounter{section}{0}
\setcounter{figure}{0}
\setcounter{table}{0}
\setcounter{equation}{0}
\renewcommand{\thesection}{S\arabic{section}}
\renewcommand{\thetable}{S\arabic{table}}
\renewcommand{\thefigure}{S\arabic{figure}}
\renewcommand{\theequation}{S\arabic{equation}}
\renewcommand{\theHsection}{supp.\arabic{section}}
\renewcommand{\theHtable}{supp.\arabic{table}}
\renewcommand{\theHfigure}{supp.\arabic{figure}}
\renewcommand{\theHequation}{supp.\arabic{equation}}

\raggedbottom

\begin{bibunit}[apalike]

\section{Supplementary Materials}
Additional methodological details, extended results, and configuration tables referenced in the main article are
provided below. This includes experimental protocols, data processing pipelines, model hyperparameters, and
further analyses supporting reproducibility and transparency.
\subsection{Related Work on Cognitive Load Measurements}
Empirical studies evaluate cognitive load through measurements selected for their sensitivity to changes in cognitive
demand while avoiding external disturbances during task execution
\cite{odonnellWorkloadAssesmentMethodology1986, orruEvolutionCognitiveLoad2019}. As introduced in the main
article, the literature identifies four categories: subjective, performance, behavioral, and physiological measures
\cite{chenMultimodalBehaviorInteraction2012}. This section provides extended background on each category.

\emph{Subjective measurements} require participants to perform introspective self-assessment of their mental demand
\cite{odonnellWorkloadAssesmentMethodology1986}. The NASA-\gls{tlx} \cite{hartDevelopmentNASATLXTask1988} and its
variants, such as the Raw-\gls{tlx} \cite{georgssonNASARTLXNovel2020}, utilize a multidimensional framework based
on performance, mental effort, frustration, mental and physical demand, and temporal demand. A key limitation is that
these questionnaires are typically answered post-task, precluding capture of temporal variations
\cite{paasInstructionalControlCognitive1994}. Even instruments designed for in-task administration, such as
\gls{isa} \cite{jordanInstantaneousSelfassessmentWorkload1992}, can interrupt ongoing performance
\cite{odonnellWorkloadAssesmentMethodology1986}.

\emph{Performance measurements} reflect cognitive load variations based on the assumption that performance declines
when capacity is overloaded \cite{paasInstructionalControlCognitive1994}. A common approach is the dual-task
paradigm, where performance on a concurrent secondary task serves as the indicator. However, these measurements can
be confounded by non-workload factors such as skill level, and the secondary task itself may intrude on primary task
performance. Performance metrics are also highly task-dependent.
For instance, reaction time to critical events serves as a metric in autonomous driving observation
\cite{oppeltADABaseMultimodalDataset2022}, whereas the number of extinguished fires is used in emergency
firefighter games \cite{sevcenkoMeasuringCognitiveLoad2021}.

\emph{Behavioral measures} are extracted from user activity during task execution. Recent studies have utilized
eye-tracker-based metrics such as fixation frequency and duration, as well as head and hand movements during video
gaming \cite{linEvaluatingUsabilityBased2006}. Facial action units have been applied in driver monitoring
\cite{yuceActionUnitsTheir2017}. Behavioral measures can be consciously concealed or may vary according to
individual character or cultural background \cite{ekmanDarwinDeceptionFacial2003}. A notable advantage is that
they can be recorded with minimal intrusion, enabling continuous monitoring during task performance.

\emph{Physiological measurements} operate on the premise that cognitive load produces alterations in physiological
processes. Signals are typically captured using non-invasive sensors: electrical cardiac activity via \gls{ecg},
muscle activation via \gls{emg}, palm sweat gland activity via \gls{eda}, mechanical cardiac activity via
\gls{ppg}, and brain activity via \gls{eeg}. Pupil dilation can be observed through specialized eye-tracker hardware
\cite{oppeltADABaseMultimodalDataset2022}. These measurements are applicable in laboratory experiments without
disrupting most tasks \cite{longoHumanMentalWorkload2022}, though extracted features may be influenced by factors
beyond mental workload \cite{cainReviewMentalWorkload2007}. Critically, biosignal changes occur on different
timescales: evoked potentials via \gls{eeg} manifest in sub-second windows, skin responses within seconds, heart
rate variability changes over 30 seconds to 10 minutes, and electrogastrography requires windows of $20$ minutes or
more \cite{oppeltADABaseMultimodalDataset2022, phamHeartRateVariability2021}. These temporal characteristics
constrain the minimum window lengths for feature extraction and classification, directly motivating the window-length
analysis presented in the main article.

No single measurement category is optimal due to limitations in predictive performance, robustness against external
factors, individual variability, and susceptibility to noise from unimodal capture. These shortcomings motivate the
multimodal fusion approach adopted in this work
\cite{chenMultimodalBehaviorInteraction2012,debieMultimodalFusionObjective2021}.
\subsection{Acquisition Setup}
Figure~\ref{app:system-overview} illustrates the system architecture used for multimodal data collection.
The setup integrates a video system, the Biopac physiological recording system, and the game system, all synchronized
via analog triggers managed by the Biopac unit. Video frame acquisition is initiated by Biopac output triggers,
while the game system communicates task phase information through dedicated trigger signals. This architecture
ensures high-precision temporal alignment across all data streams and prevents time drift between systems.
\begin{figure}[htb!]
  \centering
\tikzset{every picture/.style={line width=0.75pt}} 
\begin{tikzpicture}[x=0.75pt,y=0.75pt,yscale=-1,xscale=1]
\draw   (0,36.92) .. controls (0,33.1) and (3.1,30) .. (6.92,30) -- (27.68,30) .. controls (31.5,30) and (34.6,33.1) .. (34.6,36.92) -- (34.6,83.31) .. controls (34.6,87.13) and (31.5,90.23) .. (27.68,90.23) -- (6.92,90.23) .. controls (3.1,90.23) and (0,87.13) .. (0,83.31) -- cycle ;
\draw    (6.4,35.98) -- (28.67,36.19) ;
\draw    (0.71,45.72) -- (34.84,45.97) ;
\draw   (28.08,41.33) .. controls (28.08,41.71) and (28.44,42.02) .. (28.88,42.02) .. controls (29.32,42.02) and (29.68,41.71) .. (29.68,41.33) .. controls (29.68,40.95) and (29.32,40.65) .. (28.88,40.65) .. controls (28.44,40.65) and (28.08,40.95) .. (28.08,41.33) -- cycle ;

\draw   (303.3,67.52) .. controls (303.3,63.04) and (307.3,59.41) .. (312.23,59.41) .. controls (317.16,59.41) and (321.16,63.04) .. (321.16,67.52) .. controls (321.16,72) and (317.16,75.63) .. (312.23,75.63) .. controls (307.3,75.63) and (303.3,72) .. (303.3,67.52) -- cycle ; \draw   (308.3,64.76) .. controls (308.3,64.31) and (308.7,63.95) .. (309.2,63.95) .. controls (309.69,63.95) and (310.09,64.31) .. (310.09,64.76) .. controls (310.09,65.21) and (309.69,65.57) .. (309.2,65.57) .. controls (308.7,65.57) and (308.3,65.21) .. (308.3,64.76) -- cycle ; \draw   (314.37,64.76) .. controls (314.37,64.31) and (314.77,63.95) .. (315.27,63.95) .. controls (315.76,63.95) and (316.16,64.31) .. (316.16,64.76) .. controls (316.16,65.21) and (315.76,65.57) .. (315.27,65.57) .. controls (314.77,65.57) and (314.37,65.21) .. (314.37,64.76) -- cycle ; \draw   (307.77,70.76) .. controls (310.74,72.92) and (313.72,72.92) .. (316.7,70.76) ;
\draw   (299.62,88.51) .. controls (299.62,83.25) and (303.89,78.98) .. (309.16,78.98) -- (315.37,78.98) .. controls (320.64,78.98) and (324.91,83.25) .. (324.91,88.51) -- (324.91,98.05) .. controls (324.91,98.05) and (324.91,98.05) .. (324.91,98.05) -- (299.62,98.05) .. controls (299.62,98.05) and (299.62,98.05) .. (299.62,98.05) -- cycle ;
\draw   (191.9,125.22) -- (253.12,125.22) -- (253.12,129.68) -- (191.9,129.68) -- cycle ;
\draw   (225.45,127.45) .. controls (225.45,127.01) and (225.87,126.65) .. (226.39,126.65) .. controls (226.91,126.65) and (227.33,127.01) .. (227.33,127.45) .. controls (227.33,127.9) and (226.91,128.26) .. (226.39,128.26) .. controls (225.87,128.26) and (225.45,127.9) .. (225.45,127.45) -- cycle ;
\draw   (216.85,127.45) .. controls (216.85,127.01) and (217.27,126.65) .. (217.8,126.65) .. controls (218.32,126.65) and (218.74,127.01) .. (218.74,127.45) .. controls (218.74,127.9) and (218.32,128.26) .. (217.8,128.26) .. controls (217.27,128.26) and (216.85,127.9) .. (216.85,127.45) -- cycle ;

\draw   (190.83,56.4) .. controls (190.83,51.45) and (194.85,47.43) .. (199.81,47.43) -- (259.31,47.43) .. controls (264.27,47.43) and (268.29,51.45) .. (268.29,56.4) -- (268.29,83.33) .. controls (268.29,88.29) and (264.27,92.3) .. (259.31,92.3) -- (199.81,92.3) .. controls (194.85,92.3) and (190.83,88.29) .. (190.83,83.33) -- cycle ;
\draw   (205.71,97.72) -- (254.69,97.72) -- (254.69,96.32) -- (205.71,96.32) -- cycle ;
\draw   (228.66,97.02) -- (230.2,97.02) -- (230.2,93.21) -- (228.66,93.21) -- cycle ;
\draw   (214.93,9.31) .. controls (214.93,7.78) and (216.17,6.54) .. (217.69,6.54) -- (246.8,6.54) .. controls (248.32,6.54) and (249.56,7.78) .. (249.56,9.31) -- (249.56,17.61) .. controls (249.56,19.14) and (248.32,20.38) .. (246.8,20.38) -- (217.69,20.38) .. controls (216.17,20.38) and (214.93,19.14) .. (214.93,17.61) -- cycle ;
\draw    (261.03,100.42) .. controls (257.22,103.72) and (263.34,120.23) .. (267.82,115.81) ;
\draw    (261.03,100.42) .. controls (262.63,99.01) and (263.84,97.78) .. (265.33,98.17) ;
\draw    (265.33,98.17) .. controls (266.92,98.49) and (267.03,99.31) .. (268.48,99.31) ;
\draw    (268.48,99.31) -- (272.26,99.36) ;
\draw    (267.82,115.81) .. controls (268.87,114.93) and (267.39,112.15) .. (272.98,112.3) ;

\draw   (263.07,103.14) .. controls (263.07,101.89) and (264.14,100.88) .. (265.46,100.88) .. controls (266.77,100.88) and (267.84,101.89) .. (267.84,103.14) .. controls (267.84,104.39) and (266.77,105.4) .. (265.46,105.4) .. controls (264.14,105.4) and (263.07,104.39) .. (263.07,103.14) -- cycle ;
\draw   (274.41,99.35) .. controls (274.48,99.53) and (274.52,99.74) .. (274.52,99.95) .. controls (274.51,100.87) and (273.71,101.61) .. (272.74,101.6) .. controls (271.76,101.59) and (270.98,100.83) .. (270.99,99.91) .. controls (270.99,99.7) and (271.04,99.5) .. (271.11,99.31) -- cycle ;
\draw   (266.1,108.63) .. controls (266.1,108.49) and (266.21,108.39) .. (266.34,108.39) -- (269.95,108.43) .. controls (270.08,108.43) and (270.19,108.53) .. (270.18,108.66) -- (270.18,109.37) .. controls (270.17,109.5) and (270.07,109.61) .. (269.94,109.61) -- (266.33,109.57) .. controls (266.2,109.57) and (266.09,109.46) .. (266.09,109.33) -- cycle ;
\draw   (267.75,110.84) .. controls (267.6,110.84) and (267.49,110.72) .. (267.49,110.58) -- (267.48,107.42) .. controls (267.48,107.28) and (267.6,107.16) .. (267.74,107.16) -- (268.53,107.16) .. controls (268.67,107.16) and (268.79,107.28) .. (268.79,107.42) -- (268.8,110.58) .. controls (268.8,110.72) and (268.68,110.84) .. (268.53,110.84) -- cycle ;
\draw   (275.41,109.06) .. controls (275.41,108.13) and (276.21,107.38) .. (277.19,107.38) .. controls (278.17,107.38) and (278.96,108.13) .. (278.96,109.06) .. controls (278.96,109.99) and (278.17,110.75) .. (277.19,110.75) .. controls (276.21,110.75) and (275.41,109.99) .. (275.41,109.06) -- cycle ;
\draw   (278.36,103.26) .. controls (278.36,102.94) and (278.64,102.68) .. (278.97,102.68) .. controls (279.31,102.68) and (279.58,102.94) .. (279.58,103.26) .. controls (279.58,103.58) and (279.31,103.84) .. (278.97,103.84) .. controls (278.64,103.84) and (278.36,103.58) .. (278.36,103.26) -- cycle ;
\draw   (281.63,103.3) .. controls (281.63,102.98) and (281.91,102.72) .. (282.24,102.72) .. controls (282.58,102.72) and (282.85,102.98) .. (282.85,103.3) .. controls (282.85,103.62) and (282.58,103.88) .. (282.24,103.88) .. controls (281.91,103.88) and (281.63,103.62) .. (281.63,103.3) -- cycle ;
\draw   (280.17,105.01) .. controls (280.17,104.69) and (280.44,104.43) .. (280.77,104.43) .. controls (281.11,104.43) and (281.38,104.69) .. (281.38,105.01) .. controls (281.38,105.33) and (281.11,105.59) .. (280.77,105.59) .. controls (280.44,105.59) and (280.17,105.33) .. (280.17,105.01) -- cycle ;
\draw   (280.08,101.67) .. controls (280.08,101.35) and (280.35,101.09) .. (280.69,101.09) .. controls (281.03,101.09) and (281.3,101.35) .. (281.3,101.67) .. controls (281.3,101.99) and (281.03,102.25) .. (280.69,102.25) .. controls (280.35,102.25) and (280.08,101.99) .. (280.08,101.67) -- cycle ;
\draw    (284.38,100.42) .. controls (288.05,103.72) and (282.15,120.23) .. (277.84,115.81) ;
\draw    (284.38,100.42) .. controls (282.84,99.01) and (281.68,97.78) .. (280.24,98.17) ;
\draw    (280.24,98.17) .. controls (278.71,98.49) and (278.6,99.31) .. (277.2,99.31) ;
\draw    (277.2,99.31) -- (273.57,99.36) ;
\draw    (277.84,115.81) .. controls (276.83,114.93) and (278.25,112.15) .. (272.87,112.3) ;

\draw   (46,76.06) .. controls (46,72.24) and (49.1,69.14) .. (52.92,69.14) -- (73.68,69.14) .. controls (77.5,69.14) and (80.6,72.24) .. (80.6,76.06) -- (80.6,122.45) .. controls (80.6,126.27) and (77.5,129.37) .. (73.68,129.37) -- (52.92,129.37) .. controls (49.1,129.37) and (46,126.27) .. (46,122.45) -- cycle ;
\draw    (52.4,75.13) -- (74.67,75.33) ;
\draw    (46.71,84.86) -- (80.84,85.11) ;
\draw   (74.08,80.47) .. controls (74.08,80.85) and (74.44,81.16) .. (74.88,81.16) .. controls (75.32,81.16) and (75.68,80.85) .. (75.68,80.47) .. controls (75.68,80.1) and (75.32,79.79) .. (74.88,79.79) .. controls (74.44,79.79) and (74.08,80.1) .. (74.08,80.47) -- cycle ;

\draw   (116.86,60.21) .. controls (116.86,56.38) and (119.96,53.29) .. (123.78,53.29) -- (144.54,53.29) .. controls (148.36,53.29) and (151.46,56.38) .. (151.46,60.21) -- (151.46,106.59) .. controls (151.46,110.42) and (148.36,113.51) .. (144.54,113.51) -- (123.78,113.51) .. controls (119.96,113.51) and (116.86,110.42) .. (116.86,106.59) -- cycle ;
\draw    (123.26,59.27) -- (145.53,59.47) ;
\draw    (117.57,69) -- (151.69,69.26) ;
\draw   (144.94,64.62) .. controls (144.94,64.99) and (145.3,65.3) .. (145.74,65.3) .. controls (146.18,65.3) and (146.54,64.99) .. (146.54,64.62) .. controls (146.54,64.24) and (146.18,63.93) .. (145.74,63.93) .. controls (145.3,63.93) and (144.94,64.24) .. (144.94,64.62) -- cycle ;

\draw    (115,55.88) -- (77.11,41.1) ;
\draw [shift={(75.25,40.38)}, rotate = 21.3] [color={rgb, 255:red, 0; green, 0; blue, 0 }  ][line width=0.75]    (6.56,-1.97) .. controls (4.17,-0.84) and (1.99,-0.18) .. (0,0) .. controls (1.99,0.18) and (4.17,0.84) .. (6.56,1.97)   ;
\draw    (16,10.9) -- (16,28.5) ;
\draw [shift={(16,30.5)}, rotate = 270] [color={rgb, 255:red, 0; green, 0; blue, 0 }  ][line width=0.75]    (6.56,-1.97) .. controls (4.17,-0.84) and (1.99,-0.18) .. (0,0) .. controls (1.99,0.18) and (4.17,0.84) .. (6.56,1.97)   ;
\draw    (16,10.9) -- (212.29,10.14) ;
\draw    (76.75,32.13) -- (209.77,15.37) ;
\draw [shift={(211.75,15.13)}, rotate = 172.82] [color={rgb, 255:red, 0; green, 0; blue, 0 }  ][line width=0.75]    (6.56,-1.97) .. controls (4.17,-0.84) and (1.99,-0.18) .. (0,0) .. controls (1.99,0.18) and (4.17,0.84) .. (6.56,1.97)   ;
\draw    (312.5,35) -- (312.55,57.46) ;
\draw    (64,45.14) -- (63.74,66) ;
\draw [shift={(63.71,68)}, rotate = 270.72] [color={rgb, 255:red, 0; green, 0; blue, 0 }  ][line width=0.75]    (6.56,-1.97) .. controls (4.17,-0.84) and (1.99,-0.18) .. (0,0) .. controls (1.99,0.18) and (4.17,0.84) .. (6.56,1.97)   ;
\draw    (151.69,79.26) -- (171.25,79.46) -- (188.09,79.63) ;
\draw [shift={(190.09,79.66)}, rotate = 180.6] [color={rgb, 255:red, 0; green, 0; blue, 0 }  ][line width=0.75]    (6.56,-1.97) .. controls (4.17,-0.84) and (1.99,-0.18) .. (0,0) .. controls (1.99,0.18) and (4.17,0.84) .. (6.56,1.97)   ;
\draw    (137.56,126.78) -- (137.37,116.22) ;
\draw [shift={(137.33,114.22)}, rotate = 88.99] [color={rgb, 255:red, 0; green, 0; blue, 0 }  ][line width=0.75]    (6.56,-1.97) .. controls (4.17,-0.84) and (1.99,-0.18) .. (0,0) .. controls (1.99,0.18) and (4.17,0.84) .. (6.56,1.97)   ;
\draw    (255,106.83) -- (157.46,106.6) ;
\draw [shift={(155.46,106.59)}, rotate = 0.14] [color={rgb, 255:red, 0; green, 0; blue, 0 }  ][line width=0.75]    (6.56,-1.97) .. controls (4.17,-0.84) and (1.99,-0.18) .. (0,0) .. controls (1.99,0.18) and (4.17,0.84) .. (6.56,1.97)   ;
\draw    (137.56,126.78) -- (189.49,126.95) ;
\draw   (65.89,29.54) .. controls (65.89,29.13) and (66.22,28.8) .. (66.63,28.8) -- (68.86,28.8) .. controls (69.27,28.8) and (69.6,29.13) .. (69.6,29.54) -- (69.6,43.2) .. controls (69.6,43.61) and (69.27,43.94) .. (68.86,43.94) -- (66.63,43.94) .. controls (66.22,43.94) and (65.89,43.61) .. (65.89,43.2) -- cycle ;
\draw   (62.17,29.54) .. controls (62.17,29.13) and (62.5,28.8) .. (62.91,28.8) -- (65.14,28.8) .. controls (65.55,28.8) and (65.89,29.13) .. (65.89,29.54) -- (65.89,43.2) .. controls (65.89,43.61) and (65.55,43.94) .. (65.14,43.94) -- (62.91,43.94) .. controls (62.5,43.94) and (62.17,43.61) .. (62.17,43.2) -- cycle ;
\draw   (58.46,29.54) .. controls (58.46,29.13) and (58.79,28.8) .. (59.2,28.8) -- (61.43,28.8) .. controls (61.84,28.8) and (62.17,29.13) .. (62.17,29.54) -- (62.17,43.2) .. controls (62.17,43.61) and (61.84,43.94) .. (61.43,43.94) -- (59.2,43.94) .. controls (58.79,43.94) and (58.46,43.61) .. (58.46,43.2) -- cycle ;
\draw   (54.74,29.54) .. controls (54.74,29.13) and (55.08,28.8) .. (55.49,28.8) -- (57.71,28.8) .. controls (58.12,28.8) and (58.46,29.13) .. (58.46,29.54) -- (58.46,43.2) .. controls (58.46,43.61) and (58.12,43.94) .. (57.71,43.94) -- (55.49,43.94) .. controls (55.08,43.94) and (54.74,43.61) .. (54.74,43.2) -- cycle ;
\draw   (69.6,29.54) .. controls (69.6,29.13) and (69.93,28.8) .. (70.34,28.8) -- (72.57,28.8) .. controls (72.98,28.8) and (73.31,29.13) .. (73.31,29.54) -- (73.31,43.2) .. controls (73.31,43.61) and (72.98,43.94) .. (72.57,43.94) -- (70.34,43.94) .. controls (69.93,43.94) and (69.6,43.61) .. (69.6,43.2) -- cycle ;
\draw    (76.25,36.13) -- (315,35) ;
\draw    (315,35) -- (315.05,57.46) ;
\draw    (310,35) -- (310.05,57.46) ;
\draw  [fill={rgb, 255:red, 0; green, 0; blue, 0 }  ,fill opacity=1 ] (307.05,35) .. controls (307.05,33.62) and (309.49,32.5) .. (312.5,32.5) .. controls (315.51,32.5) and (317.95,33.62) .. (317.95,35) .. controls (317.95,36.38) and (315.51,37.5) .. (312.5,37.5) .. controls (309.49,37.5) and (307.05,36.38) .. (307.05,35) -- cycle ;
\draw   (254.78,19.93) .. controls (251.89,19.93) and (249.56,17.59) .. (249.56,14.71) -- (249.56,12.15) .. controls (249.57,9.27) and (251.9,6.93) .. (254.79,6.93) -- (260,6.94) .. controls (260,6.94) and (260,6.94) .. (260,6.94) -- (260,19.94) .. controls (260,19.94) and (260,19.94) .. (260,19.94) -- cycle ;
\draw   (204.65,133.35) -- (236.06,133.35) -- (236.06,131.88) -- (204.65,131.88) -- cycle ;
\draw   (220.35,132.62) -- (221.69,132.62) -- (221.69,130.44) -- (220.35,130.44) -- cycle ;

\draw (62.00,102.26) node  [font=\scriptsize] [align=left] {\begin{minipage}[lt]{18.59pt}\setlength\topsep{0pt}
\begin{center}
Biopac\\System
\end{center}

\end{minipage}};
\draw (16.23,64.8) node  [font=\scriptsize] [align=left] {\begin{minipage}[lt]{18.59pt}\setlength\topsep{0pt}
\begin{center}
Video\\System
\end{center}
\end{minipage}};
\draw (132.0,85.65) node  [font=\scriptsize] [align=left] {\begin{minipage}[lt]{18.59pt}\setlength\topsep{0pt}
\begin{center}
Game\\System
\end{center}
\end{minipage}};
\draw (229.5,11.46) node  [font=\scriptsize] [align=left] {\begin{minipage}[lt]{19.66pt}\setlength\topsep{0pt}
\begin{center}
Camera
\end{center}
\end{minipage}};
\draw (221.34,118.52) node  [font=\scriptsize] [align=left] {Eye Tracker};
\draw (287.91,121.83) node  [font=\scriptsize] [align=left] {Controller};
\draw (312.49,104.79) node  [font=\scriptsize] [align=left] {Subject};
\draw (63.65,19.86) node  [font=\scriptsize] [align=left] {\begin{minipage}[lt]{17.28pt}\setlength\topsep{0pt}
\begin{center}
Biopac
\end{center}
\end{minipage}};
\draw (231.19,28.67) node  [font=\scriptsize] [align=left] {capture analog biosignals};
\draw (170.63,73.01) node  [font=\scriptsize] [align=left] {displays};
\draw (158.55,125.71) node  [font=\scriptsize] [align=left] {data\\transfer};
\draw (92.67,4.8) node  [font=\scriptsize] [align=left] {data transfer};
\draw (130.07,19.27) node  [font=\scriptsize,rotate=-351.68] [align=left] {triggers};
\draw (178.29,100.81) node  [font=\scriptsize] [align=left] {controls};
\draw (227.56,68.87) node  [font=\scriptsize] [align=left] {\begin{minipage}[lt]{18.59pt}\setlength\topsep{0pt}
\begin{center}
Monitor
\end{center}
\end{minipage}};
\draw (92.88,56.01) node  [font=\scriptsize,rotate=-22.78] [align=left] {sync.};
\end{tikzpicture}
  \caption{System components used during data collection and their communication interfaces.}
\label{app:system-overview}
  \vspace{-0.15cm}
\end{figure}

\subsection{Sample Composition}
The corpus comprises 96 sessions from 83 healthy participants, colleagues and students of our institute (37
female, 45 male, one not reported, aged 18--59 years, mean $27\,\pm\,8$ years at first participation). The driving study
contributed 51 sessions from 51 participants and the gaming study 45 sessions from 37 participants. Five participants
took part in both studies and eight completed the gaming protocol twice. Table~\ref{tab:domain-composition} lists the
sessions and participants per evaluation domain.
\begin{table}[!htb]
  \centering
  \setlength{\tabcolsep}{4pt}
  \tablefont
  \caption{Sessions and participants per evaluation domain. Participants who contributed more than one session
  are counted once.}
  \label{tab:domain-composition}
  \begin{tabularx}{\linewidth}{Xlcc}
    \toprule
    Domain & Cohort & Sessions & Participants \\
    \midrule
    $n$-back (Driving Study) & driving & 51 & 51 \\
    $n$-back Dual Task & driving & 51 & 51 \\
    $n$-back (Gaming Study) & gaming & 44 & 37 \\
    $n$-back (Combined) & both & 95 & 83 \\
    Driving Application & driving & 51 & 51 \\
    \emph{Hogwarts Legacy} & gaming & 44 & 37 \\
    \emph{Overcooked! 2} & gaming & 44 & 36 \\
    Gaming & gaming & 45 & 37 \\
    Combined Applications & both & 96 & 83 \\
    Complete & both & 96 & 83 \\
    \bottomrule
  \end{tabularx}
  \begin{minipage}{\linewidth}
    \vspace{2pt}\footnotesize\textit{Note.} Each gaming application lacks one session. One
    session has no labeled $n$-back windows, one session is absent from \emph{Hogwarts Legacy}, and one session is
    absent from \emph{Overcooked! 2}. The latter is the only session of its participant, hence 36 participants for
    \emph{Overcooked! 2}.
  \end{minipage}
\end{table}

\subsection{Demographics and Personalities}
\begin{figure}[htb!]
  \centering
  \subfloat[\centering \gls{bmi}]{\includegraphics[width=0.5\linewidth]{%
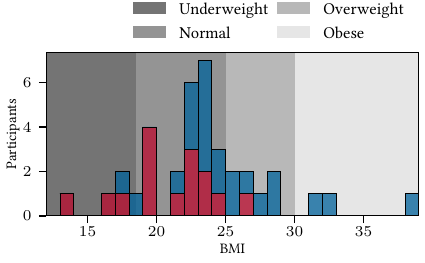}} \hfill%
  \subfloat[\centering Age Distribution]{\includegraphics[width=0.5\linewidth]{%
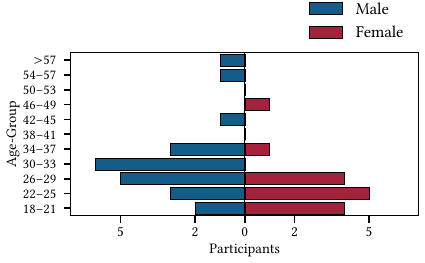}} %
  \caption{Distribution of \gls{bmi} and age of the 37 gaming-study participants, each counted once.}
\label{fig:demographics}
\end{figure}

In addition to the demographic information reported in the main article, we present participants' \gls{bmi} and age
distributions in Fig.~\ref{fig:demographics}. The narrow age and \gls{bmi} range indicates limited diversity as a
dataset limitation.
\begin{figure*}[!t]
  \centering
  \subfloat[\centering Game Experience Questionnaire Core\label{sfig:core-game-experience-questionnaire}]{%
\begin{tabular}[t]{@{}c@{}}
\includegraphics[width=0.33\linewidth]{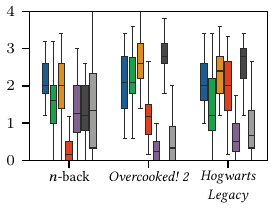} \\
\includegraphics[width=0.33\linewidth]{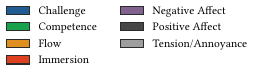}
\end{tabular}}
  \subfloat[\centering Post Game Experience Questionnaire\label{sfig:post-game-experience-questionnaire}]{%
\begin{tabular}[t]{@{}c@{}}
\includegraphics[width=0.33\linewidth]{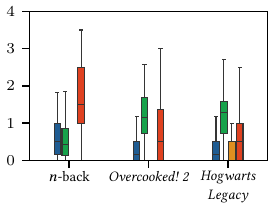} \\
\includegraphics[width=0.33\linewidth]{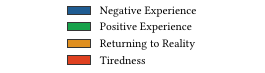}
\end{tabular}}
  \subfloat[\centering Intrinsic Motivation Inventory]{%
\begin{tabular}[t]{@{}c@{}}
\includegraphics[width=0.33\linewidth]{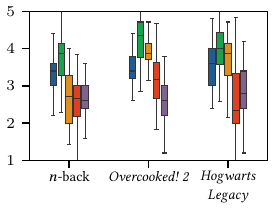} \\
\includegraphics[width=0.33\linewidth]{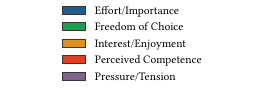}
\end{tabular}}
  \caption{Game Experience and Intrinsic Motivation Inventory.}
\label{fig:game-experience-questionnaire}
\end{figure*}
Participants' task-related experiences were assessed using the Game Experience Questionnaire 
\cite{jsselsteijnGameExperienceQuestionnaire2013}, with Core and Post scores shown in
Fig.~\ref{sfig:core-game-experience-questionnaire} and Fig.~\ref{sfig:post-game-experience-questionnaire},
respectively. Negative affect and low immersion were most pronounced during the $n$-back task, while both
\emph{Overcooked! 2} and \emph{Hogwarts Legacy} elicited predominantly positive affect. The Intrinsic
Motivation Inventory \cite{plantIntrinsicMotivationEffects1985} supports this pattern, showing lower
interest/enjoyment during $n$-back compared to the gaming tasks. These findings suggest motivational and affective
differences between standardized cognitive testing and gaming that warrant future investigation.
\subsection{Subjective Measures}
\begin{figure*}[!t]
  \centering
  \subfloat[\centering \ac{ocean} scores for all recorded subjects.\label{sfig:ocean_mean}]{%
\includegraphics[width=0.33\linewidth]{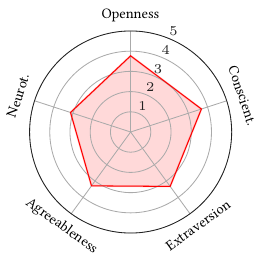}}%
  \subfloat[\centering Distribution of subjects with low and high neuroticism.\label{sfig:top-5-neuroticism}]{%
\includegraphics[width=0.33\linewidth]{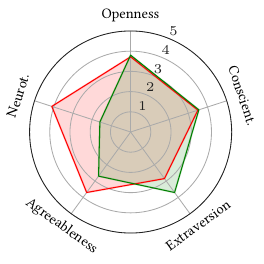}}%
  \subfloat[\centering Distributions of subjects with low and high agreeableness.\label{sfig:top-5-agreeableness}]{%
\includegraphics[width=0.33\linewidth]{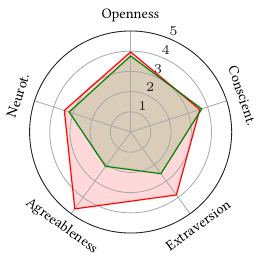}}%
  \caption{\ac{ocean} personality trait distributions of the study participants and selected subgroups.}
\label{fig:ocean_traits}
  \vspace{-0.15cm}
\end{figure*}

As noted in Section~\ref{ssec:target_operationalization} of the main article, we administered the \gls{bfik}
\cite{rammstedtKurzversionBigFive2005} to assess personality traits (\gls{ocean}) and examine their potential
influence on self-reported cognitive load. Mean \gls{ocean} scores are shown in Fig.~\ref{sfig:ocean_mean},
indicating moderate levels across all dimensions with slightly lower neuroticism and higher openness.
Figures~\ref{sfig:top-5-neuroticism} and~\ref{sfig:top-5-agreeableness} present the five participants with the
highest and lowest neuroticism and agreeableness, respectively, illustrating the personality distribution within
our sample and enabling subgroup analyses of character-based effects on subjective assessments.
\subsubsection{Correlation Between Personality Traits and Subjective Ratings}
\begin{figure}[!htb]
  \centering
  \subfloat[\centering \textbf{O}penness]{\includegraphics[width=0.48\linewidth]{%
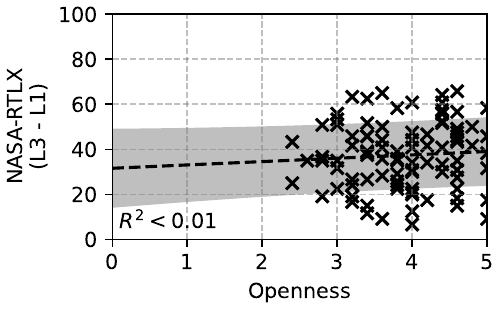}}%
  \hfill
  \subfloat[\centering \textbf{C}onscientiousness]{\includegraphics[width=0.48\linewidth]{%
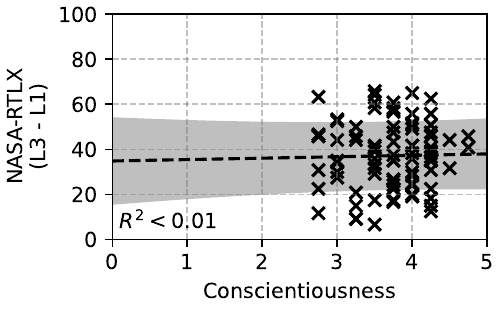}}%
  \\[4pt]
  \subfloat[\centering \textbf{E}xtraversion]{\includegraphics[width=0.48\linewidth]{%
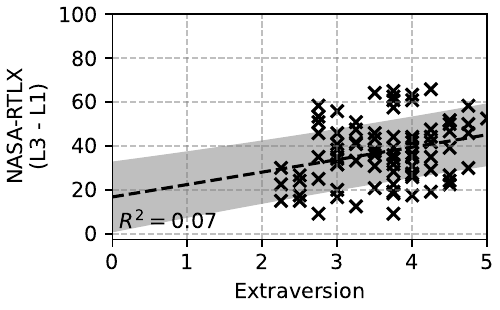}}%
  \hfill
  \subfloat[\centering \textbf{A}greeableness]{\includegraphics[width=0.48\linewidth]{%
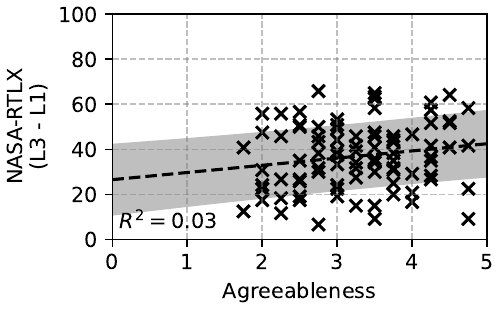}}%
  \\[4pt]
  \subfloat[\centering \textbf{N}euroticism]{\includegraphics[width=0.48\linewidth]{%
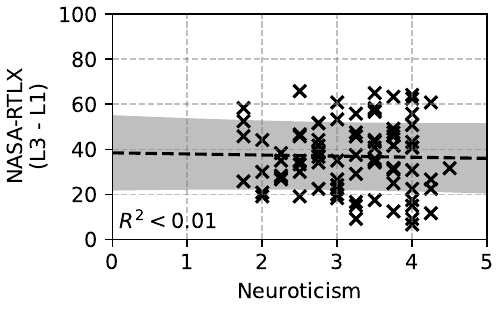}}%
  \caption{Correlation between \gls{ocean} traits and the difference in \gls{rtlx} between maximum and minimum
  $n$-back levels. No substantial association is observed.}
  \label{app:nasa-rtlx-ocean}
\end{figure}

We computed the correlation between the change in \gls{rtlx} scores from the lowest to the highest $n$-back level
and each \gls{ocean} trait. Four of the five traits showed negligible associations ($R^2 < 0.05$): Openness
$R^2 = 0.001$, Conscientiousness $R^2 = 0.001$, Agreeableness $R^2 = 0.041$, Neuroticism $R^2 = 0.002$. Only
Extraversion marginally exceeded this threshold ($R^2 = 0.083$). These results, visualized in
Figure~\ref{app:nasa-rtlx-ocean}, indicate that increases in self-reported cognitive load are not substantially
associated with individual personality traits in our sample.
\subsubsection{Weighted NASA-TLX Results}
\begin{figure}[!htb]
  \centering
  \subfloat[\centering $n$-back single]{\includegraphics[width=0.48\linewidth]{%
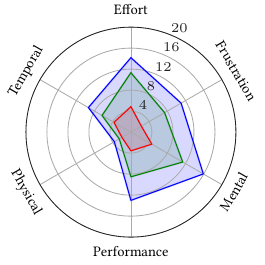}}%
  \hfill
  \subfloat[\centering $n$-back dual]{\includegraphics[width=0.48\linewidth]{%
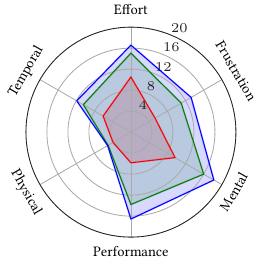}}%
  \\[4pt]
  \subfloat[\centering Driving]{\includegraphics[width=0.48\linewidth]{%
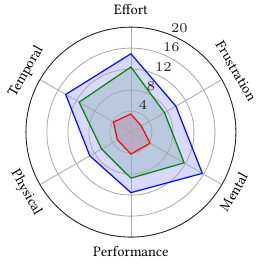}}%
  \caption{NASA-\gls{rtlx} dimension scores from \ac{adabase} for single-task $n$-back (visual stimuli),
  dual-task $n$-back (visual and auditory stimuli), and driving, with increasing task load levels
  (red: level~1, green: level~2, blue: level~3).}
  \label{fig:nasa-rtlx-adabase}
\end{figure}

\begin{figure}[!htb]
  \centering
  \subfloat[\centering \gls{adabase} $n$-back (visual)]{\includegraphics[width=0.48\linewidth]{%
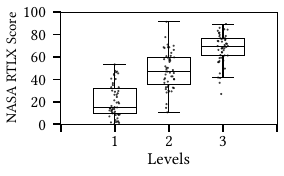}}%
  \hfill
  \subfloat[\centering \gls{adabase} $n$-back (dual)]{\includegraphics[width=0.48\linewidth]{%
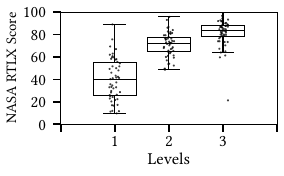}}%
  \\[4pt]
  \subfloat[\centering \gls{adabase} driving]{\includegraphics[width=0.48\linewidth]{%
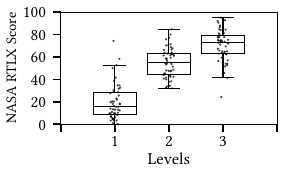}}%
  \caption{Weighted NASA-\ac{tlx} scores for the \gls{adabase} single-task and dual-task $n$-back and the
  driving application.}
  \label{app:nasa-tlx}
\end{figure}

\begin{figure}[!htb]
  \centering
  \subfloat[\centering $n$-back]{\includegraphics[width=0.48\linewidth]{%
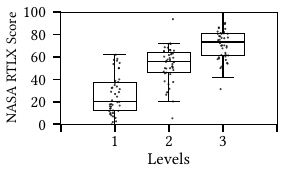}}%
  \hfill
  \subfloat[\centering \emph{Hogwarts Legacy}]{\includegraphics[width=0.48\linewidth]{%
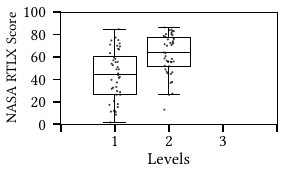}}%
  \\[4pt]
  \subfloat[\centering \emph{Overcooked! 2}]{\includegraphics[width=0.48\linewidth]{%
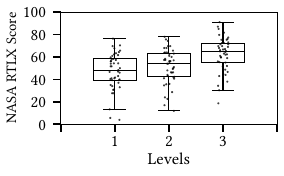}}%
  \caption{Weighted NASA-\gls{tlx} for $n$-back, \emph{Hogwarts Legacy}, and \emph{Overcooked! 2} across
  increasing task load levels, supplementing the \gls{rtlx} values reported in the main article.}
  \label{app:fig:nasa-tlx}
\end{figure}

The main article reports raw task load values as recommended by \cite{georgssonNASARTLXNovel2020}. For
comparison with studies employing weighted task load measures, we provide the corresponding weighted NASA-TLX
results in Figures~\ref{app:nasa-tlx} and~\ref{app:fig:nasa-tlx}. As reference,
Figure~\ref{fig:nasa-rtlx-adabase} shows performance metrics from the \gls{adabase} publication for participants
who completed the questionnaire following the single-task, dual-task $n$-back, and driving tests. The single-task
$n$-back results are consistent with those observed in our experiment.
For $n$-back, repeated-measures ANOVA revealed a significant main effect of level on NASA-TLX scores,
$F(2, 72) = 198.63$, $p < 0.001$, $\eta^2_G = 0.622$, $\varepsilon_{GG} = 0.78$, $p_{GG} < 0.001$. Post-hoc tests (Holm-corrected) confirmed all
pairwise differences: level~1 $<$ level~2 ($p < 0.001$), level~1 $<$ level~3 ($p < 0.001$), level~2 $<$ level~3
($p < 0.001$). For \emph{Overcooked! 2}, $F(2, 72) = 16.09$, $p < 0.001$, $\eta^2_G = 0.126$, $\varepsilon_{GG} = 0.77$, $p_{GG} < 0.001$, with
post-hoc: level~1 $<$ level~2 ($p = 0.072$), level~1 $<$ level~3 ($p < 0.001$), level~2 $<$ level~3
($p < 0.001$). For \emph{Hogwarts Legacy}, $F(1, 36) = 102.79$, $p < 0.001$, $\eta^2_G = 0.187$, with
level~1 $<$ level~2 ($p < 0.001$).

\subsubsection{Subjective Skill Level and Task Performance}
\label{app:ssec:skill-performance}
\begin{figure}[!htb]
  \centering
  \subfloat[\centering \emph{Overcooked! 2} Level 2]{\includegraphics[width=0.48\linewidth]{%
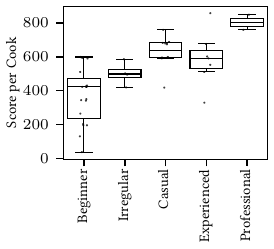}}%
  \hfill
  \subfloat[\centering \emph{Overcooked! 2} Level 3]{\includegraphics[width=0.48\linewidth]{%
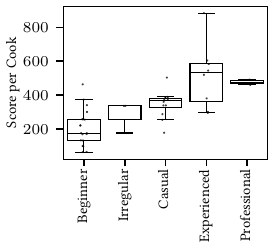}}%
  \\[4pt]
  \subfloat[\centering \emph{Hogwarts Legacy} Level 2]{\includegraphics[width=0.48\linewidth]{%
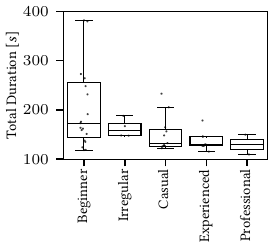}}%
  \hfill
  \subfloat[\centering $n$-back Level 2]{\includegraphics[width=0.48\linewidth]{%
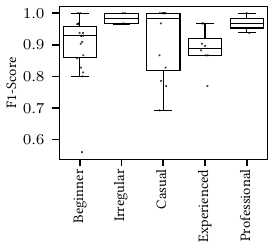}}%
  \\[4pt]
  \subfloat[\centering $n$-back Level 3]{\includegraphics[width=0.48\linewidth]{%
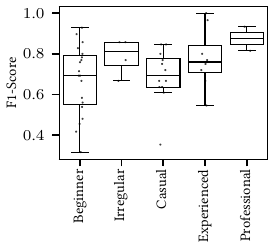}}%
  \caption{Subjective skill level vs.\ task performance.}
  \label{app:performance-vs-skill}
\end{figure}

Figure~\ref{app:performance-vs-skill} shows the relationship between self-reported skill level and task
performance. Kruskal-Wallis tests revealed significant differences for \emph{Overcooked! 2} Level~2,
$H(4) = 21.79$, $p < 0.001$, and Level~3, $H(4) = 22.88$, $p < 0.001$. No significant differences were found
for \emph{Hogwarts Legacy} Level~2, $H(4) = 9.09$, $p = 0.059$, $n$-back Level~2, $H(4) = 6.77$, $p = 0.149$,
or $n$-back Level~3, $H(4) = 6.25$, $p = 0.181$. These results suggest that prior gaming experience influenced
performance in \emph{Overcooked! 2} but not in the other tasks.

Task familiarity was operationalized separately per domain. In the driving study, 50 of 51 participants held a
driver's license, for $8.0\,\pm\,6.0$ years (range 0.4--22, $n = 49$, one license holder did not report a duration).
For the $n$-back, 8 of the 45 gaming sessions came from five participants who had completed the paradigm in the driving
study (more than a year between the two studies). All participants repeated a practice block before each level. For the
gaming tasks, 18 of the 45 sessions came from participants who reported playing (almost) never and therefore skipped
the genre questionnaire. The remaining 27 ranked eight genres by preference rather than by proficiency. Median ranks
were 2 for role-playing, 3 for action and first-person-shooter, 3 for simulation, and 4 for puzzle games. The closest
genres to our titles are role-playing for \emph{Hogwarts Legacy} and action for \emph{Overcooked! 2}.

\section{Prediction-Label Monotonicity}
\label{sec:supp-prediction-label-monotonicity}
\begin{figure}[!htb]
  \centering
  \subfloat[\centering Game $n$-back]{\includegraphics[width=0.49\linewidth]{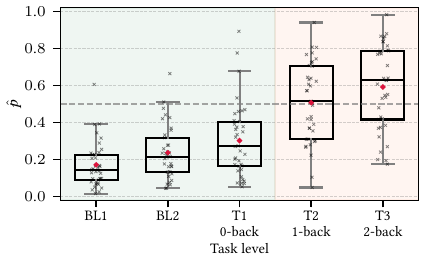}}\hfill%
  \subfloat[\centering Drive $n$-back]{\includegraphics[width=0.49\linewidth]{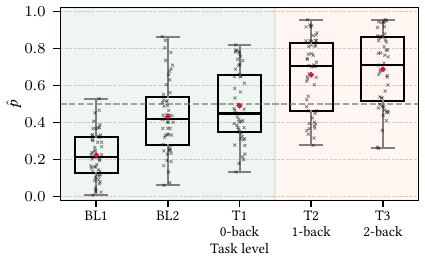}}\\
  \subfloat[\centering Combined $n$-back]{\includegraphics[width=0.49\linewidth]{./figures/monotonicity-combined-n-back.pdf}}\hfill%
  \subfloat[\centering \emph{Overcooked! 2}]{\includegraphics[width=0.49\linewidth]{./figures/monotonicity-game-s-cook.pdf}}\\
  \subfloat[\centering \emph{Hogwarts Legacy}]{\includegraphics[width=0.49\linewidth]{./figures/monotonicity-game-u-harry.pdf}}%
  \caption{Per-subject predicted probability $\hat{p}$ across ordered task levels (ConvNeXt-small trained on the $n$-back or the gaming data). Each violin shows the across-subject distribution at one level.}
  \label{fig:supp-pred-label-monotonicity}
\end{figure}

This section provides the extended per-subject statistics summarized in Section~\ref{ssec:label-monotonicity} of the
main article. For each subject we computed the mean predicted probability $\hat{p}$ per level, the Spearman rank
correlation between $\hat{p}$ and the ordered levels, and a Page's $L$ trend test across the per-level means. Predictions were averaged over the windows of a session and then over the sessions of a participant.
Outer monotonicity is defined as the mean $\hat{p}$ over the low-difficulty levels lying below the mean
$\hat{p}$ over the high-difficulty levels. Figure~\ref{fig:supp-pred-label-monotonicity} shows the per-subject
distributions and Table~\ref{tab:supp-label-monotonicity} reports group-level summary statistics.
\begin{table}[!htb]
  \centering
  \setlength{\tabcolsep}{4pt}
  \tablefont
  \caption{Per-subject monotonicity of predicted task load $\hat p$ for the ConvNeXt-small models trained on the $n$-back and on the gaming data, evaluated across all ordered levels without retraining. \% positive Spearman is the share of subjects with a positive rank correlation between level and $\hat p$; \% outer-monotone the share whose mean $\hat p$ over the lowest levels is below the mean over the highest levels. Complements the group-mean summary in Table~\ref{tab:label-monotonicity}.}
  \label{tab:supp-label-monotonicity}
  \begin{tabularx}{\linewidth}{Xccc}
    \toprule
    Group & $n$ & \% positive $\rho$ & \% outer mono. \\
    \midrule
    Game $n$-back & 37 & 86 & 95 \\
    Drive $n$-back & 51 & 94 & 96 \\
    Combined $n$-back & 83 & 90 & 95 \\
    \emph{Overcooked! 2} & 35 & 97 & 97 \\
    \emph{Hogwarts Legacy} & 36 & 92 & 97 \\
    \bottomrule
  \end{tabularx}
  \begin{minipage}{\linewidth}
    \vspace{2pt}\footnotesize\textit{Note.} Shares are computed over the same per-subject $\hat p$ arrays as Table~\ref{tab:label-monotonicity} (modality \texttt{adabase-bio-eye}, deploy-run predictions, no retraining). The low/high blocks follow each ladder: ranks 1--3 vs.\ 4--5 for the $n$-back groups and ranks 1--2 vs.\ 3--4 for \emph{Overcooked! 2} and \emph{Hogwarts Legacy}. Percentages are rounded to whole percent.
  \end{minipage}
\end{table}

\begin{figure}[!htb]
  \centering
  \subfloat[\centering Game $n$-back]{\includegraphics[width=0.49\linewidth]{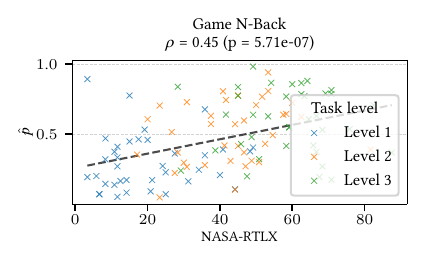}}\hfill%
  \subfloat[\centering \emph{Overcooked! 2}]{\includegraphics[width=0.49\linewidth]{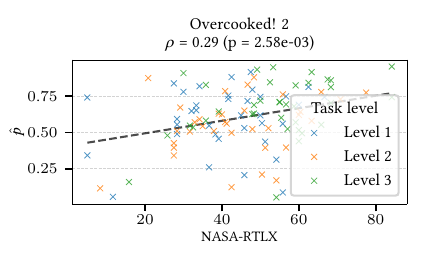}}%
  \caption{Predicted task-load probability $\hat{p}$ versus NASA-\gls{rtlx} score (test levels T1--T3 only).
  Spearman correlations: $\rho = 0.454$, $p = 5.71\times 10^{-7}$, $n = 37$ participants for $n$-back and
  $\rho = 0.288$, $p = 2.58\times 10^{-3}$, $n = 36$ participants for \emph{Overcooked! 2}.}
  \label{fig:supp-pred-label-rtlx-scatter}
\end{figure}

Page's $L$ trend tests reject the null of no monotonic association at high confidence across all five groups:
Game $n$-back $L = 1901$, $p = 4.26\times 10^{-15}$, mean $\rho = .64$;
Drive $n$-back $L = 2690$, $p = 9.57\times 10^{-29}$, $\rho = .77$;
Combined $n$-back $L = 4324$, $p = 1.52\times 10^{-38}$, $\rho = .71$;
\emph{Overcooked! 2} $L = 999$, $p = 1.93\times 10^{-13}$, $\rho = .71$;
\emph{Hogwarts Legacy} $L = 1039$, $p = 5.07\times 10^{-16}$, $\rho = .77$.
The proportion of subjects with positive Spearman correlation is 86\,\%, 94\,\%, 90\,\%, 97\,\%, and 92\,\% for the five groups, respectively. In \emph{Overcooked! 2} the Easy to Medium adjacent contrast
is non-significant ($W = 211$, $p = 0.956$ one-sided), consistent with comparable NASA-\gls{rtlx} ratings
for those two levels as discussed in Section~\ref{ssec:label-monotonicity}.
Figure~\ref{fig:supp-pred-label-rtlx-scatter} further relates the predicted task-load probability $\hat{p}$ to the
subjective NASA-\gls{rtlx} score on the test levels, showing a significant positive association for both $n$-back and
\emph{Overcooked! 2}.
\subsection{Participant-Level Mixed-Effects Analysis}
\label{ssec:supp-mixed-effects}
To separate the association of $\hat{p}$ with NASA-\gls{rtlx} within participants from differences between
participants, we fitted linear mixed models on the $n$-back. It is the paradigm shared by both cohorts and the
paradigm of the convergent validity analysis, with three to six rated and scored test levels per participant.
NASA-\gls{rtlx} was split into the participant mean and the deviation from it.
\begin{table*}[!t]
  \centering
  \setlength{\tabcolsep}{3pt}
  \tablefont
  \caption{Participant-level association of the predicted high-load probability $\hat{p}$ (logit) of the ConvNeXt-small models trained on the $n$-back data with NASA-\gls{rtlx} and visual task performance. Standardized within-participant coefficients of linear mixed models with 95\,\% participant cluster-bootstrap intervals. M1: NASA-\gls{rtlx} within and between participants. M2: M1 with reaction time and F1-score. M3: NASA-\gls{rtlx} with the difficulty level as fixed effect.}
  \label{tab:prediction-mixed-effects}
  \resizebox{\linewidth}{!}{%
  \begin{tabular}{lcccccc}
    \toprule
    Cohort & $n$ & M1 NASA-\gls{rtlx} & M3 NASA-\gls{rtlx} & M2 NASA-\gls{rtlx} & M2 Reaction time & M2 F1-score \\
    \midrule
    Drive $n$-back & 51 (306) & $0.28$ $[0.20,\,0.37]$\textsuperscript{***} & $0.04$ $[{-}0.11,\,0.18]$\textsuperscript{} & $0.27$ $[0.13,\,0.41]$\textsuperscript{***} & $0.01$ $[{-}0.09,\,0.12]$\textsuperscript{} & $0.00$ $[{-}0.09,\,0.09]$\textsuperscript{} \\
    Game $n$-back & 37 (111) & $0.55$ $[0.40,\,0.73]$\textsuperscript{***} & $0.17$ $[{-}0.15,\,0.53]$\textsuperscript{} & $0.46$ $[0.24,\,0.70]$\textsuperscript{***} & $0.22$ $[{-}0.03,\,0.46]$\textsuperscript{} & $0.09$ $[{-}0.10,\,0.30]$\textsuperscript{} \\
    \bottomrule
  \end{tabular}}
  \begin{minipage}{\linewidth}
    \vspace{2pt}\footnotesize\textit{Note.} $n$: participants (participant-levels). Random intercept per participant, maximum likelihood. Predictions averaged over the windows of a session and the sessions of a participant. Intervals from 10{{,}}000 participant cluster-bootstrap draws. $p$ from likelihood-ratio tests, Holm-corrected across the two cohorts: \textsuperscript{*}$p<.05$, \textsuperscript{**}$p<.01$, \textsuperscript{***}$p<.001$.
  \end{minipage}
\end{table*}

Within participants, $\hat{p}$ followed NASA-\gls{rtlx} in both cohorts. This held when reaction time and
F1-score were added, and neither of them was significant (Table~\ref{tab:prediction-mixed-effects}). With the
difficulty level as a fixed effect, the association was no longer significant. Thus $\hat{p}$ tracks the induced
difficulty but shows no evidence of tracking experienced load beyond it. All nine networks gave consistent
coefficients.
\section{Held-Out Monotonicity}
\label{sec:supp-held-out-monotonicity}
\begin{figure}[!htb]
  \centering
  \subfloat[\centering Combined $n$-back]{\includegraphics[width=0.48\linewidth]{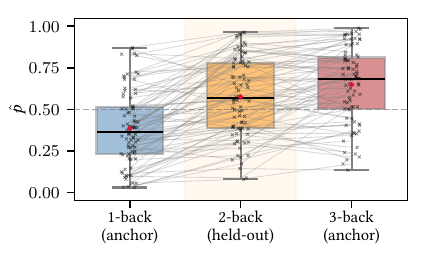}}%
  \hfill
  \subfloat[\centering \emph{Overcooked! 2}]{\includegraphics[width=0.48\linewidth]{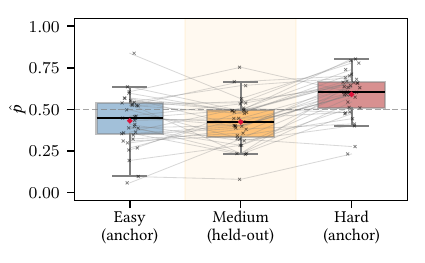}}%
  \\[4pt]
  \subfloat[\centering \emph{Hogwarts Legacy}]{\includegraphics[width=0.48\linewidth]{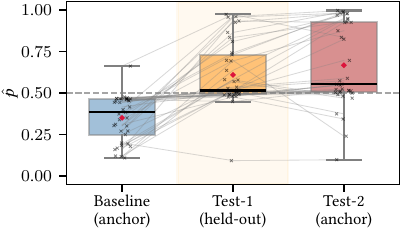}}%
  \caption{Predicted probability $\hat{p}$ across three levels (low anchor, held-out middle, high anchor)
  for ConvNeXt-small. Page's $L$: Combined $n$-back $p=5.0\times 10^{-17}$, $\rho=0.64$;
  \emph{Overcooked! 2} $p=2.6\times 10^{-7}$, $\rho=0.60$;
  \emph{Hogwarts Legacy} $p=1.7\times 10^{-11}$, $\rho=0.77$.}
  \label{fig:supp-held-out-monotonicity}
\end{figure}

Section~\ref{ssec:heldout-monotonicity} of the main article reports held-out monotonicity results per architecture in
Table~\ref{tab:heldout-monotonicity}. Here we provide the extended protocol details and per-architecture visualizations.
Models were retrained using only the two outer anchor levels, and the predicted probability for the held-out middle
level was inspected for interpolation between anchors. The protocol was run for ConvNeXt-small, xLSTM-small, and
Transformer-small across five task groups (Game $n$-back, Drive $n$-back, Combined $n$-back, \emph{Overcooked! 2},
\emph{Hogwarts Legacy}), yielding 75/75 completed training runs. Group-level monotonicity is declared if
$\hat{p}_\text{anchor low} < \hat{p}_\text{held-out} < \hat{p}_\text{anchor high}$ together with significant
one-sided Wilcoxon tests for both adjacent contrasts at the Bonferroni-corrected level ($\alpha = 0.025$).
Figure~\ref{fig:supp-held-out-monotonicity} visualizes the resulting per-level predicted probabilities for the
ConvNeXt-small encoder across the representative task groups.
Across the 15 (group $\times$ architecture) cells, group-level monotonicity holds for $n$-back and
\emph{Hogwarts Legacy} with all three architectures (12/12) and fails for \emph{Overcooked! 2} with all three (0/3).
For \emph{Overcooked! 2}, the held-out Medium level is statistically indistinguishable from the Easy anchor
($p=0.599$, $0.845$, $0.765$ for ConvNeXt-, xLSTM-, and Transformer-small, respectively). Page's $L$ trend tests
reject the null in all 15 cells. The subject-level fully monotone rate ranges from 23\,\% (\emph{Overcooked! 2},
Transformer-small) to 76\,\% (\emph{Hogwarts Legacy}, Transformer-small). These per-architecture breakdowns confirm
that the patterns reported in the main article are consistent across model families and not architecture-specific.

\section{Additional Findings}
\subsection{Multimodal Fusion vs.\ the Best Single Modality}
Table~\ref{tab:supp-multimodal-vs-unimodal} compares the multimodal model (biosignals and eye tracking) with
the best single modality of each domain, separately for each encoder. Each of the 24 domain and encoder pairs is
tested at the participant level (cluster bootstrap and two-sided sign-flip test, 10{,}000 draws each, Holm-corrected
across the 24 pairs), so folds and encoders are not treated as observations. The multimodal model is ahead in 16 of
the 24 pairs (mean $\Delta\mathrm{AUROC} = +0.014$) and significantly ahead in four, namely the complete dataset with
all three encoders ($+0.057$ to $+0.092$, $p_{\mathrm{Holm}} = .002$) and the driving application with the
x\gls{lstm} ($+0.067$, $p_{\mathrm{Holm}} = .048$). In these two domains eye movement is the best single input. In the
$n$-back and gaming domains, pupil diameter alone already achieves strong discrimination (\gls{auroc} 0.75--0.86), so
fusion adds little. The eight pairs in which fusion trails (the x\gls{lstm} on the combined $n$-back, the x\gls{lstm} and Transformer on the
driving-study and on the gaming-study $n$-back, and all three encoders on \emph{Overcooked! 2}, $-0.002$ to $-0.037$) are
not significant. Selecting the best single modality on the test data favours it, so the
comparison is conservative for fusion. These results complement the per-modality breakdown in
Table~\ref{tab:modality-multimodal-vs-unimodal} of the main article.
\begin{table*}[!t]
  \centering
  \setlength{\tabcolsep}{4pt}
  \tablefont
  \caption{Multimodal fusion (biosignals $+$ eye tracking) against the best single modality per domain and encoder (test \gls{auroc}). A positive $\Delta$ favours fusion. Brackets give participant-clustered 95\,\% confidence intervals.}
  \label{tab:supp-multimodal-vs-unimodal}
  \begin{tabularx}{\linewidth}{Xllcccc}
    \toprule
    Domain & Encoder & Best single & Fusion & Single & $\Delta$ & $p_{\mathrm{Holm}}$ \\
    \midrule
    Complete & x\gls{lstm} & Eye Mv. & 0.822 & 0.765 & ${+}0.057$ $[{+}0.03,\,{+}0.08]$ & $.002$ \\
     & ConvNeXt & Eye Mv. & 0.825 & 0.733 & ${+}0.092$ $[{+}0.07,\,{+}0.12]$ & $.002$ \\
     & Transformer & Eye Mv. & 0.833 & 0.776 & ${+}0.057$ $[{+}0.03,\,{+}0.08]$ & $.002$ \\
    \addlinespace
    $n$-back (Combined) & x\gls{lstm} & Pupil & 0.836 & 0.841 & ${-}0.005$ $[{-}0.03,\,{+}0.02]$ & ${>}.999$ \\
     & ConvNeXt & Pupil & 0.831 & 0.804 & ${+}0.027$ $[{+}0.01,\,{+}0.05]$ & $.801$ \\
     & Transformer & Pupil & 0.838 & 0.829 & ${+}0.009$ $[{-}0.02,\,{+}0.03]$ & ${>}.999$ \\
    \addlinespace
    $n$-back (Driving) & x\gls{lstm} & Pupil & 0.836 & 0.857 & ${-}0.021$ $[{-}0.06,\,{+}0.03]$ & ${>}.999$ \\
     & ConvNeXt & Pupil & 0.780 & 0.765 & ${+}0.014$ $[{-}0.04,\,{+}0.08]$ & ${>}.999$ \\
     & Transformer & Pupil & 0.821 & 0.851 & ${-}0.029$ $[{-}0.06,\,{+}0.02]$ & ${>}.999$ \\
    \addlinespace
    $n$-back (Gaming) & x\gls{lstm} & Pupil & 0.792 & 0.830 & ${-}0.037$ $[{-}0.08,\,{-}0.00]$ & ${>}.999$ \\
     & ConvNeXt & Pupil & 0.845 & 0.829 & ${+}0.016$ $[{-}0.02,\,{+}0.05]$ & ${>}.999$ \\
     & Transformer & Pupil & 0.798 & 0.818 & ${-}0.020$ $[{-}0.08,\,{+}0.05]$ & ${>}.999$ \\
    \addlinespace
    Driving Application & x\gls{lstm} & Eye Mv. & 0.868 & 0.801 & ${+}0.067$ $[{+}0.03,\,{+}0.10]$ & $.048$ \\
     & ConvNeXt & Eye Mv. & 0.851 & 0.830 & ${+}0.021$ $[{-}0.01,\,{+}0.05]$ & ${>}.999$ \\
     & Transformer & Eye Mv. & 0.877 & 0.876 & ${+}0.001$ $[{-}0.03,\,{+}0.03]$ & ${>}.999$ \\
    \addlinespace
    Gaming & x\gls{lstm} & Pupil & 0.822 & 0.795 & ${+}0.028$ $[{+}0.01,\,{+}0.06]$ & ${>}.999$ \\
     & ConvNeXt & Pupil & 0.813 & 0.792 & ${+}0.021$ $[{-}0.00,\,{+}0.05]$ & ${>}.999$ \\
     & Transformer & Pupil & 0.818 & 0.792 & ${+}0.026$ $[{+}0.00,\,{+}0.06]$ & $.690$ \\
    \addlinespace
    \emph{Hogwarts L.} & x\gls{lstm} & Pupil & 0.854 & 0.853 & ${+}0.001$ $[{-}0.04,\,{+}0.04]$ & ${>}.999$ \\
     & ConvNeXt & Pupil & 0.896 & 0.851 & ${+}0.045$ $[{+}0.01,\,{+}0.08]$ & $.122$ \\
     & Transformer & Pupil & 0.865 & 0.854 & ${+}0.011$ $[{-}0.02,\,{+}0.04]$ & ${>}.999$ \\
    \addlinespace
    \emph{Overcooked! 2} & x\gls{lstm} & Pupil & 0.734 & 0.764 & ${-}0.030$ $[{-}0.07,\,{+}0.01]$ & ${>}.999$ \\
     & ConvNeXt & Pupil & 0.774 & 0.776 & ${-}0.002$ $[{-}0.04,\,{+}0.03]$ & ${>}.999$ \\
     & Transformer & Pupil & 0.739 & 0.748 & ${-}0.009$ $[{-}0.05,\,{+}0.04]$ & ${>}.999$ \\
    \bottomrule
  \end{tabularx}
  \begin{minipage}{\linewidth}
    \vspace{2pt}\footnotesize\textit{Note.} \gls{auroc} per subject-wise fold, averaged over the five folds. The best single modality is the one with the highest test \gls{auroc} among the eight biosignal and eye-tracking inputs, which favours it. Intervals from a participant-level cluster bootstrap, $p$ from a two-sided participant-level sign-flip test (10{,}000 draws each), Holm-corrected across the 24 domain and encoder pairs.
  \end{minipage}
\end{table*}

\subsection{Cross-Domain Generalization Detail}
Table~\ref{tab:supp-generalization-detail} provides the full per-source breakdown referenced in
Section~\ref{ssec:task-load-generalization}. For each test domain $Y$ and each single-task training domain $X \neq Y$,
we compared complete-dataset training with $X$-only training on domain $Y$ (ConvNeXt-small, fold-mean
\gls{auroc}, participant-level cluster bootstrap and two-sided sign-flip test with 10{,}000 draws each,
Benjamini-Hochberg \gls{fdr}-corrected across all 42 source-target pairs). Twenty-two of the 42 pairs are significant.
Twenty favour complete-dataset training, mostly lab-to-application and cross-application pairs ($+0.26$ to
$+0.47$ on the driving application). Two favour the single source, both on the driving-study $n$-back: training on
the combined $n$-back is better by $0.057$ ($p_{\mathrm{BH}} = .005$) and training on the driving application by
$0.065$ ($p_{\mathrm{BH}} = .022$). Both sources contain $n$-back or application data of the driving cohort. Elsewhere,
complete-dataset training is never significantly worse than a single source.
\begin{table*}[!t]
  \centering
  \setlength{\tabcolsep}{4pt}
  \tablefont
  \caption{Per-source generalization detail (ConvNeXt-small, test \gls{auroc}). Each row compares complete-dataset training with one single-domain source $X\,\to Y$ on domain $Y$. A positive $\Delta$ favours complete-dataset training. Brackets give participant-clustered 95\,\% confidence intervals.}
  \label{tab:supp-generalization-detail}
  \begin{tabularx}{\linewidth}{XXcccc}
    \toprule
    Domain $Y$ & Single source $X$ & compl.\,${\to}\,Y$ & $X\,{\to}\,Y$ & $\Delta$ & $p_{\mathrm{BH}}$ \\
    \midrule
    $n$-back (Driving Study) & $n$-back (Gaming Study) & 0.713 & 0.730 & ${-}0.017$ $[{-}0.07,\,{+}0.04]$ & $.649$ \\
    $n$-back (Driving Study) & $n$-back (Combined) & 0.713 & 0.771 & ${-}0.057$ $[{-}0.09,\,{-}0.02]$ & $.005$ \\
    $n$-back (Driving Study) & Driving Application & 0.713 & 0.778 & ${-}0.065$ $[{-}0.10,\,{-}0.02]$ & $.022$ \\
    $n$-back (Driving Study) & \emph{Hogwarts Legacy} & 0.713 & 0.709 & ${+}0.005$ $[{-}0.04,\,{+}0.05]$ & $.840$ \\
    $n$-back (Driving Study) & \emph{Overcooked! 2} & 0.713 & 0.547 & ${+}0.167$ $[{+}0.12,\,{+}0.23]$ & ${<}.001$ \\
    $n$-back (Driving Study) & Gaming & 0.713 & 0.667 & ${+}0.046$ $[{-}0.01,\,{+}0.11]$ & $.124$ \\
    \addlinespace
    $n$-back (Gaming Study) & $n$-back (Driving Study) & 0.807 & 0.683 & ${+}0.124$ $[{+}0.07,\,{+}0.18]$ & ${<}.001$ \\
    $n$-back (Gaming Study) & $n$-back (Combined) & 0.807 & 0.797 & ${+}0.011$ $[{-}0.03,\,{+}0.06]$ & $.738$ \\
    $n$-back (Gaming Study) & Driving Application & 0.807 & 0.675 & ${+}0.132$ $[{+}0.08,\,{+}0.19]$ & ${<}.001$ \\
    $n$-back (Gaming Study) & \emph{Hogwarts Legacy} & 0.807 & 0.792 & ${+}0.016$ $[{-}0.02,\,{+}0.06]$ & $.606$ \\
    $n$-back (Gaming Study) & \emph{Overcooked! 2} & 0.807 & 0.767 & ${+}0.041$ $[{+}0.01,\,{+}0.08]$ & $.105$ \\
    $n$-back (Gaming Study) & Gaming & 0.807 & 0.800 & ${+}0.008$ $[{-}0.02,\,{+}0.04]$ & $.738$ \\
    \addlinespace
    $n$-back (Combined) & $n$-back (Driving Study) & 0.808 & 0.644 & ${+}0.164$ $[{+}0.12,\,{+}0.21]$ & ${<}.001$ \\
    $n$-back (Combined) & $n$-back (Gaming Study) & 0.808 & 0.786 & ${+}0.022$ $[{-}0.00,\,{+}0.04]$ & $.232$ \\
    $n$-back (Combined) & Driving Application & 0.808 & 0.716 & ${+}0.092$ $[{+}0.06,\,{+}0.13]$ & ${<}.001$ \\
    $n$-back (Combined) & \emph{Hogwarts Legacy} & 0.808 & 0.798 & ${+}0.010$ $[{-}0.01,\,{+}0.03]$ & $.459$ \\
    $n$-back (Combined) & \emph{Overcooked! 2} & 0.808 & 0.728 & ${+}0.080$ $[{+}0.06,\,{+}0.10]$ & ${<}.001$ \\
    $n$-back (Combined) & Gaming & 0.808 & 0.801 & ${+}0.008$ $[{-}0.01,\,{+}0.03]$ & $.606$ \\
    \addlinespace
    Driving Application & $n$-back (Driving Study) & 0.845 & 0.585 & ${+}0.260$ $[{+}0.18,\,{+}0.33]$ & ${<}.001$ \\
    Driving Application & $n$-back (Gaming Study) & 0.845 & 0.524 & ${+}0.321$ $[{+}0.25,\,{+}0.38]$ & ${<}.001$ \\
    Driving Application & $n$-back (Combined) & 0.845 & 0.379 & ${+}0.466$ $[{+}0.38,\,{+}0.54]$ & ${<}.001$ \\
    Driving Application & \emph{Hogwarts Legacy} & 0.845 & 0.519 & ${+}0.326$ $[{+}0.25,\,{+}0.40]$ & ${<}.001$ \\
    Driving Application & \emph{Overcooked! 2} & 0.845 & 0.435 & ${+}0.410$ $[{+}0.33,\,{+}0.48]$ & ${<}.001$ \\
    Driving Application & Gaming & 0.845 & 0.463 & ${+}0.382$ $[{+}0.30,\,{+}0.45]$ & ${<}.001$ \\
    \addlinespace
    \emph{Hogwarts Legacy} & $n$-back (Driving Study) & 0.863 & 0.604 & ${+}0.259$ $[{+}0.21,\,{+}0.31]$ & ${<}.001$ \\
    \emph{Hogwarts Legacy} & $n$-back (Gaming Study) & 0.863 & 0.849 & ${+}0.013$ $[{-}0.03,\,{+}0.05]$ & $.669$ \\
    \emph{Hogwarts Legacy} & $n$-back (Combined) & 0.863 & 0.837 & ${+}0.025$ $[{-}0.01,\,{+}0.06]$ & $.152$ \\
    \emph{Hogwarts Legacy} & Driving Application & 0.863 & 0.602 & ${+}0.261$ $[{+}0.21,\,{+}0.31]$ & ${<}.001$ \\
    \emph{Hogwarts Legacy} & \emph{Overcooked! 2} & 0.863 & 0.813 & ${+}0.050$ $[{-}0.01,\,{+}0.09]$ & $.227$ \\
    \emph{Hogwarts Legacy} & Gaming & 0.863 & 0.887 & ${-}0.024$ $[{-}0.05,\,{+}0.00]$ & $.198$ \\
    \addlinespace
    \emph{Overcooked! 2} & $n$-back (Driving Study) & 0.759 & 0.599 & ${+}0.159$ $[{+}0.09,\,{+}0.22]$ & ${<}.001$ \\
    \emph{Overcooked! 2} & $n$-back (Gaming Study) & 0.759 & 0.731 & ${+}0.028$ $[{-}0.01,\,{+}0.06]$ & $.198$ \\
    \emph{Overcooked! 2} & $n$-back (Combined) & 0.759 & 0.700 & ${+}0.059$ $[{+}0.02,\,{+}0.09]$ & $.001$ \\
    \emph{Overcooked! 2} & Driving Application & 0.759 & 0.577 & ${+}0.182$ $[{+}0.12,\,{+}0.23]$ & ${<}.001$ \\
    \emph{Overcooked! 2} & \emph{Hogwarts Legacy} & 0.759 & 0.752 & ${+}0.006$ $[{-}0.04,\,{+}0.05]$ & $.820$ \\
    \emph{Overcooked! 2} & Gaming & 0.759 & 0.763 & ${-}0.004$ $[{-}0.03,\,{+}0.01]$ & $.768$ \\
    \addlinespace
    Gaming & $n$-back (Driving Study) & 0.805 & 0.593 & ${+}0.211$ $[{+}0.16,\,{+}0.26]$ & ${<}.001$ \\
    Gaming & $n$-back (Gaming Study) & 0.805 & 0.769 & ${+}0.035$ $[{+}0.00,\,{+}0.06]$ & $.082$ \\
    Gaming & $n$-back (Combined) & 0.805 & 0.756 & ${+}0.049$ $[{+}0.02,\,{+}0.07]$ & ${<}.001$ \\
    Gaming & Driving Application & 0.805 & 0.592 & ${+}0.213$ $[{+}0.17,\,{+}0.25]$ & ${<}.001$ \\
    Gaming & \emph{Hogwarts Legacy} & 0.805 & 0.809 & ${-}0.005$ $[{-}0.03,\,{+}0.03]$ & $.820$ \\
    Gaming & \emph{Overcooked! 2} & 0.805 & 0.790 & ${+}0.015$ $[{-}0.01,\,{+}0.04]$ & $.488$ \\
    \bottomrule
  \end{tabularx}
  \begin{minipage}{\linewidth}
    \vspace{2pt}\footnotesize\textit{Note.} \gls{auroc} per subject-wise fold, averaged over the five folds. Intervals from a participant-level cluster bootstrap, $p$ from a two-sided participant-level sign-flip test (10{,}000 draws each). Benjamini--Hochberg \gls{fdr} correction across all $42$ source--target pairs, because they share the complete-dataset arm.
  \end{minipage}
\end{table*}

\subsection{Leave-One-Domain-Out: Source Diversity and Data Quantity}
\label{ssec:supp-lodo}
Table~\ref{tab:lodo-sources} breaks down the leave-one-domain-out analysis of Section~\ref{ssec:lodo} of the
main article by target and single source. For each target, the multi-source model was subsampled to the training and
validation windows per class of each single-source model. Diversity is this size-matched multi-source model minus the
single source. Quantity is the full minus the size-matched multi-source model. Both are paired \gls{auroc} differences
on the target, Holm-corrected within the 12 target-source pairs per contrast and architecture.
\begin{table}[!htb]
  \centering
  \setlength{\tabcolsep}{4pt}
  \tablefont
  \caption{Source diversity and data quantity under leave-one-domain-out. Diversity is the multi-source model subsampled to the windows per class of the single source minus that single source. Quantity is the full multi-source model minus the subsampled one. Paired \gls{auroc} differences on the target.}
  \label{tab:lodo-sources}
  \begin{tabularx}{\linewidth}{Xcccc}
    \toprule
    & \multicolumn{2}{c}{Diversity} & \multicolumn{2}{c}{Quantity} \\
    \cmidrule(lr){2-3}\cmidrule(lr){4-5}
    Source & Base & Small & Base & Small \\
    \midrule
    \multicolumn{5}{@{}l}{\emph{Target: $n$-back}} \\
    Driving Application & ${+}0.245$\textsuperscript{**} & ${+}0.250$\textsuperscript{**} & ${+}0.009$ & ${+}0.030$\textsuperscript{**} \\
    \emph{Overcooked! 2} & ${+}0.103$\textsuperscript{**} & ${+}0.113$\textsuperscript{**} & ${+}0.010$ & ${-}0.001$ \\
    \emph{Hogwarts Legacy} & ${+}0.011$ & ${-}0.002$ & ${+}0.005$ & ${+}0.011$ \\
    \addlinespace
    \multicolumn{5}{@{}l}{\emph{Target: Driving Application}} \\
    $n$-back & ${-}0.152$\textsuperscript{**} & ${-}0.167$\textsuperscript{**} & ${+}0.017$ & ${+}0.009$ \\
    \emph{Overcooked! 2} & ${-}0.052$ & ${-}0.022$ & ${+}0.005$ & ${-}0.002$ \\
    \emph{Hogwarts Legacy} & ${-}0.101$\textsuperscript{*} & ${-}0.109$\textsuperscript{*} & ${-}0.007$ & ${-}0.006$ \\
    \addlinespace
    \multicolumn{5}{@{}l}{\emph{Target: \emph{Overcooked! 2}}} \\
    $n$-back & ${+}0.040$ & ${+}0.051$ & ${-}0.004$ & ${-}0.013$ \\
    Driving Application & ${+}0.278$\textsuperscript{**} & ${+}0.322$\textsuperscript{**} & ${-}0.019$ & ${-}0.044$\textsuperscript{*} \\
    \emph{Hogwarts Legacy} & ${-}0.027$ & ${-}0.009$ & ${-}0.007$ & ${-}0.035$ \\
    \addlinespace
    \multicolumn{5}{@{}l}{\emph{Target: \emph{Hogwarts Legacy}}} \\
    $n$-back & ${+}0.044$ & ${+}0.021$ & ${+}0.014$ & ${+}0.033$ \\
    Driving Application & ${+}0.509$\textsuperscript{**} & ${+}0.513$\textsuperscript{**} & ${+}0.002$ & ${+}0.037$\textsuperscript{*} \\
    \emph{Overcooked! 2} & ${-}0.039$ & ${+}0.025$ & ${+}0.018$ & ${+}0.032$ \\
    \bottomrule
  \end{tabularx}
  \begin{minipage}{\linewidth}
    \vspace{2pt}\footnotesize\textit{Note.} Two-sided participant-level sign-flip test (10{,}000 draws), Holm-corrected within the 12 target-source pairs per contrast and capacity: \textsuperscript{*}$p<.05$, \textsuperscript{**}$p<.01$, \textsuperscript{***}$p<.001$.
  \end{minipage}
\end{table}

The diversity effect depends on whether the driving application is the source or the target. With driving as
the single source, diversity raised the \gls{auroc} on all three targets ($+0.25$ to $+0.51$, all significant). With
driving as the target, it lowered the \gls{auroc} in two of three pairs ($-0.10$ to $-0.17$). Among the six pairs
without driving, only \emph{Overcooked! 2} as the source for the $n$-back improved significantly ($+0.10$ and $+0.11$).
The quantity contrast averaged $+0.004$ with both architectures and was significant in none of the 12 pairs.

Compared with the best single source of each target, the multi-source model differed by $+0.015$ and $+0.008$
on the $n$-back, $-0.034$ and $-0.044$ on \emph{Overcooked! 2}, $-0.021$ and $+0.054$ on \emph{Hogwarts Legacy}, and
$-0.135$ and $-0.158$ on the driving application (ConvNeXt-base and ConvNeXt-small). Only the two driving differences
and the $+0.054$ on \emph{Hogwarts Legacy} ($p = .017$) excluded zero. The latter depends on which source counts as the
best. As single sources, the $n$-back and \emph{Overcooked! 2} differed by only $0.003$, and the difference is not
significant when the sources are ranked by their fold-mean \gls{auroc}.
\subsection{Cross-Domain Transfer Tables}
\label{ssec:supp-cross-domain-tables}
The following tables report the full cross-domain \gls{auroc} matrices supporting Figures~\ref{fig:supervised-distribution-shifts}--\ref{fig:supervised-distribution-shifts-applications} in the main article.
Rows denote training subsets and columns denote evaluation subsets.
\subsection{Domain Confounds: Cross-Cohort Transfer}
\label{ssec:supp-domain-confounds}
To probe whether the learned workload is task-general or carries recording-cohort artifacts, we paired same-paradigm
cohorts (Drive $n$-back vs.\ Game $n$-back) under three modality regimes: biosignals only, biosignals plus eye tracking,
and biosignals plus eye tracking plus facial action units. The in-domain minus cross-domain differences in
Table~\ref{sup:tab:domain-confounds-stats} are reported for all three regimes.

\begin{table}[!htb]
  \centering
  \caption{Cross-domain AUROC (mean $\pm$ std) for the ConvNeXt-small encoder across measurement modalities and domain pairs.}
  \label{sup:tab:domain-confounds-auroc}
  \begin{NiceTabular}[name=tabDomainConfoundsAuroc]{c|l|l|c}
  \CodeBefore
  \cellcolor[RGB]{165,0,38}{3-4, 5-4}
  \cellcolor[RGB]{171,219,109}{8-4, 12-4}
  \cellcolor[RGB]{182,16,38}{4-4}
  \cellcolor[RGB]{209,235,133}{10-4}
  \cellcolor[RGB]{231,245,159}{6-4}
  \cellcolor[RGB]{248,139,81}{2-4}
  \cellcolor[RGB]{253,204,122}{11-4}
  \cellcolor[RGB]{253,206,124}{7-4}
  \cellcolor[RGB]{254,237,161}{13-4}
  \cellcolor[RGB]{254,242,169}{9-4}
  \Body
      \hline
      \scriptsize{Modality} & \scriptsize{Train Domain} & \scriptsize{Eval Domain} & \scriptsize{AUROC ($\bar{x} \pm \sigma$)} \\
      \hline
  \scriptsize{\Block{4-1}{Biosignals}} & \scriptsize{Drive N-Back} & \scriptsize{In-Domain (Baseline)} & \scriptsize{$0.625 \pm 0.054$} \\
  \scriptsize{} & \scriptsize{Drive N-Back} & \scriptsize{Game N-Back} & \scriptsize{$0.433 \pm 0.064$} \\
  \scriptsize{} & \scriptsize{Game N-Back} & \scriptsize{In-Domain (Baseline)} & \scriptsize{$0.518 \pm 0.084$} \\
  \scriptsize{} & \scriptsize{Game N-Back} & \scriptsize{Drive N-Back} & \scriptsize{$0.457 \pm 0.086$} \\
      \hline
  \scriptsize{\Block{4-1}{Biosignals, Eye}} & \scriptsize{Drive N-Back} & \scriptsize{In-Domain (Baseline)} & \scriptsize{$0.780 \pm 0.096$} \\
  \scriptsize{} & \scriptsize{Drive N-Back} & \scriptsize{Game N-Back} & \scriptsize{$0.683 \pm 0.120$} \\
  \scriptsize{} & \scriptsize{Game N-Back} & \scriptsize{In-Domain (Baseline)} & \scriptsize{$0.845 \pm 0.035$} \\
  \scriptsize{} & \scriptsize{Game N-Back} & \scriptsize{Drive N-Back} & \scriptsize{$0.730 \pm 0.060$} \\
      \hline
  \scriptsize{\Block{4-1}{Bio., AUs, ET}} & \scriptsize{Drive N-Back} & \scriptsize{In-Domain (Baseline)} & \scriptsize{$0.808 \pm 0.080$} \\
  \scriptsize{} & \scriptsize{Drive N-Back} & \scriptsize{Game N-Back} & \scriptsize{$0.681 \pm 0.085$} \\
  \scriptsize{} & \scriptsize{Game N-Back} & \scriptsize{In-Domain (Baseline)} & \scriptsize{$0.846 \pm 0.026$} \\
  \scriptsize{} & \scriptsize{Game N-Back} & \scriptsize{Drive N-Back} & \scriptsize{$0.721 \pm 0.072$} \\
      \hline
  \end{NiceTabular}
\end{table}
\begin{table}[!htb]
  \centering
  \caption{Expected calibration error (ECE, mean $\pm$ std) for the ConvNeXt-small encoder across measurement modalities and domain pairs; lower values indicate better calibration.}
  \label{sup:tab:domain-confounds-ece}
  \begin{NiceTabular}[name=tabDomainConfoundsEce]{c|l|l|c}
  \CodeBefore
  \cellcolor[RGB]{17,136,71}{2-4}
  \cellcolor[RGB]{84,180,94}{5-4}
  \cellcolor[RGB]{99,187,98}{9-4}
  \cellcolor[RGB]{107,191,99}{10-4}
  \cellcolor[RGB]{114,194,100}{6-4, 13-4}
  \cellcolor[RGB]{117,195,100}{7-4}
  \cellcolor[RGB]{144,207,103}{4-4}
  \cellcolor[RGB]{207,234,132}{3-4}
  \cellcolor[RGB]{225,242,151}{11-4}
  \cellcolor[RGB]{249,252,183}{8-4}
  \cellcolor[RGB]{253,190,111}{12-4}
  \Body
      \hline
      \scriptsize{Modality} & \scriptsize{Train Domain} & \scriptsize{Eval Domain} & \scriptsize{ECE ($\bar{x} \pm \sigma$)} \\
      \hline
  \scriptsize{\Block{4-1}{Biosignals}} & \scriptsize{Drive N-Back} & \scriptsize{Drive N-Back} & \scriptsize{$0.035 \pm 0.023$} \\
  \scriptsize{} & \scriptsize{Drive N-Back} & \scriptsize{Game N-Back} & \scriptsize{$0.191 \pm 0.049$} \\
  \scriptsize{} & \scriptsize{Game N-Back} & \scriptsize{Drive N-Back} & \scriptsize{$0.134 \pm 0.066$} \\
  \scriptsize{} & \scriptsize{Game N-Back} & \scriptsize{Game N-Back} & \scriptsize{$0.089 \pm 0.055$} \\
      \hline
  \scriptsize{\Block{4-1}{Biosignals, Eye}} & \scriptsize{Drive N-Back} & \scriptsize{Drive N-Back} & \scriptsize{$0.110 \pm 0.040$} \\
  \scriptsize{} & \scriptsize{Drive N-Back} & \scriptsize{Game N-Back} & \scriptsize{$0.112 \pm 0.041$} \\
  \scriptsize{} & \scriptsize{Game N-Back} & \scriptsize{Drive N-Back} & \scriptsize{$0.243 \pm 0.121$} \\
  \scriptsize{} & \scriptsize{Game N-Back} & \scriptsize{Game N-Back} & \scriptsize{$0.098 \pm 0.039$} \\
      \hline
  \scriptsize{\Block{4-1}{Bio., AUs, ET}} & \scriptsize{Drive N-Back} & \scriptsize{Drive N-Back} & \scriptsize{$0.104 \pm 0.040$} \\
  \scriptsize{} & \scriptsize{Drive N-Back} & \scriptsize{Game N-Back} & \scriptsize{$0.211 \pm 0.052$} \\
  \scriptsize{} & \scriptsize{Game N-Back} & \scriptsize{Drive N-Back} & \scriptsize{$0.333 \pm 0.065$} \\
  \scriptsize{} & \scriptsize{Game N-Back} & \scriptsize{Game N-Back} & \scriptsize{$0.111 \pm 0.047$} \\
      \hline
  \end{NiceTabular}
\end{table}
\begin{table*}[!t]
  \centering
  \setlength{\tabcolsep}{4pt}
  \tablefont
  \caption{In-domain minus cross-domain test \gls{auroc} for the paired $n$-back cohorts per modality set and transfer direction. A positive $\Delta$\gls{auroc} favours in-domain evaluation.}
  \label{sup:tab:domain-confounds-stats}
  \begin{tabularx}{\linewidth}{l X X r@{.}l c c}
    \toprule
    Modality & Train domain & Eval domain & \multicolumn{2}{c}{$\Delta$\gls{auroc}} & Encoder range & Pairs $\Delta > 0$ \\
    \midrule
    Biosignals & Drive N-Back & Game N-Back & ${+}0$ & $176$ & $[{+}0.15,\,{+}0.19]$ & 15/15 \\
     & Game N-Back & Drive N-Back & ${+}0$ & $000$ & $[{-}0.06,\,{+}0.06]$ & 9/15 \\
    \addlinespace
    Biosignals, Eye & Drive N-Back & Game N-Back & ${+}0$ & $189$ & $[{+}0.10,\,{+}0.25]$ & 14/15 \\
     & Game N-Back & Drive N-Back & ${+}0$ & $124$ & $[{+}0.12,\,{+}0.13]$ & 15/15 \\
    \addlinespace
    Bio., AUs, ET & Drive N-Back & Game N-Back & ${+}0$ & $194$ & $[{+}0.13,\,{+}0.27]$ & 15/15 \\
     & Game N-Back & Drive N-Back & ${+}0$ & $105$ & $[{+}0.07,\,{+}0.12]$ & 14/15 \\
    \bottomrule
  \end{tabularx}
  \begin{minipage}{\linewidth}
    \vspace{2pt}\footnotesize\textit{Note.} $\Delta$\gls{auroc} is the mean over five subject-wise folds and three encoders. Encoder range spans the three encoder means. Pairs counts the fold and encoder pairs with an in-domain advantage. Descriptive, because folds and encoders are not independent replicates. The participant-level test of the cohort shift is reported in Section~\ref{ssec:cohort-shift} of the main article.
  \end{minipage}
\end{table*}

Cross-cohort transfer reduced the \gls{auroc} in every combination except biosignals alone from gaming to
driving (Table~\ref{sup:tab:domain-confounds-stats}). With biosignals and eye tracking, the in-domain advantage was
$+0.189$ (driving to gaming $n$-back) and $+0.124$ (gaming to driving), and with action units added it was $+0.194$ and
$+0.105$. On average it was $+0.153$, and 58 of the 60 fold and encoder pairs favoured in-domain evaluation. These
differences are descriptive, because folds and encoders are not independent. The participant-level test of
the cohort shift is the leave-one-cohort-out analysis in Section~\ref{ssec:cohort-shift} of the main article, where
all 14 contrasts are significant ($p \leq .0013$). This indicates that models partially capture domain-specific
recording artifacts rather than purely task-general task-load (Tables~\ref{sup:tab:domain-confounds-auroc} -
\ref{sup:tab:domain-confounds-stats}). The \gls{ece} counterpart (Table~\ref{sup:tab:domain-confounds-ece}) confirms
that confidence reliability degrades out of cohort: cross-cohort \gls{ece} exceeds in-cohort \gls{ece} in
comparison. This finding reinforces the importance of diverse training data as discussed in
Section~\ref{ssec:task-load-generalization} and motivates future strategies to disentangle genuine task oad signals
from cohort-specific confounds
\textit{Participants in both studies.} Five participants took part in both cohorts. Folds are grouped by
participant, so each of them is tested on both cohorts in one fold and never trained on in that fold. Excluding them
changed no leave-one-cohort-out contrast by more than $0.024$, and all contrasts remained significant ($p < .002$).
\subsection{Ablation: Male and Female Participants}
\label{ssec:supp-sex-ablation}
Independent of the driving and gaming cohorts, this ablation asks whether task-load models transfer between
male and female participants. It is not meant as a supplementary comparison with to cohort shift, that tests for the
effect of participant sex on its own. We formed two groups by reported sex and applied the leave-one-cohort-out design
of Section~\ref{ssec:cohort-shift} of the main article to them, training on one sex and testing on the other, together
with a dataset-size-matched variant. The analysis covered two encoders and six targets (24 contrasts,
fold-mean \gls{auroc}, participant-level cluster bootstrap and sign-flip test).
Twenty-two contrasts were not significant ($|\Delta| \le 0.031$). The exceptions were the male target of the
driving cohort with ConvNeXt-small, at $-0.080$ (95\,\% CI $[-0.125, -0.025]$, $p = .005$) and at $-0.093$ in the
budget-matched variant ($[-0.144, -0.047]$, $p < .001$). Only the budget-matched contrast remained significant after
Holm correction across the 24 contrasts ($p_{\mathrm{Holm}} = .014$). With ConvNeXt-base, the same contrasts were
$-0.031$ and $-0.021$ and not significant. We therefore found no consistent effect of participant sex: models trained
on one sex transfer to the other with little or no loss.
\subsection{Per-Window Input Normalization}
\label{ssec:supp-window-normalization}
To test whether the results depend on statistics computed over a complete application,
we retrained ConvNeXt-\emph{small} on the complete dataset with the same five subject-wise folds. Each input window was
normalized with its own statistics, the mean and standard deviation for the standardized channels and the median and
the spread between the 10th and 90th percentile for the robustly scaled channels. This normalization uses no data
outside the window and can therefore be applied online. With biosignals and eye tracking, the \gls{auroc} was
$0.81 \pm 0.016$, compared with $0.83 \pm 0.045$ under normalization over the application (mean and standard deviation
over the five folds).
\subsection{Calibration}
\label{ssec:supp-calibration}
\begin{figure}[htb!]
  \centering
  \subfloat[\centering Evaluated on $n$-back]{\includegraphics[width=0.49\linewidth]{%
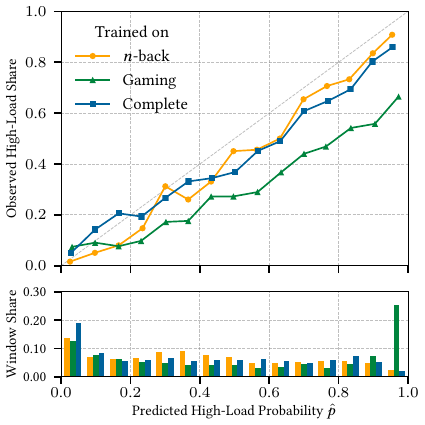}} \hfill%
  \subfloat[\centering Evaluated on gaming]{\includegraphics[width=0.49\linewidth]{%
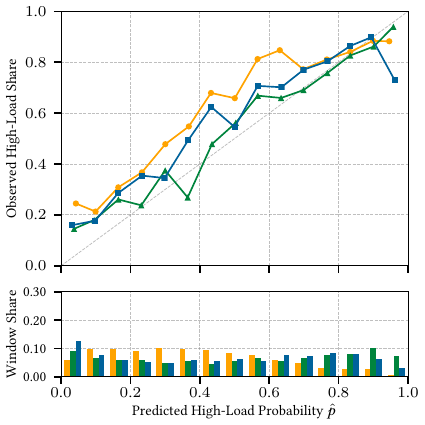}}
  \caption{Reliability diagrams of ConvNeXt-small with biosignals and eye tracking, trained on $n$-back, gaming
  or the complete dataset. Observed share of high-load windows against the predicted probability in 15 equal-width
  bins, bins with fewer than 10 windows omitted. The lower panels show the share of windows per bin. The diagonal marks
  perfect calibration.}
  \label{fig:supp-calibration}
\end{figure}

Table~\ref{tab:supp-calibration} complements \gls{auroc} and \gls{ece} of the main text, using the Brier score
and the \gls{nll}. Here, we use ConvNeXt-small with biosignals and eye tracking, trained on $n$-back,
gaming or the complete dataset and evaluated on $n$-back and gaming. Fig.~\ref{fig:supp-calibration} shows the
reliability diagrams. The models were not recalibrated. The scoring rules and the diagrams use the out-of-fold 
predictions of the five subject-wise folds. The 95\,\% intervals come from a participant-clustered bootstrap.
The $n$-back model underpredicted high load on gaming, and the gaming model overpredicted it on $n$-back.
Compared with these models, the complete model lowered the Brier score by $0.048$ on $n$-back and $0.034$ on gaming.
The \gls{nll} and the reliability term decreased as well, and all intervals of these paired differences excluded zero.
Compared with a model trained on the evaluated domain, the complete model had a slightly higher Brier score ($+0.014$
and $+0.015$). On $n$-back, this difference stems from lower resolution, not from worse reliability.
\begin{sidewaystable*}
  \centering
  \setlength{\tabcolsep}{3pt}
  \tablefont
  \caption{Calibration of ConvNeXt-small under distribution shift for the $n$-back, Gaming and Complete training subsets, evaluated on $n$-back and Gaming. AUROC and ECE as in the main text (mean $\pm$ SD over five subject-wise folds). All other scores are out-of-fold estimates pooled over the five folds with 95\,\% participant-clustered bootstrap intervals. Higher is better for AUROC and Res., lower for ECE, Brier, NLL and Rel.}
  \label{tab:supp-calibration}
  \begin{tabular*}{\linewidth}{@{\extracolsep{\fill}}llccccccc}
    \toprule
    Trained on & Evaluated on & AUROC & ECE & Brier & NLL & Rel. & Res. & $\bar{\hat{p}} - \bar{y}$ \\
    \midrule
    $n$-back & $n$-back & 0.831 $\pm$ 0.033 & 0.04 $\pm$ 0.01 & $0.163$ [$0.151$, $0.175$] & $0.489$ [$0.459$, $0.519$] & $0.006$ [$0.004$, $0.011$] & $0.071$ [$0.062$, $0.084$] & $0.063$ [$0.040$, $0.086$] \\
    $n$-back & Gaming & 0.756 $\pm$ 0.093 & 0.09 $\pm$ 0.06 & $0.227$ [$0.206$, $0.251$] & $0.663$ [$0.605$, $0.728$] & $0.029$ [$0.023$, $0.042$] & $0.050$ [$0.033$, $0.071$] & $-0.157$ [$-0.186$, $-0.128$] \\
    Gaming & $n$-back & 0.801 $\pm$ 0.044 & 0.14 $\pm$ 0.02 & $0.226$ [$0.206$, $0.246$] & $0.760$ [$0.681$, $0.843$] & $0.054$ [$0.041$, $0.071$] & $0.057$ [$0.049$, $0.068$] & $0.193$ [$0.155$, $0.230$] \\
    Gaming & Gaming & 0.813 $\pm$ 0.051 & 0.07 $\pm$ 0.03 & $0.178$ [$0.155$, $0.202$] & $0.542$ [$0.480$, $0.613$] & $0.004$ [$0.003$, $0.011$] & $0.074$ [$0.056$, $0.096$] & $-0.026$ [$-0.051$, $0.003$] \\
    Complete & $n$-back & 0.808 $\pm$ 0.041 & 0.06 $\pm$ 0.05 & $0.178$ [$0.164$, $0.192$] & $0.534$ [$0.495$, $0.576$] & $0.007$ [$0.005$, $0.014$] & $0.058$ [$0.049$, $0.071$] & $0.053$ [$0.025$, $0.081$] \\
    Complete & Gaming & 0.805 $\pm$ 0.086 & 0.08 $\pm$ 0.07 & $0.193$ [$0.166$, $0.223$] & $0.598$ [$0.509$, $0.706$] & $0.011$ [$0.006$, $0.024$] & $0.067$ [$0.049$, $0.089$] & $-0.076$ [$-0.105$, $-0.044$] \\
    \bottomrule
  \end{tabular*}
  \begin{minipage}{\linewidth}
    \vspace{2pt}\footnotesize\textit{Note.} Rel.\ and Res.\ are the calibration and discrimination terms of the Brier decomposition. $\bar{\hat{p}} - \bar{y}$ is the mean predicted probability minus the share of high-load windows, negative values underpredict high load.
  \end{minipage}
\end{sidewaystable*}

\section{Network Architecture Details}
\label{sec:supp-network-architectures}
This section provides the full configuration tables and per-architecture results referenced in Section~\ref{sec:model}
of the main article. All hyperparameters follow established literature and original codebases. Model sizes
reflect typical choices in affective computing. As stated in the main
article, the objective is not to compare architectures but to confirm that conclusions are dominated by
domain and modality rather than network choice.
\subsection{Configuration Tables}
\label{ssec:supp-configs}
For recurrent models (\gls{lstm}\footnote{LSTM:
\url{https://github.com/pytorch/pytorch/blob/v2.6.0/torch/nn/modules/rnn.py}} and
x\gls{lstm}\footnote{
x\gls{lstm}: \url{https://github.com/NX-AI/xlstm/releases/tag/v2.0.3}
}) and transformer
encoders\footnote{Transformer:
\url{https://github.com/pytorch/pytorch/blob/v2.6.0/torch/nn/modules/transformer.py}}, a shared 1D
convolutional front-end (Table~\ref{tab:input-dimensions}) increases channel dimensionality and
downsamples the input sequence. The final representation is mean-pooled over time and passed through a
linear projection for classification. LSTM and xLSTM configurations are given in
Tables~\ref{tab:lstm-dimensions} and~\ref{tab:xlstm-dimensions}. For x\gls{lstm}, established
hyperparameter conventions are limited, so we defined custom configurations in which the \emph{tiny},
\emph{small}, and \emph{base} variants use 2, 4, and 8 m\gls{lstm} and s\gls{lstm} heads, respectively. Transformer depth, width, and attention settings are given in
Table~\ref{tab:transformer-dimensions}.
\begin{table}[!htb]
  \caption{Input convolution front-end configurations.}
  \tablefont
  \label{tab:input-dimensions}
  \begin{tabularx}{\columnwidth}{l|ccc}
      \toprule
      Size         & Kernels    & Strides   & Channels     \\ \midrule
      \emph{tiny}  & 7, 7, 7   & 3, 3, 3   & 16, 32, 32   \\
      \emph{small} & 7, 7, 7   & 3, 3, 2   & 32, 64, 64   \\
      \emph{base}  & 7, 7, 7   & 3, 2, 2   & 64, 128, 128 \\
      \bottomrule
  \end{tabularx}
\end{table}
\begin{table}[!htb]
  \caption{LSTM configurations.}
  \tablefont
  \label{tab:lstm-dimensions}
  \begin{tabularx}{\columnwidth}{l|cc}
      \toprule
      Size         & Hidden Size & Layers \\ \midrule
      \emph{tiny}  & 64         & 2      \\
      \emph{small} & 128        & 4      \\
      \emph{base}  & 256        & 8      \\
      \bottomrule
  \end{tabularx}
\end{table}
\begin{table}[!htb]
  \caption{xLSTM configurations.}
  \label{tab:xlstm-dimensions}
  \tablefont
  \begin{tabularx}{\columnwidth}{l|cc}
      \toprule
      Size         & MLSTM Heads & SLSTM Heads \\ \midrule
      \emph{tiny}  & 2          & 2           \\
      \emph{small} & 4          & 4           \\
      \emph{base}  & 8          & 8           \\
      \bottomrule
  \end{tabularx}
\end{table}
\begin{table}[!htb]
  \caption{Transformer configurations.}
  \label{tab:transformer-dimensions}
  \tablefont
  \begin{tabularx}{\columnwidth}{l|ccccc}
      \toprule
      Size         & Layers & Dim. & Heads & \gls{ffn} & Dropout \\ \midrule
      \emph{tiny}  & 6     & 32       & 8     & 1024    & 0.05    \\
      \emph{small} & 8     & 64       & 8     & 2048    & 0.10    \\
      \emph{base}  & 12    & 128      & 16    & 3072    & 0.15    \\
      \bottomrule
  \end{tabularx}
\end{table}
The \gls{cnn} backbones (1D
ResNet\footnote{ResNet:
\url{https://github.com/pytorch/vision/blob/release/2.0/torchvision/models/resnet.py}} and
ConvNeXt\footnote{ConvNeXt:
\url{https://github.com/facebookresearch/ConvNeXt}}) do not use the shared front-end. We implement
1D versions of ResNet-18, ResNet-34, and ResNet-50
\cite{jinHumanCentricCognitiveState2025,wangTimeSeriesClassification2017}. For ConvNeXt, all 2D
convolutions are adapted to 1D with per-stage dimensions in Table~\ref{tab:convnext-dimensions}
\cite{liuConvNet2020s2022}.
\begin{table}[!htb]
  \caption{ConvNeXt per-stage configurations.}
  \label{tab:convnext-dimensions}
  \tablefont
  \begin{tabularx}{\columnwidth}{l|l|ccc}
      \toprule
      Size & Stage & in & out & layers \\ \midrule
      \multirow{4}{*}{\emph{tiny}}
        & 1 & 96  & 192  & 3  \\
        & 2 & 192 & 384  & 3  \\
        & 3 & 384 & 768  & 9  \\
        & 4 & 768 & -    & 3  \\ \midrule
      \multirow{4}{*}{\emph{small}}
        & 1 & 96  & 192  & 3  \\
        & 2 & 192 & 384  & 3  \\
        & 3 & 384 & 768  & 27 \\
        & 4 & 768 & -    & 3  \\ \midrule
      \multirow{4}{*}{\emph{base}}
        & 1 & 128 & 256  & 3  \\
        & 2 & 256 & 512  & 3  \\
        & 3 & 512 & 1024 & 27 \\
        & 4 & 1024& -    & 3  \\
      \bottomrule
  \end{tabularx}
\end{table}
\subsection{Per-Architecture Comparison}
\label{ssec:supp-network-comparison}
\begin{figure}[!htb]
  \centering
  \subfloat[\centering xLSTM]{\includegraphics[width=0.48\linewidth]{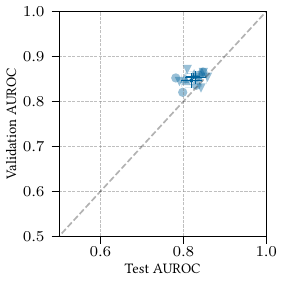}}%
  \hfill
  \subfloat[\centering ConvNeXt]{\includegraphics[width=0.48\linewidth]{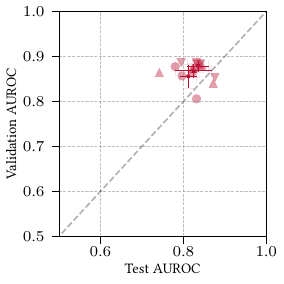}}%
  \\[4pt]
  \subfloat[\centering Transformer]{\includegraphics[width=0.48\linewidth]{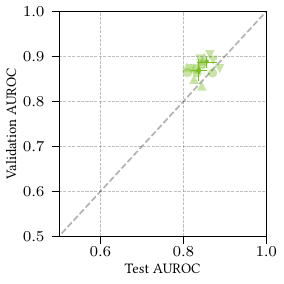}}%
  \caption{Per-architecture test \gls{auroc} (mean $\pm$ SD over five subject-wise folds) for the three
  encoder families across canonical datasets.}
  \label{fig:supp-network-comparison}
\end{figure}

Figure~\ref{fig:supp-network-comparison} confirms that the three architectures produce consistent
results across domains, supporting the main article's conclusion (Section~\ref{ssec:task-load-estimation}, Figure~\ref{fig:supervised-baselines}) that domain
and modality dominate over architecture choice. Within each family, tiny, small, and base variants are
shown side by side.
Table~\ref{tab:modalities-single} reports per-architecture single-modality \gls{auroc} across all datasets,
complementing the multimodal results in Tables~\ref{tab:modalities-multiple-convnext} and~\ref{tab:modalities-multiple} of the main article. Per-architecture modality
breakdowns are provided for xLSTM
(Tables~\ref{tab:supp-modalities-single-xlstm} and~\ref{tab:supp-modalities-multiple-xlstm}) and
Transformer (Tables~\ref{tab:supp-modalities-single-transformer}
and~\ref{tab:supp-modalities-multiple-transformer}).
\subsection{Training Protocol and Computational Cost}
\label{ssec:supp-training}
All models were trained on an Nvidia RTX 4090 GPU using the Adam optimizer with weight decay
$10^{-6}$ and learning rate $10^{-5}$. A reduce-on-plateau scheduler (factor 0.5, patience 16 epochs)
and early stopping (patience 32 epochs, maximum 128 epochs) were applied. No extensive
hyperparameter optimization was performed, as the primary objective is to provide a cross-domain
evaluation with subject-wise splitting for future work to build upon.
For the baseline experiments (5 folds $\times$ 5 dataset configurations $\times$ 3 model sizes
$\times$ 3 architectures), total runtime was approximately seven days. For multimodal experiments
(3 architectures $\times$ 5 datasets $\times$ 5 folds, 21 unique model-dataset combinations), total
runtime was approximately 25 days.
Table~\ref{tab:architecture-batch-size-memory} reports the GPU memory usage of each architecture and size at
batch sizes from 32 to 1024. Empty cells mark batch sizes that exceeded the 24~GB VRAM of the RTX 4090.

\begin{table}[!htb]
  \tablefont
  \caption{GPU memory usage (GB) at each batch size for all architecture and size combinations.}
  \label{tab:architecture-batch-size-memory}
  \centering
  \begin{NiceTabular}[width=\linewidth,name=tabArchitectureBatchSizeMemory]{l|l|Y|Y|Y|Y|Y|Y}
  \CodeBefore
      \cellcolor{yellow!25!green}{4-3,5-3,6-3,6-4,6-5,7-3,7-4,7-5,7-6,8-3,8-4,9-3,9-4,10-3,10-4}
      \cellcolor{yellow!27!green}{2-3,3-3,4-4,5-4,5-5,6-6,7-7,8-5,9-5,10-5,10-6}
      \cellcolor{yellow!30!green}{3-4,4-5,7-8,8-6,9-6}
      \cellcolor{yellow!32!green}{2-4,5-6,6-7,10-7}
      \cellcolor{yellow!35!green}{3-5,4-6,9-7}
      \cellcolor{yellow!37!green}{2-5,6-8,8-7}
      \cellcolor{yellow!42!green}{3-6,4-7,5-7,10-8}
  \Body
        \hline
        ~ & ~ & \scriptsize{32} & \scriptsize{64} & \scriptsize{128} & \scriptsize{256} & \scriptsize{512} & \scriptsize{1024} \\
        \hline
        \cline{1-8} 
\multirow{3}{*}{ConvNeXt} & \scriptsize{base} & \scriptsize{5.4} & \scriptsize{9.1} & \scriptsize{15.9} & ~ & ~ & ~ \\
        ~ & \scriptsize{small} & \scriptsize{4.0} & \scriptsize{6.8} & \scriptsize{11.8} & \scriptsize{22.3} & ~ & ~ \\
        ~ & \scriptsize{tiny} & \scriptsize{2.4} & \scriptsize{3.9} & \scriptsize{6.5} & \scriptsize{11.9} & \scriptsize{22.8} & ~ \\
        \cline{1-8} 
\multirow{3}{*}{Transformer} & \scriptsize{base} & \scriptsize{2.0} & \scriptsize{3.1} & \scriptsize{5.7} & \scriptsize{10.5} & \scriptsize{20.4} & ~ \\
        ~ & \scriptsize{small} & \scriptsize{1.1} & \scriptsize{1.6} & \scriptsize{2.8} & \scriptsize{4.8} & \scriptsize{8.8} & \scriptsize{16.9} \\
        ~ & \scriptsize{tiny} & \scriptsize{1.2} & \scriptsize{0.9} & \scriptsize{1.3} & \scriptsize{2.0} & \scriptsize{3.3} & \scriptsize{6.0} \\
        \cline{1-8} 
\multirow{3}{*}{xLSTM} & \scriptsize{base} & \scriptsize{1.6} & \scriptsize{2.5} & \scriptsize{4.6} & \scriptsize{8.5} & \scriptsize{16.2} & ~ \\
        ~ & \scriptsize{small} & \scriptsize{1.3} & \scriptsize{2.1} & \scriptsize{3.5} & \scriptsize{6.8} & \scriptsize{13.1} & ~ \\
        ~ & \scriptsize{tiny} & \scriptsize{1.2} & \scriptsize{2.0} & \scriptsize{3.1} & \scriptsize{5.5} & \scriptsize{10.9} & \scriptsize{20.9} \\

        \hline
  \end{NiceTabular}
\end{table}

%
\FloatBarrier

\section{Per-Architecture Modality and Cross-Domain Tables}
The following tables supplement the main results with the full per-modality performance of the single- and
multi-modality models across all evaluated architectures, as well as the cross-domain \gls{auroc} statistics reported
both as an average and separately per architecture. The cross-domain \gls{auroc} matrices averaged over architectures
are given in Table~\ref{tab:cross-domain-auroc}, and the per-architecture matrices for xLSTM, ConvNeXt, and
Transformer encoders in Tables~\ref{tab:cross-domain-auroc-xlstm}, \ref{tab:cross-domain-auroc-convnext},
and~\ref{tab:cross-domain-auroc-transformer}, respectively. In every matrix, rows denote training subsets and columns
denote evaluation subsets.

\begin{table*}[!t]

\caption{Mean and standard deviation of the AUROC for single modalities (xLSTM).}
  \label{tab:supp-modalities-single-xlstm}
  \centering
  \setlength{\tabcolsep}{3pt}
  \renewcommand{\arraystretch}{1.25}
  \begin{NiceTabular}{r|c|c|c|c|c|c|c|c|c|c}
  \CodeBefore
      \cellcolor[RGB]{255, 184, 0}{2-2}
      \cellcolor[RGB]{255, 187, 0}{2-3,5-4}
      \cellcolor[RGB]{255, 180, 0}{2-4}
      \cellcolor[RGB]{255, 197, 0}{2-5}
      \cellcolor[RGB]{218, 204, 9}{2-6,9-4}
      \cellcolor[RGB]{255, 205, 0}{2-7,5-10}
      \cellcolor[RGB]{255, 198, 0}{2-8,5-5,6-10}
      \cellcolor[RGB]{255, 201, 0}{2-9}
      \cellcolor[RGB]{215, 203, 10}{2-10}
      \cellcolor[RGB]{199, 201, 13}{2-11,4-8,6-7}
      \cellcolor[RGB]{243, 207, 3}{3-2}
      \cellcolor[RGB]{208, 202, 11}{3-3,4-5,4-10}
      \cellcolor[RGB]{255, 191, 0}{3-4,3-9,10-9}
      \cellcolor[RGB]{246, 208, 2}{3-5,7-8}
      \cellcolor[RGB]{132, 192, 29}{3-6,7-6}
      \cellcolor[RGB]{253, 209, 1}{3-7,6-8,7-2}
      \cellcolor[RGB]{255, 178, 0}{3-8}
      \cellcolor[RGB]{176, 198, 19}{3-10,7-9}
      \cellcolor[RGB]{186, 199, 16}{3-11,4-11,8-6,9-9}
      \cellcolor[RGB]{205, 202, 12}{4-2}
      \cellcolor[RGB]{227, 205, 7}{4-3}
      \cellcolor[RGB]{255, 200, 0}{4-4}
      \cellcolor[RGB]{192, 200, 15}{4-6}
      \cellcolor[RGB]{122, 190, 31}{4-7,8-11,10-11}
      \cellcolor[RGB]{141, 193, 27}{4-9,7-10,10-10}
      \cellcolor[RGB]{221, 204, 8}{5-2}
      \cellcolor[RGB]{237, 206, 4}{5-3,7-3}
      \cellcolor[RGB]{138, 193, 28}{5-6,9-7}
      \cellcolor[RGB]{255, 196, 0}{5-7,7-4}
      \cellcolor[RGB]{250, 208, 1}{5-8,5-9}
      \cellcolor[RGB]{255, 199, 0}{5-11,6-6,6-11}
      \cellcolor[RGB]{255, 186, 0}{6-2,10-4}
      \cellcolor[RGB]{255, 181, 0}{6-3}
      \cellcolor[RGB]{255, 172, 0}{6-4}
      \cellcolor[RGB]{255, 163, 0}{6-5}
      \cellcolor[RGB]{195, 201, 14}{6-9}
      \cellcolor[RGB]{255, 209, 0}{7-5}
      \cellcolor[RGB]{183, 199, 17}{7-7}
      \cellcolor[RGB]{167, 197, 21}{7-11}
      \cellcolor[RGB]{40, 151, 51}{8-2}
      \cellcolor[RGB]{31, 147, 53}{8-3}
      \cellcolor[RGB]{60, 161, 47}{8-4}
      \cellcolor[RGB]{54, 158, 48}{8-5}
      \cellcolor[RGB]{43, 153, 51}{8-7,10-2}
      \cellcolor[RGB]{111, 185, 34}{8-8}
      \cellcolor[RGB]{88, 174, 40}{8-9}
      \cellcolor[RGB]{154, 195, 24}{8-10}
      \cellcolor[RGB]{255, 183, 0}{9-2,10-8}
      \cellcolor[RGB]{77, 169, 43}{9-3}
      \cellcolor[RGB]{129, 191, 30}{9-5}
      \cellcolor[RGB]{82, 172, 41}{9-6}
      \cellcolor[RGB]{151, 194, 25}{9-8}
      \cellcolor[RGB]{105, 183, 36}{9-10}
      \cellcolor[RGB]{108, 184, 35}{9-11}
      \cellcolor[RGB]{17, 140, 57}{10-3}
      \cellcolor[RGB]{113, 187, 34}{10-5}
      \cellcolor[RGB]{23, 143, 56}{10-6}
      \cellcolor[RGB]{255, 176, 0}{10-7}
  \Body
      \hline
      ~ & \scriptsize{\makecell{n-back\\(Driving)}} & \scriptsize{\makecell{n-back\\Dual Task}} & \scriptsize{\makecell{n-back\\(Gaming)}} & \scriptsize{\makecell{n-back\\(Combined)}} & \scriptsize{\makecell{Driving\\Application}} & \scriptsize{\makecell{Hogwarts\\Legacy}} & \scriptsize{Overcooked! 2} & \scriptsize{Gaming} & \scriptsize{\makecell{Combined\\Applications}} & \scriptsize{Complete} \\
      \cline{2-11}
      ~ \scriptsize{RSP} & \scriptsize{0.50\,$\pm$\,0.112} & \scriptsize{0.51\,$\pm$\,0.054} & \scriptsize{0.49\,$\pm$\,0.057} & \scriptsize{0.55\,$\pm$\,0.059} & \scriptsize{0.63\,$\pm$\,0.109} & \scriptsize{0.58\,$\pm$\,0.043} & \scriptsize{0.55\,$\pm$\,0.062} & \scriptsize{0.56\,$\pm$\,0.067} & \scriptsize{0.64\,$\pm$\,0.017} & \scriptsize{0.66\,$\pm$\,0.013} \\ 
      ~ \scriptsize{EMG} & \scriptsize{0.60\,$\pm$\,0.049} & \scriptsize{0.65\,$\pm$\,0.045} & \scriptsize{0.53\,$\pm$\,0.060} & \scriptsize{0.60\,$\pm$\,0.026} & \scriptsize{0.74\,$\pm$\,0.043} & \scriptsize{0.59\,$\pm$\,0.057} & \scriptsize{0.48\,$\pm$\,0.034} & \scriptsize{0.53\,$\pm$\,0.027} & \scriptsize{0.68\,$\pm$\,0.030} & \scriptsize{0.67\,$\pm$\,0.012} \\ 
      ~ \scriptsize{EDA} & \scriptsize{0.65\,$\pm$\,0.063} & \scriptsize{0.62\,$\pm$\,0.050} & \scriptsize{0.56\,$\pm$\,0.073} & \scriptsize{0.64\,$\pm$\,0.049} & \scriptsize{0.66\,$\pm$\,0.044} & \scriptsize{0.75\,$\pm$\,0.086} & \scriptsize{0.66\,$\pm$\,0.071} & \scriptsize{0.73\,$\pm$\,0.094} & \scriptsize{0.65\,$\pm$\,0.030} & \scriptsize{0.67\,$\pm$\,0.041} \\ 
      ~ \scriptsize{SKT} & \scriptsize{0.63\,$\pm$\,0.136} & \scriptsize{0.61\,$\pm$\,0.047} & \scriptsize{0.51\,$\pm$\,0.093} & \scriptsize{0.55\,$\pm$\,0.018} & \scriptsize{0.73\,$\pm$\,0.044} & \scriptsize{0.54\,$\pm$\,0.101} & \scriptsize{0.59\,$\pm$\,0.110} & \scriptsize{0.59\,$\pm$\,0.089} & \scriptsize{0.58\,$\pm$\,0.050} & \scriptsize{0.55\,$\pm$\,0.017} \\ 
      ~ \scriptsize{PPG} & \scriptsize{0.51\,$\pm$\,0.072} & \scriptsize{0.49\,$\pm$\,0.065} & \scriptsize{0.46\,$\pm$\,0.049} & \scriptsize{0.43\,$\pm$\,0.031} & \scriptsize{0.55\,$\pm$\,0.115} & \scriptsize{0.66\,$\pm$\,0.092} & \scriptsize{0.59\,$\pm$\,0.065} & \scriptsize{0.66\,$\pm$\,0.066} & \scriptsize{0.55\,$\pm$\,0.035} & \scriptsize{0.56\,$\pm$\,0.026} \\ 
      ~ \scriptsize{ECG} & \scriptsize{0.59\,$\pm$\,0.100} & \scriptsize{0.61\,$\pm$\,0.033} & \scriptsize{0.54\,$\pm$\,0.039} & \scriptsize{0.59\,$\pm$\,0.014} & \scriptsize{0.74\,$\pm$\,0.038} & \scriptsize{0.67\,$\pm$\,0.059} & \scriptsize{0.60\,$\pm$\,0.038} & \scriptsize{0.68\,$\pm$\,0.075} & \scriptsize{0.73\,$\pm$\,0.049} & \scriptsize{0.69\,$\pm$\,0.053} \\ 
      ~ \scriptsize{ET Pupil Diameter} & \scriptsize{0.86\,$\pm$\,0.045} & \scriptsize{0.87\,$\pm$\,0.046} & \scriptsize{0.83\,$\pm$\,0.048} & \scriptsize{0.84\,$\pm$\,0.052} & \scriptsize{0.67\,$\pm$\,0.039} & \scriptsize{0.85\,$\pm$\,0.086} & \scriptsize{0.76\,$\pm$\,0.095} & \scriptsize{0.79\,$\pm$\,0.092} & \scriptsize{0.71\,$\pm$\,0.061} & \scriptsize{0.75\,$\pm$\,0.052} \\ 
      ~ \scriptsize{ET Move.} & \scriptsize{0.50\,$\pm$\,0.037} & \scriptsize{0.81\,$\pm$\,0.038} & \scriptsize{0.63\,$\pm$\,0.093} & \scriptsize{0.74\,$\pm$\,0.014} & \scriptsize{0.80\,$\pm$\,0.044} & \scriptsize{0.73\,$\pm$\,0.088} & \scriptsize{0.71\,$\pm$\,0.047} & \scriptsize{0.67\,$\pm$\,0.066} & \scriptsize{0.77\,$\pm$\,0.051} & \scriptsize{0.77\,$\pm$\,0.025} \\ 
      ~ \scriptsize{AUs} & \scriptsize{0.85\,$\pm$\,0.048} & \scriptsize{0.89\,$\pm$\,0.066} & \scriptsize{0.51\,$\pm$\,0.043} & \scriptsize{0.76\,$\pm$\,0.065} & \scriptsize{0.88\,$\pm$\,0.048} & \scriptsize{0.48\,$\pm$\,0.109} & \scriptsize{0.50\,$\pm$\,0.037} & \scriptsize{0.53\,$\pm$\,0.048} & \scriptsize{0.72\,$\pm$\,0.085} & \scriptsize{0.75\,$\pm$\,0.034} \\ 
\hline
  \end{NiceTabular}

\end{table*}

\begin{table*}[!t]

\caption{Mean and standard deviation of the AUROC for multi-modality combinations (xLSTM).}
  \label{tab:supp-modalities-multiple-xlstm}
  \centering
  \setlength{\tabcolsep}{3pt}
  \renewcommand{\arraystretch}{1.25}
  \begin{NiceTabular}{r|c|c|c|c|c|c|c|c|c|c}
  \CodeBefore
      \cellcolor[RGB]{255, 208, 0}{2-2,6-4}
      \cellcolor[RGB]{255, 177, 0}{2-3}
      \cellcolor[RGB]{255, 178, 0}{2-4}
      \cellcolor[RGB]{255, 197, 0}{2-5,6-5,8-6,8-8,8-9}
      \cellcolor[RGB]{215, 203, 10}{2-6,2-10}
      \cellcolor[RGB]{218, 204, 9}{2-7,2-9}
      \cellcolor[RGB]{255, 202, 0}{2-8,6-9,7-8,8-10,9-8}
      \cellcolor[RGB]{192, 200, 15}{2-11,3-10,12-9}
      \cellcolor[RGB]{186, 199, 16}{3-2}
      \cellcolor[RGB]{164, 196, 22}{3-3}
      \cellcolor[RGB]{255, 188, 0}{3-4}
      \cellcolor[RGB]{224, 205, 7}{3-5}
      \cellcolor[RGB]{138, 193, 28}{3-6,5-5}
      \cellcolor[RGB]{148, 194, 25}{3-7}
      \cellcolor[RGB]{221, 204, 8}{3-8,9-5,9-10,13-8}
      \cellcolor[RGB]{141, 193, 27}{3-9,5-11}
      \cellcolor[RGB]{157, 195, 23}{3-11}
      \cellcolor[RGB]{57, 159, 47}{4-2,10-2,10-5}
      \cellcolor[RGB]{34, 148, 53}{4-3,11-6}
      \cellcolor[RGB]{71, 166, 44}{4-4}
      \cellcolor[RGB]{45, 154, 50}{4-5,13-6}
      \cellcolor[RGB]{62, 162, 46}{4-6,11-10}
      \cellcolor[RGB]{26, 144, 55}{4-7}
      \cellcolor[RGB]{88, 174, 40}{4-8,10-4,13-7,13-10}
      \cellcolor[RGB]{37, 150, 52}{4-9}
      \cellcolor[RGB]{51, 157, 49}{4-10,4-11,5-3,13-3}
      \cellcolor[RGB]{60, 161, 47}{5-2,5-6}
      \cellcolor[RGB]{234, 206, 5}{5-4}
      \cellcolor[RGB]{255, 198, 0}{5-7}
      \cellcolor[RGB]{255, 196, 0}{5-8,8-5}
      \cellcolor[RGB]{253, 209, 1}{5-9,6-7,6-11,7-9}
      \cellcolor[RGB]{122, 190, 31}{5-10,11-8,12-2}
      \cellcolor[RGB]{255, 189, 0}{6-2}
      \cellcolor[RGB]{255, 173, 0}{6-3}
      \cellcolor[RGB]{255, 204, 0}{6-6,6-10,9-9,12-8}
      \cellcolor[RGB]{255, 201, 0}{6-8,9-7}
      \cellcolor[RGB]{173, 197, 19}{7-2}
      \cellcolor[RGB]{151, 194, 25}{7-3}
      \cellcolor[RGB]{246, 208, 2}{7-4}
      \cellcolor[RGB]{208, 202, 11}{7-5}
      \cellcolor[RGB]{125, 191, 31}{7-6}
      \cellcolor[RGB]{237, 206, 4}{7-7}
      \cellcolor[RGB]{183, 199, 17}{7-10}
      \cellcolor[RGB]{199, 201, 13}{7-11}
      \cellcolor[RGB]{255, 185, 0}{8-2}
      \cellcolor[RGB]{255, 200, 0}{8-3,8-11}
      \cellcolor[RGB]{255, 190, 0}{8-4}
      \cellcolor[RGB]{255, 192, 0}{8-7}
      \cellcolor[RGB]{176, 198, 19}{9-2,13-4}
      \cellcolor[RGB]{195, 201, 14}{9-3,12-7}
      \cellcolor[RGB]{255, 206, 0}{9-4}
      \cellcolor[RGB]{170, 197, 20}{9-6}
      \cellcolor[RGB]{202, 202, 13}{9-11}
      \cellcolor[RGB]{28, 146, 54}{10-3}
      \cellcolor[RGB]{31, 147, 53}{10-6,11-7}
      \cellcolor[RGB]{43, 153, 51}{10-7}
      \cellcolor[RGB]{132, 192, 29}{10-8}
      \cellcolor[RGB]{65, 164, 45}{10-9,10-11}
      \cellcolor[RGB]{54, 158, 48}{10-10,11-9}
      \cellcolor[RGB]{77, 169, 43}{11-2,13-5}
      \cellcolor[RGB]{20, 142, 56}{11-3}
      \cellcolor[RGB]{119, 190, 32}{11-4}
      \cellcolor[RGB]{68, 165, 45}{11-5}
      \cellcolor[RGB]{40, 151, 51}{11-11}
      \cellcolor[RGB]{113, 187, 34}{12-3}
      \cellcolor[RGB]{231, 206, 6}{12-4}
      \cellcolor[RGB]{144, 193, 26}{12-5,13-9}
      \cellcolor[RGB]{102, 181, 36}{12-6}
      \cellcolor[RGB]{154, 195, 24}{12-10}
      \cellcolor[RGB]{135, 192, 28}{12-11}
      \cellcolor[RGB]{96, 179, 38}{13-2}
      \cellcolor[RGB]{82, 172, 41}{13-11}
  \Body
      \hline
      ~ & \scriptsize{\makecell{n-back\\(Driving)}} & \scriptsize{\makecell{n-back\\Dual Task}} & \scriptsize{\makecell{n-back\\(Gaming)}} & \scriptsize{\makecell{n-back\\(Combined)}} & \scriptsize{\makecell{Driving\\Application}} & \scriptsize{\makecell{Hogwarts\\Legacy}} & \scriptsize{Overcooked! 2} & \scriptsize{Gaming} & \scriptsize{\makecell{Combined\\Applications}} & \scriptsize{Complete} \\
      \cline{2-11}
      ~ \scriptsize{Heart Comb.} & \scriptsize{0.58\,$\pm$\,0.039} & \scriptsize{0.48\,$\pm$\,0.104} & \scriptsize{0.48\,$\pm$\,0.112} & \scriptsize{0.55\,$\pm$\,0.046} & \scriptsize{0.64\,$\pm$\,0.044} & \scriptsize{0.63\,$\pm$\,0.150} & \scriptsize{0.56\,$\pm$\,0.052} & \scriptsize{0.63\,$\pm$\,0.046} & \scriptsize{0.64\,$\pm$\,0.030} & \scriptsize{0.67\,$\pm$\,0.026} \\ 
      ~ \scriptsize{Biosignals} & \scriptsize{0.67\,$\pm$\,0.053} & \scriptsize{0.70\,$\pm$\,0.040} & \scriptsize{0.52\,$\pm$\,0.053} & \scriptsize{0.63\,$\pm$\,0.004} & \scriptsize{0.73\,$\pm$\,0.028} & \scriptsize{0.72\,$\pm$\,0.055} & \scriptsize{0.63\,$\pm$\,0.104} & \scriptsize{0.72\,$\pm$\,0.068} & \scriptsize{0.66\,$\pm$\,0.065} & \scriptsize{0.71\,$\pm$\,0.015} \\ 
      ~ \scriptsize{Eye Comb.} & \scriptsize{0.83\,$\pm$\,0.043} & \scriptsize{0.86\,$\pm$\,0.025} & \scriptsize{0.82\,$\pm$\,0.068} & \scriptsize{0.85\,$\pm$\,0.023} & \scriptsize{0.83\,$\pm$\,0.053} & \scriptsize{0.88\,$\pm$\,0.071} & \scriptsize{0.79\,$\pm$\,0.052} & \scriptsize{0.86\,$\pm$\,0.054} & \scriptsize{0.84\,$\pm$\,0.029} & \scriptsize{0.84\,$\pm$\,0.021} \\ 
      ~ \scriptsize{Head Rotation} & \scriptsize{0.83\,$\pm$\,0.059} & \scriptsize{0.84\,$\pm$\,0.062} & \scriptsize{0.62\,$\pm$\,0.089} & \scriptsize{0.73\,$\pm$\,0.028} & \scriptsize{0.83\,$\pm$\,0.055} & \scriptsize{0.55\,$\pm$\,0.060} & \scriptsize{0.54\,$\pm$\,0.091} & \scriptsize{0.59\,$\pm$\,0.026} & \scriptsize{0.75\,$\pm$\,0.051} & \scriptsize{0.73\,$\pm$\,0.047} \\ 
      ~ \scriptsize{Head Position} & \scriptsize{0.52\,$\pm$\,0.062} & \scriptsize{0.46\,$\pm$\,0.086} & \scriptsize{0.58\,$\pm$\,0.043} & \scriptsize{0.55\,$\pm$\,0.044} & \scriptsize{0.57\,$\pm$\,0.030} & \scriptsize{0.59\,$\pm$\,0.080} & \scriptsize{0.56\,$\pm$\,0.073} & \scriptsize{0.56\,$\pm$\,0.048} & \scriptsize{0.57\,$\pm$\,0.038} & \scriptsize{0.59\,$\pm$\,0.060} \\ 
      ~ \scriptsize{Head Comb.} & \scriptsize{0.69\,$\pm$\,0.062} & \scriptsize{0.71\,$\pm$\,0.054} & \scriptsize{0.60\,$\pm$\,0.042} & \scriptsize{0.65\,$\pm$\,0.041} & \scriptsize{0.74\,$\pm$\,0.065} & \scriptsize{0.61\,$\pm$\,0.038} & \scriptsize{0.57\,$\pm$\,0.066} & \scriptsize{0.59\,$\pm$\,0.049} & \scriptsize{0.68\,$\pm$\,0.055} & \scriptsize{0.66\,$\pm$\,0.036} \\ 
      ~ \scriptsize{Body Position} & \scriptsize{0.51\,$\pm$\,0.080} & \scriptsize{0.56\,$\pm$\,0.067} & \scriptsize{0.52\,$\pm$\,0.064} & \scriptsize{0.54\,$\pm$\,0.051} & \scriptsize{0.54\,$\pm$\,0.034} & \scriptsize{0.53\,$\pm$\,0.097} & \scriptsize{0.55\,$\pm$\,0.105} & \scriptsize{0.55\,$\pm$\,0.080} & \scriptsize{0.57\,$\pm$\,0.042} & \scriptsize{0.56\,$\pm$\,0.035} \\ 
      ~ \scriptsize{Move. Comb.} & \scriptsize{0.68\,$\pm$\,0.091} & \scriptsize{0.66\,$\pm$\,0.088} & \scriptsize{0.58\,$\pm$\,0.068} & \scriptsize{0.63\,$\pm$\,0.060} & \scriptsize{0.69\,$\pm$\,0.022} & \scriptsize{0.56\,$\pm$\,0.044} & \scriptsize{0.56\,$\pm$\,0.030} & \scriptsize{0.57\,$\pm$\,0.037} & \scriptsize{0.63\,$\pm$\,0.012} & \scriptsize{0.65\,$\pm$\,0.045} \\ 
      ~ \scriptsize{Biosignals, Eye} & \scriptsize{0.84\,$\pm$\,0.041} & \scriptsize{0.87\,$\pm$\,0.024} & \scriptsize{0.79\,$\pm$\,0.074} & \scriptsize{0.84\,$\pm$\,0.028} & \scriptsize{0.87\,$\pm$\,0.028} & \scriptsize{0.85\,$\pm$\,0.069} & \scriptsize{0.73\,$\pm$\,0.045} & \scriptsize{0.82\,$\pm$\,0.043} & \scriptsize{0.84\,$\pm$\,0.012} & \scriptsize{0.82\,$\pm$\,0.019} \\ 
      ~ \scriptsize{Bio., AUs, ET} & \scriptsize{0.81\,$\pm$\,0.045} & \scriptsize{0.88\,$\pm$\,0.037} & \scriptsize{0.75\,$\pm$\,0.069} & \scriptsize{0.82\,$\pm$\,0.037} & \scriptsize{0.87\,$\pm$\,0.031} & \scriptsize{0.87\,$\pm$\,0.047} & \scriptsize{0.75\,$\pm$\,0.044} & \scriptsize{0.84\,$\pm$\,0.044} & \scriptsize{0.83\,$\pm$\,0.027} & \scriptsize{0.86\,$\pm$\,0.026} \\ 
      ~ \scriptsize{Move., AUs, ET} & \scriptsize{0.75\,$\pm$\,0.049} & \scriptsize{0.76\,$\pm$\,0.079} & \scriptsize{0.62\,$\pm$\,0.014} & \scriptsize{0.72\,$\pm$\,0.048} & \scriptsize{0.78\,$\pm$\,0.032} & \scriptsize{0.66\,$\pm$\,0.086} & \scriptsize{0.57\,$\pm$\,0.067} & \scriptsize{0.66\,$\pm$\,0.052} & \scriptsize{0.71\,$\pm$\,0.034} & \scriptsize{0.73\,$\pm$\,0.022} \\ 
      ~ \scriptsize{Bio, AUs, ET, Move.} & \scriptsize{0.78\,$\pm$\,0.056} & \scriptsize{0.84\,$\pm$\,0.036} & \scriptsize{0.68\,$\pm$\,0.147} & \scriptsize{0.81\,$\pm$\,0.031} & \scriptsize{0.85\,$\pm$\,0.047} & \scriptsize{0.79\,$\pm$\,0.072} & \scriptsize{0.63\,$\pm$\,0.104} & \scriptsize{0.72\,$\pm$\,0.059} & \scriptsize{0.79\,$\pm$\,0.038} & \scriptsize{0.80\,$\pm$\,0.023} \\ 
\hline
  \end{NiceTabular}

\end{table*}

\begin{table*}[!t]

\caption{Mean and standard deviation of the AUROC for single modalities (Transformer).}
  \label{tab:supp-modalities-single-transformer}
  \centering
  \setlength{\tabcolsep}{3pt}
  \renewcommand{\arraystretch}{1.25}
  \begin{NiceTabular}{r|c|c|c|c|c|c|c|c|c|c}
  \CodeBefore
      \cellcolor[RGB]{255, 183, 0}{2-2,2-4,10-7}
      \cellcolor[RGB]{255, 191, 0}{2-3,3-8,6-2}
      \cellcolor[RGB]{255, 209, 0}{2-5,3-5}
      \cellcolor[RGB]{215, 203, 10}{2-6,4-5,4-6}
      \cellcolor[RGB]{255, 184, 0}{2-7,5-4}
      \cellcolor[RGB]{255, 200, 0}{2-8,5-8}
      \cellcolor[RGB]{255, 206, 0}{2-9}
      \cellcolor[RGB]{224, 205, 7}{2-10,9-4}
      \cellcolor[RGB]{205, 202, 12}{2-11}
      \cellcolor[RGB]{234, 206, 5}{3-2}
      \cellcolor[RGB]{208, 202, 11}{3-3}
      \cellcolor[RGB]{255, 196, 0}{3-4,9-2}
      \cellcolor[RGB]{122, 190, 31}{3-6,8-8}
      \cellcolor[RGB]{255, 208, 0}{3-7}
      \cellcolor[RGB]{255, 189, 0}{3-9}
      \cellcolor[RGB]{160, 196, 22}{3-10,7-11}
      \cellcolor[RGB]{186, 199, 16}{3-11}
      \cellcolor[RGB]{211, 203, 10}{4-2,4-3,4-8}
      \cellcolor[RGB]{255, 188, 0}{4-4}
      \cellcolor[RGB]{113, 187, 34}{4-7}
      \cellcolor[RGB]{144, 193, 26}{4-9,5-6}
      \cellcolor[RGB]{202, 202, 13}{4-10}
      \cellcolor[RGB]{189, 200, 16}{4-11}
      \cellcolor[RGB]{227, 205, 7}{5-2}
      \cellcolor[RGB]{253, 209, 1}{5-3}
      \cellcolor[RGB]{255, 186, 0}{5-5}
      \cellcolor[RGB]{255, 185, 0}{5-7}
      \cellcolor[RGB]{250, 208, 1}{5-9}
      \cellcolor[RGB]{255, 205, 0}{5-10,6-7}
      \cellcolor[RGB]{255, 197, 0}{5-11,6-11,10-9}
      \cellcolor[RGB]{255, 178, 0}{6-3}
      \cellcolor[RGB]{255, 172, 0}{6-4}
      \cellcolor[RGB]{255, 171, 0}{6-5}
      \cellcolor[RGB]{255, 201, 0}{6-6}
      \cellcolor[RGB]{221, 204, 8}{6-8}
      \cellcolor[RGB]{173, 197, 19}{6-9}
      \cellcolor[RGB]{255, 199, 0}{6-10}
      \cellcolor[RGB]{255, 203, 0}{7-2}
      \cellcolor[RGB]{246, 208, 2}{7-3,7-5}
      \cellcolor[RGB]{255, 190, 0}{7-4}
      \cellcolor[RGB]{148, 194, 25}{7-6}
      \cellcolor[RGB]{132, 192, 29}{7-7,9-5}
      \cellcolor[RGB]{255, 202, 0}{7-8}
      \cellcolor[RGB]{183, 199, 17}{7-9,9-9}
      \cellcolor[RGB]{141, 193, 27}{7-10}
      \cellcolor[RGB]{45, 154, 50}{8-2,10-2}
      \cellcolor[RGB]{31, 147, 53}{8-3}
      \cellcolor[RGB]{71, 166, 44}{8-4}
      \cellcolor[RGB]{62, 162, 46}{8-5}
      \cellcolor[RGB]{154, 195, 24}{8-6}
      \cellcolor[RGB]{43, 153, 51}{8-7}
      \cellcolor[RGB]{88, 174, 40}{8-9}
      \cellcolor[RGB]{157, 195, 23}{8-10,9-8}
      \cellcolor[RGB]{119, 190, 32}{8-11,10-10}
      \cellcolor[RGB]{74, 168, 43}{9-3,9-10}
      \cellcolor[RGB]{26, 144, 55}{9-6,10-6}
      \cellcolor[RGB]{167, 197, 21}{9-7}
      \cellcolor[RGB]{102, 181, 36}{9-11}
      \cellcolor[RGB]{23, 143, 56}{10-3}
      \cellcolor[RGB]{255, 174, 0}{10-4}
      \cellcolor[RGB]{116, 188, 33}{10-5,10-11}
      \cellcolor[RGB]{255, 193, 0}{10-8}
  \Body
      \hline
      ~ & \scriptsize{\makecell{n-back\\(Driving)}} & \scriptsize{\makecell{n-back\\Dual Task}} & \scriptsize{\makecell{n-back\\(Gaming)}} & \scriptsize{\makecell{n-back\\(Combined)}} & \scriptsize{\makecell{Driving\\Application}} & \scriptsize{\makecell{Hogwarts\\Legacy}} & \scriptsize{Overcooked! 2} & \scriptsize{Gaming} & \scriptsize{\makecell{Combined\\Applications}} & \scriptsize{Complete} \\
      \cline{2-11}
      ~ \scriptsize{RSP} & \scriptsize{0.50\,$\pm$\,0.061} & \scriptsize{0.53\,$\pm$\,0.035} & \scriptsize{0.50\,$\pm$\,0.076} & \scriptsize{0.59\,$\pm$\,0.025} & \scriptsize{0.64\,$\pm$\,0.102} & \scriptsize{0.50\,$\pm$\,0.088} & \scriptsize{0.56\,$\pm$\,0.040} & \scriptsize{0.58\,$\pm$\,0.065} & \scriptsize{0.62\,$\pm$\,0.032} & \scriptsize{0.65\,$\pm$\,0.022} \\ 
      ~ \scriptsize{EMG} & \scriptsize{0.62\,$\pm$\,0.050} & \scriptsize{0.64\,$\pm$\,0.051} & \scriptsize{0.54\,$\pm$\,0.076} & \scriptsize{0.59\,$\pm$\,0.052} & \scriptsize{0.75\,$\pm$\,0.045} & \scriptsize{0.58\,$\pm$\,0.091} & \scriptsize{0.53\,$\pm$\,0.068} & \scriptsize{0.52\,$\pm$\,0.009} & \scriptsize{0.70\,$\pm$\,0.026} & \scriptsize{0.67\,$\pm$\,0.010} \\ 
      ~ \scriptsize{EDA} & \scriptsize{0.64\,$\pm$\,0.069} & \scriptsize{0.64\,$\pm$\,0.028} & \scriptsize{0.52\,$\pm$\,0.034} & \scriptsize{0.64\,$\pm$\,0.032} & \scriptsize{0.64\,$\pm$\,0.034} & \scriptsize{0.76\,$\pm$\,0.077} & \scriptsize{0.64\,$\pm$\,0.064} & \scriptsize{0.72\,$\pm$\,0.099} & \scriptsize{0.65\,$\pm$\,0.027} & \scriptsize{0.67\,$\pm$\,0.044} \\ 
      ~ \scriptsize{SKT} & \scriptsize{0.62\,$\pm$\,0.104} & \scriptsize{0.59\,$\pm$\,0.042} & \scriptsize{0.50\,$\pm$\,0.116} & \scriptsize{0.51\,$\pm$\,0.051} & \scriptsize{0.72\,$\pm$\,0.037} & \scriptsize{0.50\,$\pm$\,0.040} & \scriptsize{0.56\,$\pm$\,0.054} & \scriptsize{0.60\,$\pm$\,0.103} & \scriptsize{0.58\,$\pm$\,0.047} & \scriptsize{0.55\,$\pm$\,0.018} \\ 
      ~ \scriptsize{PPG} & \scriptsize{0.53\,$\pm$\,0.078} & \scriptsize{0.48\,$\pm$\,0.125} & \scriptsize{0.46\,$\pm$\,0.107} & \scriptsize{0.46\,$\pm$\,0.070} & \scriptsize{0.56\,$\pm$\,0.079} & \scriptsize{0.58\,$\pm$\,0.177} & \scriptsize{0.63\,$\pm$\,0.098} & \scriptsize{0.69\,$\pm$\,0.070} & \scriptsize{0.55\,$\pm$\,0.017} & \scriptsize{0.55\,$\pm$\,0.024} \\ 
      ~ \scriptsize{ECG} & \scriptsize{0.57\,$\pm$\,0.069} & \scriptsize{0.60\,$\pm$\,0.034} & \scriptsize{0.52\,$\pm$\,0.033} & \scriptsize{0.60\,$\pm$\,0.034} & \scriptsize{0.72\,$\pm$\,0.072} & \scriptsize{0.74\,$\pm$\,0.024} & \scriptsize{0.56\,$\pm$\,0.045} & \scriptsize{0.67\,$\pm$\,0.031} & \scriptsize{0.73\,$\pm$\,0.058} & \scriptsize{0.70\,$\pm$\,0.050} \\ 
      ~ \scriptsize{ET Pupil Diameter} & \scriptsize{0.85\,$\pm$\,0.047} & \scriptsize{0.87\,$\pm$\,0.051} & \scriptsize{0.82\,$\pm$\,0.064} & \scriptsize{0.83\,$\pm$\,0.043} & \scriptsize{0.71\,$\pm$\,0.051} & \scriptsize{0.85\,$\pm$\,0.091} & \scriptsize{0.75\,$\pm$\,0.090} & \scriptsize{0.79\,$\pm$\,0.081} & \scriptsize{0.71\,$\pm$\,0.066} & \scriptsize{0.75\,$\pm$\,0.045} \\ 
      ~ \scriptsize{ET Move.} & \scriptsize{0.54\,$\pm$\,0.035} & \scriptsize{0.81\,$\pm$\,0.042} & \scriptsize{0.63\,$\pm$\,0.062} & \scriptsize{0.74\,$\pm$\,0.013} & \scriptsize{0.88\,$\pm$\,0.036} & \scriptsize{0.69\,$\pm$\,0.034} & \scriptsize{0.70\,$\pm$\,0.045} & \scriptsize{0.68\,$\pm$\,0.047} & \scriptsize{0.81\,$\pm$\,0.025} & \scriptsize{0.78\,$\pm$\,0.026} \\ 
      ~ \scriptsize{AUs} & \scriptsize{0.85\,$\pm$\,0.053} & \scriptsize{0.88\,$\pm$\,0.072} & \scriptsize{0.47\,$\pm$\,0.053} & \scriptsize{0.76\,$\pm$\,0.061} & \scriptsize{0.88\,$\pm$\,0.048} & \scriptsize{0.50\,$\pm$\,0.032} & \scriptsize{0.53\,$\pm$\,0.038} & \scriptsize{0.55\,$\pm$\,0.046} & \scriptsize{0.75\,$\pm$\,0.036} & \scriptsize{0.75\,$\pm$\,0.035} \\ 
\hline
  \end{NiceTabular}

\end{table*}

\begin{table*}[!t]

\caption{Mean and standard deviation of the AUROC for multi-modality combinations (Transformer).}
  \label{tab:supp-modalities-multiple-transformer}
  \centering
  \setlength{\tabcolsep}{3pt}
  \renewcommand{\arraystretch}{1.25}
  \begin{NiceTabular}{r|c|c|c|c|c|c|c|c|c|c}
  \CodeBefore
      \cellcolor[RGB]{255, 201, 0}{2-2}
      \cellcolor[RGB]{255, 167, 0}{2-3}
      \cellcolor[RGB]{255, 172, 0}{2-4}
      \cellcolor[RGB]{255, 188, 0}{2-5}
      \cellcolor[RGB]{215, 203, 10}{2-6,9-5,12-8}
      \cellcolor[RGB]{180, 198, 18}{2-7}
      \cellcolor[RGB]{255, 202, 0}{2-8}
      \cellcolor[RGB]{234, 206, 5}{2-9}
      \cellcolor[RGB]{199, 201, 13}{2-10}
      \cellcolor[RGB]{192, 200, 15}{2-11,7-11}
      \cellcolor[RGB]{224, 205, 7}{3-2}
      \cellcolor[RGB]{173, 197, 19}{3-3}
      \cellcolor[RGB]{255, 190, 0}{3-4,6-5,8-5}
      \cellcolor[RGB]{246, 208, 2}{3-5,5-4,6-4}
      \cellcolor[RGB]{148, 194, 25}{3-6,3-9}
      \cellcolor[RGB]{119, 190, 32}{3-7,11-8}
      \cellcolor[RGB]{218, 204, 9}{3-8,7-5,13-8}
      \cellcolor[RGB]{208, 202, 11}{3-10}
      \cellcolor[RGB]{176, 198, 19}{3-11,9-3}
      \cellcolor[RGB]{77, 169, 43}{4-2,13-11}
      \cellcolor[RGB]{31, 147, 53}{4-3,11-6}
      \cellcolor[RGB]{60, 161, 47}{4-4,5-2,5-3,10-11,11-5}
      \cellcolor[RGB]{37, 150, 52}{4-5}
      \cellcolor[RGB]{28, 146, 54}{4-6}
      \cellcolor[RGB]{34, 148, 53}{4-7,4-10,10-7}
      \cellcolor[RGB]{79, 170, 42}{4-8}
      \cellcolor[RGB]{40, 151, 51}{4-9}
      \cellcolor[RGB]{43, 153, 51}{4-11,11-10}
      \cellcolor[RGB]{144, 193, 26}{5-5,7-2}
      \cellcolor[RGB]{57, 159, 47}{5-6,10-10}
      \cellcolor[RGB]{243, 207, 3}{5-7}
      \cellcolor[RGB]{253, 209, 1}{5-8}
      \cellcolor[RGB]{255, 200, 0}{5-9,8-2}
      \cellcolor[RGB]{125, 191, 31}{5-10}
      \cellcolor[RGB]{135, 192, 28}{5-11}
      \cellcolor[RGB]{255, 181, 0}{6-2,6-8}
      \cellcolor[RGB]{255, 178, 0}{6-3}
      \cellcolor[RGB]{255, 205, 0}{6-6}
      \cellcolor[RGB]{255, 195, 0}{6-7,8-10,8-11}
      \cellcolor[RGB]{250, 208, 1}{6-9,6-10}
      \cellcolor[RGB]{255, 204, 0}{6-11}
      \cellcolor[RGB]{141, 193, 27}{7-3,13-9}
      \cellcolor[RGB]{231, 206, 6}{7-4}
      \cellcolor[RGB]{138, 193, 28}{7-6}
      \cellcolor[RGB]{255, 203, 0}{7-7}
      \cellcolor[RGB]{255, 187, 0}{7-8,8-4,8-9}
      \cellcolor[RGB]{255, 192, 0}{7-9,8-3,8-6,9-8}
      \cellcolor[RGB]{189, 200, 16}{7-10}
      \cellcolor[RGB]{255, 199, 0}{8-7}
      \cellcolor[RGB]{255, 180, 0}{8-8}
      \cellcolor[RGB]{160, 196, 22}{9-2}
      \cellcolor[RGB]{240, 207, 4}{9-4}
      \cellcolor[RGB]{154, 195, 24}{9-6}
      \cellcolor[RGB]{255, 206, 0}{9-7}
      \cellcolor[RGB]{255, 191, 0}{9-9}
      \cellcolor[RGB]{205, 202, 12}{9-10,9-11}
      \cellcolor[RGB]{68, 165, 45}{10-2,12-3}
      \cellcolor[RGB]{20, 142, 56}{10-3,11-7}
      \cellcolor[RGB]{85, 173, 40}{10-4}
      \cellcolor[RGB]{54, 158, 48}{10-5,11-9}
      \cellcolor[RGB]{26, 144, 55}{10-6}
      \cellcolor[RGB]{129, 191, 30}{10-8}
      \cellcolor[RGB]{71, 166, 44}{10-9}
      \cellcolor[RGB]{74, 168, 43}{11-2}
      \cellcolor[RGB]{23, 143, 56}{11-3}
      \cellcolor[RGB]{96, 179, 38}{11-4,12-6}
      \cellcolor[RGB]{51, 157, 49}{11-11}
      \cellcolor[RGB]{105, 183, 36}{12-2,13-7}
      \cellcolor[RGB]{164, 196, 22}{12-4}
      \cellcolor[RGB]{113, 187, 34}{12-5}
      \cellcolor[RGB]{183, 199, 17}{12-7}
      \cellcolor[RGB]{186, 199, 16}{12-9}
      \cellcolor[RGB]{116, 188, 33}{12-10}
      \cellcolor[RGB]{132, 192, 29}{12-11}
      \cellcolor[RGB]{82, 172, 41}{13-2,13-5}
      \cellcolor[RGB]{45, 154, 50}{13-3}
      \cellcolor[RGB]{167, 197, 21}{13-4}
      \cellcolor[RGB]{17, 140, 57}{13-6}
      \cellcolor[RGB]{88, 174, 40}{13-10}
  \Body
      \hline
      ~ & \scriptsize{\makecell{n-back\\(Driving)}} & \scriptsize{\makecell{n-back\\Dual Task}} & \scriptsize{\makecell{n-back\\(Gaming)}} & \scriptsize{\makecell{n-back\\(Combined)}} & \scriptsize{\makecell{Driving\\Application}} & \scriptsize{\makecell{Hogwarts\\Legacy}} & \scriptsize{Overcooked! 2} & \scriptsize{Gaming} & \scriptsize{\makecell{Combined\\Applications}} & \scriptsize{Complete} \\
      \cline{2-11}
      ~ \scriptsize{Heart Comb.} & \scriptsize{0.56\,$\pm$\,0.058} & \scriptsize{0.44\,$\pm$\,0.092} & \scriptsize{0.46\,$\pm$\,0.085} & \scriptsize{0.51\,$\pm$\,0.018} & \scriptsize{0.64\,$\pm$\,0.056} & \scriptsize{0.68\,$\pm$\,0.067} & \scriptsize{0.56\,$\pm$\,0.050} & \scriptsize{0.62\,$\pm$\,0.059} & \scriptsize{0.66\,$\pm$\,0.041} & \scriptsize{0.66\,$\pm$\,0.066} \\ 
      ~ \scriptsize{Biosignals} & \scriptsize{0.62\,$\pm$\,0.018} & \scriptsize{0.69\,$\pm$\,0.037} & \scriptsize{0.52\,$\pm$\,0.124} & \scriptsize{0.60\,$\pm$\,0.042} & \scriptsize{0.72\,$\pm$\,0.033} & \scriptsize{0.75\,$\pm$\,0.126} & \scriptsize{0.63\,$\pm$\,0.080} & \scriptsize{0.72\,$\pm$\,0.073} & \scriptsize{0.65\,$\pm$\,0.067} & \scriptsize{0.68\,$\pm$\,0.033} \\ 
      ~ \scriptsize{Eye Comb.} & \scriptsize{0.81\,$\pm$\,0.034} & \scriptsize{0.87\,$\pm$\,0.038} & \scriptsize{0.83\,$\pm$\,0.035} & \scriptsize{0.86\,$\pm$\,0.022} & \scriptsize{0.87\,$\pm$\,0.031} & \scriptsize{0.87\,$\pm$\,0.061} & \scriptsize{0.80\,$\pm$\,0.050} & \scriptsize{0.86\,$\pm$\,0.053} & \scriptsize{0.87\,$\pm$\,0.021} & \scriptsize{0.85\,$\pm$\,0.028} \\ 
      ~ \scriptsize{Head Rotation} & \scriptsize{0.83\,$\pm$\,0.073} & \scriptsize{0.83\,$\pm$\,0.040} & \scriptsize{0.60\,$\pm$\,0.066} & \scriptsize{0.72\,$\pm$\,0.044} & \scriptsize{0.84\,$\pm$\,0.053} & \scriptsize{0.60\,$\pm$\,0.030} & \scriptsize{0.59\,$\pm$\,0.047} & \scriptsize{0.56\,$\pm$\,0.058} & \scriptsize{0.74\,$\pm$\,0.036} & \scriptsize{0.73\,$\pm$\,0.033} \\ 
      ~ \scriptsize{Head Position} & \scriptsize{0.49\,$\pm$\,0.071} & \scriptsize{0.48\,$\pm$\,0.051} & \scriptsize{0.60\,$\pm$\,0.016} & \scriptsize{0.52\,$\pm$\,0.025} & \scriptsize{0.58\,$\pm$\,0.041} & \scriptsize{0.54\,$\pm$\,0.129} & \scriptsize{0.49\,$\pm$\,0.073} & \scriptsize{0.60\,$\pm$\,0.073} & \scriptsize{0.59\,$\pm$\,0.024} & \scriptsize{0.57\,$\pm$\,0.048} \\ 
      ~ \scriptsize{Head Comb.} & \scriptsize{0.72\,$\pm$\,0.086} & \scriptsize{0.72\,$\pm$\,0.087} & \scriptsize{0.62\,$\pm$\,0.064} & \scriptsize{0.63\,$\pm$\,0.048} & \scriptsize{0.73\,$\pm$\,0.050} & \scriptsize{0.57\,$\pm$\,0.105} & \scriptsize{0.51\,$\pm$\,0.088} & \scriptsize{0.53\,$\pm$\,0.053} & \scriptsize{0.67\,$\pm$\,0.048} & \scriptsize{0.67\,$\pm$\,0.040} \\ 
      ~ \scriptsize{Body Position} & \scriptsize{0.56\,$\pm$\,0.030} & \scriptsize{0.53\,$\pm$\,0.040} & \scriptsize{0.51\,$\pm$\,0.026} & \scriptsize{0.52\,$\pm$\,0.045} & \scriptsize{0.53\,$\pm$\,0.040} & \scriptsize{0.55\,$\pm$\,0.076} & \scriptsize{0.49\,$\pm$\,0.041} & \scriptsize{0.51\,$\pm$\,0.075} & \scriptsize{0.54\,$\pm$\,0.031} & \scriptsize{0.54\,$\pm$\,0.019} \\ 
      ~ \scriptsize{Move. Comb.} & \scriptsize{0.70\,$\pm$\,0.058} & \scriptsize{0.68\,$\pm$\,0.066} & \scriptsize{0.61\,$\pm$\,0.031} & \scriptsize{0.64\,$\pm$\,0.033} & \scriptsize{0.71\,$\pm$\,0.027} & \scriptsize{0.58\,$\pm$\,0.058} & \scriptsize{0.53\,$\pm$\,0.026} & \scriptsize{0.53\,$\pm$\,0.027} & \scriptsize{0.65\,$\pm$\,0.050} & \scriptsize{0.65\,$\pm$\,0.042} \\ 
      ~ \scriptsize{Biosignals, Eye} & \scriptsize{0.82\,$\pm$\,0.044} & \scriptsize{0.89\,$\pm$\,0.018} & \scriptsize{0.80\,$\pm$\,0.041} & \scriptsize{0.84\,$\pm$\,0.018} & \scriptsize{0.88\,$\pm$\,0.028} & \scriptsize{0.86\,$\pm$\,0.052} & \scriptsize{0.74\,$\pm$\,0.099} & \scriptsize{0.82\,$\pm$\,0.055} & \scriptsize{0.83\,$\pm$\,0.041} & \scriptsize{0.83\,$\pm$\,0.021} \\ 
      ~ \scriptsize{Bio., AUs, ET} & \scriptsize{0.81\,$\pm$\,0.071} & \scriptsize{0.88\,$\pm$\,0.027} & \scriptsize{0.78\,$\pm$\,0.067} & \scriptsize{0.83\,$\pm$\,0.021} & \scriptsize{0.87\,$\pm$\,0.012} & \scriptsize{0.89\,$\pm$\,0.033} & \scriptsize{0.75\,$\pm$\,0.058} & \scriptsize{0.84\,$\pm$\,0.052} & \scriptsize{0.85\,$\pm$\,0.010} & \scriptsize{0.84\,$\pm$\,0.048} \\ 
      ~ \scriptsize{Move., AUs, ET} & \scriptsize{0.77\,$\pm$\,0.039} & \scriptsize{0.82\,$\pm$\,0.019} & \scriptsize{0.70\,$\pm$\,0.074} & \scriptsize{0.76\,$\pm$\,0.042} & \scriptsize{0.78\,$\pm$\,0.047} & \scriptsize{0.68\,$\pm$\,0.053} & \scriptsize{0.64\,$\pm$\,0.060} & \scriptsize{0.67\,$\pm$\,0.061} & \scriptsize{0.75\,$\pm$\,0.029} & \scriptsize{0.73\,$\pm$\,0.025} \\ 
      ~ \scriptsize{Bio, AUs, ET, Move.} & \scriptsize{0.80\,$\pm$\,0.033} & \scriptsize{0.85\,$\pm$\,0.034} & \scriptsize{0.69\,$\pm$\,0.041} & \scriptsize{0.80\,$\pm$\,0.030} & \scriptsize{0.89\,$\pm$\,0.016} & \scriptsize{0.77\,$\pm$\,0.058} & \scriptsize{0.63\,$\pm$\,0.080} & \scriptsize{0.72\,$\pm$\,0.079} & \scriptsize{0.79\,$\pm$\,0.030} & \scriptsize{0.81\,$\pm$\,0.034} \\ 
\hline
  \end{NiceTabular}

\end{table*}

\begin{table*}[!t]
\caption{Mean and standard deviation of the AUROC for single-modality inputs across datasets. The complete dataset consists of $n$-back, driving and gaming experiments.}
  \label{tab:modalities-single}
  \centering
  \setlength{\tabcolsep}{6.0pt}
  \renewcommand{\arraystretch}{1.1}
  \begin{NiceTabular}{r|ccc|ccc|ccc|ccc}
  \CodeBefore 
      \cellcolor[RGB]{199, 201, 13}{3-2}
      \cellcolor[RGB]{205, 202, 12}{3-3,3-4}
      \cellcolor[RGB]{255, 197, 0}{3-5,6-4,7-4,11-13}
      \cellcolor[RGB]{255, 203, 0}{3-6}
      \cellcolor[RGB]{255, 209, 0}{3-7,4-7,8-5}
      \cellcolor[RGB]{218, 204, 9}{3-8,5-9,7-12}
      \cellcolor[RGB]{202, 202, 13}{3-9,5-6,9-3}
      \cellcolor[RGB]{215, 203, 10}{3-10,5-7,5-10}
      \cellcolor[RGB]{255, 201, 0}{3-11,7-10}
      \cellcolor[RGB]{237, 206, 4}{3-12}
      \cellcolor[RGB]{255, 206, 0}{3-13}
      \cellcolor[RGB]{186, 199, 16}{4-2,4-4,5-2,9-8,10-11}
      \cellcolor[RGB]{211, 203, 10}{4-3,9-9,10-12}
      \cellcolor[RGB]{246, 208, 2}{4-5,8-7}
      \cellcolor[RGB]{255, 208, 0}{4-6}
      \cellcolor[RGB]{132, 192, 29}{4-8,8-8,10-7}
      \cellcolor[RGB]{151, 194, 25}{4-9,8-12}
      \cellcolor[RGB]{122, 190, 31}{4-10,9-2,11-2}
      \cellcolor[RGB]{255, 191, 0}{4-11,11-11}
      \cellcolor[RGB]{255, 195, 0}{4-12,6-3,6-6}
      \cellcolor[RGB]{255, 189, 0}{4-13}
      \cellcolor[RGB]{192, 200, 15}{5-3,5-8}
      \cellcolor[RGB]{189, 200, 16}{5-4}
      \cellcolor[RGB]{208, 202, 11}{5-5}
      \cellcolor[RGB]{141, 193, 27}{5-11}
      \cellcolor[RGB]{160, 196, 22}{5-12,8-4,10-6}
      \cellcolor[RGB]{144, 193, 26}{5-13,6-10}
      \cellcolor[RGB]{255, 199, 0}{6-2,7-2,7-8}
      \cellcolor[RGB]{255, 198, 0}{6-5}
      \cellcolor[RGB]{255, 186, 0}{6-7}
      \cellcolor[RGB]{138, 193, 28}{6-8,8-3}
      \cellcolor[RGB]{170, 197, 20}{6-9}
      \cellcolor[RGB]{250, 208, 1}{6-11,6-13}
      \cellcolor[RGB]{255, 202, 0}{6-12}
      \cellcolor[RGB]{255, 187, 0}{7-3}
      \cellcolor[RGB]{255, 163, 0}{7-5}
      \cellcolor[RGB]{255, 184, 0}{7-6}
      \cellcolor[RGB]{255, 171, 0}{7-7}
      \cellcolor[RGB]{255, 193, 0}{7-9}
      \cellcolor[RGB]{195, 201, 14}{7-11}
      \cellcolor[RGB]{173, 197, 19}{7-13}
      \cellcolor[RGB]{167, 197, 21}{8-2}
      \cellcolor[RGB]{243, 207, 3}{8-6}
      \cellcolor[RGB]{108, 184, 35}{8-9,10-2}
      \cellcolor[RGB]{148, 194, 25}{8-10}
      \cellcolor[RGB]{176, 198, 19}{8-11}
      \cellcolor[RGB]{183, 199, 17}{8-13,10-13}
      \cellcolor[RGB]{119, 190, 32}{9-4}
      \cellcolor[RGB]{54, 158, 48}{9-5}
      \cellcolor[RGB]{79, 170, 42}{9-6}
      \cellcolor[RGB]{62, 162, 46}{9-7}
      \cellcolor[RGB]{154, 195, 24}{9-10}
      \cellcolor[RGB]{88, 174, 40}{9-11,9-12,9-13}
      \cellcolor[RGB]{135, 192, 28}{10-3}
      \cellcolor[RGB]{102, 181, 36}{10-4}
      \cellcolor[RGB]{129, 191, 30}{10-5,11-3}
      \cellcolor[RGB]{82, 172, 41}{10-8}
      \cellcolor[RGB]{60, 161, 47}{10-9}
      \cellcolor[RGB]{26, 144, 55}{10-10,11-9,11-10}
      \cellcolor[RGB]{116, 188, 33}{11-4,11-6,11-7}
      \cellcolor[RGB]{113, 187, 34}{11-5}
      \cellcolor[RGB]{23, 143, 56}{11-8}
      \cellcolor[RGB]{255, 200, 0}{11-12}
  \Body
      \hline
      ~ & \multicolumn{3}{c}{Complete} & \multicolumn{3}{c}{n-Back} & \multicolumn{3}{c}{Driving} & \multicolumn{3}{c}{Gaming} \\
      \hline
      ~ & \scriptsize{xLSTM} & \scriptsize{\makecell{Conv-\\Next}} & \scriptsize{\makecell{Trans-\\former}} & \scriptsize{xLSTM} & \scriptsize{\makecell{Conv-\\Next}} & \scriptsize{\makecell{Trans-\\former}} & \scriptsize{xLSTM} & \scriptsize{\makecell{Conv-\\Next}} & \scriptsize{\makecell{Trans-\\former}} & \scriptsize{xLSTM} & \scriptsize{\makecell{Conv-\\Next}} & \scriptsize{\makecell{Trans-\\former}} \\
      \cline{2-13}
      ~ \scriptsize{RSP} & \scriptsize{\makecell{0.66\\$\pm$\,0.013}} & \scriptsize{\makecell{0.65\\$\pm$\,0.013}} & \scriptsize{\makecell{0.65\\$\pm$\,0.022}} & \scriptsize{\makecell{0.55\\$\pm$\,0.059}} & \scriptsize{\makecell{0.57\\$\pm$\,0.047}} & \scriptsize{\makecell{0.59\\$\pm$\,0.025}} & \scriptsize{\makecell{0.63\\$\pm$\,0.109}} & \scriptsize{\makecell{0.65\\$\pm$\,0.063}} & \scriptsize{\makecell{0.64\\$\pm$\,0.102}} & \scriptsize{\makecell{0.56\\$\pm$\,0.067}} & \scriptsize{\makecell{0.61\\$\pm$\,0.054}} & \scriptsize{\makecell{0.58\\$\pm$\,0.065}} \\ 
      ~ \scriptsize{EMG} & \scriptsize{\makecell{0.67\\$\pm$\,0.012}} & \scriptsize{\makecell{0.64\\$\pm$\,0.016}} & \scriptsize{\makecell{0.67\\$\pm$\,0.010}} & \scriptsize{\makecell{0.60\\$\pm$\,0.026}} & \scriptsize{\makecell{0.58\\$\pm$\,0.031}} & \scriptsize{\makecell{0.59\\$\pm$\,0.052}} & \scriptsize{\makecell{0.74\\$\pm$\,0.043}} & \scriptsize{\makecell{0.71\\$\pm$\,0.032}} & \scriptsize{\makecell{0.75\\$\pm$\,0.045}} & \scriptsize{\makecell{0.53\\$\pm$\,0.027}} & \scriptsize{\makecell{0.54\\$\pm$\,0.034}} & \scriptsize{\makecell{0.52\\$\pm$\,0.009}} \\ 
      ~ \scriptsize{EDA} & \scriptsize{\makecell{0.67\\$\pm$\,0.041}} & \scriptsize{\makecell{0.66\\$\pm$\,0.038}} & \scriptsize{\makecell{0.67\\$\pm$\,0.044}} & \scriptsize{\makecell{0.64\\$\pm$\,0.049}} & \scriptsize{\makecell{0.65\\$\pm$\,0.040}} & \scriptsize{\makecell{0.64\\$\pm$\,0.032}} & \scriptsize{\makecell{0.66\\$\pm$\,0.044}} & \scriptsize{\makecell{0.63\\$\pm$\,0.042}} & \scriptsize{\makecell{0.64\\$\pm$\,0.034}} & \scriptsize{\makecell{0.73\\$\pm$\,0.094}} & \scriptsize{\makecell{0.70\\$\pm$\,0.078}} & \scriptsize{\makecell{0.72\\$\pm$\,0.099}} \\ 
      ~ \scriptsize{SKT} & \scriptsize{\makecell{0.55\\$\pm$\,0.017}} & \scriptsize{\makecell{0.54\\$\pm$\,0.048}} & \scriptsize{\makecell{0.55\\$\pm$\,0.018}} & \scriptsize{\makecell{0.55\\$\pm$\,0.018}} & \scriptsize{\makecell{0.54\\$\pm$\,0.076}} & \scriptsize{\makecell{0.51\\$\pm$\,0.051}} & \scriptsize{\makecell{0.73\\$\pm$\,0.044}} & \scriptsize{\makecell{0.69\\$\pm$\,0.028}} & \scriptsize{\makecell{0.72\\$\pm$\,0.037}} & \scriptsize{\makecell{0.59\\$\pm$\,0.089}} & \scriptsize{\makecell{0.57\\$\pm$\,0.123}} & \scriptsize{\makecell{0.60\\$\pm$\,0.103}} \\ 
      ~ \scriptsize{PPG} & \scriptsize{\makecell{0.56\\$\pm$\,0.026}} & \scriptsize{\makecell{0.51\\$\pm$\,0.043}} & \scriptsize{\makecell{0.55\\$\pm$\,0.024}} & \scriptsize{\makecell{0.43\\$\pm$\,0.031}} & \scriptsize{\makecell{0.50\\$\pm$\,0.048}} & \scriptsize{\makecell{0.46\\$\pm$\,0.070}} & \scriptsize{\makecell{0.55\\$\pm$\,0.115}} & \scriptsize{\makecell{0.53\\$\pm$\,0.085}} & \scriptsize{\makecell{0.56\\$\pm$\,0.079}} & \scriptsize{\makecell{0.66\\$\pm$\,0.066}} & \scriptsize{\makecell{0.64\\$\pm$\,0.041}} & \scriptsize{\makecell{0.69\\$\pm$\,0.070}} \\ 
      ~ \scriptsize{ECG} & \scriptsize{\makecell{0.69\\$\pm$\,0.053}} & \scriptsize{\makecell{0.73\\$\pm$\,0.026}} & \scriptsize{\makecell{0.70\\$\pm$\,0.050}} & \scriptsize{\makecell{0.59\\$\pm$\,0.014}} & \scriptsize{\makecell{0.60\\$\pm$\,0.042}} & \scriptsize{\makecell{0.60\\$\pm$\,0.034}} & \scriptsize{\makecell{0.74\\$\pm$\,0.038}} & \scriptsize{\makecell{0.77\\$\pm$\,0.032}} & \scriptsize{\makecell{0.72\\$\pm$\,0.072}} & \scriptsize{\makecell{0.68\\$\pm$\,0.075}} & \scriptsize{\makecell{0.71\\$\pm$\,0.032}} & \scriptsize{\makecell{0.67\\$\pm$\,0.031}} \\ 
      ~ \scriptsize{ET Pupil Diameter} & \scriptsize{\makecell{0.75\\$\pm$\,0.052}} & \scriptsize{\makecell{0.65\\$\pm$\,0.037}} & \scriptsize{\makecell{0.75\\$\pm$\,0.045}} & \scriptsize{\makecell{0.84\\$\pm$\,0.052}} & \scriptsize{\makecell{0.80\\$\pm$\,0.034}} & \scriptsize{\makecell{0.83\\$\pm$\,0.043}} & \scriptsize{\makecell{0.67\\$\pm$\,0.039}} & \scriptsize{\makecell{0.64\\$\pm$\,0.038}} & \scriptsize{\makecell{0.71\\$\pm$\,0.051}} & \scriptsize{\makecell{0.79\\$\pm$\,0.092}} & \scriptsize{\makecell{0.79\\$\pm$\,0.063}} & \scriptsize{\makecell{0.79\\$\pm$\,0.081}} \\ 
      ~ \scriptsize{ET Move.} & \scriptsize{\makecell{0.77\\$\pm$\,0.025}} & \scriptsize{\makecell{0.73\\$\pm$\,0.019}} & \scriptsize{\makecell{0.78\\$\pm$\,0.026}} & \scriptsize{\makecell{0.74\\$\pm$\,0.014}} & \scriptsize{\makecell{0.70\\$\pm$\,0.020}} & \scriptsize{\makecell{0.74\\$\pm$\,0.013}} & \scriptsize{\makecell{0.80\\$\pm$\,0.044}} & \scriptsize{\makecell{0.83\\$\pm$\,0.035}} & \scriptsize{\makecell{0.88\\$\pm$\,0.036}} & \scriptsize{\makecell{0.67\\$\pm$\,0.066}} & \scriptsize{\makecell{0.64\\$\pm$\,0.046}} & \scriptsize{\makecell{0.68\\$\pm$\,0.047}} \\ 
      ~ \scriptsize{AUs} & \scriptsize{\makecell{0.75\\$\pm$\,0.034}} & \scriptsize{\makecell{0.74\\$\pm$\,0.043}} & \scriptsize{\makecell{0.75\\$\pm$\,0.035}} & \scriptsize{\makecell{0.76\\$\pm$\,0.065}} & \scriptsize{\makecell{0.76\\$\pm$\,0.041}} & \scriptsize{\makecell{0.76\\$\pm$\,0.061}} & \scriptsize{\makecell{0.88\\$\pm$\,0.048}} & \scriptsize{\makecell{0.88\\$\pm$\,0.051}} & \scriptsize{\makecell{0.88\\$\pm$\,0.048}} & \scriptsize{\makecell{0.53\\$\pm$\,0.048}} & \scriptsize{\makecell{0.56\\$\pm$\,0.075}} & \scriptsize{\makecell{0.55\\$\pm$\,0.046}} \\ 
\hline
  \end{NiceTabular}
\end{table*}

\begin{sidewaystable*}
\centering
  \setlength{\tabcolsep}{3pt}
  \tablefont
  \caption{Cross-domain AUROC averaged across xLSTM-small, ConvNeXt-small, and Transformer-small (mean$\pm$SD, 5 folds $\times$ 3 architectures).}
  \label{tab:cross-domain-auroc}
  \begin{tabularx}{\linewidth}{X r@{.}l@{$\,{\pm}\,$}r@{.}l r@{.}l@{$\,{\pm}\,$}r@{.}l r@{.}l@{$\,{\pm}\,$}r@{.}l r@{.}l@{$\,{\pm}\,$}r@{.}l r@{.}l@{$\,{\pm}\,$}r@{.}l r@{.}l@{$\,{\pm}\,$}r@{.}l r@{.}l@{$\,{\pm}\,$}r@{.}l r@{.}l@{$\,{\pm}\,$}r@{.}l r@{.}l@{$\,{\pm}\,$}r@{.}l r@{.}l@{$\,{\pm}\,$}r@{.}l}
    \toprule
    Train & \multicolumn{4}{c}{\shortstack{n-back\\(Driving)}} & \multicolumn{4}{c}{\shortstack{n-back\\Dual\\Task}} & \multicolumn{4}{c}{\shortstack{n-back\\(Gaming)}} & \multicolumn{4}{c}{\shortstack{n-back\\(Combined)}} & \multicolumn{4}{c}{\shortstack{Driving\\Application}} & \multicolumn{4}{c}{\shortstack{Hogwarts\\Legacy}} & \multicolumn{4}{c}{\shortstack{Overcooked!\\2}} & \multicolumn{4}{c}{\shortstack{Gaming}} & \multicolumn{4}{c}{\shortstack{Combined\\Applications}} & \multicolumn{4}{c}{\shortstack{Complete}} \\
    \midrule
    n-back (Driving) & 0 & 812 & 0 & 066 & 0 & 733 & 0 & 115 & 0 & 623 & 0 & 105 & 0 & 694 & 0 & 076 & 0 & 531 & 0 & 108 & 0 & 613 & 0 & 101 & 0 & 571 & 0 & 088 & 0 & 583 & 0 & 077 & 0 & 561 & 0 & 073 & 0 & 571 & 0 & 044 \\
    n-back Dual Task & 0 & 747 & 0 & 063 & 0 & 875 & 0 & 028 & 0 & 726 & 0 & 093 & 0 & 812 & 0 & 032 & 0 & 376 & 0 & 085 & 0 & 835 & 0 & 062 & 0 & 681 & 0 & 096 & 0 & 742 & 0 & 071 & 0 & 493 & 0 & 044 & 0 & 568 & 0 & 031 \\
    n-back (Gaming) & 0 & 687 & 0 & 068 & 0 & 774 & 0 & 063 & 0 & 812 & 0 & 055 & 0 & 763 & 0 & 040 & 0 & 528 & 0 & 103 & 0 & 818 & 0 & 080 & 0 & 702 & 0 & 081 & 0 & 742 & 0 & 063 & 0 & 635 & 0 & 053 & 0 & 650 & 0 & 033 \\
    n-back (Combined) & 0 & 779 & 0 & 054 & 0 & 860 & 0 & 042 & 0 & 768 & 0 & 057 & 0 & 835 & 0 & 025 & 0 & 387 & 0 & 057 & 0 & 804 & 0 & 090 & 0 & 672 & 0 & 101 & 0 & 727 & 0 & 089 & 0 & 489 & 0 & 045 & 0 & 575 & 0 & 025 \\
    Driving Application & 0 & 781 & 0 & 062 & 0 & 778 & 0 & 090 & 0 & 621 & 0 & 088 & 0 & 726 & 0 & 058 & 0 & 866 & 0 & 026 & 0 & 632 & 0 & 096 & 0 & 584 & 0 & 096 & 0 & 604 & 0 & 087 & 0 & 761 & 0 & 035 & 0 & 747 & 0 & 029 \\
    Hogwarts Legacy & 0 & 697 & 0 & 057 & 0 & 824 & 0 & 036 & 0 & 751 & 0 & 088 & 0 & 784 & 0 & 039 & 0 & 512 & 0 & 087 & 0 & 871 & 0 & 055 & 0 & 729 & 0 & 058 & 0 & 784 & 0 & 051 & 0 & 635 & 0 & 053 & 0 & 656 & 0 & 029 \\
    Overcooked! 2 & 0 & 580 & 0 & 077 & 0 & 723 & 0 & 081 & 0 & 744 & 0 & 083 & 0 & 721 & 0 & 058 & 0 & 458 & 0 & 088 & 0 & 784 & 0 & 102 & 0 & 749 & 0 & 066 & 0 & 760 & 0 & 069 & 0 & 592 & 0 & 061 & 0 & 613 & 0 & 039 \\
    Gaming & 0 & 687 & 0 & 062 & 0 & 833 & 0 & 025 & 0 & 782 & 0 & 069 & 0 & 804 & 0 & 034 & 0 & 477 & 0 & 045 & 0 & 882 & 0 & 059 & 0 & 770 & 0 & 042 & 0 & 818 & 0 & 046 & 0 & 612 & 0 & 024 & 0 & 650 & 0 & 015 \\
    Combined Applications & 0 & 649 & 0 & 049 & 0 & 762 & 0 & 046 & 0 & 745 & 0 & 071 & 0 & 741 & 0 & 039 & 0 & 860 & 0 & 031 & 0 & 844 & 0 & 056 & 0 & 747 & 0 & 057 & 0 & 791 & 0 & 047 & 0 & 831 & 0 & 026 & 0 & 800 & 0 & 023 \\
    Complete & 0 & 705 & 0 & 060 & 0 & 823 & 0 & 050 & 0 & 765 & 0 & 081 & 0 & 799 & 0 & 035 & 0 & 865 & 0 & 034 & 0 & 850 & 0 & 079 & 0 & 747 & 0 & 066 & 0 & 793 & 0 & 068 & 0 & 836 & 0 & 037 & 0 & 827 & 0 & 029 \\
    \bottomrule
  \end{tabularx}
  \begin{minipage}{\linewidth}
    \vspace{2pt}\footnotesize\textit{Note.} Cells report the mean $\pm$ standard deviation of the test AUROC across the five folds. Rows index the training subset and columns the evaluation subset; the diagonal is in-distribution performance.
  \end{minipage}
\vspace{24pt}
\centering
  \setlength{\tabcolsep}{3pt}
  \tablefont
  \caption{Cross-domain AUROC for xLSTM-small (mean$\pm$SD, 5 folds).}
  \label{tab:cross-domain-auroc-xlstm}
  \begin{tabularx}{\linewidth}{X r@{.}l@{$\,{\pm}\,$}r@{.}l r@{.}l@{$\,{\pm}\,$}r@{.}l r@{.}l@{$\,{\pm}\,$}r@{.}l r@{.}l@{$\,{\pm}\,$}r@{.}l r@{.}l@{$\,{\pm}\,$}r@{.}l r@{.}l@{$\,{\pm}\,$}r@{.}l r@{.}l@{$\,{\pm}\,$}r@{.}l r@{.}l@{$\,{\pm}\,$}r@{.}l r@{.}l@{$\,{\pm}\,$}r@{.}l r@{.}l@{$\,{\pm}\,$}r@{.}l}
    \toprule
    Train & \multicolumn{4}{c}{\shortstack{n-back\\(Driving)}} & \multicolumn{4}{c}{\shortstack{n-back\\Dual\\Task}} & \multicolumn{4}{c}{\shortstack{n-back\\(Gaming)}} & \multicolumn{4}{c}{\shortstack{n-back\\(Combined)}} & \multicolumn{4}{c}{\shortstack{Driving\\Application}} & \multicolumn{4}{c}{\shortstack{Hogwarts\\Legacy}} & \multicolumn{4}{c}{\shortstack{Overcooked!\\2}} & \multicolumn{4}{c}{\shortstack{Gaming}} & \multicolumn{4}{c}{\shortstack{Combined\\Applications}} & \multicolumn{4}{c}{\shortstack{Complete}} \\
    \midrule
    n-back (Driving) & 0 & 836 & 0 & 041 & 0 & 772 & 0 & 078 & 0 & 583 & 0 & 083 & 0 & 699 & 0 & 062 & 0 & 518 & 0 & 159 & 0 & 596 & 0 & 108 & 0 & 542 & 0 & 129 & 0 & 559 & 0 & 089 & 0 & 550 & 0 & 090 & 0 & 560 & 0 & 043 \\
    n-back Dual Task & 0 & 723 & 0 & 073 & 0 & 874 & 0 & 024 & 0 & 690 & 0 & 110 & 0 & 801 & 0 & 032 & 0 & 367 & 0 & 096 & 0 & 853 & 0 & 062 & 0 & 647 & 0 & 075 & 0 & 734 & 0 & 061 & 0 & 482 & 0 & 046 & 0 & 557 & 0 & 032 \\
    n-back (Gaming) & 0 & 665 & 0 & 055 & 0 & 742 & 0 & 062 & 0 & 792 & 0 & 074 & 0 & 746 & 0 & 024 & 0 & 586 & 0 & 127 & 0 & 813 & 0 & 034 & 0 & 682 & 0 & 096 & 0 & 734 & 0 & 054 & 0 & 664 & 0 & 052 & 0 & 664 & 0 & 026 \\
    n-back (Combined) & 0 & 786 & 0 & 048 & 0 & 864 & 0 & 049 & 0 & 768 & 0 & 068 & 0 & 836 & 0 & 028 & 0 & 385 & 0 & 068 & 0 & 793 & 0 & 094 & 0 & 688 & 0 & 108 & 0 & 725 & 0 & 093 & 0 & 487 & 0 & 060 & 0 & 571 & 0 & 028 \\
    Driving Application & 0 & 776 & 0 & 069 & 0 & 784 & 0 & 095 & 0 & 587 & 0 & 071 & 0 & 733 & 0 & 049 & 0 & 868 & 0 & 028 & 0 & 665 & 0 & 111 & 0 & 613 & 0 & 098 & 0 & 637 & 0 & 094 & 0 & 776 & 0 & 048 & 0 & 757 & 0 & 034 \\
    Hogwarts Legacy & 0 & 729 & 0 & 071 & 0 & 828 & 0 & 036 & 0 & 751 & 0 & 104 & 0 & 785 & 0 & 038 & 0 & 511 & 0 & 072 & 0 & 854 & 0 & 069 & 0 & 723 & 0 & 069 & 0 & 774 & 0 & 057 & 0 & 634 & 0 & 048 & 0 & 653 & 0 & 030 \\
    Overcooked! 2 & 0 & 598 & 0 & 106 & 0 & 730 & 0 & 046 & 0 & 729 & 0 & 081 & 0 & 725 & 0 & 045 & 0 & 510 & 0 & 095 & 0 & 808 & 0 & 094 & 0 & 734 & 0 & 045 & 0 & 759 & 0 & 072 & 0 & 625 & 0 & 057 & 0 & 637 & 0 & 036 \\
    Gaming & 0 & 682 & 0 & 027 & 0 & 841 & 0 & 024 & 0 & 773 & 0 & 090 & 0 & 803 & 0 & 033 & 0 & 463 & 0 & 037 & 0 & 886 & 0 & 061 & 0 & 775 & 0 & 036 & 0 & 822 & 0 & 043 & 0 & 602 & 0 & 014 & 0 & 643 & 0 & 011 \\
    Combined Applications & 0 & 674 & 0 & 037 & 0 & 766 & 0 & 044 & 0 & 739 & 0 & 096 & 0 & 748 & 0 & 038 & 0 & 864 & 0 & 018 & 0 & 871 & 0 & 050 & 0 & 752 & 0 & 076 & 0 & 805 & 0 & 057 & 0 & 839 & 0 & 012 & 0 & 807 & 0 & 014 \\
    Complete & 0 & 689 & 0 & 065 & 0 & 832 & 0 & 060 & 0 & 745 & 0 & 075 & 0 & 792 & 0 & 039 & 0 & 860 & 0 & 015 & 0 & 849 & 0 & 070 & 0 & 738 & 0 & 072 & 0 & 789 & 0 & 068 & 0 & 833 & 0 & 021 & 0 & 822 & 0 & 019 \\
    \bottomrule
  \end{tabularx}
  \begin{minipage}{\linewidth}
    \vspace{2pt}\footnotesize\textit{Note.} Cells report the mean $\pm$ standard deviation of the test AUROC across the five folds. Rows index the training subset and columns the evaluation subset; the diagonal is in-distribution performance.
  \end{minipage}
\end{sidewaystable*}

\begin{sidewaystable*}
\centering
  \setlength{\tabcolsep}{3pt}
  \tablefont
  \caption{Cross-domain AUROC for ConvNeXt-small (mean$\pm$SD, 5 folds).}
  \label{tab:cross-domain-auroc-convnext}
  \begin{tabularx}{\linewidth}{X r@{.}l@{$\,{\pm}\,$}r@{.}l r@{.}l@{$\,{\pm}\,$}r@{.}l r@{.}l@{$\,{\pm}\,$}r@{.}l r@{.}l@{$\,{\pm}\,$}r@{.}l r@{.}l@{$\,{\pm}\,$}r@{.}l r@{.}l@{$\,{\pm}\,$}r@{.}l r@{.}l@{$\,{\pm}\,$}r@{.}l r@{.}l@{$\,{\pm}\,$}r@{.}l r@{.}l@{$\,{\pm}\,$}r@{.}l r@{.}l@{$\,{\pm}\,$}r@{.}l}
    \toprule
    Train & \multicolumn{4}{c}{\shortstack{n-back\\(Driving)}} & \multicolumn{4}{c}{\shortstack{n-back\\Dual\\Task}} & \multicolumn{4}{c}{\shortstack{n-back\\(Gaming)}} & \multicolumn{4}{c}{\shortstack{n-back\\(Combined)}} & \multicolumn{4}{c}{\shortstack{Driving\\Application}} & \multicolumn{4}{c}{\shortstack{Hogwarts\\Legacy}} & \multicolumn{4}{c}{\shortstack{Overcooked!\\2}} & \multicolumn{4}{c}{\shortstack{Gaming}} & \multicolumn{4}{c}{\shortstack{Combined\\Applications}} & \multicolumn{4}{c}{\shortstack{Complete}} \\
    \midrule
    n-back (Driving) & 0 & 780 & 0 & 096 & 0 & 634 & 0 & 138 & 0 & 683 & 0 & 120 & 0 & 644 & 0 & 102 & 0 & 585 & 0 & 051 & 0 & 604 & 0 & 118 & 0 & 599 & 0 & 071 & 0 & 593 & 0 & 078 & 0 & 603 & 0 & 046 & 0 & 589 & 0 & 049 \\
    n-back Dual Task & 0 & 753 & 0 & 072 & 0 & 865 & 0 & 039 & 0 & 787 & 0 & 069 & 0 & 824 & 0 & 045 & 0 & 394 & 0 & 051 & 0 & 823 & 0 & 066 & 0 & 711 & 0 & 105 & 0 & 754 & 0 & 084 & 0 & 505 & 0 & 035 & 0 & 581 & 0 & 019 \\
    n-back (Gaming) & 0 & 730 & 0 & 060 & 0 & 813 & 0 & 071 & 0 & 845 & 0 & 035 & 0 & 786 & 0 & 052 & 0 & 524 & 0 & 093 & 0 & 849 & 0 & 095 & 0 & 731 & 0 & 068 & 0 & 769 & 0 & 068 & 0 & 648 & 0 & 042 & 0 & 660 & 0 & 036 \\
    n-back (Combined) & 0 & 771 & 0 & 078 & 0 & 843 & 0 & 046 & 0 & 797 & 0 & 041 & 0 & 831 & 0 & 033 & 0 & 379 & 0 & 046 & 0 & 837 & 0 & 089 & 0 & 700 & 0 & 095 & 0 & 756 & 0 & 093 & 0 & 509 & 0 & 036 & 0 & 592 & 0 & 015 \\
    Driving Application & 0 & 778 & 0 & 047 & 0 & 748 & 0 & 090 & 0 & 675 & 0 & 080 & 0 & 716 & 0 & 058 & 0 & 851 & 0 & 018 & 0 & 602 & 0 & 097 & 0 & 577 & 0 & 094 & 0 & 592 & 0 & 083 & 0 & 738 & 0 & 021 & 0 & 733 & 0 & 015 \\
    Hogwarts Legacy & 0 & 709 & 0 & 047 & 0 & 831 & 0 & 041 & 0 & 792 & 0 & 081 & 0 & 798 & 0 & 041 & 0 & 519 & 0 & 107 & 0 & 896 & 0 & 046 & 0 & 752 & 0 & 033 & 0 & 809 & 0 & 029 & 0 & 647 & 0 & 060 & 0 & 665 & 0 & 029 \\
    Overcooked! 2 & 0 & 547 & 0 & 034 & 0 & 750 & 0 & 055 & 0 & 767 & 0 & 083 & 0 & 728 & 0 & 048 & 0 & 435 & 0 & 063 & 0 & 813 & 0 & 105 & 0 & 774 & 0 & 050 & 0 & 790 & 0 & 049 & 0 & 588 & 0 & 039 & 0 & 607 & 0 & 010 \\
    Gaming & 0 & 667 & 0 & 061 & 0 & 826 & 0 & 027 & 0 & 800 & 0 & 059 & 0 & 801 & 0 & 044 & 0 & 463 & 0 & 047 & 0 & 887 & 0 & 047 & 0 & 763 & 0 & 042 & 0 & 813 & 0 & 051 & 0 & 608 & 0 & 020 & 0 & 646 & 0 & 012 \\
    Combined Applications & 0 & 666 & 0 & 037 & 0 & 748 & 0 & 043 & 0 & 748 & 0 & 059 & 0 & 739 & 0 & 042 & 0 & 840 & 0 & 015 & 0 & 839 & 0 & 066 & 0 & 758 & 0 & 060 & 0 & 794 & 0 & 045 & 0 & 821 & 0 & 014 & 0 & 790 & 0 & 012 \\
    Complete & 0 & 713 & 0 & 058 & 0 & 805 & 0 & 057 & 0 & 807 & 0 & 070 & 0 & 808 & 0 & 041 & 0 & 845 & 0 & 050 & 0 & 863 & 0 & 113 & 0 & 759 & 0 & 068 & 0 & 805 & 0 & 086 & 0 & 830 & 0 & 057 & 0 & 825 & 0 & 045 \\
    \bottomrule
  \end{tabularx}
  \begin{minipage}{\linewidth}
    \vspace{2pt}\footnotesize\textit{Note.} Cells report the mean $\pm$ standard deviation of the test AUROC across the five folds. Rows index the training subset and columns the evaluation subset; the diagonal is in-distribution performance.
  \end{minipage}
\vspace{24pt}
\centering
  \setlength{\tabcolsep}{3pt}
  \tablefont
  \caption{Cross-domain AUROC for Transformer-small (mean$\pm$SD, 5 folds).}
  \label{tab:cross-domain-auroc-transformer}
  \begin{tabularx}{\linewidth}{X r@{.}l@{$\,{\pm}\,$}r@{.}l r@{.}l@{$\,{\pm}\,$}r@{.}l r@{.}l@{$\,{\pm}\,$}r@{.}l r@{.}l@{$\,{\pm}\,$}r@{.}l r@{.}l@{$\,{\pm}\,$}r@{.}l r@{.}l@{$\,{\pm}\,$}r@{.}l r@{.}l@{$\,{\pm}\,$}r@{.}l r@{.}l@{$\,{\pm}\,$}r@{.}l r@{.}l@{$\,{\pm}\,$}r@{.}l r@{.}l@{$\,{\pm}\,$}r@{.}l}
    \toprule
    Train & \multicolumn{4}{c}{\shortstack{n-back\\(Driving)}} & \multicolumn{4}{c}{\shortstack{n-back\\Dual\\Task}} & \multicolumn{4}{c}{\shortstack{n-back\\(Gaming)}} & \multicolumn{4}{c}{\shortstack{n-back\\(Combined)}} & \multicolumn{4}{c}{\shortstack{Driving\\Application}} & \multicolumn{4}{c}{\shortstack{Hogwarts\\Legacy}} & \multicolumn{4}{c}{\shortstack{Overcooked!\\2}} & \multicolumn{4}{c}{\shortstack{Gaming}} & \multicolumn{4}{c}{\shortstack{Combined\\Applications}} & \multicolumn{4}{c}{\shortstack{Complete}} \\
    \midrule
    n-back (Driving) & 0 & 821 & 0 & 044 & 0 & 792 & 0 & 053 & 0 & 603 & 0 & 100 & 0 & 738 & 0 & 027 & 0 & 489 & 0 & 083 & 0 & 639 & 0 & 092 & 0 & 572 & 0 & 057 & 0 & 598 & 0 & 073 & 0 & 531 & 0 & 070 & 0 & 563 & 0 & 042 \\
    n-back Dual Task & 0 & 765 & 0 & 047 & 0 & 886 & 0 & 018 & 0 & 701 & 0 & 078 & 0 & 812 & 0 & 017 & 0 & 369 & 0 & 113 & 0 & 830 & 0 & 067 & 0 & 685 & 0 & 115 & 0 & 739 & 0 & 080 & 0 & 493 & 0 & 056 & 0 & 567 & 0 & 040 \\
    n-back (Gaming) & 0 & 667 & 0 & 079 & 0 & 766 & 0 & 044 & 0 & 798 & 0 & 041 & 0 & 758 & 0 & 038 & 0 & 472 & 0 & 068 & 0 & 791 & 0 & 100 & 0 & 693 & 0 & 088 & 0 & 723 & 0 & 070 & 0 & 592 & 0 & 045 & 0 & 626 & 0 & 027 \\
    n-back (Combined) & 0 & 779 & 0 & 042 & 0 & 874 & 0 & 032 & 0 & 738 & 0 & 052 & 0 & 838 & 0 & 018 & 0 & 397 & 0 & 065 & 0 & 783 & 0 & 099 & 0 & 627 & 0 & 106 & 0 & 701 & 0 & 092 & 0 & 472 & 0 & 038 & 0 & 561 & 0 & 024 \\
    Driving Application & 0 & 789 & 0 & 080 & 0 & 802 & 0 & 096 & 0 & 602 & 0 & 099 & 0 & 727 & 0 & 078 & 0 & 877 & 0 & 028 & 0 & 629 & 0 & 091 & 0 & 561 & 0 & 109 & 0 & 584 & 0 & 092 & 0 & 769 & 0 & 022 & 0 & 752 & 0 & 034 \\
    Hogwarts Legacy & 0 & 655 & 0 & 022 & 0 & 812 & 0 & 038 & 0 & 710 & 0 & 074 & 0 & 768 & 0 & 039 & 0 & 507 & 0 & 099 & 0 & 865 & 0 & 052 & 0 & 713 & 0 & 069 & 0 & 767 & 0 & 061 & 0 & 626 & 0 & 059 & 0 & 651 & 0 & 033 \\
    Overcooked! 2 & 0 & 596 & 0 & 080 & 0 & 690 & 0 & 125 & 0 & 738 & 0 & 098 & 0 & 709 & 0 & 084 & 0 & 428 & 0 & 095 & 0 & 733 & 0 & 107 & 0 & 739 & 0 & 099 & 0 & 732 & 0 & 083 & 0 & 563 & 0 & 076 & 0 & 596 & 0 & 052 \\
    Gaming & 0 & 712 & 0 & 088 & 0 & 832 & 0 & 027 & 0 & 773 & 0 & 069 & 0 & 808 & 0 & 032 & 0 & 505 & 0 & 044 & 0 & 872 & 0 & 077 & 0 & 774 & 0 & 055 & 0 & 818 & 0 & 055 & 0 & 627 & 0 & 032 & 0 & 660 & 0 & 017 \\
    Combined Applications & 0 & 606 & 0 & 048 & 0 & 773 & 0 & 057 & 0 & 748 & 0 & 069 & 0 & 737 & 0 & 046 & 0 & 877 & 0 & 043 & 0 & 821 & 0 & 052 & 0 & 730 & 0 & 040 & 0 & 773 & 0 & 041 & 0 & 835 & 0 & 041 & 0 & 803 & 0 & 036 \\
    Complete & 0 & 714 & 0 & 068 & 0 & 832 & 0 & 038 & 0 & 744 & 0 & 094 & 0 & 798 & 0 & 029 & 0 & 888 & 0 & 015 & 0 & 838 & 0 & 062 & 0 & 745 & 0 & 070 & 0 & 784 & 0 & 061 & 0 & 846 & 0 & 029 & 0 & 833 & 0 & 021 \\
    \bottomrule
  \end{tabularx}
  \begin{minipage}{\linewidth}
    \vspace{2pt}\footnotesize\textit{Note.} Cells report the mean $\pm$ standard deviation of the test AUROC across the five folds. Rows index the training subset and columns the evaluation subset; the diagonal is in-distribution performance.
  \end{minipage}
\end{sidewaystable*}

\FloatBarrier

\clearpage
\ifdefined\forceputbib
  \putbib[bibliography]
\else
  \IfFileExists{bu1.bbl}{\putbib[bibliography]}{
}
\fi
\end{bibunit}

\clearpage
\onecolumn
\section{Datasheets}
\label{sec:supp-datasheets}
Figures~\ref{fig:data-card-gaming} and~\ref{fig:data-card-driving} summarize the gaming and driving datasets of
\gls{revelio} as data cards styled after nutrition labels.

\begin{figure}[H]
\begin{minipage}[t]{0.485\textwidth}\vspace{0pt}
    \centering
    \resizebox{\linewidth}{!}{
    
    \sffamily
    {
    \fbox{%
    \begin{tabular}{@{}p{\NFwidth}@{}}
    
    \multicolumn{1}{c}{\huge\bfseries Cognitive Load: Gaming}\\
    \NFrule
    \NFtext{\textbf{Dataset} Cognitive Load Gaming}\\
    \NFtext{\textbf{Year of Creation} 2023}\\
    \NFtext{\textbf{Instances Per Dataset} 30 Public, 15 Private}\\
    \NFtext{\textbf{DOI (Public)} https://doi.org/10.5281/zenodo.24060698}\\
    \NFtext{\textbf{DOI (Private)} https://doi.org/10.5281/zenodo.24076078}\\
    
    \NFRULE
    \NFline{Motivation}\\
    \NFrule
    \NFelement{Original Authors}{}{Maximilian P. Oppelt et al.}\\
    \NFelement{Original Use Case}{}{Cognitive Load Detection}\\
    \NFelement{Original Funding}{}{Digital Health and Analytics, Fraunhofer IIS}\\
    
    \NFRule
    \NFline{Composition}\\
    \NFrule
    \NFelement{Sample or Complete}{}{Sample (Public and Private Split)}\\ 
    \NFelement{Missing Data}{}{Some records with noise, lead falloffs and closed eyes}\\
    \NFelement{Sensitive Information }{}{No image/video data publicly available}\\

    \NFRule
    \NFline{Collection}\\
    \NFrule
    \NFelement{Ethical Review}{}{129\_21 B on 21.04.2021}\\
    \NFelement{Ethics Committee}{}{Friedrich-Alexander-University Erlangen-Nuremberg}\\

    \NFRule
    \NFline{Cleaning and Labeling}\\
    \NFrule
    \NFelement{Cleaning Done}{}{2024-01-01}\\
    \NFelement{Labeling Done}{}{2024-01-01}\\

    \NFRule
    \NFline{Uses and Distribution}\\
    \NFrule
    \NFelement{Notable Uses}{}{Cognitive Load detection systems}\\
    \NFelement{Other Uses}{}{None}\\

    \NFRule
    \NFline{Maintenance and Evolution}\\
    \NFrule
    \NFelement{Corrections or Erratum}{}{None}\\
    \NFelement{Methods to Extend}{}{None}\\
    \NFelement{Replicate Maintainers}{}{Maximilian P. Oppelt}\\
    
    \NFRULE
    \NFline{Breakdown}\\
    \NFrule
    \NFelement{Physiological Measurements}{}{\gls{ecg} lead I and II, \gls{ppg}, \gls{emg}, respiration from chest belt, skin temperature}\\
    \NFrule
    \NFelement{Fiducial Points}{}{Extracted \gls{ecg} R-peaks, \gls{ppg} peaks, respiration peaks and throughts, \gls{eda} onsets, peaks and recovery}\\
    \NFrule
    \NFelement{Action Units}{}{Estimations of activation of the action units of 1, 2, 4, 5, 6, 7, 9, 10, 11, 12, 14, 15, 17, 20, 23, 24, 25, 26, 28, 43}\\
    \NFrule
    \NFelement{Basic Emotions}{}{Anger, Fear, Happiness, Sadness, Surprise, neutral, pitch}\\
    \NFrule
    \NFelement{Facial Properties}{}{Pitch, Roll, Yaw, X, Y, Width, Height, Detection score}\\
    \NFrule
    \NFelement{Body Posture and Movement}{}{Estimations of X, Y, Z, visibility and presence of Nose, Left and right eye, Left and right ear, Left and right shoulder,  Left and right elbow, Left and right wrist, Left and right hip, Left and right knee, Left and right ankle,}\\
    \NFrule
    \NFelement{Eye Tracker}{}{Estimation of X, Y, Z and validity of the gaze in user and tracebox coordinates for both eyes and diameter and validity of the left and right pupil}\\
    \NFrule
    \NFelement{Meta-Data}{}{PSS, BFI-K, POGQ-SF, ADS-L, Fear of flying, claustrophobia and acrophobia, age, sex, height, weight, marital status, job}\\
    \NFrule
    \NFelement{Further Questionnaires}{}{Gaming Habits, favorite game genre rating, gaming self-assessment abilities}\\
    \NFrule
    \NFelement{Performance}{}{Recall for n-back scores, missed rings, race time, total time, successful recipes, true and false positive hit rate, mean reaction time}\\
    \NFrule
    \NFelement{Subjective Rating}{}{NASA-TLX rating for effort, mental, performance, physical and temporal domains and the corresponding weights.}\\
  \end{tabular}}
    
    \par} 
  }
        \caption{A data card styled after nutrition labels. Description of newly released gaming use case for cognitive load detection.}
    \label{fig:data-card-gaming}
\end{minipage}\hfill
\begin{minipage}[t]{0.485\textwidth}\vspace{0pt}
  \centering
  \resizebox{\linewidth}{!}{
  
  \sffamily
  {
  \fbox{%
  \begin{tabular}{@{}p{\NFwidth}@{}}
  
  \multicolumn{1}{c}{\huge\bfseries Cognitive Load: Driving}\\
  \NFrule
  \NFtext{\textbf{Dataset} Cognitive Load Driving}\\
  \NFtext{\textbf{Year of Creation} 2022}\\
  \NFtext{\textbf{Instances Per Dataset} 30 Public, 21 Private}\\
  \NFtext{\textbf{DOI (Public)} https://doi.org/10.5281/zenodo.24051552}\\
  \NFtext{\textbf{DOI (Private)} https://doi.org/10.5281/zenodo.24059812}\\
  
  \NFRULE
  \NFline{Motivation}\\
  \NFrule
  \NFelement{Original Authors}{}{Maximilian P. Oppelt et al.}\\
  \NFelement{Original Use Case}{}{Cognitive Load Detection}\\
  \NFelement{Original Funding}{}{Digital Health and Analytics, Fraunhofer IIS}\\
  
  \NFRule
  \NFline{Composition}\\
  \NFrule
  \NFelement{Sample or Complete}{}{Sample (Public and Private Split)}\\ 
  \NFelement{Missing Data}{}{Some records with noise, lead falloffs and closed eyes}\\
  \NFelement{Sensitive Information }{}{No image/video data publicly available}\\

  \NFRule
  \NFline{Collection}\\
  \NFrule
  \NFelement{Ethical Review}{}{129\_21 B on 21.04.2021}\\
  \NFelement{Ethics Committee}{}{Friedrich-Alexander-University Erlangen-Nuremberg}\\

  \NFRule
  \NFline{Cleaning and Labeling}\\
  \NFrule
  \NFelement{Cleaning Done}{}{2023-01-01}\\
  \NFelement{Labeling Done}{}{2023-01-01}\\

  \NFRule
  \NFline{Uses and Distribution}\\
  \NFrule
  \NFelement{Notable Uses}{}{https://doi.org/10.3390/s23010340}\\
  \NFelement{Other Uses}{}{https://doi.org/10.3389/fcomp.2024.1371181}\\

  \NFRule
  \NFline{Maintenance and Evolution}\\
  \NFrule
  \NFelement{Corrections or Erratum}{}{None}\\
  \NFelement{Methods to Extend}{}{None}\\
  \NFelement{Replicate Maintainers}{}{Maximilian P. Oppelt}\\
  \NFelement{Updates}{}{\\ - Added estimation of body posture and movement}\\
  
  \NFRULE
  \NFline{Breakdown}\\
  \NFrule
  \NFelement{Physiological Measurements}{}{\gls{ecg} lead I and II, \gls{ppg}, \gls{emg}, respiration from chest belt, skin temperature}\\
  \NFrule
  \NFelement{Fiducial Points}{}{Extracted \gls{ecg} R-peaks, \gls{ppg} peaks, respiration peaks and throughts, \gls{eda} onsets, peaks and recovery}\\
  \NFrule
  \NFelement{Action Units}{}{Estimations of activation of the action units of 1, 2, 4, 5, 6, 7, 9, 10, 11, 12, 14, 15, 17, 20, 23, 24, 25, 26, 28, 43}\\
  \NFrule
  \NFelement{Basic Emotions}{}{Anger, Fear, Happiness, Sadness, Surprise, neutral, pitch}\\
  \NFrule
  \NFelement{Facial Properties}{}{Pitch, Roll, Yaw, X, Y, Width, Height, Detection score}\\
  \NFrule
  \NFelement{Body Posture and Movement}{}{Estimations of X, Y, Z, visibility and presence of Nose, Left and right eye, Left and right ear, Left and right shoulder,  Left and right elbow, Left and right wrist, Left and right hip, Left and right knee, Left and right ankle,}\\
  \NFrule
  \NFelement{Eye Tracker}{}{Estimation of X, Y, Z and validity of the gaze in user and tracebox coordinates for both eyes and diameter and validity of the left and right pupil}\\
  \NFrule
  \NFelement{Meta-Data}{}{PSS, BFI-K, ADS-L, Driver's license and experience, age, sex, height, weight, marital status, job}\\
  \NFrule
  \NFelement{Performance}{}{Recall for n-back scores, true and false positive hit rate, mean reaction time for both visual and auditive responses, number of added songs}\\
  \NFrule
  \NFelement{Subjective Rating}{}{NASA-TLX rating for effort, mental, performance, physical and temporal domains and the corresponding weights.}\\
\end{tabular}}
  
  \par} 
}
  
  \caption{Dataset card for the driving use case for cognitive load detection, originally released in early 2023. Added body posture and movement estimation in 2023.}
  \label{fig:data-card-driving}
\end{minipage}
\end{figure}

\end{document}